\documentclass{article}

\PassOptionsToPackage{numbers}{natbib} % For numeric citations with et al.
\usepackage[preprint]{neurips_2025}
\usepackage{listings}
\usepackage{blindtext}
\usepackage{enumitem}
\usepackage{subcaption}

\usepackage[utf8]{inputenc} % allow utf-8 input
\usepackage[T1]{fontenc}    % use 8-bit T1 fonts
\usepackage{hyperref}       % hyperlinks
\usepackage{url}            % simple URL typesetting
\usepackage{booktabs}       % professional-quality tables
\usepackage{amsfonts}       % blackboard math symbols
\usepackage{nicefrac}       % compact symbols for 1/2, etc.
\usepackage{microtype}      % microtypography
\usepackage{xcolor}         % colors
\usepackage{float}          % for placing figures

\usepackage{makecell}
\usepackage{hhline}
\usepackage{multirow}
\usepackage{xcolor}
\usepackage{amsthm, amsmath}
\usepackage{tcolorbox} %% for vlm box 

\usepackage{pifont}    % For symbols
\newcommand{\cmark}{\ding{51}}  % Check mark
\newcommand{\xmark}{\ding{55}}  % X mark

\newtcolorbox{vlmcasebox}{
  colback=gray!10,
  colframe=black!50,
  boxrule=0.4pt,
  arc=2pt,
  left=6pt, right=6pt, top=6pt, bottom=6pt
}

\usepackage{tcolorbox}
\newtcolorbox{mycase}[1]{
  colback=#1,
  colframe=white,
  boxrule=1pt, % Adjust the border thickness
  left=1pt,
  right=1pt,
  top=1pt,
  bottom=1pt,
}
\newtcolorbox{question}[1]{
  colback=#1, % Set background color to the passed parameter
  colframe=#1, % Set frame color to the passed parameter
  boxrule=0pt,
  left=1pt,
  right=1pt,
  top=1pt,
  bottom=1pt,
  fonttitle=\bfseries\color{black}, % Set title font color to black
  title=Question
}
\newtcolorbox{answer}[1]{
  colback=#1, % Set background color to the passed parameter
  colframe=#1, % Set frame color to the passed parameter
  boxrule=0pt,
  left=1pt,
  right=1pt,
  top=1pt,
  bottom=1pt,
  fonttitle=\bfseries\color{black}, % Set title font color to black
  title=Answer
}

\definecolor{my_green}{RGB}{51,102,0}
\definecolor{my_yellow}{RGB}{255,165,0}
\definecolor{my_red}{RGB}{204, 0, 0}
\definecolor{light_pink}{RGB}{250,245,247}
\definecolor{light_blue}{RGB}{240,245,255}
\definecolor{light_green}{RGB}{240,255,240}
\definecolor{light_yellow}{RGB}{255,255,240}
\definecolor{light_grey}{RGB}{240,240,240}

\usepackage[colorinlistoftodos,textwidth=2.3cm]{todonotes}

\title{BYO-Eval: Build Your Own Dataset for Fine-Grained Visual Assessment of Multimodal Language Models}

\author{%
  Ludovic Arnould$^{*}$ \\
  R\&D Center, Talan, Paris \\
  \texttt{ludovic.arnould@talan.com} \\
  \And
  Salim Khazem$^{*}$ \\
  R\&D Center, Talan, Paris \\
  \And
   Hugues Ali Mehenni \\
   R\&D Center, Talan, Paris \\
}

\begin{document}

\maketitle

\begingroup
\renewcommand\thefootnote{*}
\footnotetext{Main contributor}
\endgroup

\begin{abstract}
Visual Language Models (VLMs) are now sufficiently advanced to support a broad range of applications, including answering complex visual questions, and are increasingly expected to interact with images in varied ways.
To evaluate them, current benchmarks often focus on specific domains (e.g., reading charts), constructing datasets of annotated real images paired with pre-defined Multiple Choice Questions (MCQs) to report aggregate accuracy scores. 
However, such benchmarks entail high annotation costs, risk information leakage, and do not clarify whether failures stem from limitations in visual perception, reasoning, or general knowledge. 
We propose a new evaluation methodology, inspired by ophthalmologic diagnostics, leveraging procedural generation of synthetic images to obtain control over visual attributes and precisely reveal perception failures in VLMs.
Specifically, we build collections of images with gradually more challenging variations in the content of interest (e.g., number of objects in a counting task) while holding other visual parameters constant. This diagnostic allows systematic stress testing and fine-grained failure analysis, shifting the focus from coarse benchmarking toward targeted and interpretable assessment of VLM capabilities. Our code is available at \hyperlink{https://github.com/byoeval/BYO-EVAL}{https://github.com/byoeval/BYO-EVAL}.

%Visual Language Models (vLMs) are now sufficiently advanced to support a broad range of applications, including answering complex visual questions, and are increasingly expected to interpret and interact with images in varied and sophisticated ways. To evaluate them, current benchmarks often focus on specific domains (e.g., reading charts), constructing datasets of annotated real images paired with pre-defined Multiple Choice Questions (MCQs) to report aggregate accuracy scores. However, such benchmarks entail high annotation costs, risk information leakage, and do not clarify whether failures stem from limitations in visual perception, reasoning, or general knowledge. We propose a new evaluation methodology, inspired by ophthalmologic diagnostics, leveraging procedural generation of synthetic images to obtain control over visual attributes and precisely reveal perception failures in vLMs. Specifically, we construct collections of images with gradually more challenging variations in the content of interest (e.g., number of objects in a counting task) while holding other visual parameters constant. This diagnostic allows systematic stress testing and fine-grained failure analysis, shifting the focus from coarse benchmarking toward targeted and interpretable assessment of vLM capabilities.

% TL DR : We propose an ophthalmology-inspired evaluation method using procedurally-generated images to systematically diagnose and precisely pinpoint visual perception failures in visual language models (VLMs)

\end{abstract}

\section{Introduction}
\label{sec:intro}
\begin{figure}[t]
  \centering
  \includegraphics[width=\linewidth]{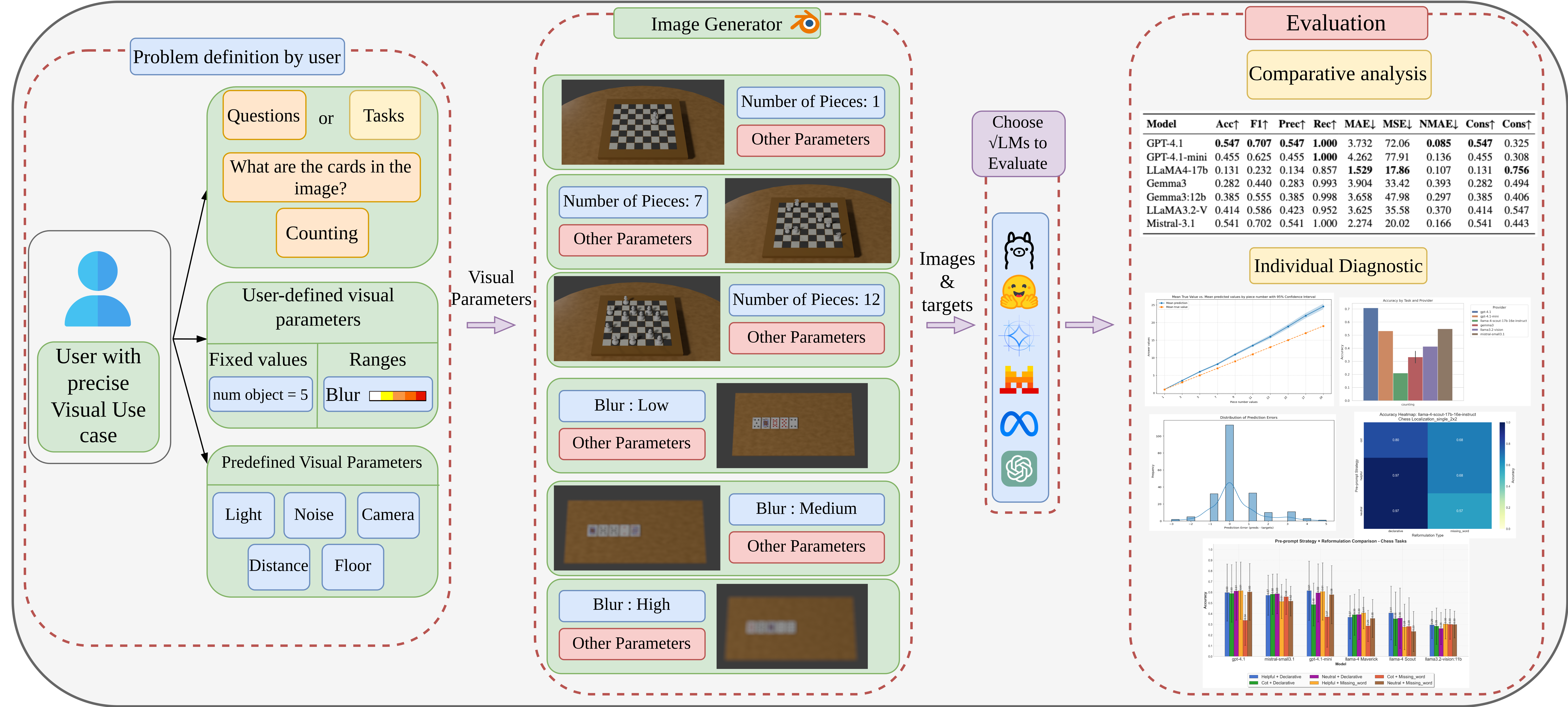}
  \caption{
    Overview of our evaluation framework. To start, a user indicates in \texttt{yaml} configurations the variables to test and their range (e.g., number of objects, blur level) and other fixed values of visual parameters (camera distance, etc.) (first panel). 
    Via Blender Python API (\texttt{bpy}), we generate from these input configurations synthetic scenes with their associated legends (middle panel).
    The resulting images and legends are input into various VLMs (e.g., GPT-4.1, Gemma, ...) for evaluation.
    Outputs are assessed through multi-dimensional analyses and detailed statistical metrics (right panel).
  }
  \label{fig:VLM_framework}
\end{figure}

%%%% Proposition de reformulation (Ludovic qu'est ce que t'en penses ? ) 
Vision-language models have achieved remarkable success in a variety of multi-modal tasks \cite{bai2024survey}. They can generate fluent image descriptions and answer complex visual questions, demonstrating capabilities far beyond early systems~\cite{campbell2024understanding}. %These models are even being deployed in diverse applications ranging from robotics and autonomous agents to image editing tools~\cite{chatterjee2024revision}. 
%However, despite this impressive progress, there remain clear gaps in their visual reasoning abilities. In fact, even state-of-the-art VLMs still exhibit surprising failures on basic multi-object reasoning tasks, for example, miscounting objects in a scene or misidentifying simple spatial relationships tasks that humans find trivial~\cite{campbell2024understanding}. %Such failure modes have been observed across various studies and benchmarks, which report that contemporary models often cannot correctly reason about spatial arrangements or perform precise counting in images. \sk{Maybe include an illustrative example of a model’s failure here?} 
Despite their progress, State-of-The-Art (SoTA) VLMs still show notable shortcomings in basic visual reasoning tasks, such as miscounting objects or incorrectly identifying simple spatial relationships \cite{rahmanzadehgervi2024vision, campbell2024understanding}, limiting their adoption in industrial contexts \cite{radsch2025bridging}. 

%The assessment of such failures, and the evaluation of VLM in general remains very challenging. Several factors contribute to this challenge, the main one being the absence of metrics (or labels) to quantify the alignment quality of outputs across modalities such as text and images with respect to provided instructions. 
Evaluating these flaws—and VLM performance more broadly—remains difficult, primarily due to the lack of reliable metrics to quantify cross-modal alignment between images, text, and task instructions.
Other reasons, including the complexity of the model or the diversity of tasks to evaluate, have led the community to design comprehensive evaluation datasets that provide a one-size-fits-all score, averaging performance across many examples \cite{duan2024VLMevalkit, al2024unibench}.

These benchmarks often aim to globally measure the performance of VLMs against human expectations across numerous use cases \cite{al2024unibench, fu2024mme, li2025benchmark}, assessing high-level abilities of VLMs without precisely distinguishing the visual, textual, or reasoning abilities. Indeed, basic perception weaknesses are masked when looking only at aggregate performance: a model might achieve high overall accuracy on broad benchmarks while consistently stumbling on certain types of questions. The success of the Arena Leaderboard \cite{chiang2024chatbot} is symptomatic of the lack of standardized criteria or precised, tailored evaluation, as users compare two anonymized Large Language Model (LLM) or VLM outputs based on a subjective choice of the "best" answer.
Moreover, general benchmarks can suffer from several flaws: \cite{chen2024we} showed that many visual datasets are corrupted in the sense that VLMs can answer many questions without seeing the image. Moreover, results of proprietary methods are hard to reproduce~\cite{schuhmann2022laion5b,cherti2023scaling} and benchmarks can be outdated rapidly due to the fast progress of VLMs \cite{ott2022benchmark,hsieh2023sugarcrepe}.

%These weaknesses are masked when looking only at aggregate performance: a model might achieve high overall accuracy on broad benchmarks while consistently stumbling on certain types of questions.

%One reason these blind spots persist is that current evaluation benchmarks are not sufficiently diagnostic. Most popular vision-language benchmarks (e.g. Visual Question Answering and image captioning datasets) provide a one-size-fits-all score that averages performance across many examples. While useful as coarse measures, such evaluations often fail to isolate specific reasoning skills. As a result, models can exploit dataset biases or spurious correlations to “succeed” without truly understanding the visual content~\cite{wang2025pulsecheck457}. For instance, real-world image datasets tend to exhibit strong biases in object viewpoints and contexts: over $70\%$ of objects in some 3D scene datasets appear in a single dominant orientation. A VLM might learn to rely on these statistical regularities (e.g. assuming a certain object is always upright or in a typical position) rather than developing a genuine 3D understanding. Similarly, many VQA models have been shown to overly rely on language priors in questions, leading to incorrect answers when faced with unusual phrasing or less common scenarios. In summary, existing benchmarks often paint an incomplete picture of a model’s capabilities, making it difficult to pinpoint what a VLM truly knows or where it fails.

To address these shortcomings, we propose a new evaluation framework of VLM that is both diagnostic and task-specific, leveraging procedurally generated synthetic images, using Blender \cite{blender}, to systematically test them on targeted visual skills.
Blender allows us to generate many image variations with precise control over scene attributes—such as object types, counts, and positions—enabling a dynamic, scalable benchmark with built-in ground-truth annotations.
%By using a 3D rendering pipeline, we can create infinite variations of images with precise control over scene content and attributes (object types, counts, positions, lighting, etc.). 
%This gives us a dynamic and scalable benchmark where each test image comes with ground-truth knowledge of the scene.
Crucially, we design each set of image to evaluate a particular capability of the model, varying the variable of interest while keeping other visual parameters constant, akin to ophthalmologic diagnostics.
For example, one task may challenge the model’s counting ability by varying the number of objects and asking “How many objects are in the image?”, while another task focuses on spatial relations by asking about the relative positions of objects.
%We can ensure a wide coverage of scenarios – including edge cases and rare situations – which would be nearly impossible to curate at scale with real images~\cite{wang2025pulsecheck457}.
Our framework thus serves as a diagnostic toolkit: by examining performance on each task in isolation, we can precisely reveal a profile of a VLM's visual abilities, unlike a single aggregate score. 
%Unlike conventional benchmarks that yield a single aggregate score, our task-specific evaluation reveals a profile of the model’s abilities (for instance, a model might excel at color recognition but struggle with counting). 

%The results show that our approach highlights significant model limitations that are not evident from traditional evaluations. For example, we find that models with near-human performance on standard benchmarks can still drastically underperform on tasks like fine-grained spatial reasoning.
%\la{Mettre des insights cool ici. Indiquer claire différence en perf entre Poker et Chess, ce qui joue en faveur de datasets spécialisés d'évaluation.}

%% Sk: Premiere proposition 
%Overall, our diagnostic evaluation methodology reveals capability gaps hiding by traditional benchmarks and offers  a principled way to assess real-world readiness. The difference in model behavior across Poker and Chess tasks, for instance, demonstrates that general benchmarks miss task and context-specific failures, further reinforcing the value of controlled, low-level assessments. 

%% Deuxieme proposition 
To validate the design and implementation of our evaluation framework, we present an in-depth analysis of several leading VLMs. We generate a diverse set of Poker and Chess images in Blender for a range of evaluation scenarios.
%Overall, our diagnostic evaluation methodology reveals significant capability gaps that are obscured by traditional aggregate benchmarks and offers a principled way to assess real-word readiness.
Overall, we observe similar limitations in terms of counting, localizing or identifying objects compared to previous work \cite{campbell2024understanding, rahmanzadehgervi2024vision,wang2024picture, zhang2024good}, for tasks that are easy for humans. Strikingly, performance varies substantially across domains: models that perform well on chess tasks often fail to generalize to visually similar poker scenes, where occlusion and less rigid layouts are introduced. Such findings support our central claim: general-purpose benchmarks mask brittle behaviors that only emerge under controlled diagnostic stress. We summarize a few key observations regarding the behavior of GPT-4.1 \cite{gpt4.1} and LLaMA-4-Scout \cite{llama4} to illustrate the insights of a fine-grained analysis (\textit{cf}. Appendix \ref{sec:diagnostic_additional}):
\begin{itemize}
\item \textbf{Counting}: GPT-4.1 is very accurate up to 5 pieces but tends to slightly overestimate the count of large object numbers in the Chess dataset, while significantly underestimating the count of Poker cards. Notably, these biases appear to be model-dependent, as LLaMA-4-Scout demonstrates a minor tendency to underestimate object counts in the Chess dataset.
\item \textbf{Counting with blur}: Both GPT-4.1 and LLaMA-4-Scout exhibit an overestimation pattern in object counting as blur increases, which may be explained by the fact that higher levels of blur can merge pixels, creating the illusion of objects where none exist.
\item \textbf{Localization of a single Chess piece on a 4x4 board}: This experiment highlights the contextual bias exhibited by the LLaMA-4-Scout model, as it consistently predicts out-of-bounds rows and columns, likely inferring a standard 8x8 chessboard.
\item \textbf{Localization of a single card on a 3x3 Poker grid}: Both models demonstrate a decline in performance when the element is positioned in the middle column of the poker scene.
\item \textbf{Relative localization of two Chess pieces on an 8x8 grid}: Both models' performance declines rapidly as the row distance between the two pieces increases, with a clear underestimation of their relative distance.
\item \textbf{Identification with camera distance (Chess)}: As anticipated, the performance appears to decline as the distance to the chess scene increases.
\item \textbf{Counting with Horizontal Overlap (Poker cards)}: Both models tend to underestimate the number of cards with horizontal overlap. Additionally, LLaMA-4-Scout seems to particularly struggle with the concept of cards encroaching upon one another, as its underestimation steadily increases with the level of overlap.
\end{itemize}

%These findings highlight the urgent need for more fine-grained evaluation: without it, progress in VLM development may be misinterpreted or overstated. 

%We hope that our benchmark will not only serve as a more rigorous testbed for current models, but also guide the community towards building more robust and truly understanding vision-language systems.

\paragraph{Contributions} Our work makes the following contributions: (1) We propose a novel evaluation framework for VLM that uses Blender-generated synthetic images to achieve controlled, scalable testing of specific visual skills. (2) We develop a suite of diagnostic tasks (object recognition, counting, localization, etc.) along with an open-source pipeline to generate labeled images and questions for each task. (3) Through extensive experiments on eight SoTA VLMs, we reveal failure modes and fine-grained performance differences between models, providing new insights into the current state of VLM capabilities. 
(4) Finally, we release our evaluation toolkit publicly, enabling researchers to create new test cases and extend our tasks for probing additional capabilities.
We believe this task-specific diagnostic approach will complement existing benchmarks and facilitate the identification of VLM shortcomings in real-life scenarios.

\section{Background: evaluation of VLMs}
\label{sec:background}
VLMs are generally composed of a visual encoder, a textual decoder and a module to align both modalities in the textual space \cite{liu2023visual, bai2024survey}. The combination of modalities adds an additional layer of complexity towards a precise assessment of performances, as it becomes harder to detect where the model fails. While the training metrics used for foundational VLMs—typically the accuracy of predicting masked tokens—are straightforward, the broad spectrum of inference applications, combined with a widespread adoption beyond deep learning practitioners, has increased the number of tasks to evaluate.
This has prompted the development of comprehensive question-answering benchmarks, such as MMMU \cite{mmmu} rather than "low-level" evaluations (e.g., counting objects).
%e.g.\ whether the visual information required to answer a question is indeed included in the visual representation, properly transmitted to the textual space, and finally properly processed by the textual module.
%Recently, multi-modal foundation models have been designed from scratch \textbf{[refs]}, but this does not solve the problem as ...
%that we use today are fine-tuned version of self-supervised foundation models which acquire a broad understanding of data modalities and subsequently apply this knowledge across diverse tasks \cite{bai2024survey}. They

We define \textbf{low-level questions} as those focused on basic perceptual tasks—counting, localization, or identification \footnote{Identification could be broken down into more fundamental visual skills like measuring, etc. }—which form the foundation of image understanding. However, ambiguity may arise: a seemingly low-level question like "How many dogs are in the image?" becomes cross-task if the image includes visually similar objects (e.g., wolves). Conversely, \textbf{high-level tasks} combine multiple perceptual abilities and may require reasoning or external knowledge. Examples include "celebrity recognition", "future prediction", "image topic", or "natural relation" \cite{liu2023mmbench}. We also classify long-form or cross-task questions (e.g., "What is the time on bottom middle phone?" \cite{singh2019towards}) as high-level.

\paragraph{High-level benchmarks}
We refer to \cite{fu2024mme, li2025benchmark} for comprehensive surveys and provide a summarized analysis. High-level benchmarks aim to globally measure the performance of VLMs against human expectations across numerous use cases.  
Images are generally paired with multiple-choice \cite{li2023seed} or short-answer
%(20 benchmarks out of the 45 representative ones analyzed in \cite{fu2024mme}, 22 out of 37 in \cite{li2025benchmark}, including yes-no questions) 
questions \cite{goyal2017making, singh2019towards}. They report high-level metrics such as the proportion of correct answers but offer limited insight into each model’s visual capabilities and weaknesses. Aggregation of benchmarks \cite{al2024unibench, duan2024VLMevalkit} further decrease the interpretability of the overall scores. 
%Due to the popularity of certain benchmarks \textbf{[refs]}, VLM releases often simply cite them while claiming State-of-the-Art (SoTA) results in their weight class\textbf{[refs]}. Yet, they often lack detailed explanations of the models' strengths and limitations, hindering their practical use \cite{radsch2025bridging}.
Furthermore, this dynamic reinforces a winner-takes-all logic, where dominant models overshadow alternatives that may offer better trade-offs in efficiency, domain adaptation, or task specialization. The centralization of attention on a handful of leading models also raises concerns about the reproducibility and transparency of evaluations, as the reported results of leading models can depend on undocumented fine-tuning processes or proprietary datasets that are inaccessible \cite{guo2025deepseek, grattafiori2024llama, yang2024qwen2, achiam2023gpt}.

%Recently, the emergence of specialized benchmarks tailored to specific domains such as mathematics or programming tackles the problem of choosing a VLM or an LLM for a given application \textbf{[refssssss]}. These domain-specific evaluations aim to gauge the capabilities of generative models within practical and realistic industrial contexts, but still at a "high-level", like Q\&A in image understanding, or "debugging" in coding, without focusing on a textual, visual, or thinking VLM ability.

High-level benchmarks often face limitations that undermine their reliability. As noted by \cite{chen2024we}, VLMs can achieve over 40\% accuracy on multiple-choice benchmarks like MMMU \cite{mmmu} \textit{without} processing the image—due to either data leakage (e.g., test samples seen during training) or questions that can be answered without visual input. Surprisingly, few papers question the bias introduced by the multiple-choice format itself, which may already guides the model toward the correct answer.
%—especially when other choices are either too similar or too dissimilar—typically among only four options. 
Moreover, since it's impossible to fully constrain a VLM's answer format, some methods use another LLM to match answers to MCQ choices \cite{li2023seed, liu2023mmbench} or compare sequences for open-closed question \cite{cao2025video, peng2024inst}, adding complexity to the method and a potential source of wrong assessment. Finally, the scope of these datasets is limited by the cost of human annotation and their static nature, which can quickly make them obsolete given the rapid progress of VLMs.

%The HuggingFace leaderboard \cite{open-llm-leaderboard-v2}\sk{à checker, je ne suis pas sur que c'est le bon} and UniBench~\cite{al2024unibench}, 

%The rapid evolution of visual Language Models (VLMs) has led to the development of numerous benchmarking efforts, aiming to provide systematic evaluations of model capabilities. 

%Existing benchmarks, such asaggregate results from various tasks but often fail to provide fine-grained insights into the strengths and weaknesses of individual models.
%The Arena Leaderboard, for example, has gained traction for its crowd-sourced evaluation framework, yet it primarily assesses broad general-purpose capabilities rather than domain-specific performance~\cite{achiam2023gpt, hurst2024gpt}. This lack of granularity makes it difficult for users to determine which VLM is most suited for a particular task.

%Additionally, the ability to generate labeled data with precise control over object placement and attributes makes synthetic datasets particularly useful for evaluating perception tasks like counting, localization, and object identification.  \sk{Peut être ajouter un peu plus d'élements}

\paragraph{Low-level task benchmarks}
Recent works have highlighted significant challenges in the performance of VLMs concerning tasks that require precise numerical reasoning \cite{qharabagh2024lVLM, amini2024countgd, jeon2025mutually, huang2024point}, spatial understanding \cite{zhang2024good, wang2024picture, stogiannidis2025mind} or basic identification skills \cite{rahmanzadehgervi2024vision}. For instance, \cite{zhang2024good} evaluated GPT-4V on earth observation data and found that, despite the model's proficiency in generating descriptive captions, it exhibited substantial limitations in object counting and localization tasks. Counting benchmarks can contain general \cite{acharya2019tallyqa, chattopadhyay2017counting, wang2021pixel} or specialized content (e.g., crowd-counting \cite{wang2021pixel}, detections from aerial views \cite{gao2024nwpu, zhang2024good}). However, they mostly provide a basic overview of the model counting abilities (average MAE and accuracy) over images that are too diversified (and that sometimes require reasoning, e.g. "How many giraffes are sitting down?" \cite{acharya2019tallyqa}) to precisely identify failure cases. Most spatial benchmarks focus on spatial reasoning instead of simple localization \cite{wang2024picture, shiri2024empirical, stogiannidis2025mind, cheng2024spatialrgpt}.
SPatialEval \cite{wang2024picture} or SpatialRGPT \cite{cheng2024spatialrgpt} assess various aspects of spatial reasoning, including understanding relationships, revealing significant shortcomings of state-of-the-art models in tasks requiring detailed spatial comprehension. Regarding low-level identification tasks, \cite{rahmanzadehgervi2024vision} show that VLMs struggle to solve seven very specific basic tasks such as spotting overlapping lines or circles.

\paragraph{Related works} Most evaluation papers contribute a static dataset with an aggregation score, whereas our work focuses on dataset generation tailored to assessing VLM performance with respect to a few controlled variables. Closer to our work, 
\cite{campbell2024understanding} use synthetic image variations to reveal VLM failures in object detection as distractors increase, but they focus solely on this "binding problem" without offering a systematic data generation or evaluation framework.
\cite{bao2024autobench} propose a pipeline that leverages VLMs and text-to-image models \cite{zelaszczyk2024text} to automatically generate descriptions, legends, and questions. However, their approach depends on generative AI, which may lack robustness \cite{liu2024survey}.
\cite{radsch2025bridging} manipulate images from standard datasets to generate new tasks and questions using segmentation tools, but this approach lacks fine control over visual parameters.
The closest to our work is \cite{zhang2024task}, which generates adaptive datasets from scene graphs using both external images and Blender-rendered objects. However, the task-image link (e.g., for counting) is less explicit due to intermediate representations, loosing control over key visual parameters like lighting or viewpoint. Similarly, \cite{zhang2024provision} automatically build scene graphs with computer vision tools and generate Q\&A pairs from existing images, without addressing precise task-level control.
%Overall, although they provide a way to rapidly generate more targeted questions, other papers still focus on high-level tasks, displaying global accuracy results, without isolating the assessment of a low-level visual skill.

%\la{Faire un tableau comparatif de toutes les méthodes, indiquant lesquelles ont quels features parmi image gen, question gen, static, low level, etc.}

%%%%%%%%%%%%%%%%%%%%%%%%%%%%%%%%%%%%%%%%%%%%%%%%%%%%%%%%%%%%%%%%%
\section{Diagnostic methodology and framework}
\label{sec:method}
\begin{figure} [ht]
    \centering
    \includegraphics[width=\linewidth]{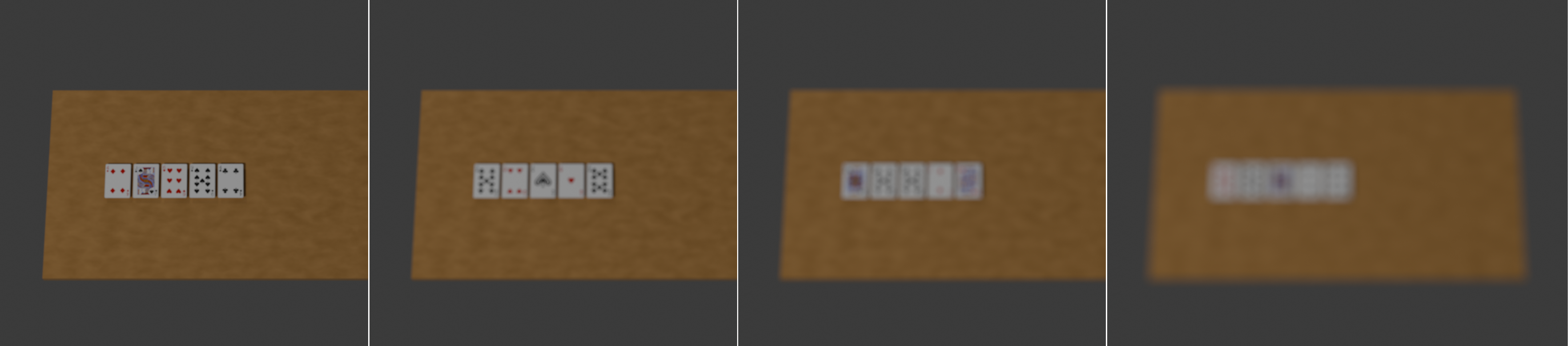}
    \caption{Examples of Poker images from light (left) to very high (right) blur. Images are cropped and overlap for display purposes.}
    \label{fig:example_poker_blur}
\end{figure}

We propose a methodology to obtain a detailed explanation of the performance of a VLM with respect to a given task and visual settings. To this end, we build a set of images with preset levels w.r.t the task at hand, and measure the evolution of the VLM score as the difficulty increases, keeping other parameters constant. Hence, the resulting score evolutions depend only on the difficulty increase. For instance, to test the ability of a VLM to count under blurry conditions, we slightly increase both the number of objects in the image and the level of blur, maintaining the camera view, the location of the objects, etc (as seen in Figure \ref{fig:example_poker_blur}). Finally, we retrieve many statistics about the predictions as illustrated in Figure \ref{fig:gpt4-1_results}.

From a user perspective, our framework enables someone to generate a set of images based on a freely customizable configuration of visual settings \ref{subsection:img_gen}, choose among a set of tasks or questions \ref{subsection:VLM_eval}, resulting in a customized image dataset along with diagnostic insights across multiple VLM services (e.g., HuggingFace, Ollama, Groq API) as seen in Figure \ref{fig:gpt4-1_results}.

%select specific tasks for (cross-)evaluation, define visual settings, and %adjust difficulty levels,. We present the image generation process \ref{subsection:img_gen}

\subsection{Image generation}
\label{subsection:img_gen}
We leverage synthetic image generation with Blender to keep full control over the content of the image and legend, in particular over the variables to test and the visual parameters of interest.
%As such, we can progressively increase the difficulty of the image w.r.t.\ the task at hand, or make slight variations of the scene setup such as light, blur, etc. %to evaluate the robustness of a VLM and obtain a precise understanding of its abilities.

\paragraph{Visual parameters} 
Blender's 3D rendering engine allows us to modify a lot of visual parameters (via the python API, \texttt{bpy}). In our framework, we have implemented a control over blur, camera distance and angle, table shape and texture, light or resolution (see Appendix~\ref{sec:detailed_framework}). As we experiment with Poker and Chess images, we also have specific parameters that can vary to change the geometry of the chessboard, the pieces, the cards, etc. These settings are defined in \texttt{yaml} configuration files, where a user sets for each parameter either a value or a range of values as shown in Appendix~\ref{sec:detailed_framework}.

\paragraph{Generation configuration content} Our framework emphasizes the difference between the variables of interest, which vary within a predefined range, and the other constant parameters. In detail, consider $n \in \mathbb{N}$ variables $v_1, \dots, v_n$, where each variable $v_i$ can take values either from a discrete set $V_i = \{v_{i,1}, v_{i,2}, \dots, v_{i,k_i}\}$ or from a discrete interval $V_i = [v_{i,\min}, v_{i,\max}]$. To construct our dataset, we systematically generate all possible combinations of these variable values:
\[
    \mathcal{D} = V_1 \times V_2 \times \dots \times V_n
\]
where $\times$ denotes the Cartesian product. We also duplicate configurations according to the (expected, or measured from a few tries) robustness of the VLM on the given task. For instance, to assess how well a VLM can perceive under blurry conditions, we define one count variable, ranging from 1 to 4, and one blur variable, varying among 5 values from a minor to very blurry image. Then, for each count and each blur level, we generate $k \in \mathbb{N}$ images ($k$ often equals 5 or 10).

\paragraph{Synthetic images} Synthetic data has been widely used in AI research to provide controlled environments for model training or evaluation \cite{tang2023new, cho2025perceptionlm, mumuni2024survey, campbell2024understanding, chow2025physbench}. Regarding evaluation, previous works have demonstrated the advantage and relevance of synthetic datasets in testing object recognition, captioning, and reasoning capabilities in multimodal models~\cite{zhang2024task, chow2025physbench, rahmanzadehgervi2024vision}. In this context, ~\cite{zhang2024task} observed a strong correlation between the results obtained on real and synthetic images. As detailed in Section \ref{subsection:benchmark}, we also observed a very strong correlation between the predictions of the VLMs on real and synthetic images for a sample of 80 images ($>0.9$ for 6 models out of 8). We refer the reader to Section 5 of \cite{mumuni2024survey} for a review focused on the challenges and methods of computer graphic modeling.

\subsection{Evaluation of VLMs}
\label{subsection:VLM_eval}

Similarly to our dataset generation, our evaluation process is task-focused. In order to conceive our questions, we first determine low-level tasks to assess the visual abilities of VLMs.

\paragraph{Low-level visual tasks} We identify 3 key visual skills to foster the adoption of VLMs within an industrial context: counting, identifying, and locating objects. One benefit of dividing an image analysis based on these tasks is that we can establish reality-grounded metrics, as described below.
\begin{itemize}
    \item \textbf{Counting}. Refers to finding the number of instances of a given object. We measure accuracy, Mean Absolute Error (MAE), Mean Squared Error (MSE) and Normalized MAE (NMAE, which is MAE divided by the target if it is different than 0 or MAE). 
    %\begin{equation*}
    %\text{NMAE}(n_{\text{real}}, n_{\text{pred}}) :=
    %\begin{cases}
    %\frac{|n_{\text{real}} - n_{\text{pred}}|}{n_{\text{real}}}, & \text{if } n_{\text{real}} \neq 0 \\
    %0, & \text{if } n_{\text{real}} = 0 \text{ and } n_{\text{pred}} = 0 \\
    %1, & \text{otherwise}
    %\end{cases}
    %\end{equation*}
    \item \textbf{Localization}. Includes two sub-tasks, absolute and relative localizations. In the first case, a grid is drawn on the image and the VLM must identify in which part of the grid an object is located. As we experiment on chess images, we use the absolute positions of the chessboard considered as a grid. Therefore, we use both accuracy and the count metrics to evaluate this skill by measuring the distance in the grid in both axes:  
    \begin{equation*}
    \mathcal{L}_{\text{LOC}} \big(p_{\text{real}}, p_{\text{pred}}\big) := |x_{p_{\text{real}}} - x_{p_{\text{pred}}}| + |y_{p_{\text{real}}} - y_{p_{\text{pred}}}|.
    \end{equation*}
    Relative localization involves finding the position of an object w.r.t.\ another object. Similarly, we rely on both accuracy and the grid to quantify the relative distance. %If ever the two objects are placed within the same cell of the grid, we either leverage an intrinsic metric of the game (card position in a hand, ...) or simply use accuracy.
    \item \textbf{Identification}. We simply ask the VLM to identify an object satisfying a condition. The metrics are all accuracy-based. %Weighted classification loss %(depending on object and distractor similarities).
    %\item \textbf{Hallucination}. We also assess the hallucination behavior of a VLM depending on the visual complexity of the image. As in \textbf{[ref]}, we use the following metrics: ??
\end{itemize}

Beyond these core metrics, we also report standard classification scores (F1, Precision, Recall).
%and two higher-level indicators: (i) \textit{Reliability} which indicates the proportion of consistent predictions over repeated inputs and (ii) \textit{Consistency} which evaluates the stability of answers across equivalent scenes. 
Finally, we define \textit{cross tasks} that combine multiple skills, such as identifying and counting a specific object type. Examples and generation strategies for all tasks are detailed in Section~\ref{subsection:diagnose}, with full question templates available in Appendix~\ref{sec:questions}.

%We also define cross tasks as the ones that combine two or three low-level tasks, such as counting a specific type of objects. Concrete examples of questions and image generation tailored to assess a given task can be found in Section \ref{subsection:diagnose} (see Appendix~\ref{} for a full list of questions). 

\paragraph{Mitigating the linguistic bias}

In order to enhance the focus on the perception capacities of a VLM instead of textual or reasoning abilities, we introduce several instruction formats as well as preprompts. Each image can thus be used several times as an input for the VLM with different question prompts. We use either no preprompt ("Neutral"), a "Helpful" preprompt (with general indications about the scene), a "Chain-of-Thought (CoT)" preprompt (to ask the model to first think carefully about the task) and two format instructions, "Declarative" (in the shape of "The answer is:") or "Missing\_Word" (details in Appendix~\ref{sec:questions}). Their effects are analyzed in Section \ref{subsection:benchmark}.

%For instance, in the case of chess, we can indicate that it is not a real game and that it is possible to see an arbitrary number of any pieces. Details are given in Appendix \textbf{[ref]}.
%\la{Mettre ça ailleurs}

%Another strategy is to first evaluate the VLM on the legends only, without the image, to make sure the VLM textual decoder is able to count up to a given number (e.g.\ if a chess piece appears a high number of times in the legend). For this purpose, obvious hints in the annotation file are first removed (the indexes of each element in the file are pseudonymized). The annotation text content is then added to the question prompt instead of the image in json format (as LLM tend to show good performance on this type of data [REF]). Results show that VLM perform clearly better with this format instead of the image on the counting task (cf Annexe). This seems to confirm that counting errors depend on visual capabilities and not just on a bias in the VLM's textual decoder counting capabilities. Finally, it also offers a glimpse of the difficulty and the relevance of this counting task, as VLM are still struggling with both modalities. %our hypothesis being that the VLM's performance is likely to be limited by that of its textual decoder.

%%%%%%%%%%%%%%%%%%%%%%%%%%%%%%%%%%%%%%%%%%%%%%%%
\section{Diagnostics of different VLMs on poker and chess images}
\label{sec:exp}
In order to validate our methodology, we build a dataset of chess and poker images and we diagnose several SoTA VLMs on a few representative configurations. Because industrial use-cases demand mastery of low-level vision tasks, we first sought an environment that exposes such challenges while remaining fully controllable. The clean geometric layouts of Chess and Poker let us generate large numbers of scenes and scale task complexity with precision. This design choice is further motivated by the persistent difficulties that VLMs and LLMs display when vision guides downstream reasoning in games \cite{paglieri2024balrog}. Finally, the popularity and rigorously defined rules of both games translate into readily available image corpora, making them ideal for reproducible, real-world evaluation. 

The full list of tasks is presented in Table~\ref{table:paper_tasks}. For each task and game, we generate a tailored set of images. For example, in counting, the number of pieces ranges from 1 to 20 and cards from 2 to 15, with 5 to 40 repetitions per level depending on the setup. We also vary visual conditions such as blur, camera distance, and object occlusion (e.g., card overlap). Examples are shown in Figures~\ref{fig:example_poker_blur} and~\ref{fig:chess_count_var}, with additional visuals in Appendix~\ref{sec:diagnostic_additional}. Each image is paired with a question, and we compute several metrics from the model responses. Section~\ref{subsection:diagnose} focuses on VLM counting abilities, while Section~\ref{subsection:benchmark} presents a broader benchmark across SoTA models.

\begin{table}[ht]
\centering
\caption{Diagnostic tasks, questions, targets, and variables for evaluation of VLM capabilities. We only display the core of the questions; the full content can be found in Appendix~\ref{sec:questions}.}
\scriptsize	
\begin{tabular}{@{}p{4.6cm} p{3cm} p{2.2cm} p{2.2cm}@{}}
\toprule
\textbf{Task} & \textbf{Question main content} & \textbf{Target} & \textbf{Variable to test} \\
\midrule
\makecell[l]{Counting (basic)} 
& How many cards/pieces?
& Number of objects
& Number of objects \\
\hline

\makecell[l]{Counting against blur} 
& \makecell[l]{How many cards/pieces?}
& \makecell[l]{Number of objects}
& \makecell[l]{Number of objects \\ Blur intensity} \\
\hline

\makecell[l]{Counting and identification} 
& \makecell[l]{How many objects X?}
& \makecell[l]{Number of objects X}
& \makecell[l]{Number of objects}\\
\hline

\makecell[l]{Counting against card overlap\\} 
& \makecell[l]{How many cards?}
& \makecell[l]{Number of cards}
& \makecell[l]{Degree of overlap \\ Number of objects} \\
\hline

\makecell[l]{Identification against camera distance} 
& \makecell[l]{What is the card/piece?}
& \makecell[l]{Object class}
& \makecell[l]{Camera distance} \\
\hline

\makecell[l]{Absolute localization of a single object} 
& \makecell[l]{Where is the card/piece?}
& \makecell[l]{Object position}
& \makecell[l]{Object type, Grid size} \\
\hline

\makecell[l]{Relative localization of two objects (Chess)} 
& \makecell[l]{Distance between the pieces?}
& \makecell[l]{Number of rows/cols}
& \makecell[l]{Piece positions} \\
\bottomrule
\end{tabular}
\label{table:paper_tasks}
\end{table}

\subsection{Individual diagnostic}
\label{subsection:diagnose}
\begin{figure}[ht]
    \centering
    \includegraphics[width=\linewidth]{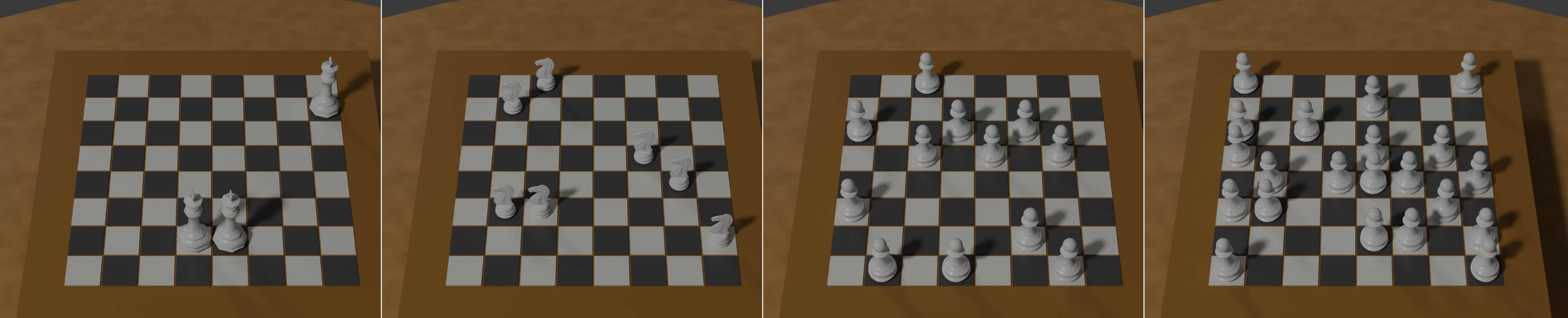}
    \caption{Variations of the number of pieces. Images are cropped and overlap for display purposes. }
    \label{fig:chess_count_var}
\end{figure}

In this section, we conduct extensive analyses of one proprietary model, GPT-4.1 \cite{gpt4.1} and one open-source model, LLaMA-4-Scout \cite{llama4}. We challenge the two models on all the aforementioned tasks (Table~\ref{table:paper_tasks}), for both Poker and Chess, using a declarative instruction as well as both a debiased and debiased-CoT preprompt (see Appendix~\ref{sec:questions} or the concrete example below). We only report the results of GPT-4.1 on Chess for the counting task (1 to 21 pieces, see Figure \ref{fig:chess_count_var}), with a declarative answer format and a debiased preprompt, which gives the following instruction:
\begin{center}
    "This is not a real chess game. The number of each piece and their position can vary arbitrary. \\ Just focus on answering the following question based on the visual content. How many pieces are there in the image? The number of pieces in the image is:"
\end{center}

\begin{figure}[ht]
    \centering
    \includegraphics[width=0.95\linewidth]{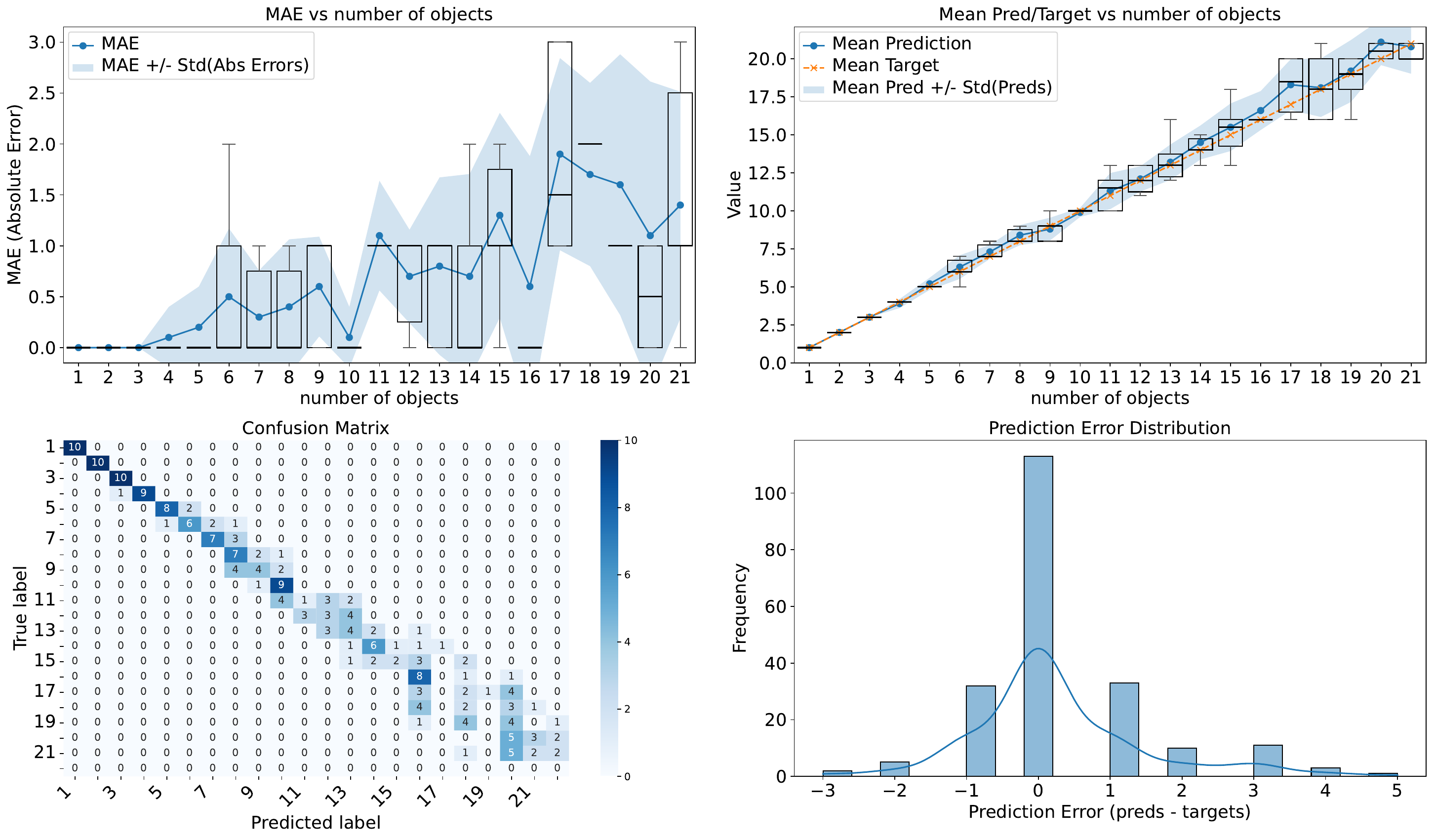}
    \caption{Results of GPT-4.1 on the counting task for the Chess dataset. Mean and std over 10 samples per level. More plots for this experiment can be found in Appendix~\ref{subsec:counting_chess}.} 
    \label{fig:gpt4-1_results}
\end{figure}

The key takeaways from this experiment are (1) a clear increase of the MAE w.r.t.\ the number of pieces on the chessboard and (2) a rather centered distribution of errors, with a slight tendency to overcount (Figure \ref{fig:gpt4-1_results}). Up to 5 pieces, the model reaches 100\% accuracy; then it miscounts one element on average up to 15 pieces, where it starts making more important mistakes.

For each task, we obtain a detailed profile of the performance of each VLM (see Appendix~\ref{sec:diagnostic_additional}). For instance, we note that counting cards is more difficult than counting chess pieces. This might be due to the split of cards over the river and players, but both models struggle more with counting Poker cards. Surprisingly, GPT-4.1 almost always predicts a number of cards below the ground truth (by more than 1 element in average from 10 to 15 cards).

Through the OpenAI API, this experiment over 210 images used 210,630 input tokens (1003 tokens in average) and 2713 output tokens (13 tokens in average) for a total cost of 0.44\$ and a total time of 734 seconds (3.5s in average).

\subsection{Comparative benchmarks}
\label{subsection:benchmark}

Beyond individual diagnostics, we conduct a comparative evaluation of the following SoTA VLMs: GPT-4.1 \cite{gpt4.1} and GPT-4.1-mini \cite{gpt4.1} (via the OpenAI API), LLaMA-4-Scout (109B, 17B active) \cite{llama4} and LLaMA-4-Maverick (400B, 17B active) \cite{llama4} (via the Groq API \hyperlink{https://groq.com/}{https://groq.com/}), Mistral 3.1 (24B) \cite{mistral31}, Gemma3 (4B, 12B) \cite{team2025gemma}, and LLaMA3.2-Vision \cite{llama32, dubey2024llama} (via Ollama \hyperlink{https://ollama.com/}{https://ollama.com/} on a local machine with 2 NVIDIA GPUs RTX 5000 ADA and one NVIDIA GPU RTX 6000 ADA).
All models are evaluated on the tasks listed in Table~\ref{table:paper_tasks}, using both chess and poker datasets with variations in object count, type, camera distance, blur, and card overlap. Averaged chess results are reported in Tables~\ref{tab:chess_localization_x_identification}, \ref{tab:counting_per_difficulty_chess} and Figure~\ref{fig:chess_radar_detailed}. Poker and full results appear in Appendix~\ref{sec:benchmark_additional}.
% Unlike conventional benchmarks that aggregate performance across heterogeneous tasks, our setup enables fine-grained comparisons under systematically varied visual conditions. 
%We assess models on a suite of low-level visual tasks such as counting, identification, and localization across both chess and poker datasets.
%Each model is evaluated under identical image conditions, prompt formulations, and task variants, allowing us to isolate performance differences attributable to visual perception rather than dataset or prompt biases. %This section reports the results across tasks and prompt configurations, highlighting the strengths and limitations of each model family under controlled stress-testing conditions.

In details, for the counting task (Table~\ref{tab:counting_per_difficulty_chess} and \texttt{Count} in Figure \ref{fig:chess_radar_detailed}), the number of objects ranges from 1 to 21, or 1 to 4 when adding blur with 5 levels of blur (\texttt{Count Blur} in Figure \ref{fig:chess_radar_detailed}). Regarding localization, we ask the VLM to detect the position of a single piece on a grid of size 4x4 (\texttt{Loc 4x4} in Figure \ref{fig:chess_radar_detailed}) or the distance in terms of rows and columns of two pieces on 4x4 and 8x8 boards (\texttt{Loc 4x4 (2)} and \texttt{Loc 8x8 (2)} in Figure \ref{fig:chess_radar_detailed}). Finally, the VLM has to identify a single piece or card for the identification task (\texttt{Identification} in Figure \ref{fig:chess_radar_detailed}). We also add a variation of the camera distance from 1.7 to 7.5 meters for identification (\texttt{Identification Distance} in Figure \ref{fig:chess_radar_detailed}).

%To ensure a fine-grained and systematic evaluation, we define a set of diagnostic tasks targeting core visual competencies under controlled conditions. 
%These include (i) counting the number of objects in an image, with variants that introduce visual stressors such as blur, occlusion, or increased object density. (ii) Identifying the category of a single object, with or without variations in camera distance and (iii) localization of the objects, requiring the model to predict a grid position (eg., chess square) or the spatial relation between two objects. Across all tasks, we systematically vary key visual parameters while holding others constant, enabling controlled stress testing and precise analysis of model robustness and failure modes. 

\begin{figure}[ht!]
  \centering
    \centering
    \includegraphics[width=\linewidth]{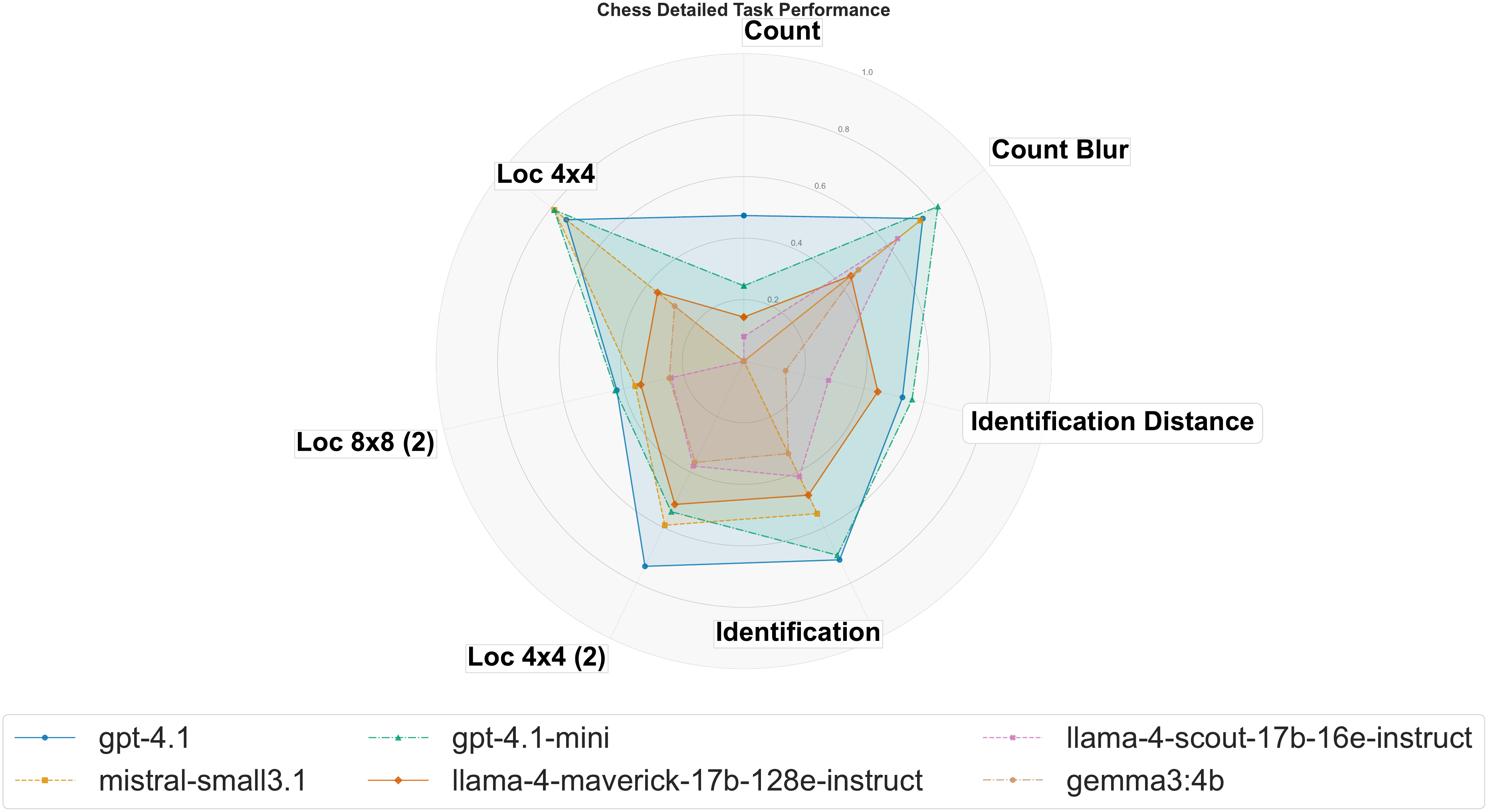}
    \label{fig:}
  \caption{
  Model performance on chess tasks. Averaged accuracy across all samples; sample sizes are indicated per task: \texttt{Loc 8x8 (2)} (n=50), \texttt{Loc 4x4} (n=50), \texttt{Loc 4x4 (2)} (n=50), \texttt{Count} (n=850), \texttt{Count Blur} (n=191), \texttt{Identification Distance} (n=140), and \texttt{Identification} (n=50).%For each model, mean accuracy is computed with each category by averaging over its associated subtasks. %Both visualizations are based on the same evaluation data but aggregated differently to reveal complementary aspects of model behavior across visual skills. 
  }
  \label{fig:chess_radar_detailed}
\end{figure}

Overall, confirming previous works \cite{campbell2024understanding, zhang2024good, wang2024picture, rahmanzadehgervi2024vision}, SoTA VLMs still face challenges to solve very basic visual tasks that are trivial to humans (e.g., counting pieces on a chessboard) (Figure \ref{fig:chess_radar_detailed} and Tables \ref{tab:chess_localization_x_identification},\ref{tab:counting_per_difficulty_chess}). Even high-performing models degrade under compositional task load or perceptual distortion. For instance, Llama-4-Scout exhibits sharp drops in accuracy on high-density chess boards (down to $29\%$ at $>$ 9 pieces, see Table~\ref{tab:counting_per_difficulty_chess}) and struggles with card overlap in poker images. These findings suggest that current VLMs may rely on superficial visual heuristics rather than robust object-level understanding.

%%% Modifer la phrase vu qu'on a enlever la table couting 
Results across all tasks are summarized in Tables~\ref{tab:chess_localization_x_identification} \ref{tab:counting_per_difficulty_chess} and Figure \ref{fig:chess_radar_detailed}.
GPT-4.1 and GPT-4.1-mini consistently outperform other models across tasks, GPT-4.1 showing slightly higher robustness in localization performances (MAE of $0.210$ compared to GPT-4.1-mini's $0.480$, Table~\ref{tab:chess_localization_x_identification}). Among open-sourced models, Mistral-3.1 \cite{mistral31} and Llama-4-Scout show stronger performance on counting tasks, but still lag in localization (Table~\ref{tab:chess_localization_x_identification}) and identification under distance variation (Table~\ref{tab:chess_localization_x_identification}). Notably, performance disparities are amplified under controlled stress conditions such as blur or increasing object count (Table~\ref{tab:counting_per_difficulty_chess}). 

\begin{table}[ht]
\centering
\caption{Average performance of VLMs on the Chess dataset. Mean and standard deviation over 50 images for each task. Localization is performed on 4x4 chessboards containing a single piece.}
\renewcommand{\arraystretch}{1.1}
\setlength{\tabcolsep}{3pt}
\scriptsize
\begin{tabular}{l|ccccc|cccc}
\toprule
& \multicolumn{5}{c|}{\textbf{Localization}} & \multicolumn{4}{c}{\textbf{Identification}} \\
\cmidrule(lr){2-6} \cmidrule(lr){7-10} 
\textbf{Model} & \textbf{Acc\,$\uparrow$} & \textbf{F1\,$\uparrow$} & \textbf{MAE\,$\downarrow$} & \textbf{MSE\,$\downarrow$} & \textbf{NMAE\,$\downarrow$} & \multicolumn{2}{c|}{\textbf{Distance Variation}} & \multicolumn{2}{c}{\textbf{No Variation}} \\
\cmidrule(lr){7-8} \cmidrule(lr){9-10}
& & & & & & \textbf{Acc\,$\uparrow$} & \textbf{F1\,$\uparrow$} & \textbf{Acc\,$\uparrow$} & \textbf{F1\,$\uparrow$}  \\
\midrule
gpt-4.1  & 0.83 {\tiny (0.33)} & 0.91 {\tiny (0.12)} & 0.17 {\tiny (0.33)} & 0.17 {\tiny (0.33)} & 0.13 {\tiny (0.23)} & \textbf{0.56} {\tiny (0.34)} & \textbf{0.72} {\tiny (0.37)}  & 0.68 {\tiny (0.40)} & 0.81 {\tiny (0.39)} \\
gpt-4.1-mini    & \textbf{0.98} {\tiny (0.09)} & \textbf{0.99} {\tiny (0.06)} & \textbf{0.02} {\tiny (0.1)} & \textbf{0.02} {\tiny (0.09)} & \textbf{0.01} {\tiny (0.05)} & 0.59 {\tiny (0.36)} & 0.74 {\tiny (0.35)}  & \textbf{0.70} {\tiny (0.33)} & \textbf{0.82} {\tiny (0.30)}  \\
llama-4-scout   & 0.29 {\tiny (0.34)} & 0.45 {\tiny (0.36)} & 2.63 {\tiny (1.54)} & 10.4 {\tiny (8.16)} & 1.45 {\tiny (1.09)} & 0.45 {\tiny (0.32)} & 0.62 {\tiny (0.35)}  & 0.43 {\tiny (0.36)} & 0.61 {\tiny (0.38)}  \\
llama-4-maverick& 0.52 {\tiny (0.40)} & 0.68 {\tiny (0.34)} & 0.77 {\tiny (0.70)} & 1.47 {\tiny (1.68)} & 0.55 {\tiny (0.48)} & 0.47 {\tiny (0.34)} & 0.62 {\tiny (0.36)}  & 0.48 {\tiny (0.39)} & 0.65 {\tiny (0.40)} \\
gemma3          & 0.42 {\tiny (0.23)} & 0.60 {\tiny (0.33)} & 1.93 {\tiny (0.96)} & 5.37 {\tiny (4.71)} & 0.94 {\tiny (0.51)} & 0.14 {\tiny (0.23)} & 0.24 {\tiny (0.30)}  & 0.35 {\tiny (0.32)} & 0.52 {\tiny (0.37)}  \\
gemma3:12b      & 0.48 {\tiny (0.27)} & 0.49 {\tiny (0.22)} & 0.97 {\tiny (0.46)} & 1.60 {\tiny (1.22)} & 0.61 {\tiny (0.40)} & 0.48 {\tiny (0.37)} & 0.65 {\tiny (0.39)}  & 0.45 {\tiny (0.35)} & 0.62 {\tiny (0.37)}  \\
llama3.2-vision & 0.30 {\tiny (0.28)} & 0.46 {\tiny (0.33)} & 1.10 {\tiny (0.51)} & 2.10 {\tiny (1.56)} & 0.69 {\tiny (0.46)} & 0.50 {\tiny (0.32)} & 0.67 {\tiny (0.33)}  & 0.48 {\tiny (0.38)} & 0.65 {\tiny (0.39)} \\
mistral-3.1     & 0.82 {\tiny (0.30)} & 0.90 {\tiny (0.22)} & 0.18 {\tiny (0.30)} & 0.18 {\tiny (0.30)} & 0.15 {\tiny (0.25)} & 0.44 {\tiny (0.37)} & 0.61 {\tiny (0.39)}  & 0.57 {\tiny (0.38)} & 0.72 {\tiny (0.38)}  \\
\bottomrule
\end{tabular}
\label{tab:chess_localization_x_identification}
\end{table}

\begin{table}[ht]
\centering
\caption{Accuracy by object count range on the Chess dataset (Declarative instruction and Helpful preprompt). Mean and standard deviation across aggregations of levels, with 40 samples per level.}

\label{tab:counting_per_difficulty_chess}
\renewcommand{\arraystretch}{1.1}
\setlength{\tabcolsep}{4pt}
\scriptsize

\begin{tabular}{lccc}
\toprule
\textbf{Model} & \textbf{Low (1–4 pcs)} & \textbf{Medium (5–9 pcs)} & \textbf{High (10-21 pcs)} \\
\midrule
GPT-4.1                        & 0.94 (0.24) & 0.75 (0.50) & 0.54 (0.05) \\
GPT-4.1-mini                   & 0.87 (0.33) & 0.58 (0.46) & 0.40 (0.23) \\
LLaMA-4-scout                 & 0.75 (0.43) & 0.61 (0.41) & 0.27 (0.11) \\
LLaMA-4-Maverick              & 0.81 (0.39) & 0.64 (0.34) & 0.32 (0.21) \\
Gemma3                         & 0.50 (0.50) & 0.12 (0.26) & 0.07 (0.34) \\
Gemma3:12b                     & 0.63 (0.48) & 0.34 (0.11) & 0.17 (0.21) \\
LLaMA3.2-Vision                & 0.65 (0.43) & 0.51 (0.17) & 0.24 (0.32) \\
Mistral-small3.1              & 0.75 (0.46) & 0.57 (0.18) & 0.25 (0.19) \\
\bottomrule
\end{tabular}
\end{table}

The radar chart (Figure \ref{fig:chess_radar_detailed}) reveals fine-grained skill profiles: for instance, LLaMA-4-Scout performs relatively well on basic counting but drops sharply on $8\times8$ localization tasks, indicating poor spatial generalization. GPT-4.1 variants consistently outperform open-source models, especially in localization. While Gemma3-12B achieves high consistency in identification, it underperforms drastically in compositional and spatially complex setups, suggesting a lack of robust visual grounding.

\begin{figure}[ht]
  \centering
    \centering
    \includegraphics[width=\columnwidth]{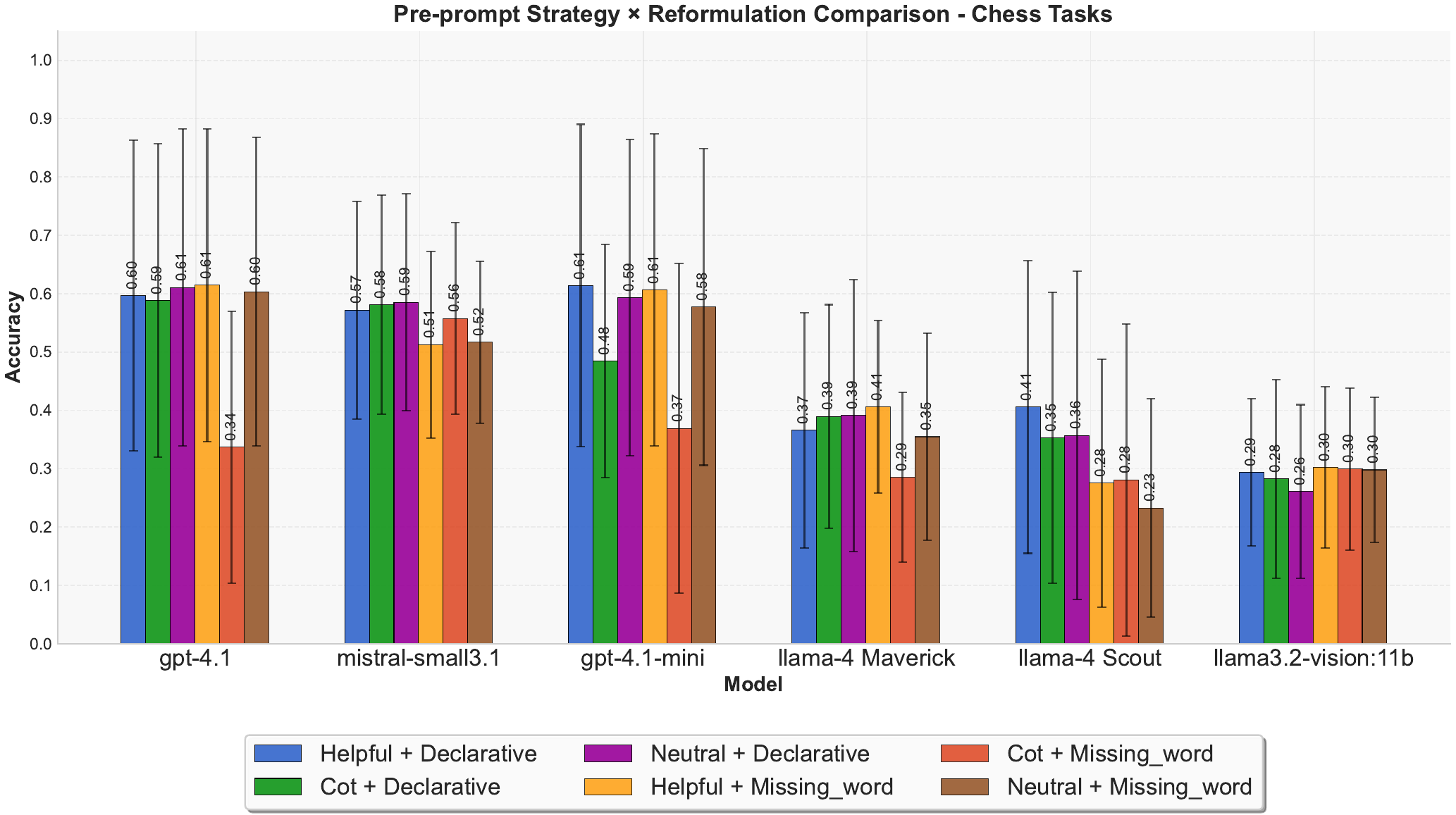}
  \caption{
    Impact of the choice of preprompt and instructions on accuracy (mean over all chess tasks, see Appendix~\ref{sec:benchmark_additional} for a comparison per task).
  }
  \label{fig:combined_preprompt}
\end{figure}

To further understand the interplay between linguistic prompt design and visual task performance, we systematically evaluate combinations of preprompt strategies (Helpful, CoT, Neutral) and instruction types (Declarative, Missing Word) across all tasks. As shown in Figure~\ref{fig:combined_preprompt}, declarative instructions yield improvements over close-style prompts across all models (up to $+5\%$ accuracy on Poker counting for GPT-4.1-mini when combined with a Helpful preprompt, see Appendix~\ref{sec:benchmark_additional}).

Finally, we examine the correlation using a sample of 80 real images. We recreate 80 Chess scenes from our chess count dataset, with the number of pieces varying from 1 to 8 (10 samples per image, as detailed in Appendix~\ref{sec:correlation_synthetic_real}). For each combination of model, preprompt, and instruction, we conduct our diagnostic analysis. After averaging predictions for each level over the 10 samples, preprompts, and instructions, we find Spearman and Pearson correlation scores exceeding 0.99 for all models, except for Gemma3-4B (0.93 and 0.91) and Gemma3-12B (0.64 and 0.75).

%Radar chart of model performance across fine-grained chess subtasks (e.g., Counting, Identification, Localization).
%Line plot showing accuracy trends across high-level task types: Count, Identification, and Localization.

%%%%%%%%%%%%%%%%%%%%%%%%%%%%%%%%%%%%%%%%%%%%%%%%%%%%%%%%%%%%%%%%%%%%%%%%%%%%%%%%%%%%%%%%%%%%%%%%%%%

%%%%%%%%%%%%%%%%%%%%%%%%%%%%%%%%%%%%%%%%%%%%%%%%%%%%%%%%%%%%%%%%%%%%%%%%%%%%%%%%%%%%%%%%%%%%%%%%%%%

\section{Discussion and future work}

\label{sec:conclusion}
We introduce both a new methodology and a framework to precisely diagnose the perception abilities of VLMs. Our approach leverages procedurally generated synthetic data to isolate specific visual tasks and measure model performance across controlled difficulty levels. By disentangling low-level perceptual skills from high-order reasoning, our approach sheds light on VLM failure modes that are otherwise obscured in traditional benchmarks.

\paragraph{Discussion} We believe that when selecting a vision-language model, the primary concern is how reliably it will handle the concrete tasks it will encounter in production—counting items on a conveyor belt, locating cracks, or interpreting color codes on labels. Therefore, although high-level evaluations are very useful to measure the progress of VLMs and get a rough idea of their abilities, a decoupled and interpretable assessment of their abilities is crucial to gain insights about what they can or cannot do in real-world scenarios \cite{ott2022mapping, reuel2024betterbench}. This implies the decomposition of high-level instruction following or image understanding into low-level functions. Focusing on the perception skills, a further decomposition could include  measuring, brightness and color identification, etc. In this context, synthetic data are very useful as images and annotations of specific use cases are too costly to obtain.

%Both approaches are not contradictory. In contrary, we could conceive an evaluation scheme that applies a fine-grained diagnostic to the failure cases of a VLM on a comprehensive benchmarks. 

%Comprehensive leaderboards still play a valuable supporting role—providing a quick gauge of overall maturity and helping to narrow the field—but it is the targeted, scenario-driven probes that ultimately reveal whether a candidate model can be trusted in the real world.

%that the complexity of VLMs and the diversity of use-cases make it illusory to determine the "best" model from comprehensive scores and that it is necessary to focus on specialized evaluation that stick to real-world scenarios. 

\paragraph{Limitations and future work} Our method and framework could benefit from several improvements. First of all, a better formalism of the evaluation process could lead to a smoother pipeline task-to-test$\to$configuration$\to$image and question generation$\to$metrics. We plan to improve the abstraction classes to offer an easier integration of additional use cases (visual parameters, scene content, questions, etc). A more direct improvement would be to generate more realistic images, fitting real industrial scenarios, using tools from \cite{roberts2021hypersim, denninger2023blenderproc2}. Finally, for LLaVA-like architectures \cite{liu2023visual} with open-sourced parameters, identifying precisely the failing module (visual, alignment or textual module) could be done by training linear probes on top of the visual or merged representations as in \cite{rahmanzadehgervi2024vision}. 

%\blue{Remettre une couche sur les data synthétiques ici en disant que les performances du modèles varient selon la difficulté de l'image, donc c'est clairement pertinent}

%\blue{Dire aussi qu'un tel framerwork permettrait de valider les claims d'autres papiers plus facilement (future work)}

\clearpage
\bibliography{bibliography}
\bibliographystyle{ieeetr}

%%%%%%%%%%%%%%%%%%%%%%%%%%%%%%%%%%%%%%%%%%%%%%%%%%%%%%%%%%%%

%%%%%%%%%%%%%%%%%%%%%%%%%%%%%%%%%%%%%%%%%%%%%%%%%%%%%%%%%%%%
\clearpage
\appendix

\section{Framework overview}
\label{sec:framework_overview}

In this section, we give an overview of our framework in practice and we refer the reader to Appendix \ref{sec:detailed_framework} for details.

\subsection{Overview of the Dataset Construction Process}
\label{sec:dataset_overview}

This subsection provides an overview of our synthetic dataset generation pipeline, from task specification to annotated image production. We illustrate the main stages through a chess example, and highlight the underlying structure and key parameters. Readers seeking a complete technical specification may refer to Appendix~\ref{sec:detailed_framework}.

\paragraph{Step 1: Task Definition and Parameter Space Design.}
The process begins by identifying the core task, selecting the relevant variables to control and their range or value. For example, for a chess piece counting task, variables might include the number and type of pieces, board size, camera viewpoint, and lighting conditions. Each parameter is formally encoded in configuration schemas, typically as Python dataclasses or structured YAML/JSON dictionaries.

\paragraph{Step 2: Scene Configuration via Strongly-Typed Models.}
All scene components (e.g., board, pieces, camera, lighting, materials) are described by hierarchical configuration models. For instance, a \texttt{CameraModel} encodes camera distance, elevation angle, and randomization settings, while a \texttt{PieceModel} specifies piece type, position, color, and size. This design enables programmatic control, type safety, and easy extensibility. An illustrative Python configuration snippet for a simple chess scene:

\begin{mycase}{light_grey}
\begin{lstlisting}[language=Python]
scene_config = {
    "board": {
        "rows": 8,
        "columns": 8,
        "board_material": {"color": (0.7, 0.6, 0.5, 1.0)}
    },
    "pieces": [
        {"type": "king", "position": (0, 4), 
        "color": (0.9, 0.9, 0.9, 1.0)},
        {"type": "queen", "position": (7, 3),
        "color": (0.1, 0.1, 0.1, 1.0)}
    ],
    "camera": {"distance": "medium", "angle": 60.0},
    "lighting": {"lighting": "high"},
    "noise": {"blur": "low", "table_texture": "medium"}
}
\end{lstlisting}
\end{mycase}

Alternatively, the same setup can be specified in YAML for experiment management:

\begin{mycase}{light_grey}
\begin{lstlisting}[]
board:
  rows: 8
  columns: 8
  board_material:
    color: [0.7, 0.6, 0.5, 1.0]
pieces:
  - type: king
    position: [0, 4]
    color: [0.9, 0.9, 0.9, 1.0]
  - type: queen
    position: [7, 3]
    color: [0.1, 0.1, 0.1, 1.0]
camera:
  distance: medium
  angle: 60.0
lighting:
  lighting: high
noise:
  blur: low
  table_texture: medium
\end{lstlisting}
\end{mycase}

\paragraph{Step 3: 3D Scene Construction in Blender.}
At the core of our pipeline is Blender, an open-source 3D engine, controlled programmatically through its Python API (\texttt{bpy}). Using our configuration models, the pipeline builds each scene element:

\begin{itemize}
    \item \textbf{Board and Pieces:} Geometry and material are either procedurally generated or loaded from pre-existing assets. Piece positioning, scale, and orientation are set via configuration.
    \item \textbf{Camera and Lighting:} Camera parameters (distance, angle, random jitter) and lighting setups (key, fill, back light) are instantiated from their model descriptions.
    \item \textbf{Materials and Noise:} Surface materials, textures, and noise effects (e.g., blur, table surface complexity) are applied based on parameter presets or custom values.
\end{itemize}

\begin{figure}[h]
    \centering
    \includegraphics[width=\linewidth]{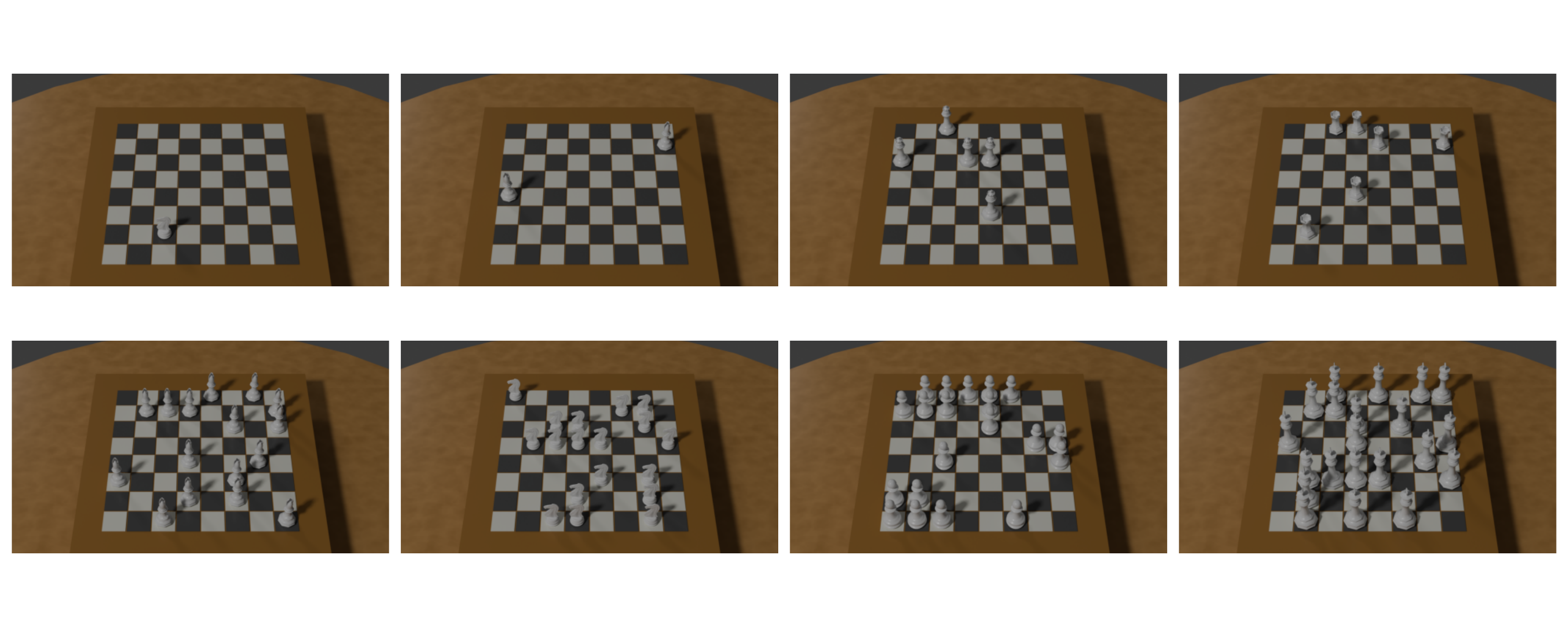}
    \caption{Examples of generated chess images with variation of the number of pieces on the board.}
    \label{fig:example_chess_count}
\end{figure}

This approach provides full flexibility: as illustrated below (Figure \ref{fig:example_chess_count}), one can systematically vary the number of pieces on the chess board. It is possible to do the same with all other visual parameters introduced in the framework: camera distance, etc.

\paragraph{Step 4: Automated Rendering and Legend Generation.}
Once the scene is constructed, images are rendered at the specified resolution and file format. Crucially, alongside each image, the pipeline generates a \textit{legend}---a comprehensive metadata file (text and JSON) capturing every scene parameter, object attribute, and applied transformation. This ensures reproducibility and enables automated extraction of ground-truth answers for downstream evaluation.

For example, the legend for a chess image includes: board geometry, piece types and positions, camera settings, noise levels, and any random seeds used. This metadata underpins both the traceability of the dataset and the automated benchmarking of model predictions.

\begin{mycase}{light_grey}
\begin{lstlisting}[]
{
  "board": {"rows": 8, "columns": 8},
  "pieces": [
    {"type": "king", "position": [0, 4],
    "color": [0.9, 0.9, 0.9, 1.0]},
    {"type": "queen", "position": [7, 3],
    "color": [0.1, 0.1, 0.1, 1.0]}
  ],
  "camera": {"distance": "medium", "angle": 60.0},
  "noise": {"blur": "low"}
}
\end{lstlisting}
\end{mycase}

\paragraph{Step 5: Question/Answer Generation and Extraction.}
For each rendered image, the system instantiates visual reasoning questions (e.g., ``How many pieces are there in the image?'') and uses the legend to automatically extract ground-truth answers. The question and answer generation logic is domain-agnostic, supporting both chess and poker (see Section~\ref{sec:questions} for details).

\paragraph{Advanced Features and Extensibility.}
\begin{itemize}
    \item \textbf{Parameter Randomization:} The framework supports controlled randomization of variables (e.g., piece placement, lighting intensity), enabling statistical evaluation of model robustness.
    \item \textbf{Presets and Customization:} Named presets (e.g., \texttt{"high"} blur, \texttt{"low"} lighting) simplify experiment setup, while custom values allow fine-grained control.
    \item \textbf{Scalable Combinatorial Generation:} For large-scale experiments, variable combination generators exhaustively or randomly instantiate all possible scene configurations, ensuring coverage of the experimental space.
    \item \textbf{Domain Extension:} Although illustrated here for chess, the modular architecture extends naturally to other domains (e.g., poker, go), supporting new object types, rules, and scene structures with minimal changes.
\end{itemize}

\paragraph{End-to-End Example.}
A minimal working example of the pipeline might involve:
\begin{enumerate}[leftmargin=2em]
    \item Defining a YAML file specifying a range of piece counts for chess images.
    \item Running the dataset generator, which builds each scene in Blender, applies the specified configurations, and renders the images.
    \item Saving, for each image, a legend file containing all parameters, and extracting answers for a suite of questions.
    \item Using the resulting image-question-answer triplets for downstream model training or evaluation.
\end{enumerate}

\vspace{0.5em}
In summary, our pipeline provides a fully automated, extensible, and transparent process for generating annotated visual reasoning datasets, with precise experimental control over all scene elements. For full technical details, schema definitions, and advanced usage, please refer to Appendix~\ref{sec:detailed_framework}.

%%%%%%%%%%%%%%%%%%%%%%%%%%%%%%%%%%%%%%%%%%%%%%%%%%%%%%%%%%%%%%%%%%%%%
%%%%%%%%%%%%%%%%%%%%%%%%%%%%%%%%%%%%%%%%%%%%%%%%%%%%%%%%%%%%%%%%%%%%%
%%%%%%%%%%%%%%%%%%%%%%%%%%%%%%%%%%%%%%%%%%%%%%%%%%%%%%%%%%%%%%%%%%%%%
%%%%%%%%%%%%%%%%%%%%%%%%%%%%%%%%%%%%%%%%%%%%%%%%%%%%%%%%%%%%%%%%%%%%%
\section{Diagnostic - additional experiments}
\label{sec:diagnostic_additional}

\subsection{General observations}
\label{sec:general_observations}
We present additional results from the diagnostic experiment on the task set introduced in Section \ref{sec:exp}. We do not systematically include all results for every task, but instead provide representative diagnostics for both GPT-4.1 and LLaMA-4-Scout.

Overall, the following key observations can be made:
\begin{itemize}
\item \textbf{Counting}: These diagnostics reveal model biases in specific domains. GPT-4.1 tends to slightly overestimate the count of large object numbers in the Chess dataset, while significantly underestimating the count of Poker cards. Notably, these biases appear to be model-dependent, as LLaMA-4-Scout demonstrates a minor tendency to underestimate object counts in the Chess dataset, in contrast to GPT-4.1.
\item \textbf{Counting with blur}: Both GPT-4.1 and LLaMA-4-Scout exhibit an overestimation pattern in object counting as blur increases, which may be explained by the fact that higher levels of blur can merge pixels, creating the illusion of objects where none exist.
\item \textbf{Localization of a single Chess piece on a 4x4 grid}: This experiment highlights the contextual bias exhibited by the LLaMA-4-Scout model, as it consistently predicts out-of-bounds rows and columns, likely inferring a standard 8x8 chessboard.
\item \textbf{Localization of a single card on a 3x3 Poker grid}: Both models demonstrate a decline in performance when the element is positioned in the middle column of the poker scene. While LLaMA-4-Scout tends to overestimate column locations, making again out-of-bounds predictions, its performance remains more accurate for rows. This may be attributed to the rectangular shape of the table, leading to a grid where rows are more compact than columns, and highlights the importance of considering the geometry of the scene.
\item \textbf{Relative localization of two Chess pieces on an 8x8 grid}: Both models' performance declines rapidly as the row distance between the two pieces increases, with a clear underestimation of their relative distance. These tendencies contrast with those observed when locating a single chess piece on a 4x4 grid, emphasizing the importance of conducting both diagnostics.
\item \textbf{Identification with camera distance (Chess)}: As anticipated, the performance appears to decline as the distance to the chess scene increases.
\item \textbf{Counting with Horizontal Overlap (Poker cards)}: Both models tend to underestimate the number of cards with horizontal overlap. Additionally, LLaMA-4-Scout seems to particularly struggle with the concept of cards encroaching upon one another, as its underestimation steadily increases with the level of overlap.
\end{itemize}

%%%%%%%%%%%%%%%%%%%%%%%%%%%%%%%%%%%%%%%%%%%%%%%%%%%%%%%%%%%%%%% 
\subsection{Counting (Chess)}
\label{subsec:counting_chess}

We present the diagnostic results for the \emph{counting} task on the Chess dataset. The following question is asked with a debiased preprompt and a declarative instruction:
\begin{center}
    "This is not a real chess game. The number of each piece and their position can vary arbitrarily. Just focus on answering the following question based on the visual content. How many pieces are there in the image? Respond in a declarative format: 'The number of pieces in the image is:'"
\end{center}

Examples of images can be found below in Figure \ref{fig:img_count_chess}:

\begin{figure}[ht]
    \centering
    \includegraphics[width=\linewidth]{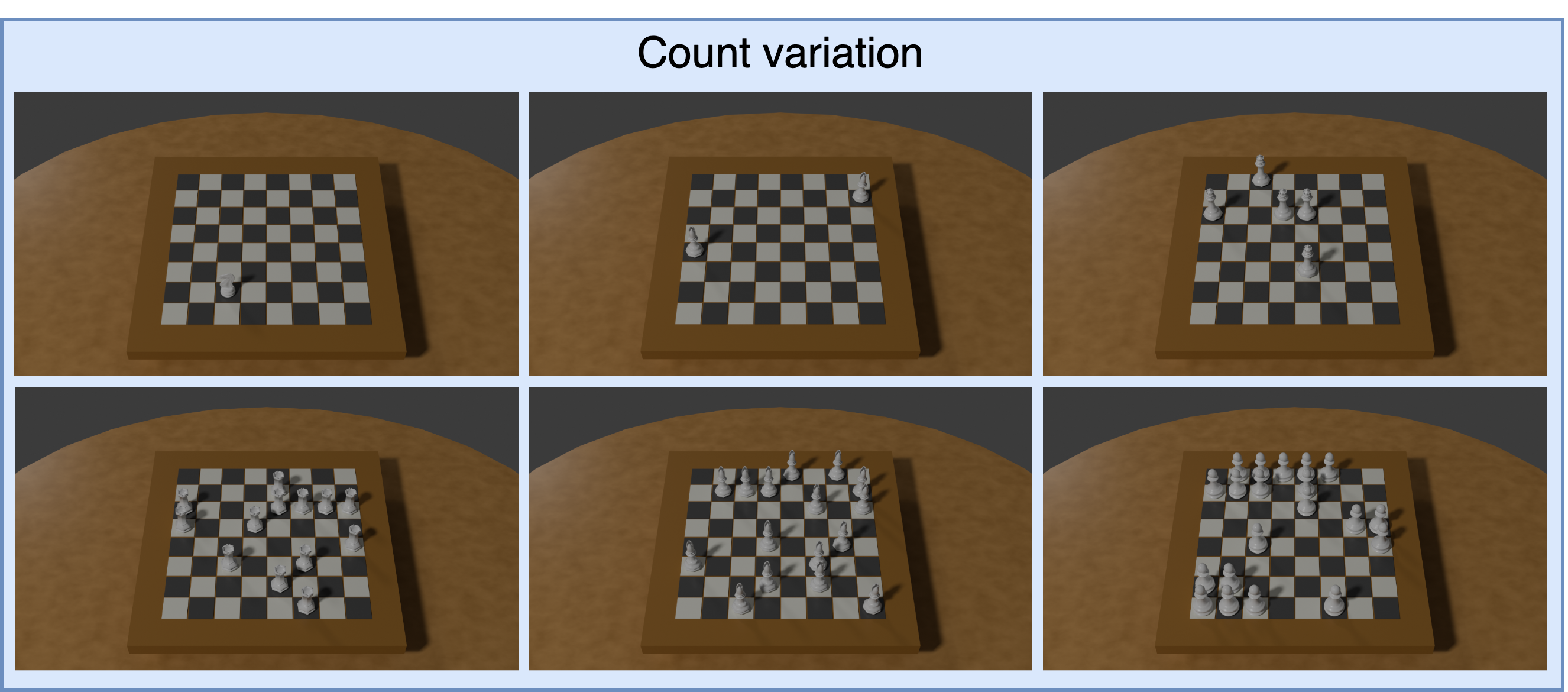}
    \caption{Variations of the number of Chess pieces (cropped images with overlap for display purposes)}
    \label{fig:img_count_chess}
\end{figure}

%%%%%%%%%%%%%%%%%%%%%%%%%%%%%%%%%%%%%%%%%%%%%%%%%%%%%%%%%%%%%%%%%%%%%%%%%%%%%%%%%%%%

\paragraph{Results}

\begin{figure}[H]
    \centering
    \includegraphics[width=0.95\columnwidth]{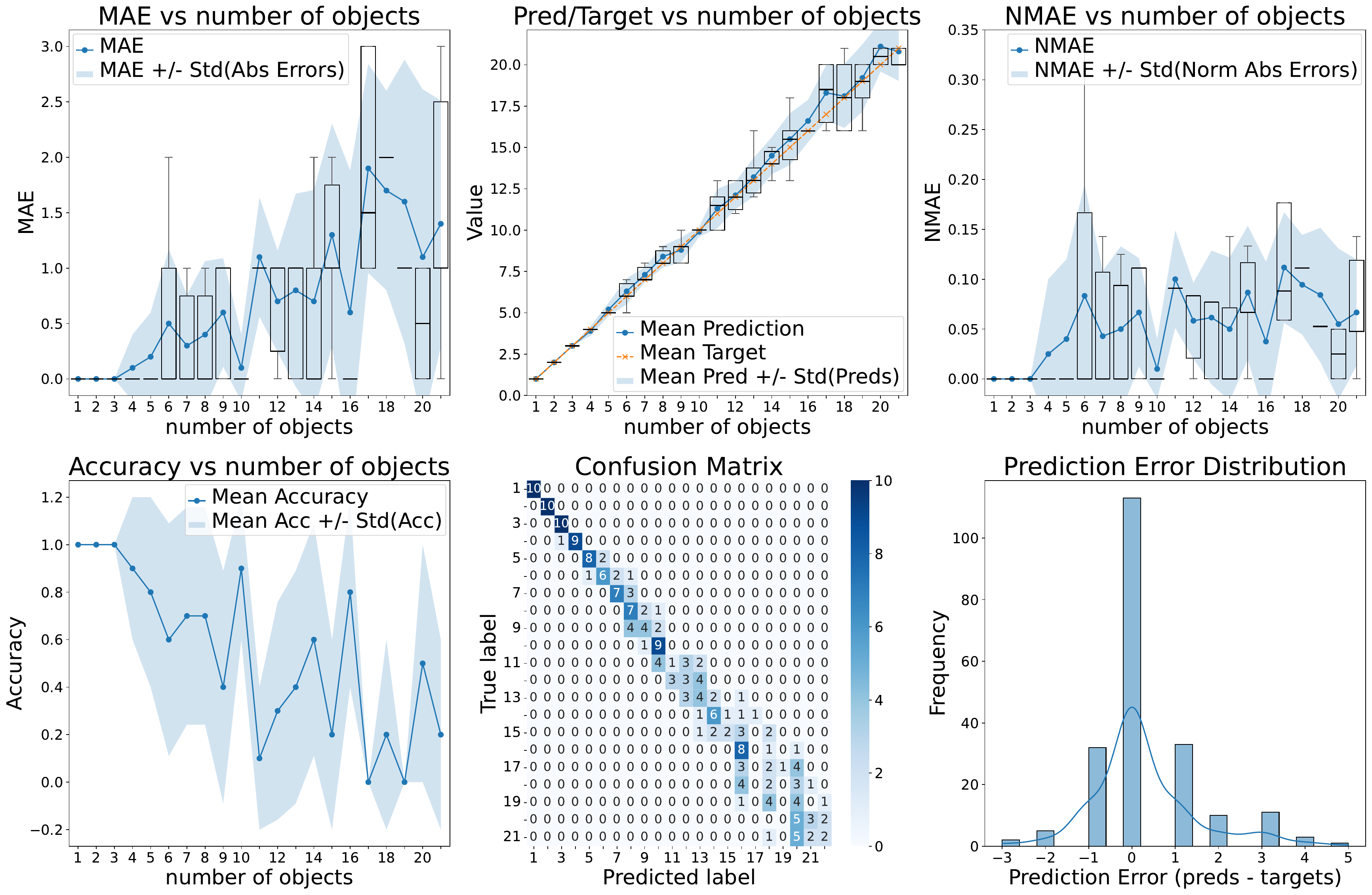}
    \caption{Results of GPT-4.1 for the counting question on the Chess dataset. Mean and standard deviation are computed over 10 samples per level.}
    \label{fig:gpt4-1_results_count_chess}
\end{figure}

\begin{figure}[H]
    \centering
    \includegraphics[width=0.95\linewidth]{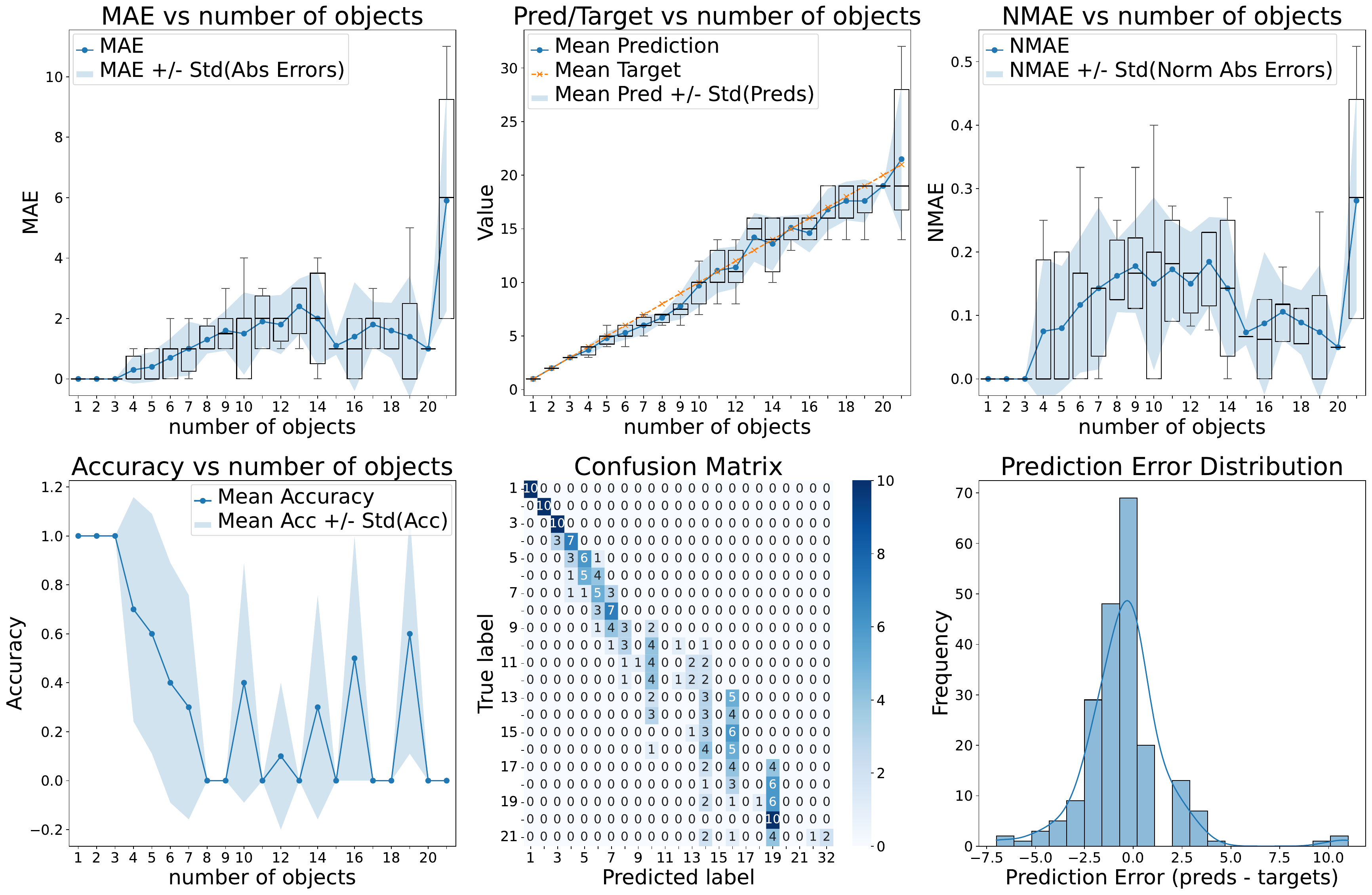}
    \caption{Results of LLaMA 4 for the counting question on the Chess dataset. Mean and standard deviation are computed over 10 samples per level.}
    \label{fig:LLaMA4_results_count_chess}
\end{figure}

\begin{itemize}
    \item GPT-4.1 (Figure \ref{fig:gpt4-1_results_count_chess}). The model achieves 100\% accuracy up to 3 objects, beyond which it begins to make errors. The Mean Absolute Error (MAE) remains near 1 up to 14 objects, after which it starts to increase. The prediction error appears to be approximately centered, with the model showing a slight tendency to overestimate the number of objects.
    \item LLaMA-4-Scout (Figure \ref{fig:LLaMA4_results_count_chess}). The model also remains at 100\% accuracy up to 3 chess pieces, beyond which it begins to make errors. However, the Mean Absolute Error (MAE) rapidly increases, reaching an average of 2, with outliers that significantly impact the average score. The distribution of prediction errors is approximately centered, with frequent slight underestimates of the object count and a few notable overestimates at higher levels.
\end{itemize}

While both models achieve perfect scores up to 3 objects, these diagnostics help to reveal a slight bias in their counting abilities. GPT-4.1 exhibits a tendency to slightly overestimate the number of objects, whereas LLaMA-4-Scout tends to slightly underestimate the count.

%%%%%%%%%%%%%%%%%%%%%%%%%%%%%%%%%%%%%%%%%%%%%%%%%%%%%%%%%%%%
\subsection{Counting (Poker)}
\label{subsec:counting_poker}

We present the diagnostic results for the \emph{counting} task on the Poker dataset. The following question is asked with a debiased preprompt and a declarative instruction:
\begin{center}
    "This is not a real poker game. The cards and their position can vary arbitrarily. Just focus on answering the following question based on the visual content. How many cards are present in the entire scene (including all hands and community cards)? Respond in a declarative format: 'The number of cards in the image is:'"
\end{center}

Examples of images can be found below in Figure \ref{fig:poker_count}:

\begin{figure}[H]
    \centering
    \includegraphics[width=\linewidth]{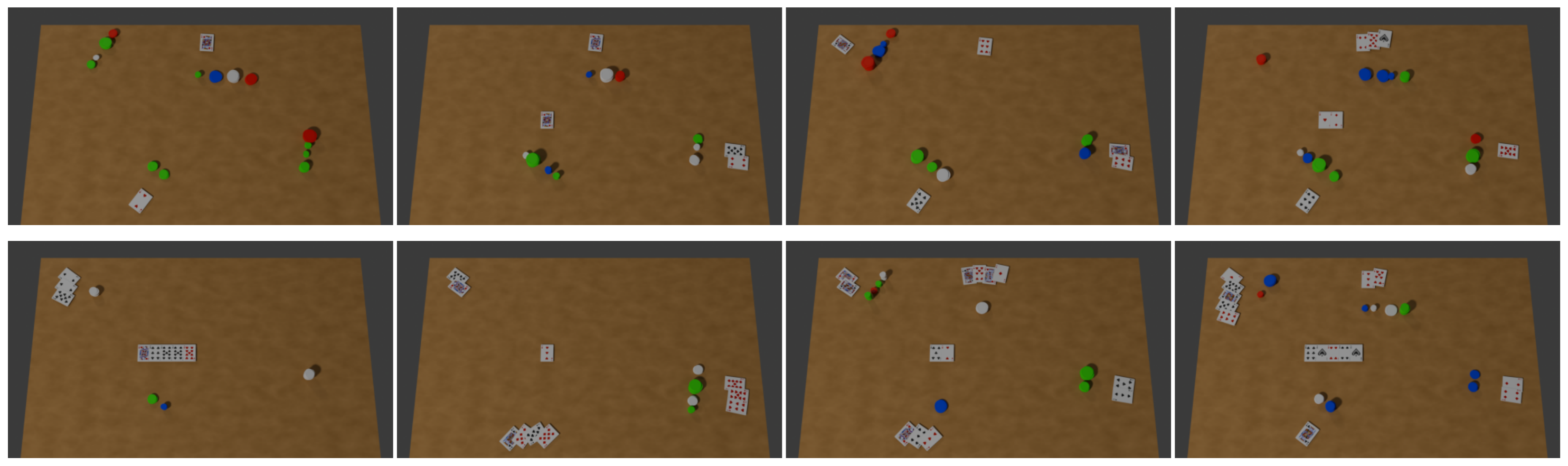}
    \caption{Variations of the number of cards in a Poker scenes (cropped images with overlap for display purposes) }
    \label{fig:poker_count}
\end{figure}

\begin{figure}[H]
    \centering
    \includegraphics[width=0.95\linewidth]{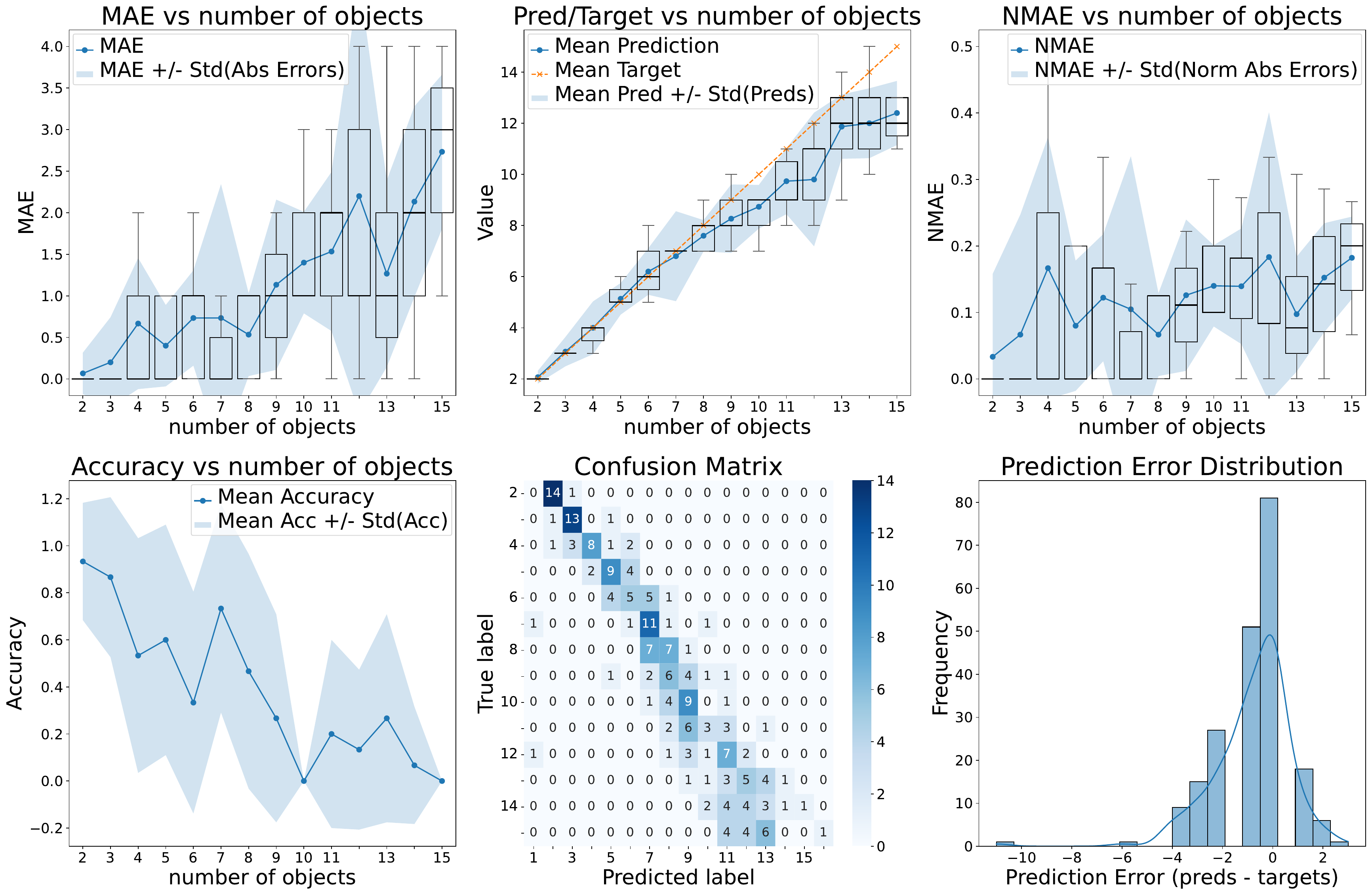}
    \caption{Results of GPT-4.1 for the counting question (\textit{How many cards are present in the entire scene?}) on the Poker dataset. Mean and standard deviation over 15 samples per level.}
    \label{fig:gpt4-1_results_count_poker}
\end{figure}

\paragraph{Results} GPT-4.1 (Figure \ref{fig:gpt4-1_results_count_poker}). Across 15 poker scenes, the model makes only one or two errors when there are up to two cards on the table. With up to 9 cards, it accurately counts approximately half of the time (accuracy of around 50\%), with an average Mean Absolute Error (MAE) close to 1. Beyond this point, the model consistently underestimates the number of cards by 2 or 3 on average, leading to a significant decline in accuracy. 
The counting task appears to be more challenging, which may be attributed to the fact that the poker game setup exhibits less regular geometries compared to the chess setup.

%Up to 9 cards, it counts properly almost half of the times (except for 6 cards), with an MAE close to 1 in average. Then it systematically underestimates the number of cards, by up to more than 5 elements. 

%%%%%%%%%%%%%%%%%%%%%%%%%%%%%%%%%%%%%%%%%%%%%%%%%%%%%%%%%%%%
\subsection{Counting with blur (Chess)}
\label{subsec:counting_blur_chess}

We present the diagnostic results for the \emph{counting with blur} task on the Chess dataset. The following question is asked with a debiased preprompt and a declarative instruction:
\begin{center}
    "This is not a real chess game. The number of each piece and their position can vary arbitrarily. Just focus on answering the following question based on the visual content. How many pieces are there in the image? Respond in a declarative format: 'The number of pieces in the image is:'"
\end{center}

Examples of images can be found below in Figure \ref{fig:img_count_blur}:

\begin{figure}[h]
    \centering
    \includegraphics[width=\linewidth]{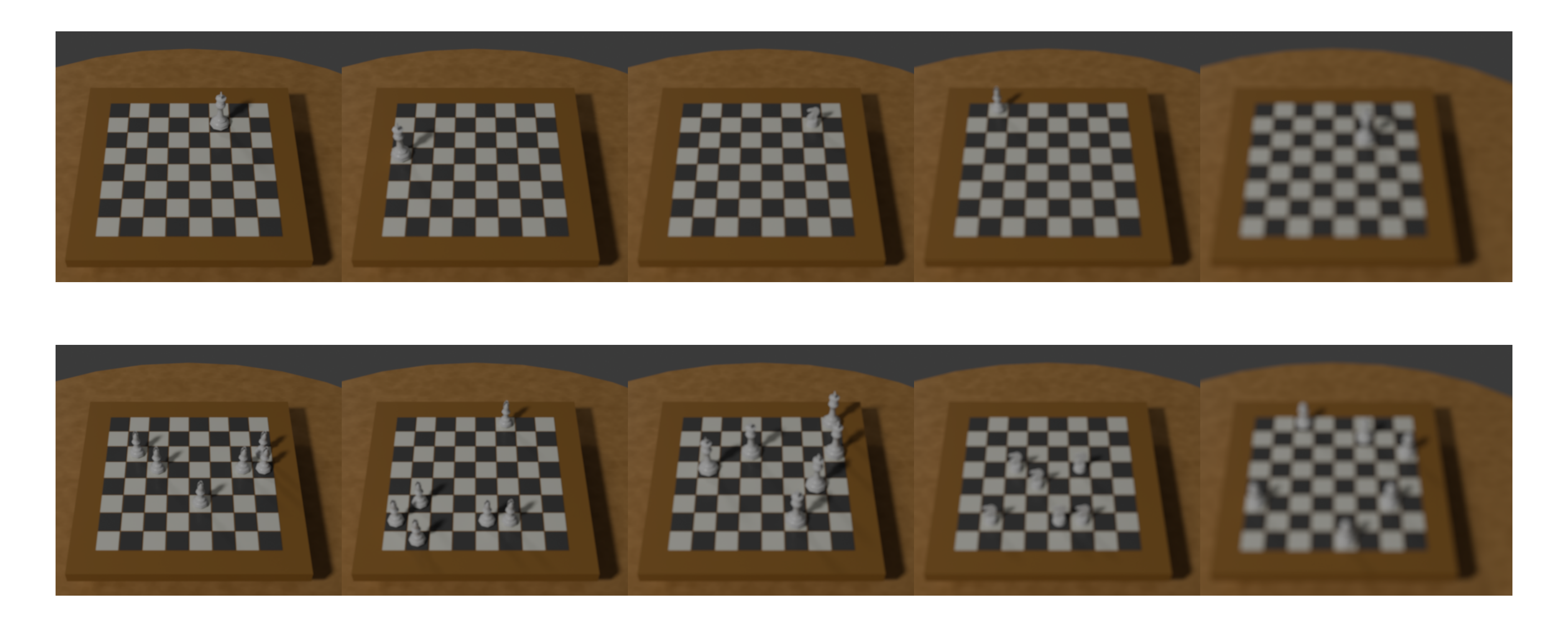}
    \caption{Variations of the five levels of blur for a chess scene, ranging from low (left) to high (right) levels. The images on the top display a single piece, while those on the bottom show multiple pieces. }
    \label{fig:img_count_blur}
\end{figure}

\begin{figure}[hb]
    \centering
    \includegraphics[width=0.95\linewidth]{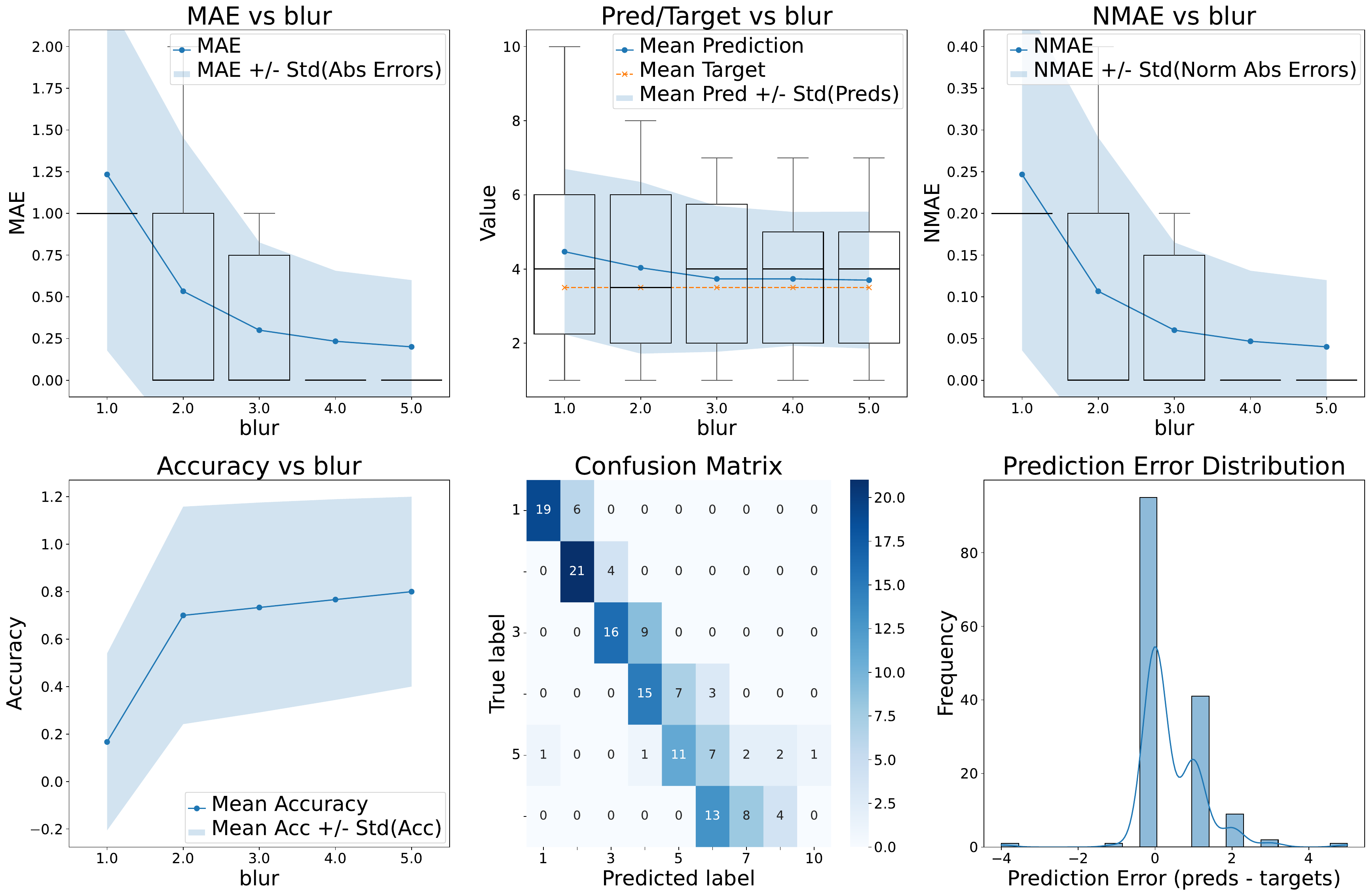}
    \caption{Results of GPT-4.1 for the counting question under blur decrease (higher values meaning less blur) on the Chess dataset. Mean and standard deviation over 5 levels (from 1 to 5), with 5 samples per level.}
    \label{fig:gpt4-1_results_count_blur_chess_gpt}
\end{figure}

\begin{figure}[ht]
    \centering
    \includegraphics[width=0.95\linewidth]{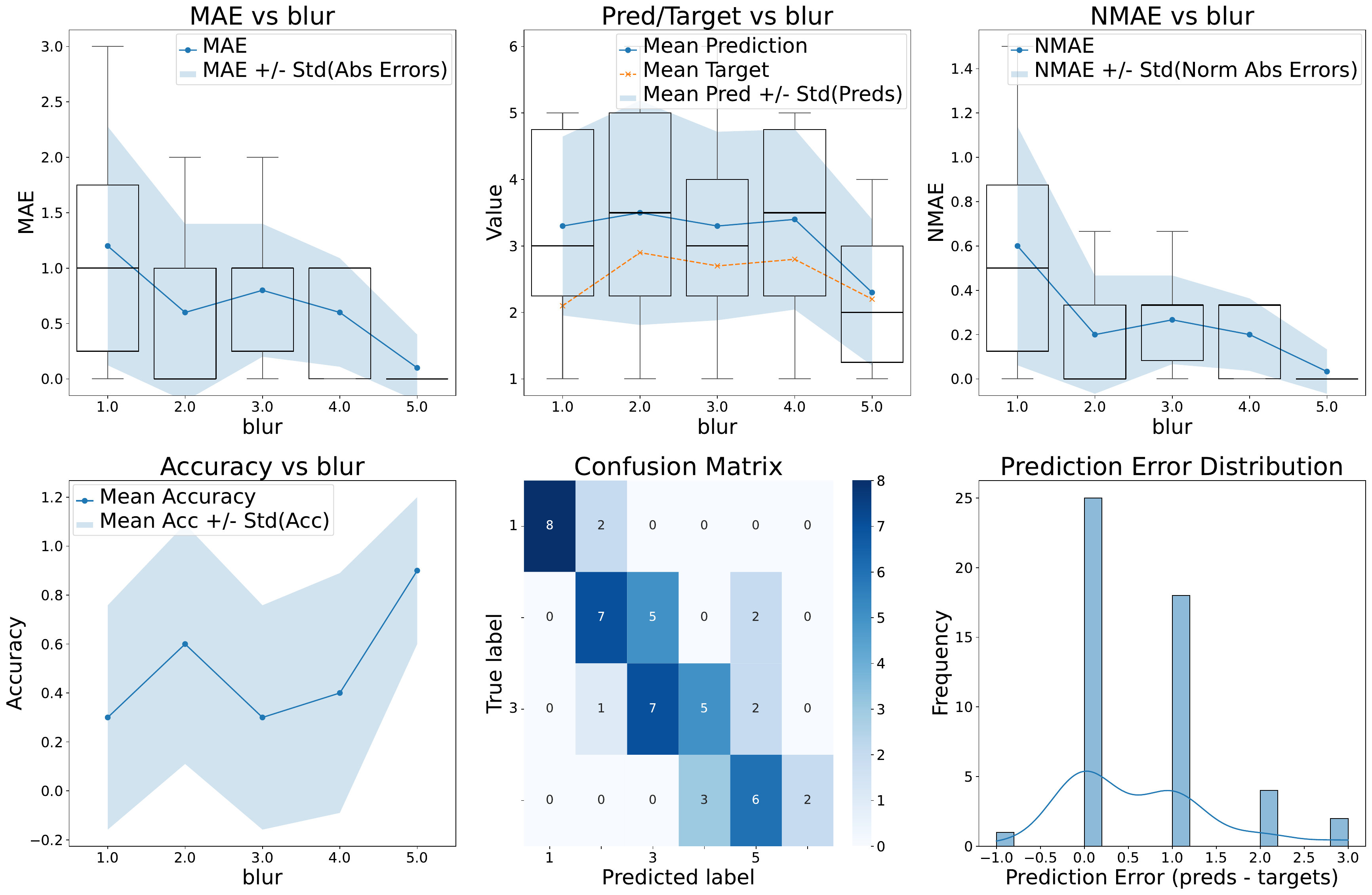}
    \caption{Results of LLaMA-4-Scout for the counting question under blur decrease (higher values meaning less blur) on the Chess dataset. Mean and standard deviation over 5 levels (from 1 to 5), with 5 samples per level.}
    \label{fig:LLaMA4_results_count_blur_chess}
\end{figure}

\begin{figure}[H]
    \centering
    \begin{subfigure}[b]{0.48\linewidth}
        \centering
        \includegraphics[width=\linewidth]{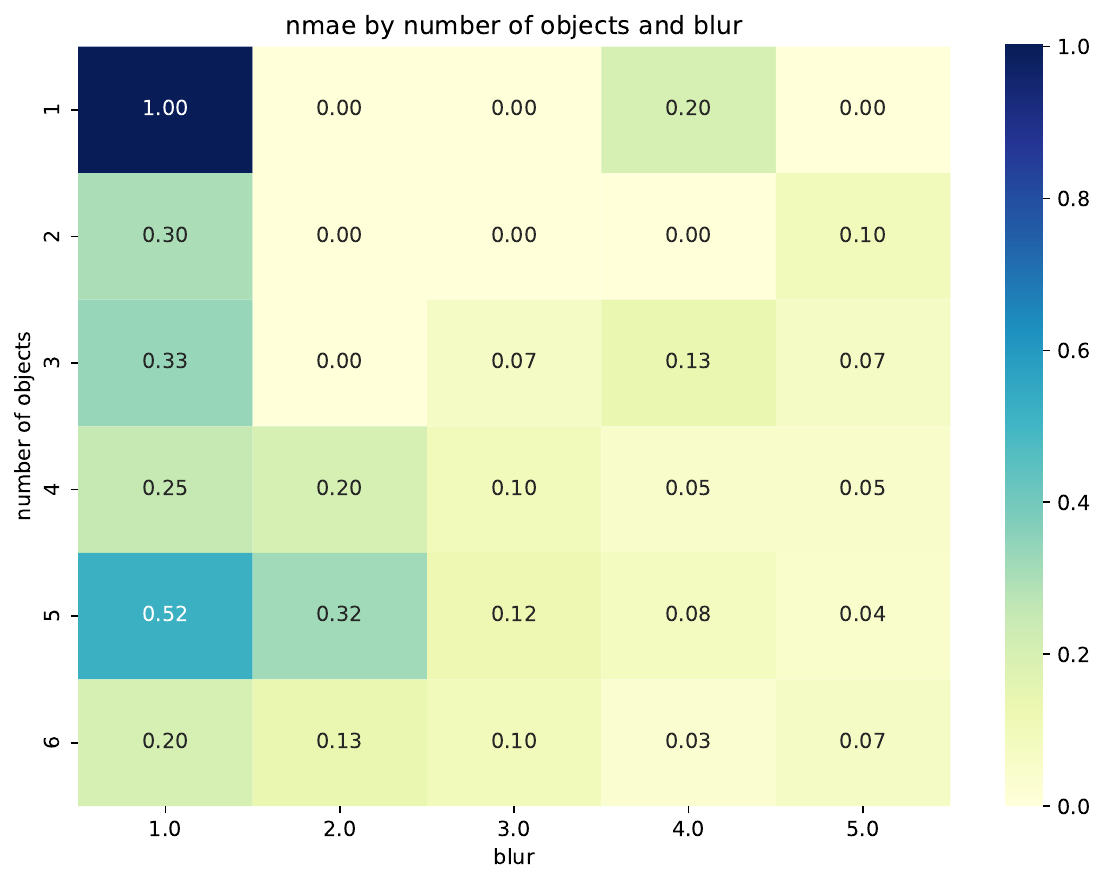}
        \caption{GPT-4.1}
        \label{fig:gpt4-1_crossplot_blur_chess_gpt}
    \end{subfigure}
    \hspace{0.02\linewidth}
    \begin{subfigure}[b]{0.48\linewidth}
        \centering
        \includegraphics[width=\linewidth]{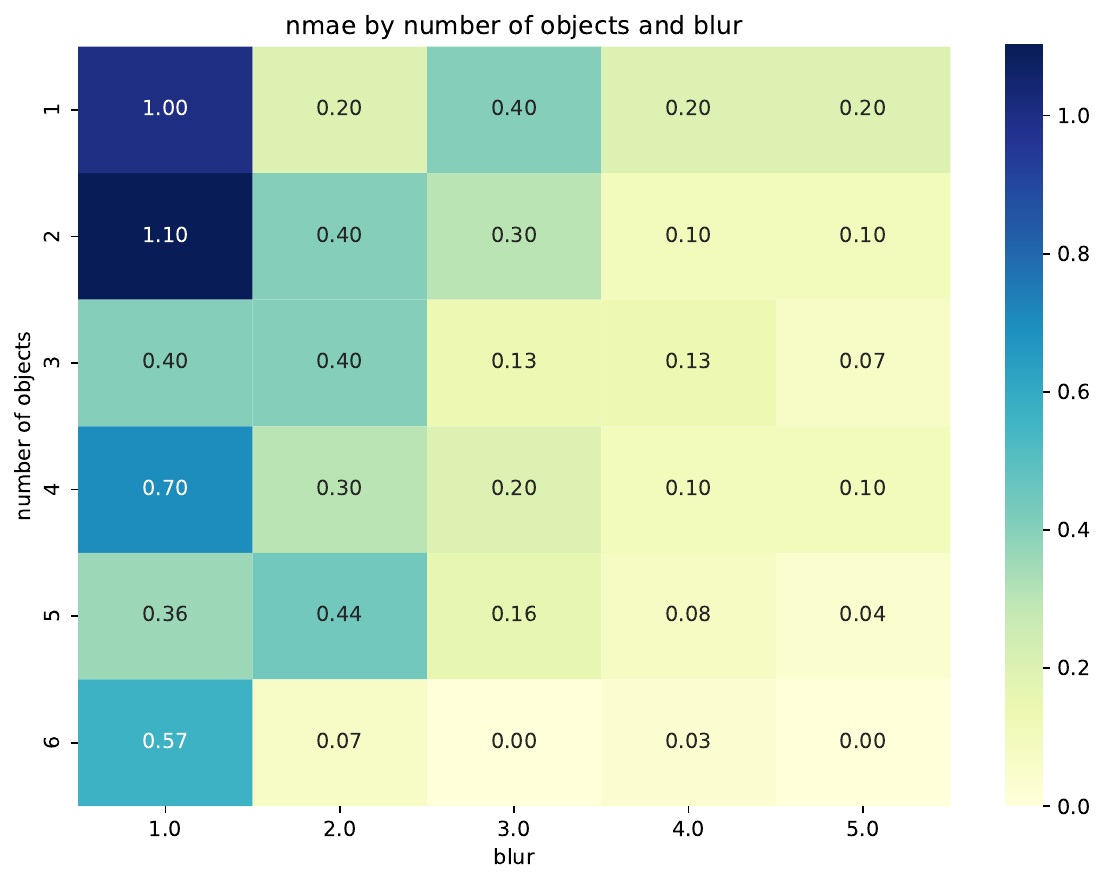}
        \caption{LLaMA-4-Scout}
        \label{fig:LLaMA4_crossplot_blur_chess}
    \end{subfigure}
    \caption{Normalized MAE of (a) GPT-4.1 and (b) LLaMA-4-Scout on cross-variations of blur and number of pieces. Higher values mean less blur. Average over 5 samples per cell.}
    \label{fig:crossplot_blur_chess_gpt_LLaMA}
\end{figure}

\paragraph{Results}
\begin{itemize}
    \item GPT-4.1 (Figure \ref{fig:gpt4-1_results_count_blur_chess_gpt}). The model's performance clearly deteriorates as the level of blur increases, with a sharp decline in accuracy at the highest blur level (level 1). GPT-4.1 tends to overestimate the number of pieces as blur increases, which is expected, as higher levels of blur can merge pixels, creating the illusion of objects where none exist. Figure \ref{fig:gpt4-1_crossplot_blur_chess_gpt} illustrates that the primary source of difficulty is the increase in blur, rather than the rise in the number of pieces, as the Normalized Mean Absolute Error (NMAE) remains relatively stable (or even decreases for larger numbers) for a fixed blur value. 
    \item LLaMA-4-Scout (Figure \ref{fig:LLaMA4_results_count_blur_chess}). LLaMA-4-Scout exhibits a similar overestimation pattern, with a marginally higher MAE and the distribution of prediction errors slightly shifted to the right. Figure \ref{fig:LLaMA4_results_count_blur_chess} demonstrates that this model experiences greater difficulty with larger numbers of objects under high blur compared to GPT-4.1.
\end{itemize}

%%%%%%%%%%%%%%%%%%%%%%%%%%%%%%%%%%%%%%%%%%
\subsection{Localization of a single piece on a 4x4 grid (Chess)}
\label{subsec:localization_single_chess}

We present the diagnostic results for the \emph{localization of a single piece} task on a 4x4 Chess board. The following questions are asked with a debiased preprompt and a declarative instruction:
\begin{center}
    "This is not a real chess game. The number and position of the pieces can vary arbitrarily. Just focus on answering the following question based on the visual content. Numbering the columns from left to right, starting with 0, on which column is the piece on the board? Respond in a declarative format."
\end{center}
\begin{center}
    "This is not a real chess game. The number and position of the pieces can vary arbitrarily. Just focus on answering the following question based on the visual content. Numbering the rows from top to bottom, starting with 0, on which row is the piece on the board? Respond in a declarative format."
\end{center}

Examples of images can be found below in Figure \ref{fig:chess_loc_4x4}:

\begin{figure}[H]
    \centering
    \includegraphics[width=\linewidth, trim=0cm 2.7cm 0cm 3cm, clip]{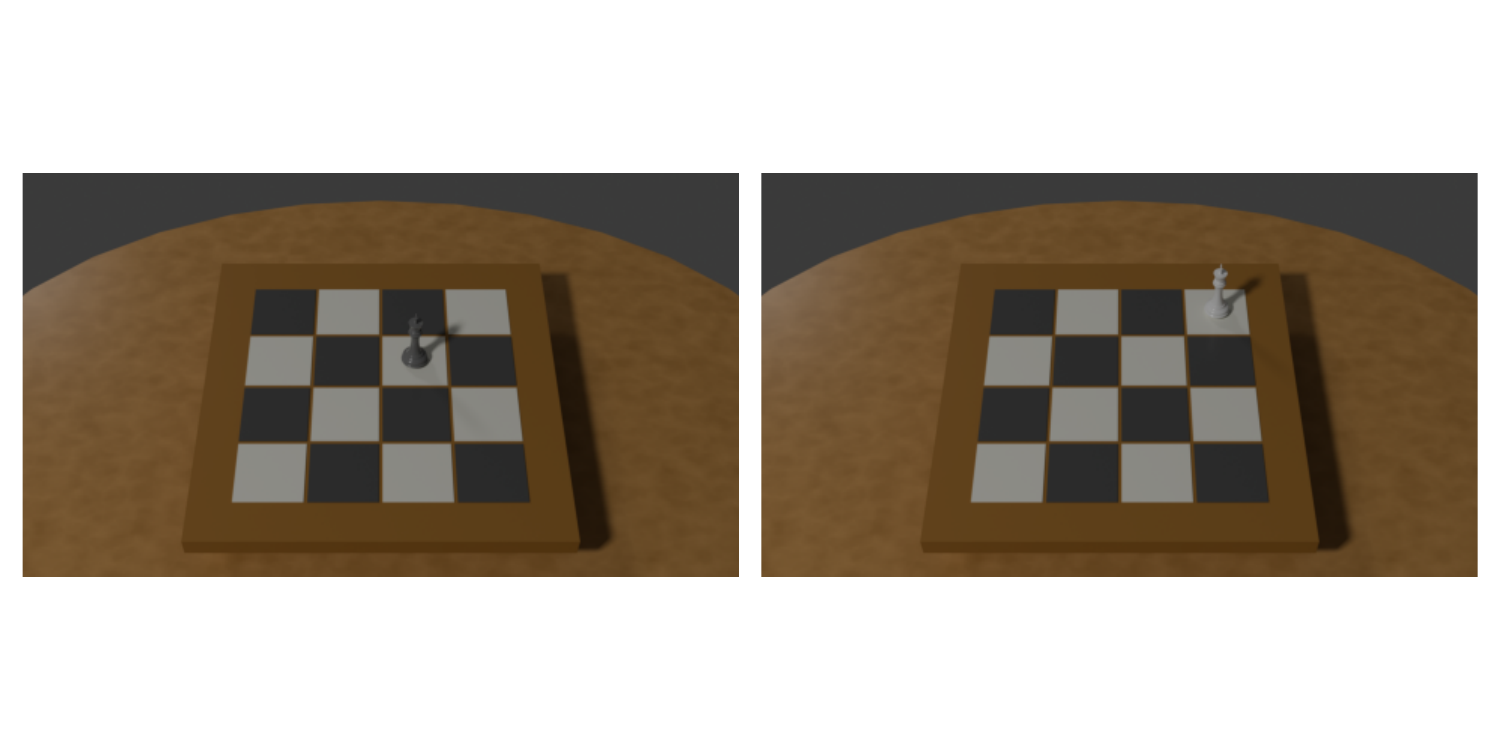}
    \caption{Illustration of the localization task for a single Chess piece on a 4x4 grid}
    \label{fig:chess_loc_4x4}
\end{figure}

\begin{figure}[H]
    \centering
    \includegraphics[width=0.8\linewidth]{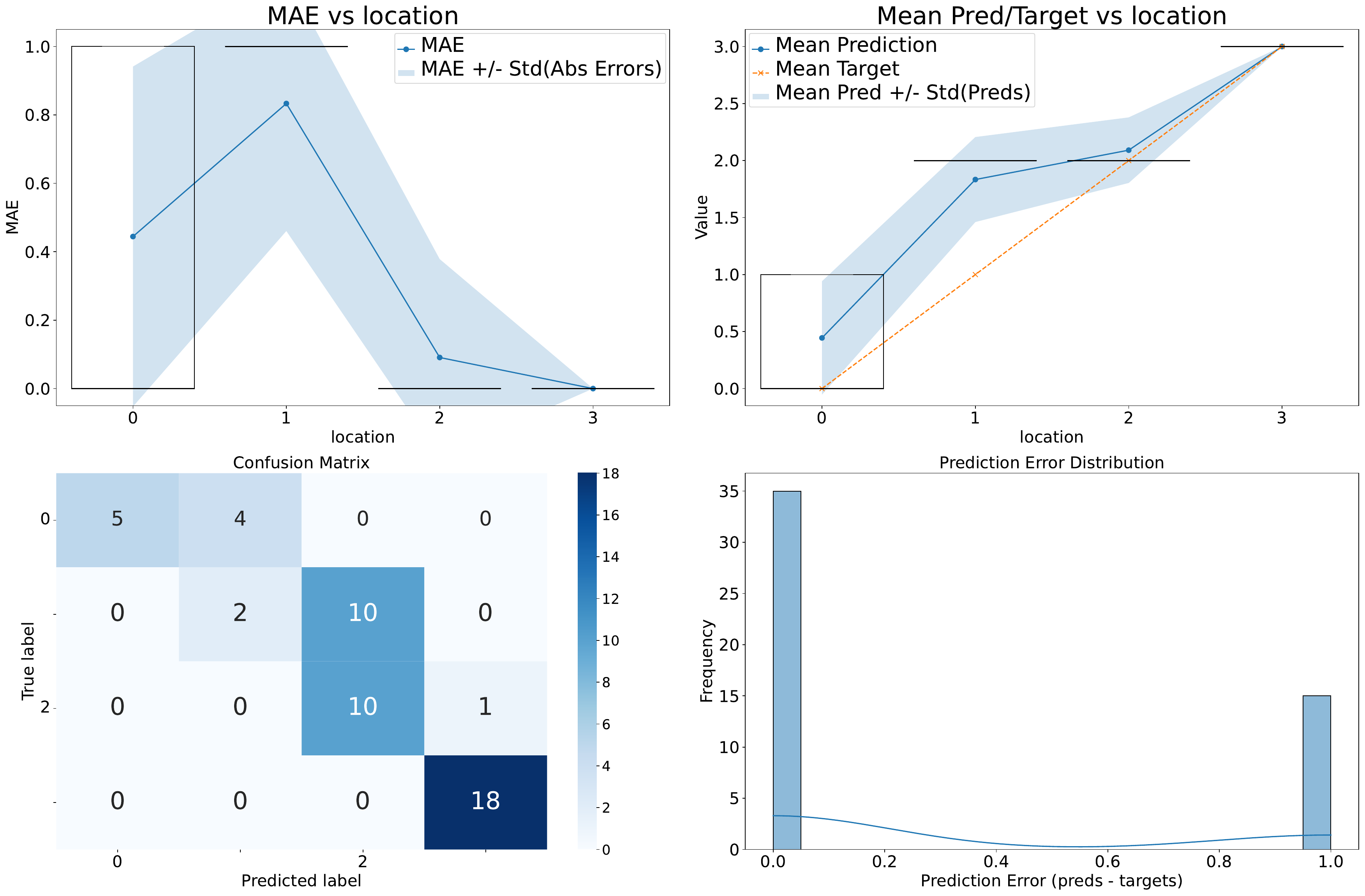}
    \caption{GPT-4.1 results for horizontal (column) localization of a single chess piece on a 4x4 grid. 50 samples overall.}
    \label{fig:gpt4-1_results_localization_col_4x4_chess}
\end{figure}

\begin{figure}[H]
    \centering
    \includegraphics[width=\linewidth]{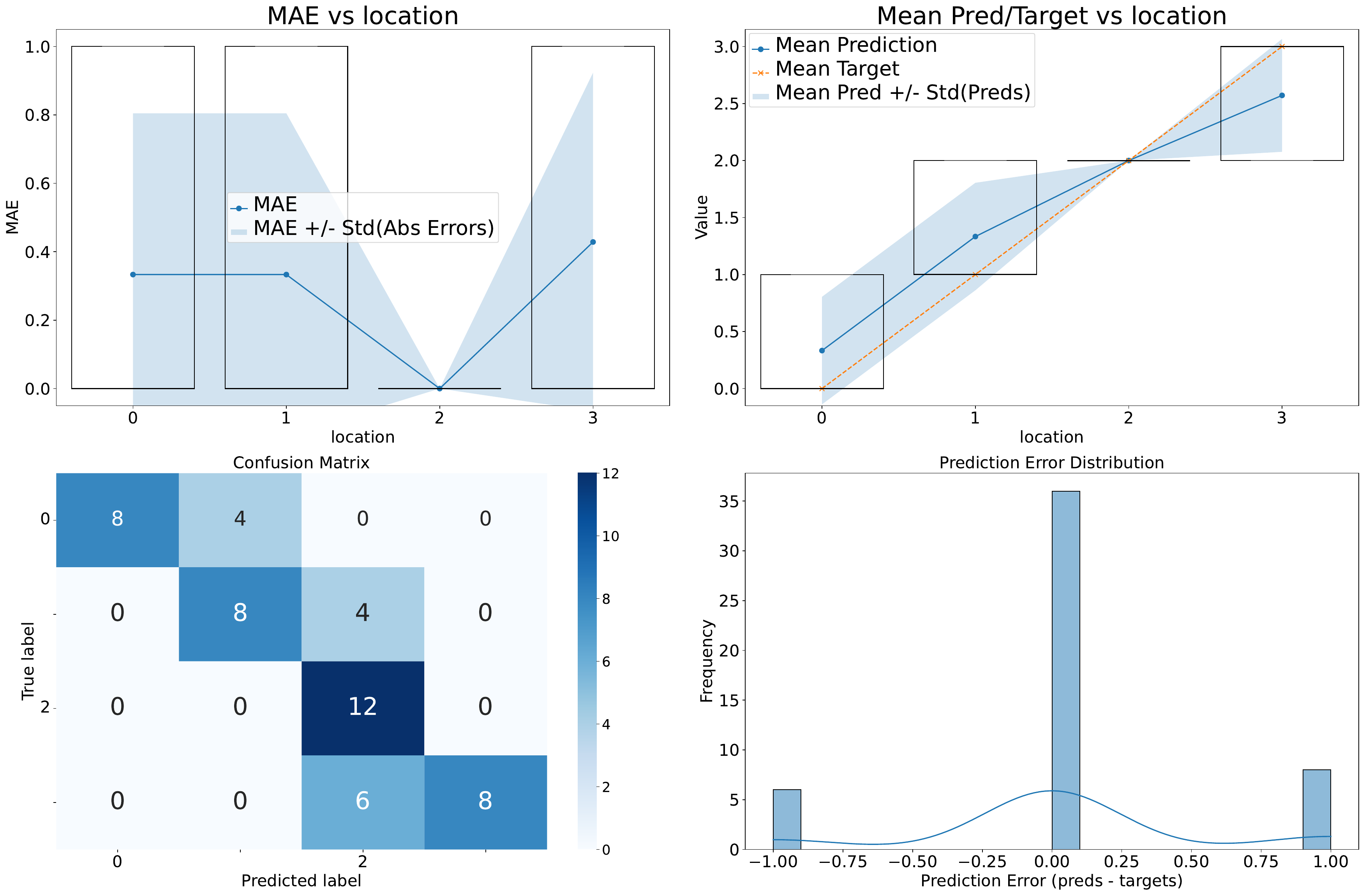}
    \caption{GPT-4.1 results for vertical (row) localization of a single chess piece on a 4x4 grid. 50 samples overall.}
    \label{fig:gpt4-1_results_localization_row_4x4_chess}
\end{figure}

\begin{figure}[H]
    \centering
    \includegraphics[width=\linewidth]{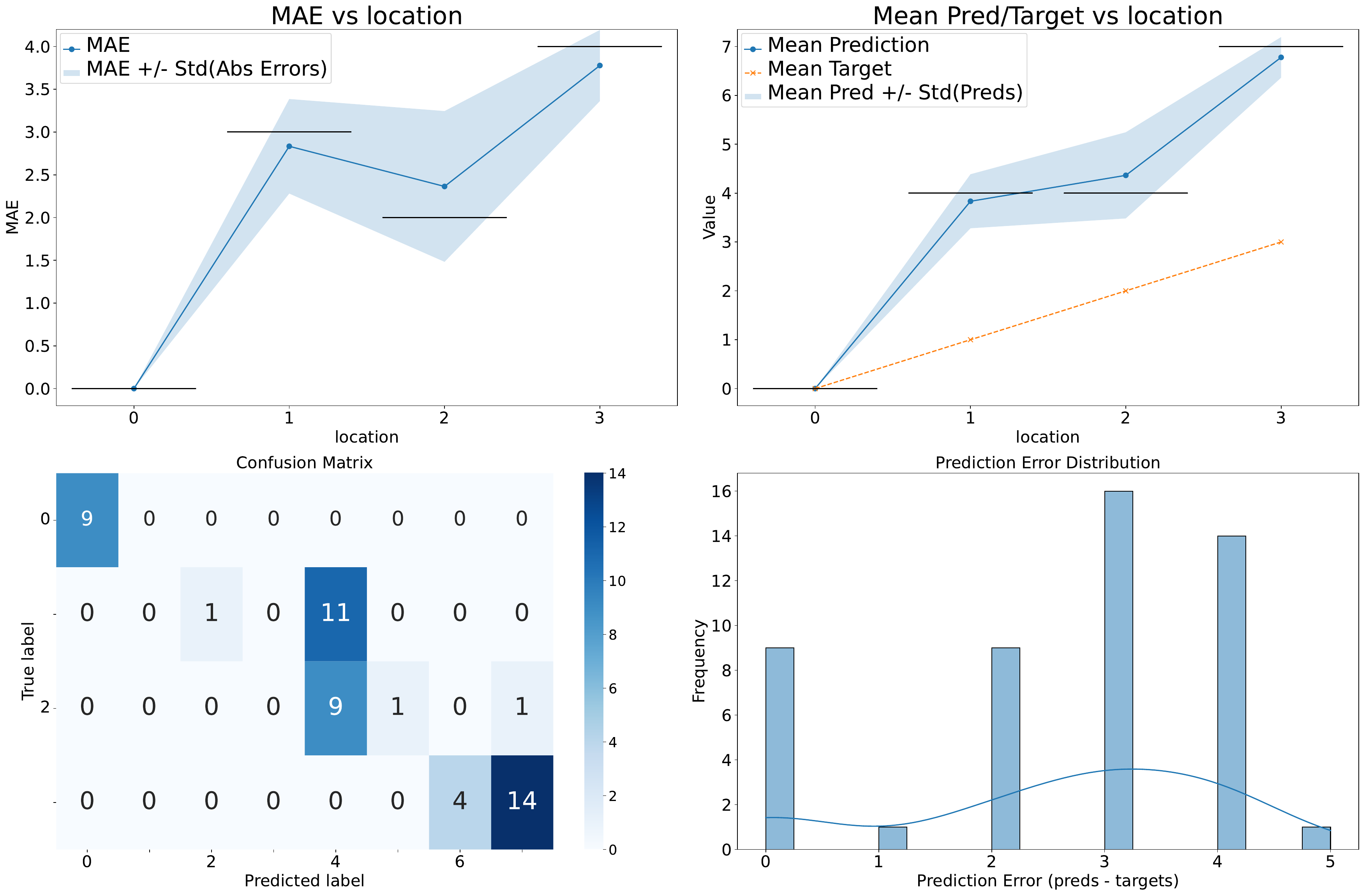}
    \caption{LLaMA-4-Scout results for horizontal (column) localization of a single chess piece on a 4x4 grid.}
    \label{fig:LLaMA_results_localization_col_4x4_chess}
\end{figure}

\begin{figure}[H]
    \centering
    \includegraphics[width=0.8\linewidth]{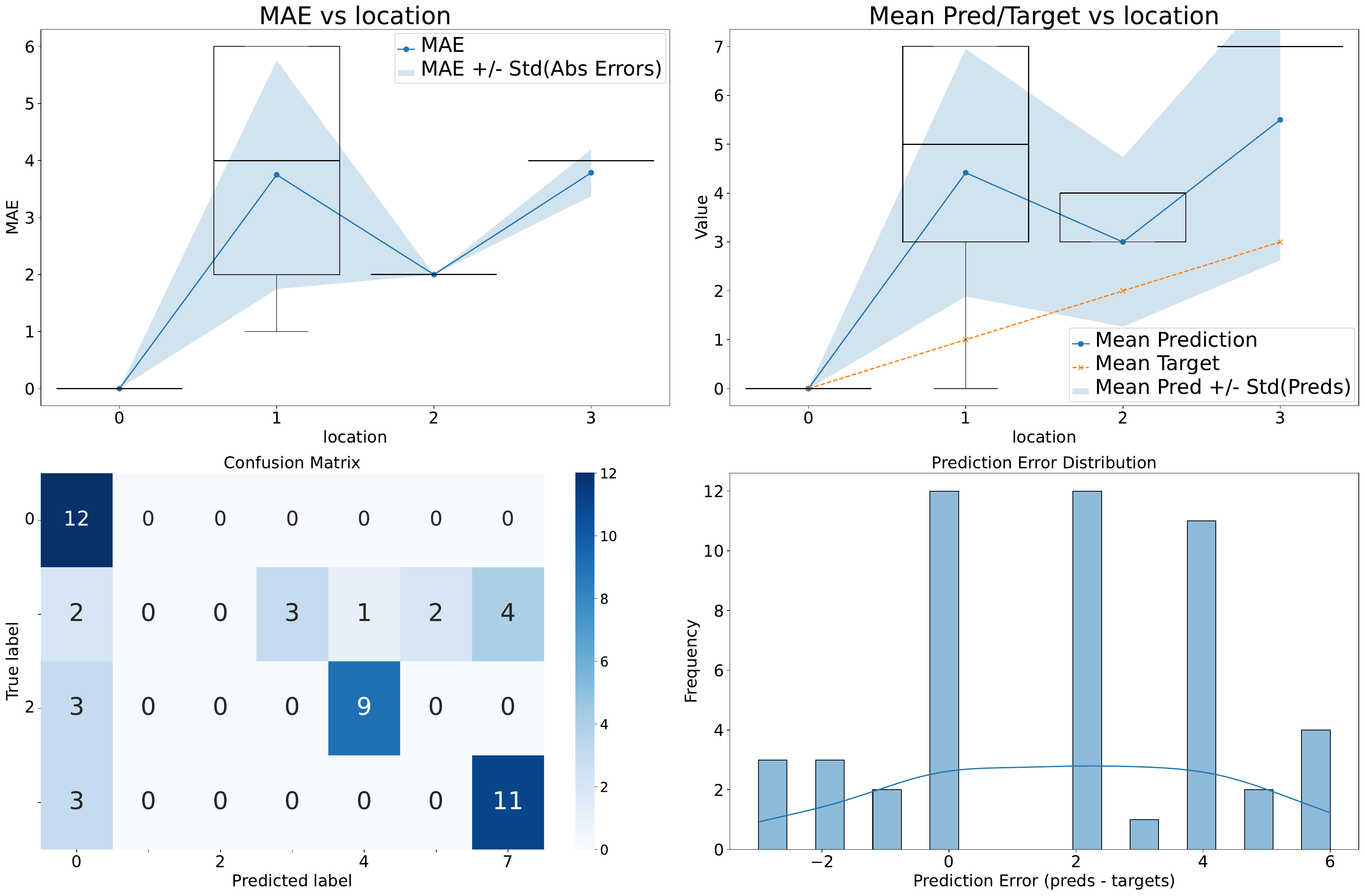}
    \caption{LLaMA-4-Scout results for vertical (row) localization of a single chess piece on a 4x4 grid. 50 samples overall.}
    \label{fig:LLaMA_results_localization_row_4x4_chess}
\end{figure}

\paragraph{Results}
\begin{itemize}
    \item GPT-4.1 (Figures \ref{fig:gpt4-1_results_localization_col_4x4_chess}, \ref{fig:gpt4-1_results_localization_row_4x4_chess}). The performance of the model is rather stable across the piece position, with an error of at most one column or row.
    \item LLaMA-4-Scout (Figure \ref{fig:LLaMA_results_localization_col_4x4_chess}, \ref{fig:LLaMA_results_localization_row_4x4_chess}). 
    %The model performs slightly better at finding the column rather than the row. 
    The model performs worse than GPT-4.1, exhibiting a clear overestimation bias for rows and columns beyond the first. In both cases, it frequently predicts row or column "7," likely reflecting a bias towards identifying an 8x8 chessboard and its extremities. This leads to a right-shifted distribution of prediction errors for both row and column localization.  
\end{itemize}

%%%%%%%%%%%%%%%%%%%%%%%%%%%%%%%%%%%%%%%%%%
\subsection{Localization of a single card on a 3x3 grid (Poker)}
\label{subsec:localization_single_poker}

We present the diagnostic results for the \emph{localization of a single card} task on a 3x3 Poker grid. The following questions are asked with a debiased preprompt and a declarative instruction:
\begin{center}
    "This is not a real poker game. The cards and their position can vary arbitrarily. Just focus on answering the following question based on the visual content. Numbering the rows from top to bottom, starting with 0, on which row is the card? Respond in a declarative format: 'The row number of the grid where card X is located is:'"
\end{center}

\begin{center}
    "This is not a real poker game. The cards and their position can vary arbitrarily. Just focus on answering the following question based on the visual content. Numbering the columns from left to right, starting with 0, on which column is the card? Respond in a declarative format: 'The column number of the grid where card X is located is:'"
\end{center}

Examples of images can be found below in Figure \ref{fig:poker_grid_viz}:

\begin{figure}[H]
    \centering
    \includegraphics[width=\linewidth, trim=0cm 2.7cm 0cm 3cm, clip]{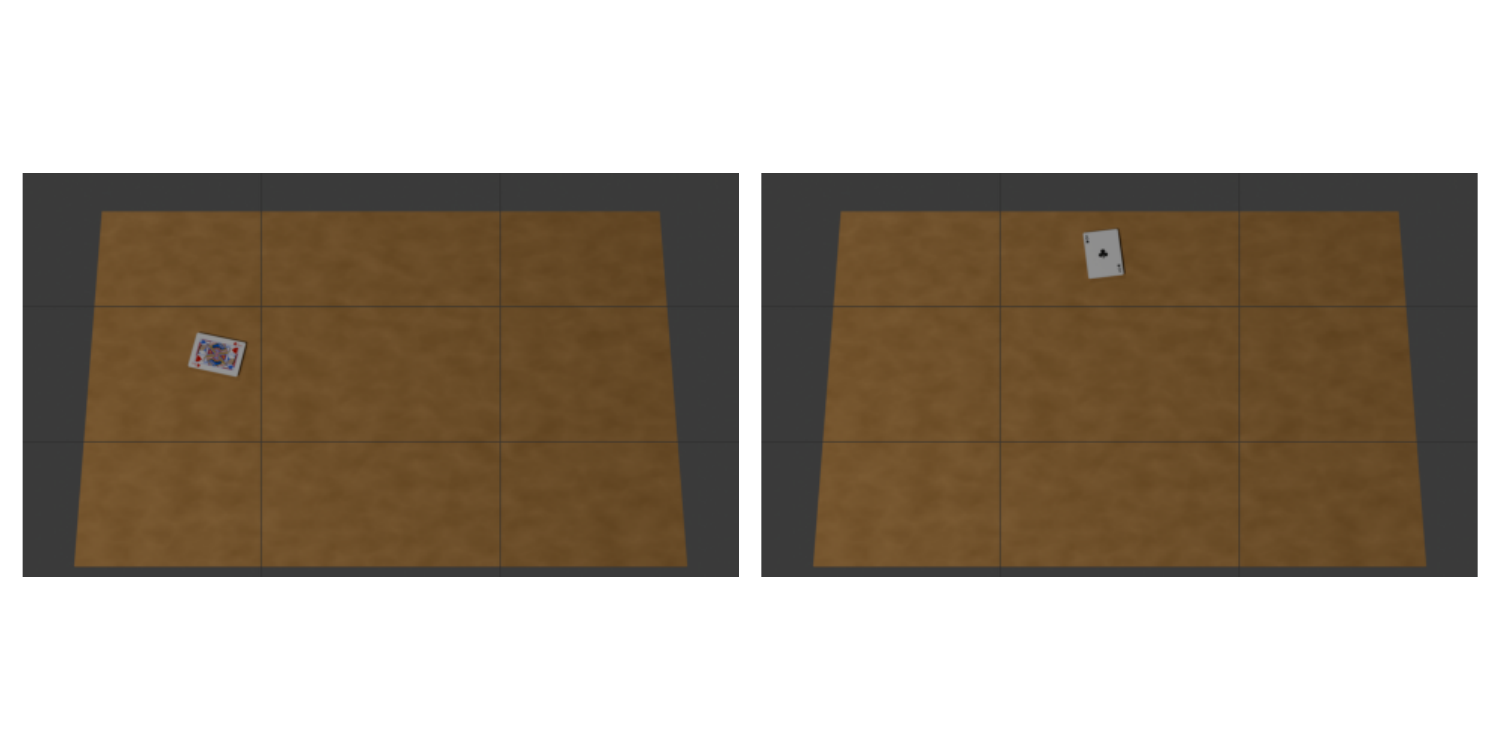}
    \caption{Illustration of the localization task of a single Poker card on a 3x3 grid}
    \label{fig:poker_grid_viz}
\end{figure}

\begin{figure}[H]
    \centering
    \includegraphics[width=0.8\linewidth]{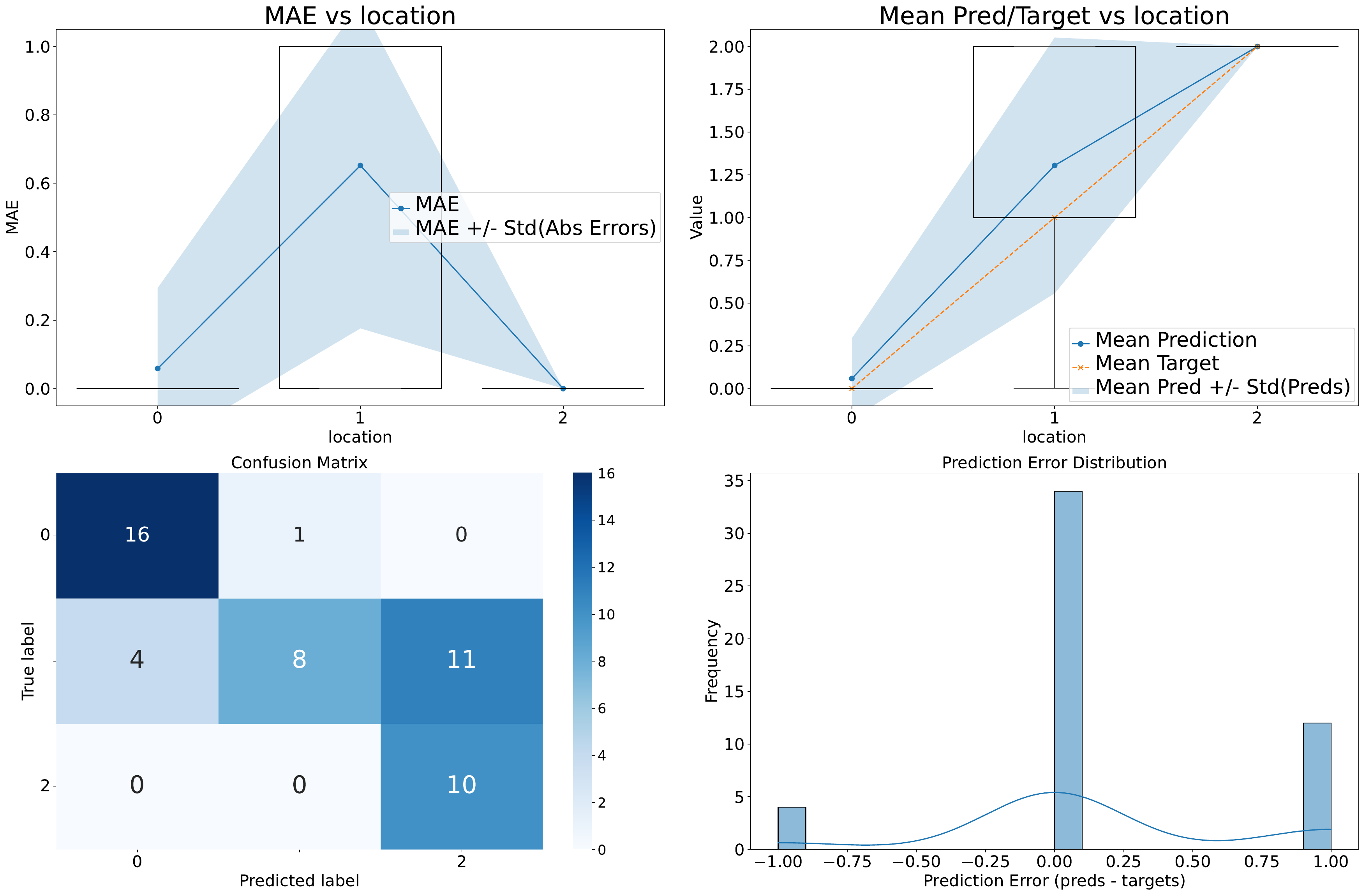}
    \caption{GPT-4.1 results for horizontal (column) localization of a single card on a 3x3 Poker grid. 50 samples overall.}
    \label{fig:gpt4-1_results_localization_col_3x3_poker}
\end{figure}

\begin{figure}[H]
    \centering
    \includegraphics[width=0.8\linewidth]{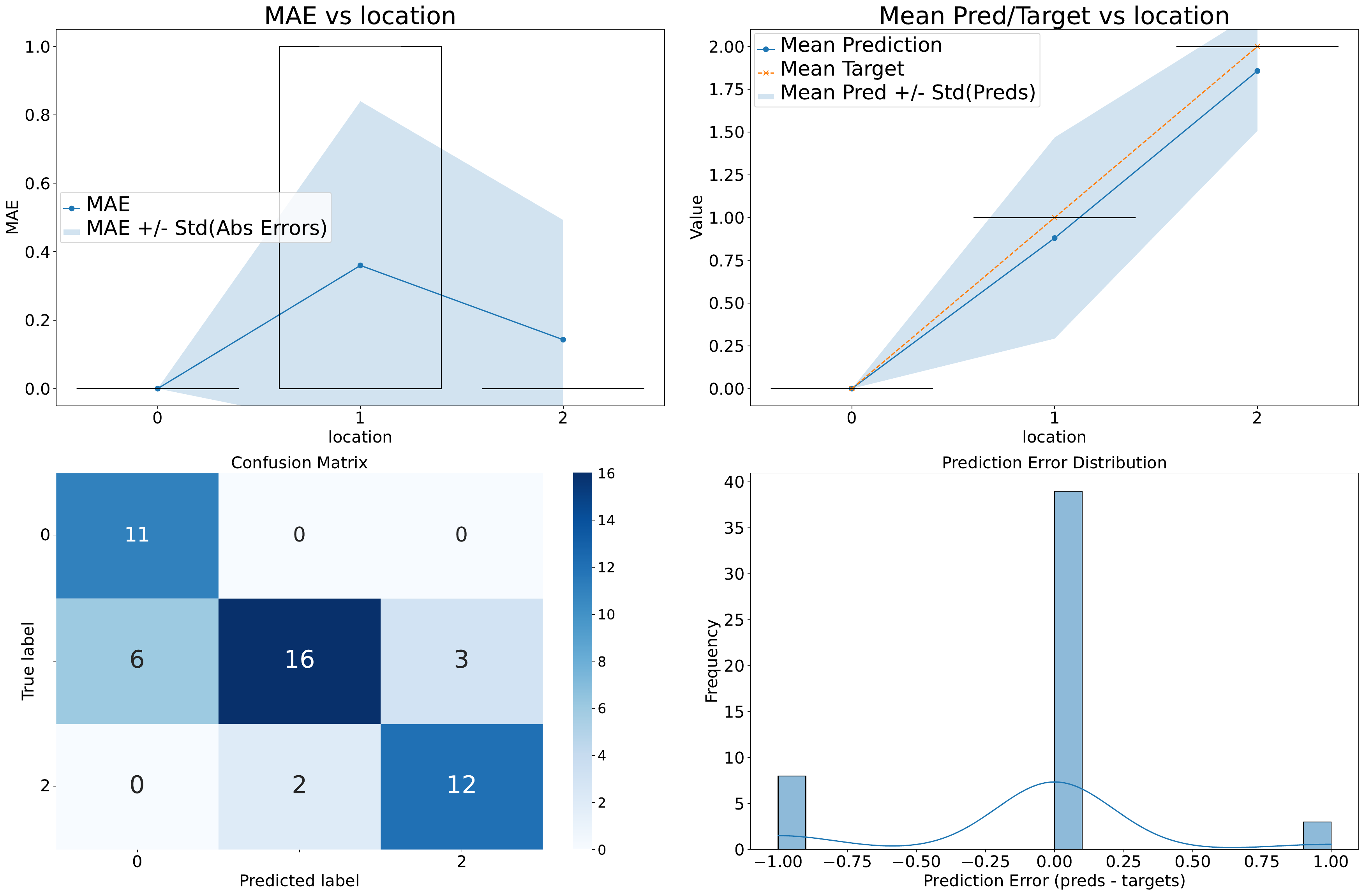}
    \caption{GPT-4.1 results for vertical (row) localization of a single card on a 3x3 Poker grid. 50 samples overall.}
    \label{fig:gpt4-1_results_localization_row_3x3_poker}
\end{figure}

\begin{figure}[H]
    \centering
    \includegraphics[width=0.8\linewidth]{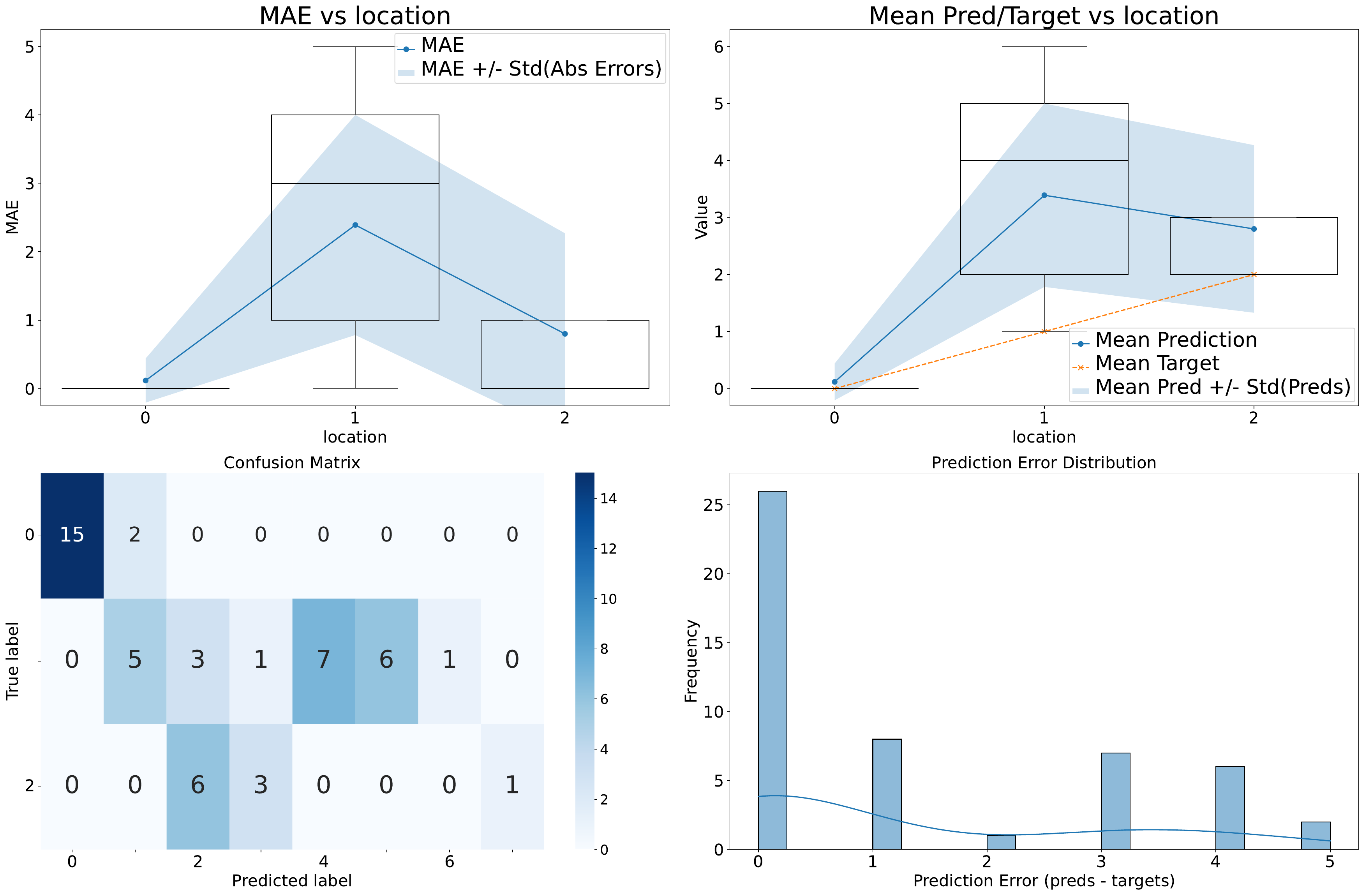}
    \caption{LLaMA results for horizontal (column) localization of a single card on a 3x3 Poker grid. 50 samples overall.}
    \label{fig:LLaMA_results_localization_col_3x3_poker}
\end{figure}

\begin{figure}[H]
    \centering
    \includegraphics[width=0.8\linewidth]{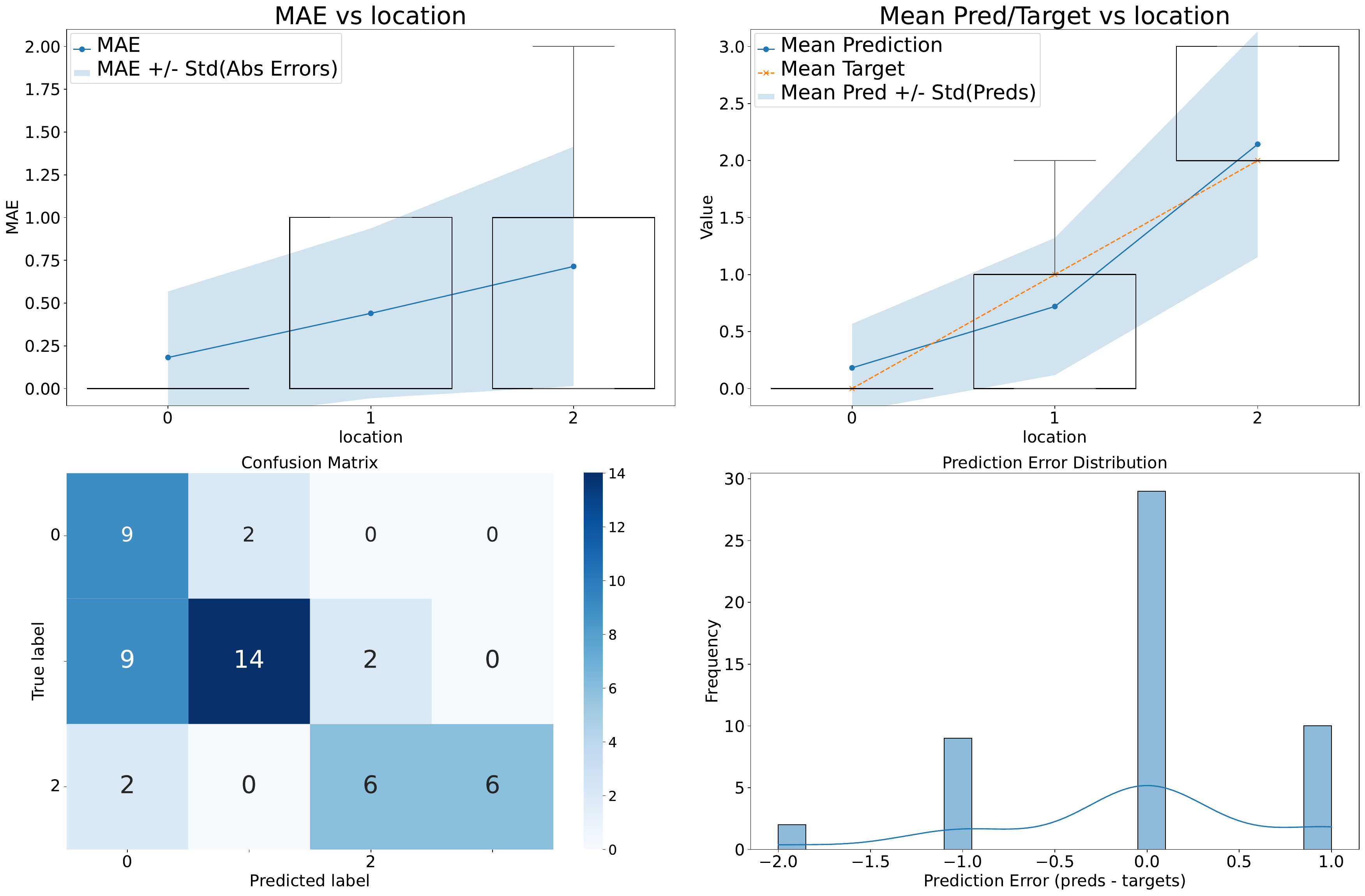}
    \caption{LLaMA results for vertical (row) localization of a single card on a 3x3 Poker grid. 50 samples overall.}
    \label{fig:LLaMA_results_localization_row_3x3_poker}
\end{figure}

\paragraph{Results}
\begin{itemize}
    \item GPT-4.1 (Figures \ref{fig:gpt4-1_results_localization_col_3x3_poker}, \ref{fig:gpt4-1_results_localization_row_3x3_poker}). The model performs well at the edges, but its performance notably declines when the card is positioned in the middle row or column. However, the model maintains a centered distribution of prediction errors.
    \item LLaMA-4-Scout (Figure \ref{fig:LLaMA_results_localization_col_3x3_poker}, \ref{fig:LLaMA_results_localization_row_3x3_poker}). Similarly, LLaMA-4-Scout performs worse when the card is positioned in the middle column, although with a significantly higher MAE. There is also a clear overestimation of the column number when the card is centrally located, leading to a prediction error distribution shifted to the right, without a clearly normalized distribution. Notably, the model tends to identify more rows and columns than are present in the grid. 
    In terms of row localization, performance remains stable across all row positions, with a centralized prediction error distribution and lower MAE. Notably, the model exhibits fewer instances of overestimating higher row numbers, which may be attributed to the rectangular geometry of the grid, where rows are more condensed than columns.
\end{itemize}

%%%%%%%%%%%%%%%%%%%%%%%%%%%%%%%%%%%%%%%%%%
\subsection{Relative localization of two pieces on an 8x8 grid (Chess)}
\label{subsec:relative_localization_chess}

We present the diagnostic results for the \emph{relative localization of two pieces} task on an 8x8 Chess grid. The following question is asked with a debiased preprompt and a declarative instruction:
\begin{center}
    "This is not a real chess game. The number and position of the pieces can vary arbitrarily. Just focus on answering the following question based on the visual content. Numbering the rows from top to bottom, starting with 0, on which row is the piece on the board? Respond in a declarative format."
\end{center}

Examples of images can be found below in Figure \ref{fig:chess_loc_8x8_double}:

\begin{figure}[H]
    \centering
    \includegraphics[width=\linewidth, trim=0cm 2.7cm 0cm 3cm, clip]{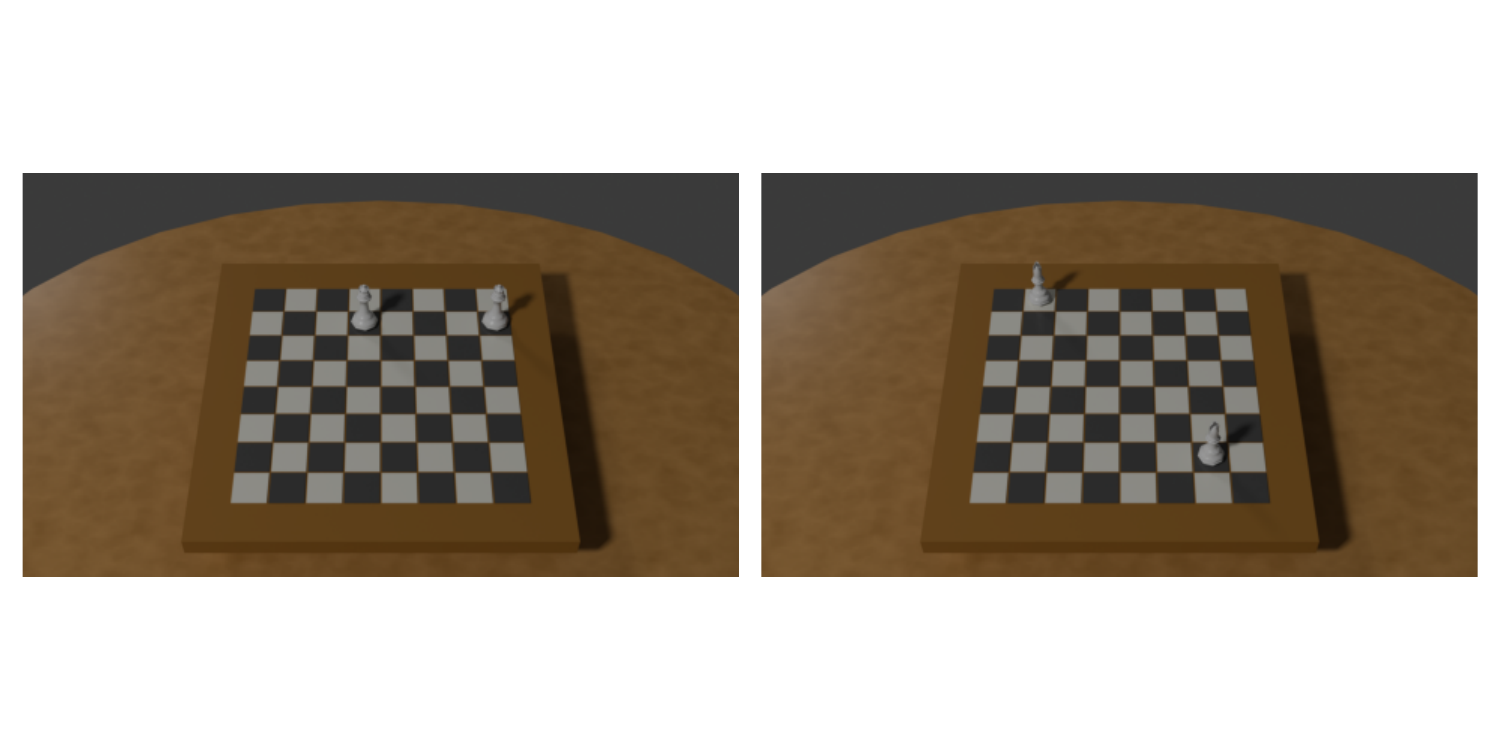}
    \caption{Illustration of the relative localization of two Chess pieces on an 8x8 grid}
    \label{fig:chess_loc_8x8_double}
\end{figure}

\begin{figure}[H]
    \centering
    \includegraphics[width=0.95\linewidth]{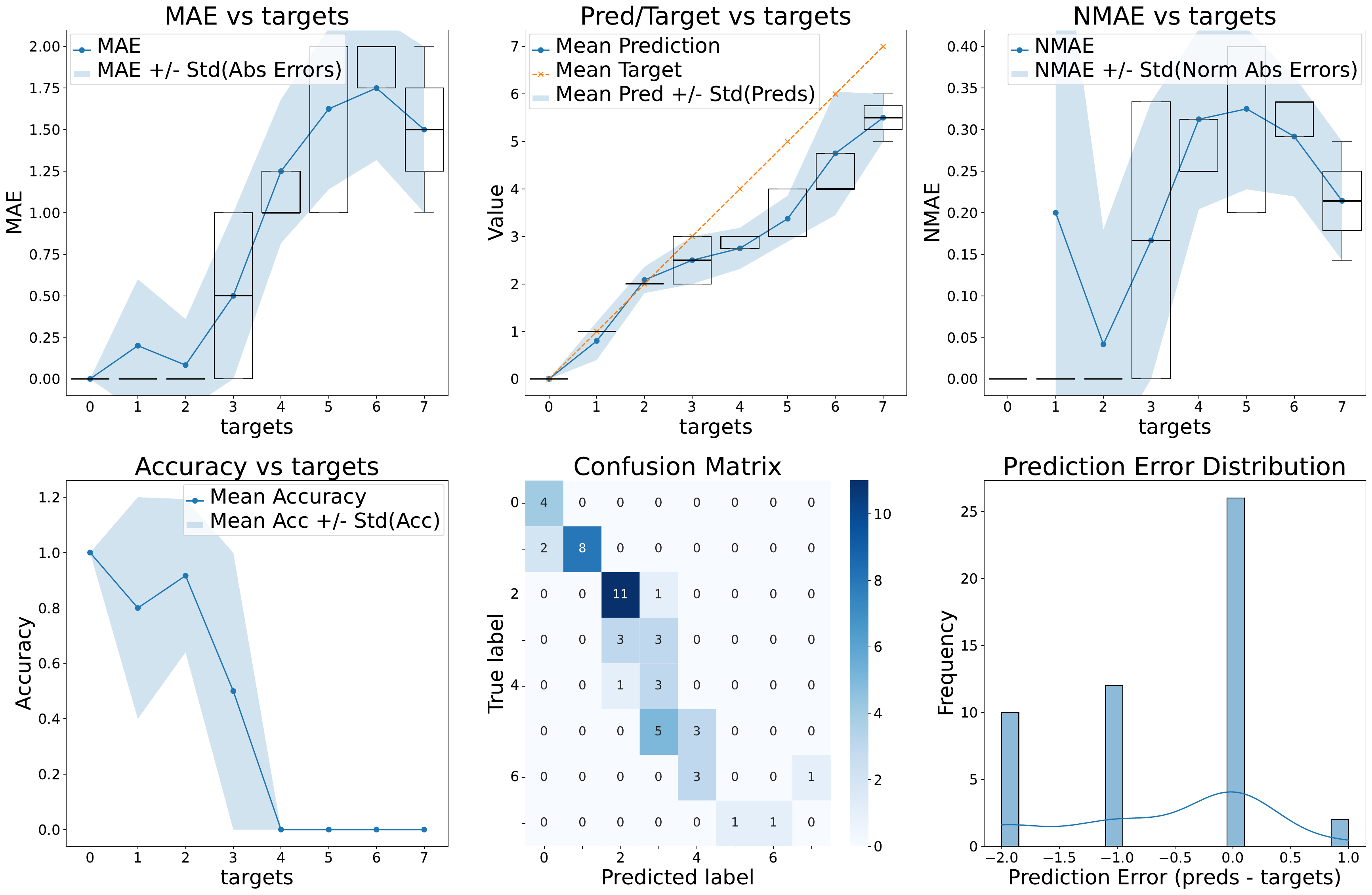}
    \caption{GPT-4.1 results for vertical distance between two pieces on an 8x8 Chess grid. 50 samples overall.}
    \label{fig:gpt4-1_results_rel_localization_row_8x8_chess}
\end{figure}

\begin{figure}[H]
    \centering
    \includegraphics[width=0.95\linewidth]{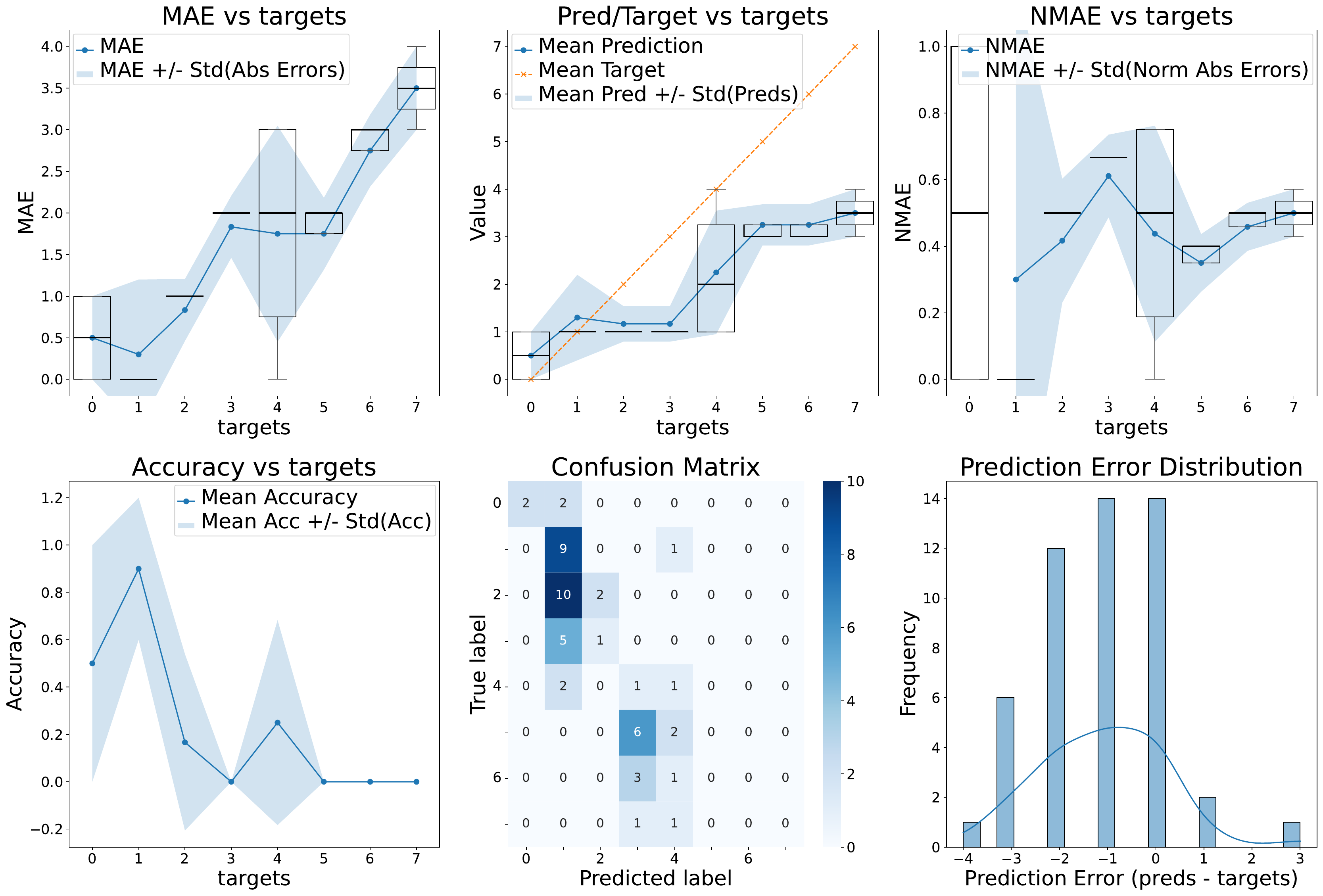}
    \caption{LLaMA-4-Scout results for vertical distance between two pieces on an 8x8 Chess grid. 50 samples overall.}
    \label{fig:LLaMA4_results_rel_localization_row_8x8_chess}
\end{figure}

\paragraph{Results}
\begin{itemize}
    \item GPT-4.1 (Figure \ref{fig:gpt4-1_results_rel_localization_row_8x8_chess}). The model accurately identifies when the two pieces are on the same row, but its performance declines rapidly as the distance between the pieces increases, reaching zero accuracy when the distance is 4 rows or greater. The distribution of prediction errors appear to be shifted to the left. GPT-4.1 demonstrates a clear underestimation of the distance when it exceeds 2 rows, with an MAE of approximately 1.5 for larger distances.  
    \item LLaMA-4-Scout (Figure \ref{fig:LLaMA4_results_rel_localization_row_8x8_chess}). Similarly, LLaMA-4-Scout's performance declines as the distance between the two pieces increases, with accuracy dropping to zero for distances greater than 4 rows. However, the underestimation bias is more pronounced, with an average prediction error of approximately 3 rows for distances of 6 or 7 rows. It consistently predicts a row distance of no more than 4. The distribution of prediction errors is distinctly shifted to the left, centered around -1.
\end{itemize}

While LLaMA-4-Scout exhibited an overestimation bias when localizing a single piece on a 4x4 grid (cf. Figure \ref{fig:LLaMA_results_localization_row_4x4_chess}), it demonstrates a clear underestimation bias for the relative localization of two pieces on an 8x8 grid. This underscores the distinction between the two tasks and emphasizes the importance of conducting both diagnostics. In the first case, the model seems to infer knowledge of a standard 8x8 chessboard, while in the second, it consistently predicts a row distance of no more than 4, failing to account for the full size of the grid.

%\begin{figure}[H]
%    \centering
%    \includegraphics[width=0.95\linewidth]{figures/Appendix_B/localization_chess_4x4/summary_grid_targets_chess_gpt-4_1_declarative_debiased__gpt-4_1_loc_two_pieces_b8x8_col.pdf}
%    \caption{GPT-4.1 results for horizontal distance between two pieces on an 8x8 Chess grid. Mean and standard deviation over 10 samples per level.}
%    \label{fig:gpt4-1_results_rel_localization_col_8x8_chess}
%\end{figure}

%\begin{figure}[H]
%    \centering
%    \includegraphics[width=0.95\linewidth]{figures/Appendix_B/localization_chess_4x4/summary_grid_targets_chess_LLaMA-4-scout_declarative_debiased__localization_2pieces_b8x8_columns.pdf}
%    \caption{LLaMA 4 results for horizontal distance between two pieces on an 8x8 Chess grid. Mean and standard deviation over 10 samples per level.}
%    \label{fig:LLaMA4_results_rel_localization_col_8x8_chess}
%\end{figure}

%%%%%%%%%%%%%%%%%%%%%%%%%%%%%%%%%%%%%
\subsection{Identification with camera distance (Chess)}
\label{subsec:identification_camera_distance_chess}

We present the diagnostic results for the \emph{identification of a single piece} task on the Chess dataset, with camera distance increasing. The following question is asked with a debiased preprompt and a declarative instruction:
\begin{center}
    "This is not a real chess game. The number and position of the pieces can vary arbitrarily. Just focus on answering the following question based on the visual content. What pieces are on the board (among pawn, rook, knight, bishop, king and queen)? Respond in a declarative format."
\end{center}

Examples of images can be found below in Figure \ref{fig:chess_id_distance}:

\begin{figure}[H]
    \centering
    \includegraphics[width=\linewidth]{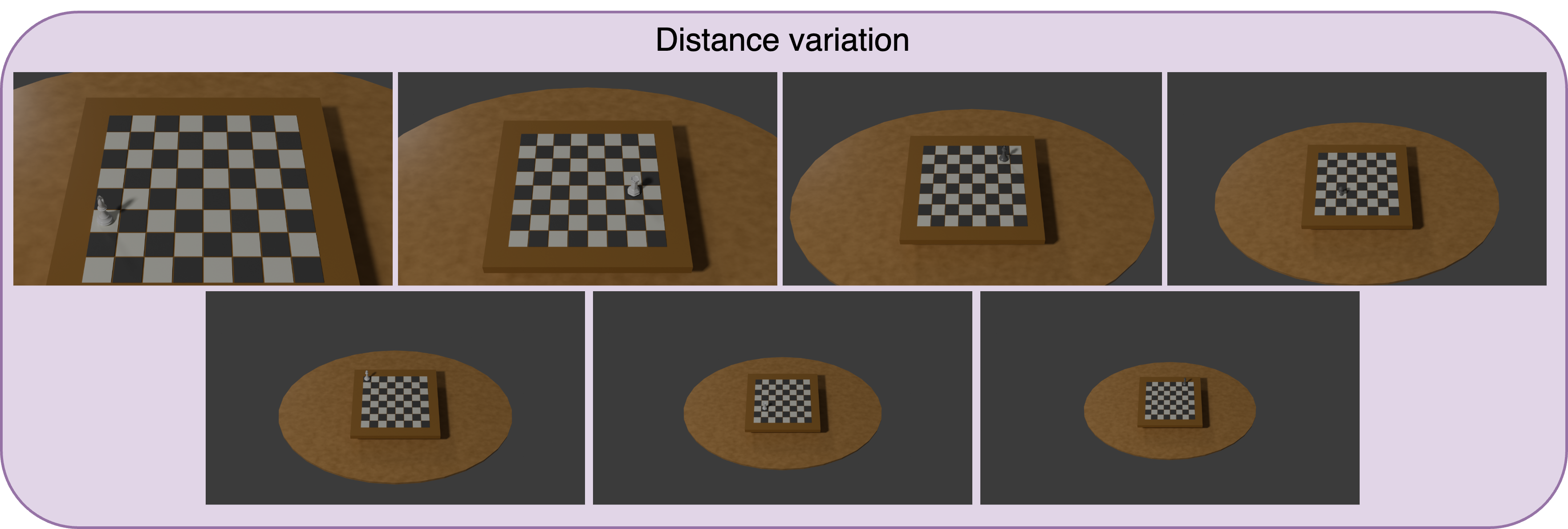}
    \caption{Variations of the 7 distance levels in the chess scene, ranging from closest (top left) to furthest (bottom right), for identifying a single chess piece.}
    \label{fig:chess_id_distance}
\end{figure}

\begin{figure}[H]
    \centering
    \begin{subfigure}[b]{0.48\linewidth}
        \centering
        \includegraphics[width=\linewidth]{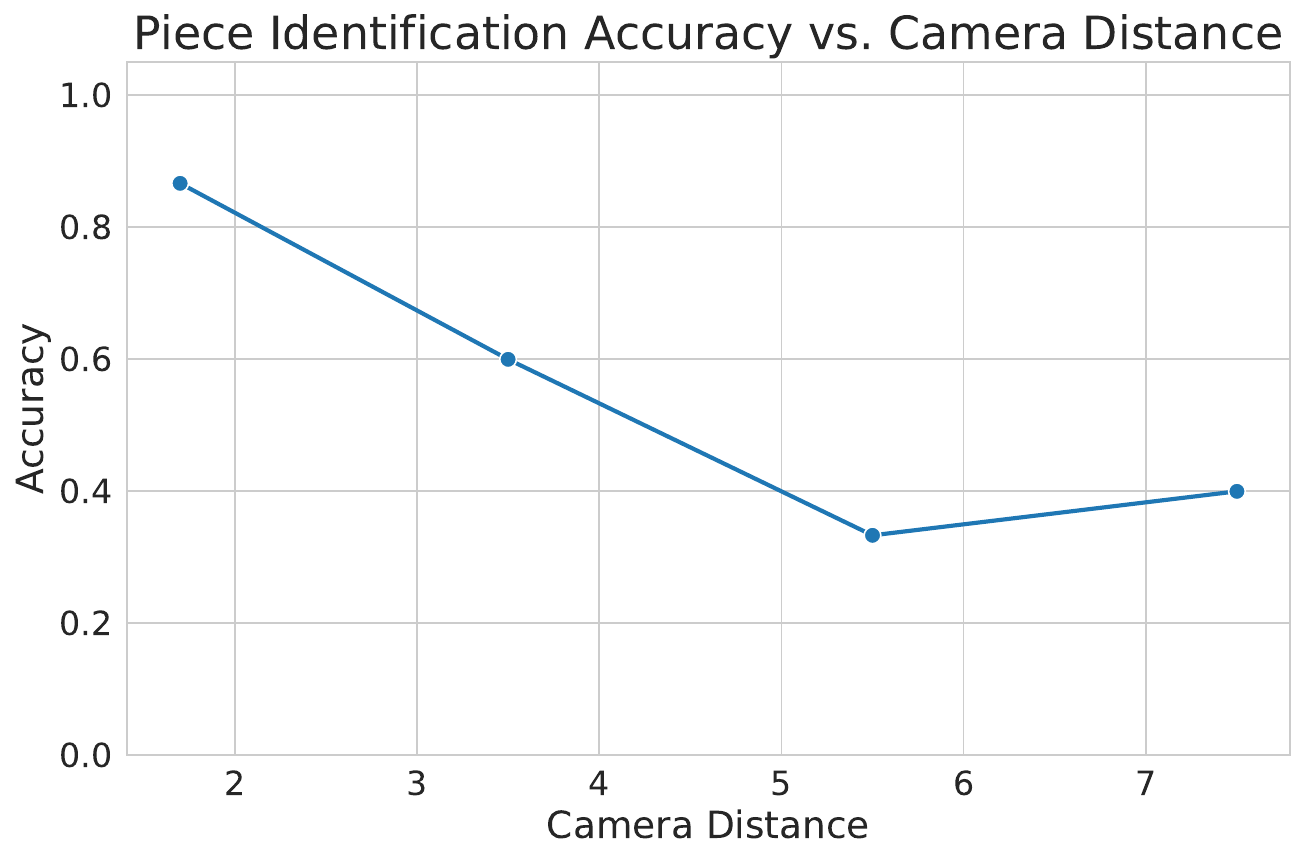}
        \caption{GPT-4.1}
        \label{fig:gpt4-1_results_identification_distance_chess}
    \end{subfigure}
    \hspace{0.02\linewidth}
    \begin{subfigure}[b]{0.48\linewidth}
        \centering
        \includegraphics[width=\linewidth]{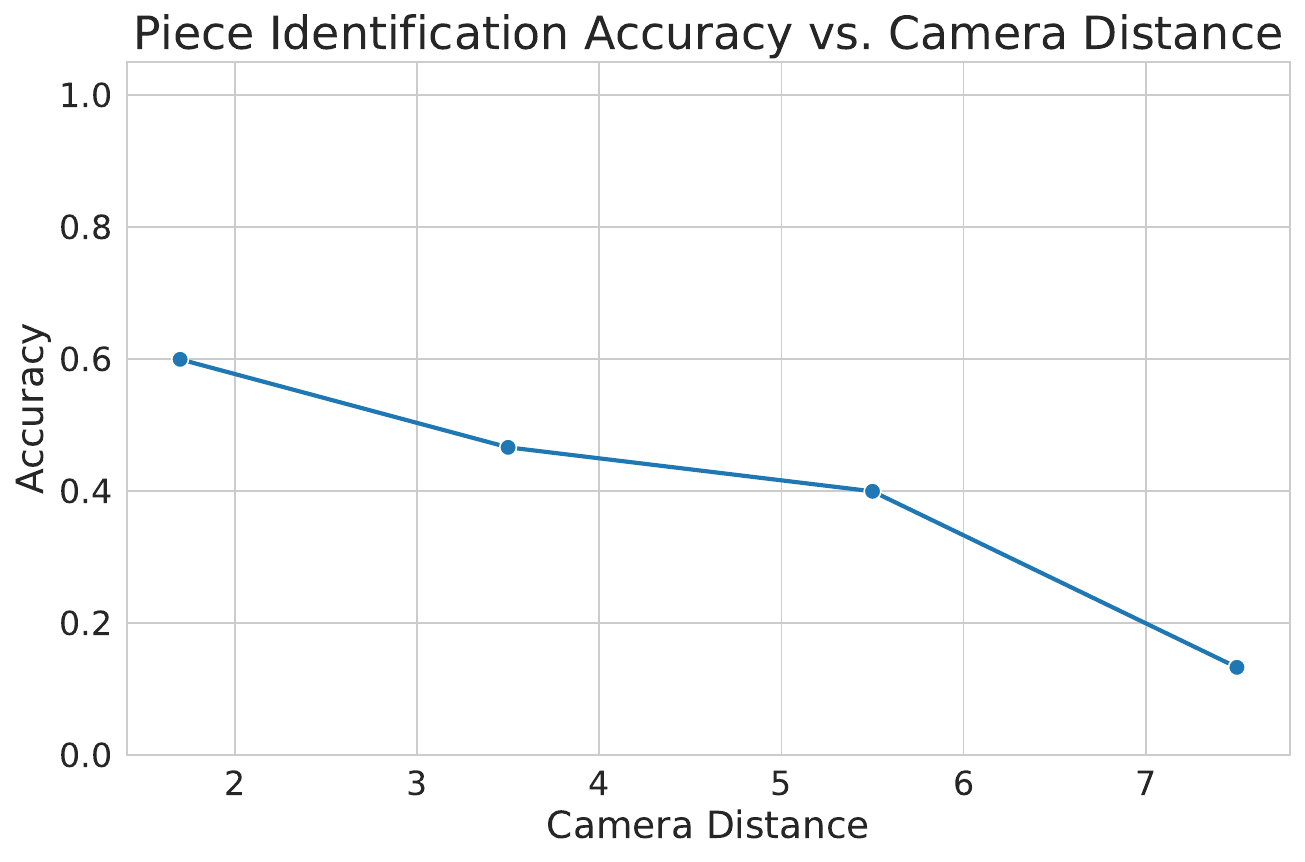}
        \caption{LLaMA 4}
        \label{fig:LLaMA4_results_identification_distance_chess}
    \end{subfigure}
    \caption{Accuracy for piece type identification at varying camera distances on the Chess dataset, for (a) GPT-4.1 and (b) LLaMA 4. Results are reported over 10 samples per distance level.}
    \label{fig:identification_distance_chess_gpt4_LLaMA4}
\end{figure}

\paragraph{Results} (Figures \ref{fig:LLaMA4_results_identification_distance_chess} and \ref{fig:gpt4-1_results_identification_distance_chess}) As anticipated, the performance of both models declines as the camera distance, i.e. the distance to the chess piece, increases. Nevertheless, GPT-4.1 appears to outperform LLaMA-4-Scout slightly in this task.

%%%%%%%%%%%%%%%%%%%%%%%%%%%%
\subsection{Counting with Overlap (Poker)}
\label{subsec:counting_overlap_poker}

We present the diagnostic results for the \emph{counting with overlap} task on the Poker dataset. The following question is asked with a debiased preprompt and a declarative instruction:
\begin{center}
    "This is not a real poker game. The cards and their position can vary arbitrarily. Just focus on answering the following question based on the visual content. How many cards are on the table? Be aware that some cards might be overlapping. Respond in a declarative format: 'The number of cards on the table is:'"
\end{center}

Examples of images can be found below in Figures \ref{fig:poker_overlap_h} and \ref{fig:poker_overlap_v}:

\begin{figure}[H]
    \centering
    \includegraphics[width=\linewidth]{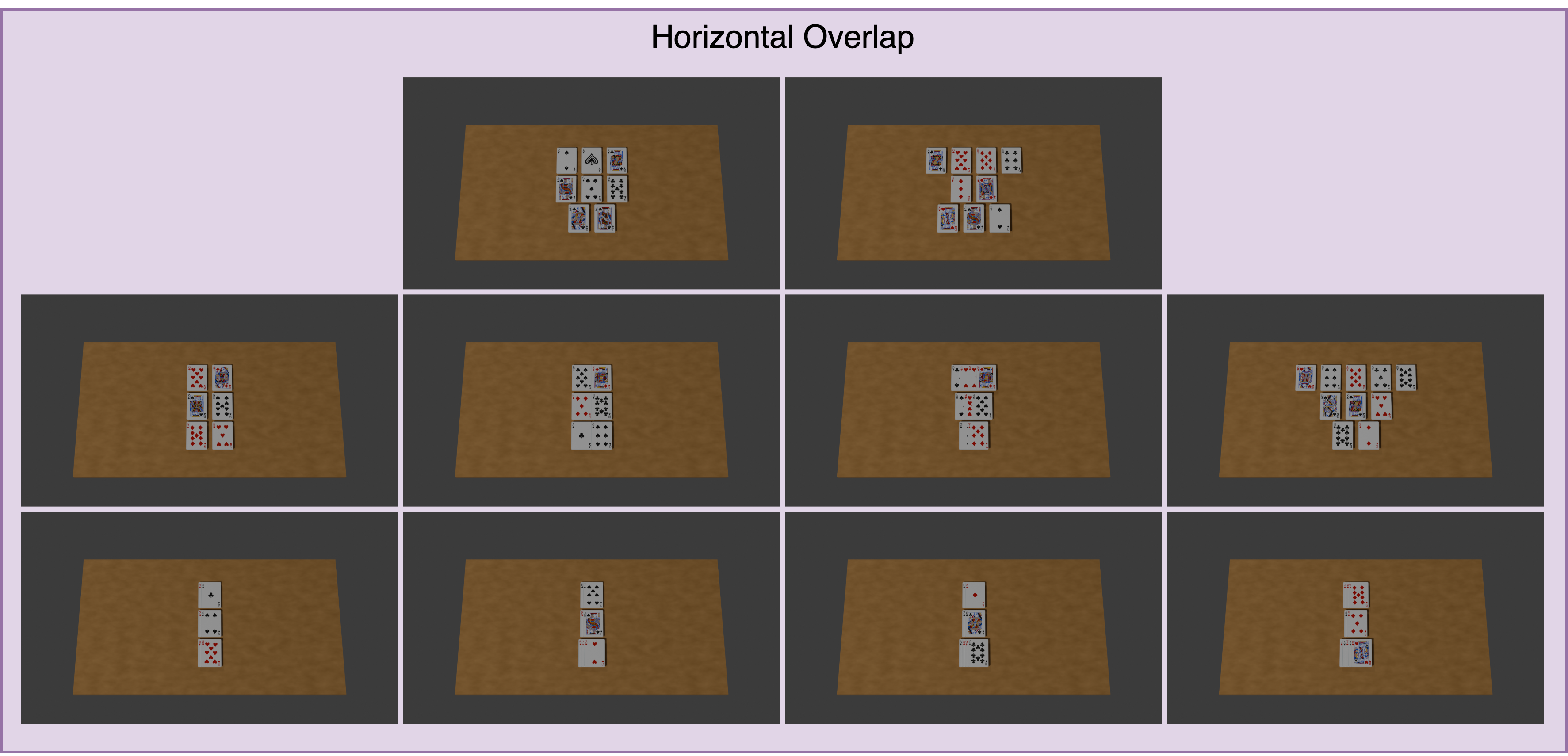}
    \caption{Variations of the horizontal overlap for Poker cards for the \emph{counting with overlap} task}
    \label{fig:poker_overlap_h}
\end{figure}

\begin{figure}[H]
    \centering
    \includegraphics[width=\linewidth]{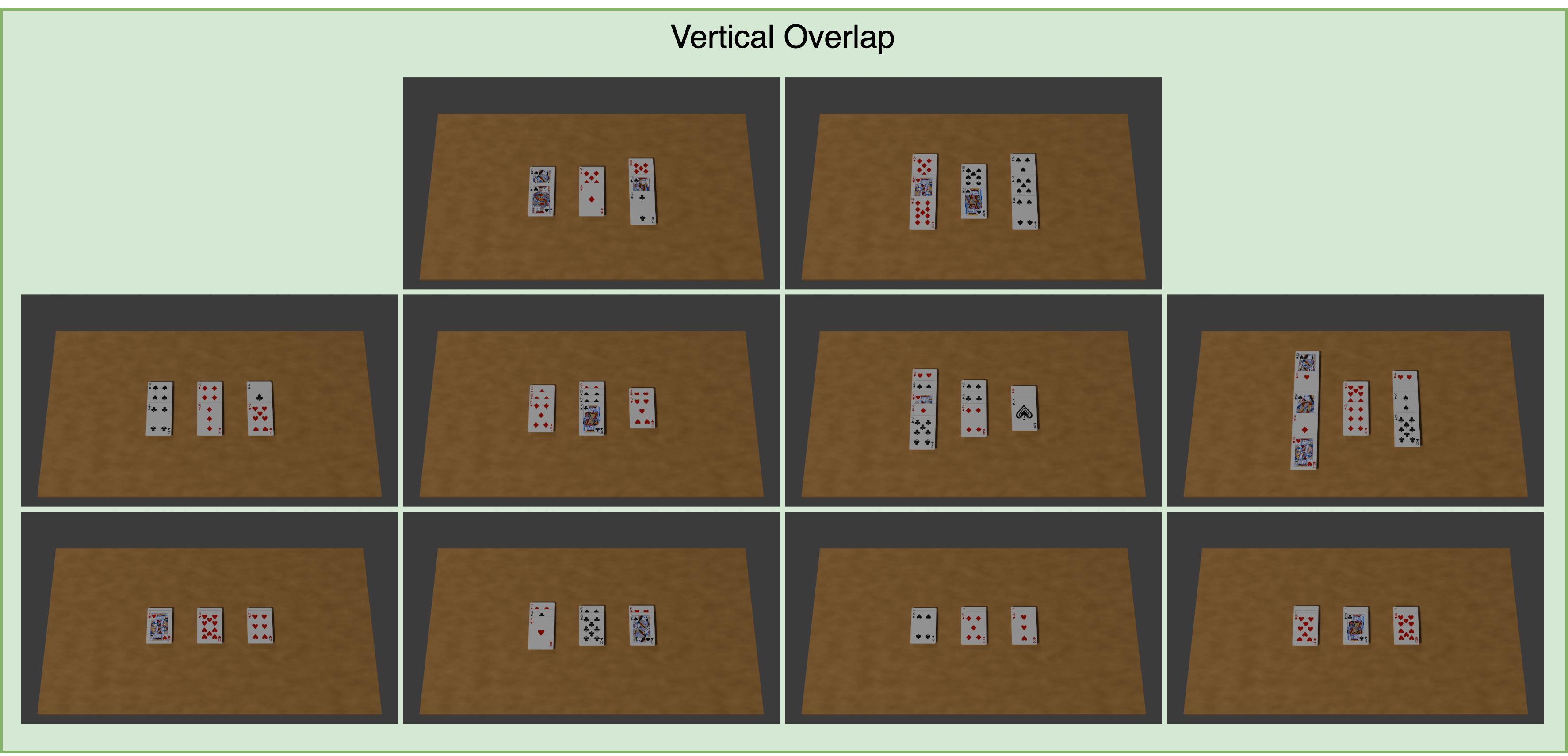}
    \caption{Variations of the vertical overlap for Poker cards for the \emph{counting with overlap} task}
    \label{fig:poker_overlap_v}
\end{figure}

\begin{figure}[H]
    \centering
    \includegraphics[width=\linewidth]{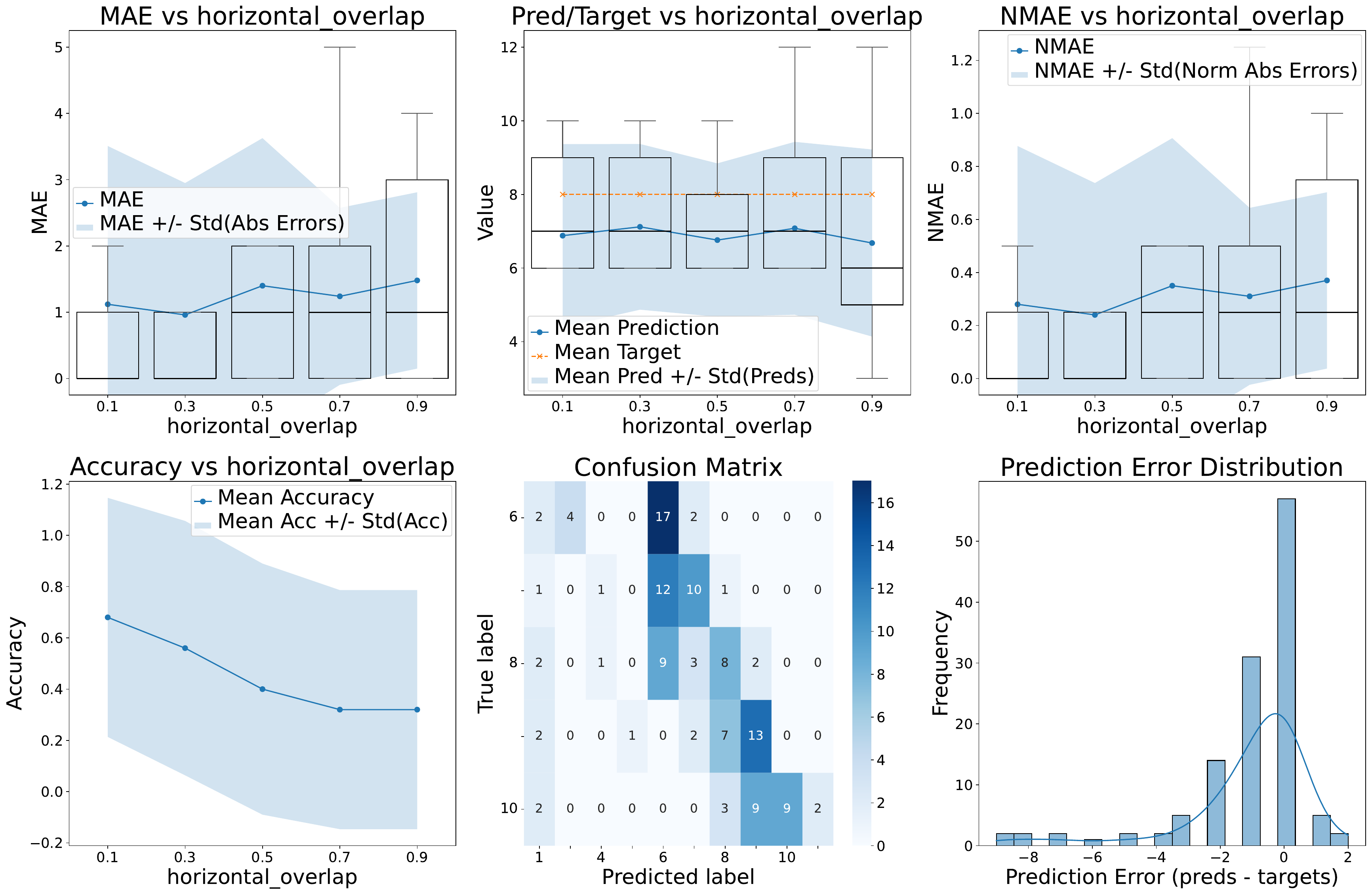}
    \caption{GPT-4.1 results for the counting question under overlapping cards on the Poker dataset. Mean and standard deviation over 5 samples per level.}
    \label{fig:gpt4-1_results_count_overlap_poker}
\end{figure}

\begin{figure}[H]
    \centering
    \includegraphics[width=\linewidth]{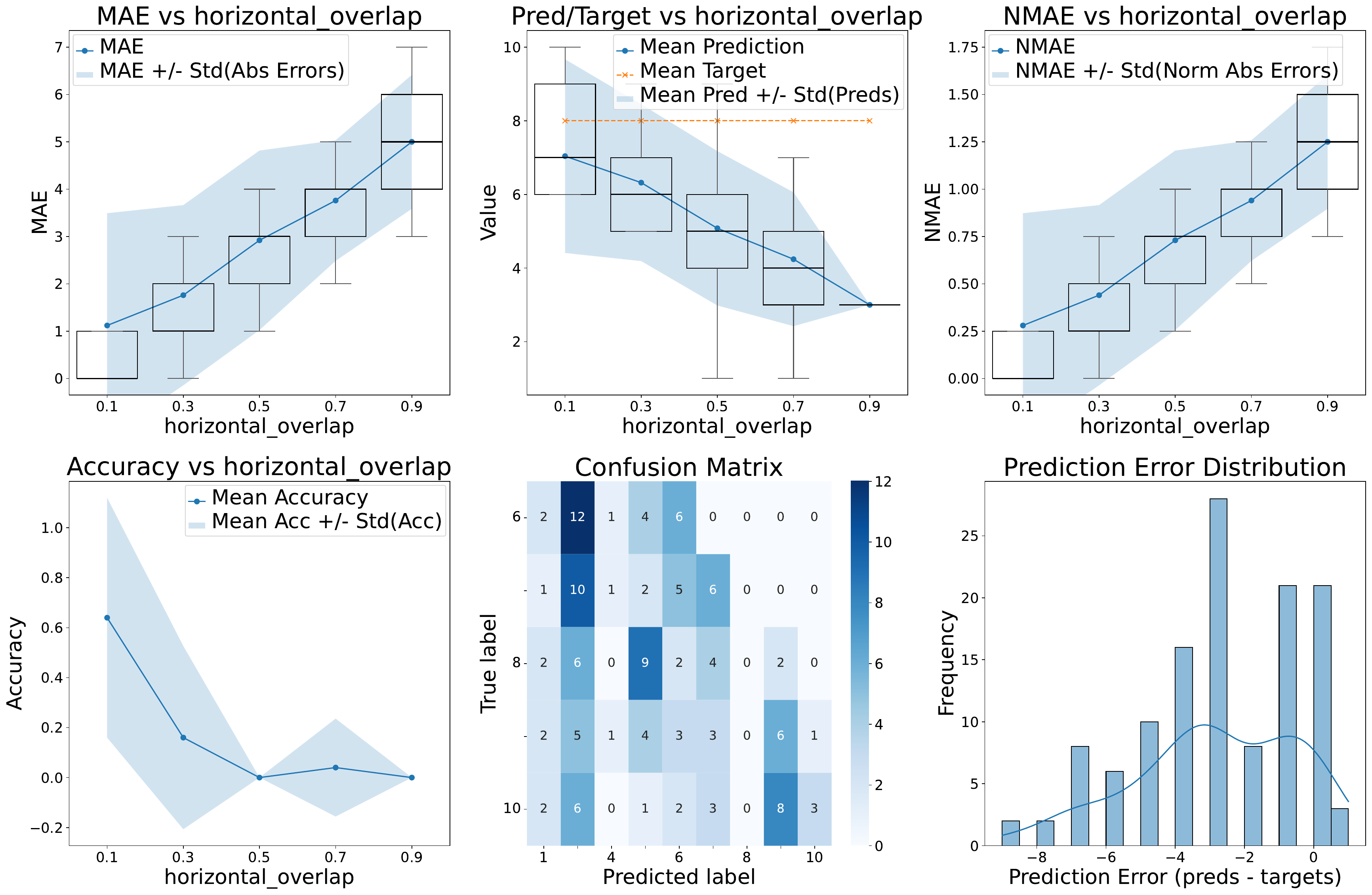}
    \caption{LLaMA results for the counting question under overlapping cards on the Poker dataset. Mean and standard deviation over 5 samples per level.}
    \label{fig:LLaMA4_results_count_overlap_poker}
\end{figure}

\begin{figure}[H]
    \centering
    \begin{subfigure}[b]{0.48\linewidth}
        \centering
        \includegraphics[width=\linewidth]{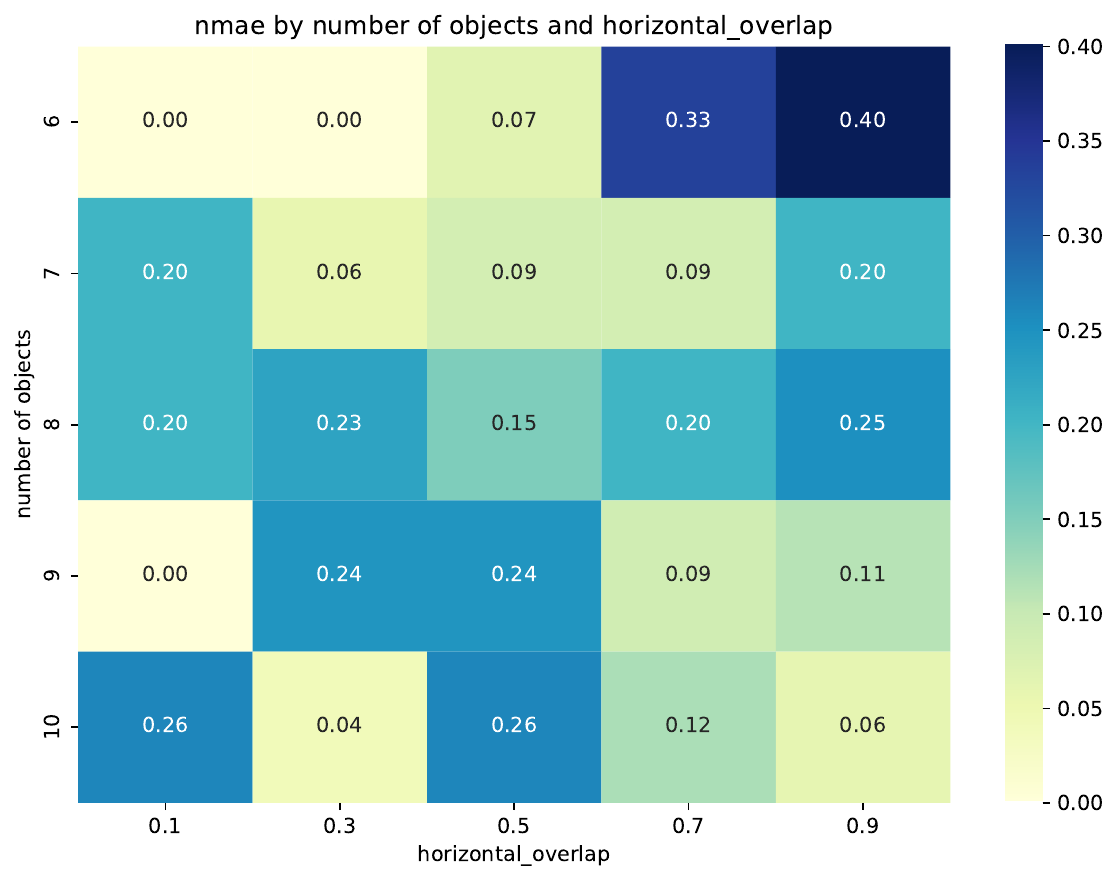}
        \caption{GPT-4.1}
        \label{fig:gpt4-1_crossplot_overlap_poker}
    \end{subfigure}
    \hspace{0.02\linewidth}
    \begin{subfigure}[b]{0.48\linewidth}
        \centering
        \includegraphics[width=\linewidth]{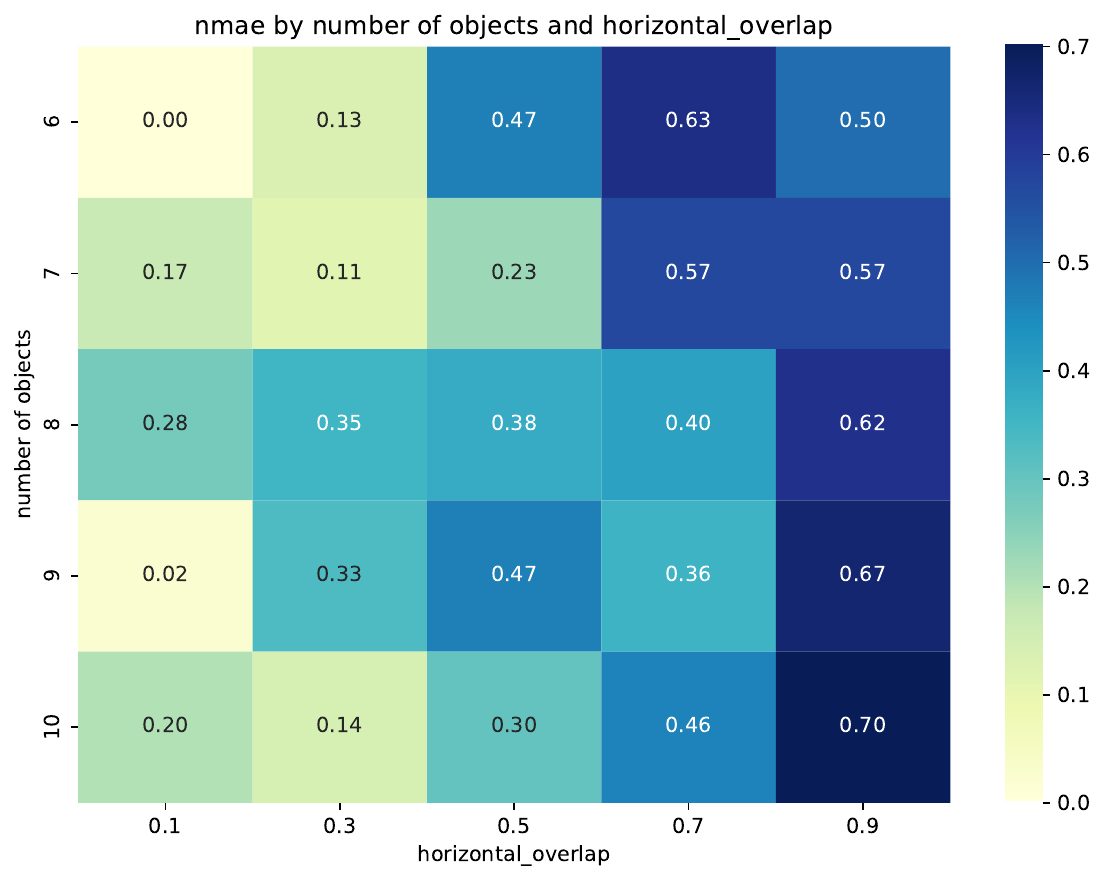}
        \caption{LLaMA-4-Scout}
        \label{fig:LLaMA4_crossplot_overlap_poker}
    \end{subfigure}
    \caption{Error analysis for the Poker dataset (overlap counting task): absolute error as a function of overlap level and number of cards, for (a) GPT-4.1 and (b) LLaMA.}
    \label{fig:crossplot_overlap_poker_gpt_LLaMA}
\end{figure}

\paragraph{Results} 
\begin{itemize}
\item GPT-4.1 (Figures \ref{fig:gpt4-1_results_count_overlap_poker},\ref{fig:crossplot_overlap_poker_gpt_LLaMA}) As expected, accuracy decreases with increasing horizontal overlap, which reduces the distinguishability of the cards. The model tends to underestimate the number of cards at every level of horizontal overlap. Figure \ref{fig:crossplot_overlap_poker_gpt_LLaMA} confirms that the difficulty is primarily driven by the overlap, as the highest NMAE values for the highest levels of horizontal overlap occur with a low number of objects 
\item LLaMA-4-Scout (Figures \ref{fig:LLaMA4_results_count_overlap_poker}, \ref{fig:crossplot_overlap_poker_gpt_LLaMA}) Accuracy declines as horizontal overlap increases, with the MAE steadily rising alongside the overlap. The model appears to struggle significantly with this aspect, as the difference between the mean prediction and the target grows with increasing overlap, reaching an average discrepancy of up to 5 objects at the highest level. As the overlap between cards increases, the model increasingly underestimates the number of cards, with the distribution of prediction errors clearly shifted to the left and centered around -2. The model also shows a bias toward predicting values of 1-3 cards for each true label, as shown in the confusion matrix, which suggests its difficulty in distinguishing cards that encroach upon each other.
Figure \ref{fig:crossplot_overlap_poker_gpt_LLaMA} further confirms that the primary source of difficulty is the level of overlap, with NMAE increasing as overlap intensifies. 
\end{itemize}

\section{Benchmark - additional experiments}
\label{sec:benchmark_additional}
In this section, we present complementary results that strengthen the core findings of the main paper. These additional experiments explore: 
\begin{itemize}
    \item Model robustness under controlled perturbations introduced in synthetic scenes. 
    \item The sensitivity of vision-language models (VLMs) to prompting strategies across tasks.
    \item Extended comparisons across multiple families of VLMs, including both proprietary (e.g., GPT-4.1) and open-source (e.g., LLaMA-4) models.
\end{itemize}

All experiments are conducted using a unified evaluation methodology, which ensures consistent sampling, annotation parsing, and metric computation across counting, identification, and localization tasks. This diagnostic protocol enables a fine-grained analysis of model behavior under realistic yet controlled visual conditions.

%In this section, we report extended experimental results that complement the main benchmark findings and provide additional evidence for the strengths and limitations of current Vision-Language Models (VLMs). These analyses are conducted using our diagnostic evaluation framework, which enables controlled, repeatable assessments of model performance under carefully manipulated scene conditions.

%Specifically, we investigate:

%\begin{enumerate}[leftmargin=1.5em]
%\item \textbf{Robustness to controlled perturbations}, including structured noise such as lighting variation, blur, and distractor clutter. These conditions are designed to simulate real-world visual variability while preserving full annotation and control.
%\item \textbf{Prompt sensitivity and transferability}, by systematically evaluating the effect of preprompt design (e.g., declarative vs. neutral instructions) across different task categories and domains (e.g., Chess vs. Poker).
%\item \textbf{Extended model comparisons}, incorporating additional baselines beyond those reported in the main text, including both API-accessible proprietary models (e.g., GPT-4.1, GPT-4.1-mini) and open-source models hosted via Groq and OLLaMA.
%\end{enumerate}

All experiments are executed using a unified evaluation pipeline, which integrates:
\begin{itemize}
\item Standardized data ingestion from our diagnostic dataset, including task-specific image folders and structured annotations,
\item Modular VLM querying interfaces supporting OpenAI, Groq, and OLLaMA endpoints,
\item A decoding and scoring pipeline tailored to each task type (e.g., MAE for counting, token-level F1 for identification, accuracy for localization),
\item Reproducible configuration management, enabling consistent experimental tracking across noise levels, prompt variations, and model families.
\end{itemize}

This diagnostic setup enables a fine-grained understanding of model generalization under distribution shifts that are difficult to control in real-world data. The additional results presented below confirm the necessity of such synthetic stress tests in assessing the real-world readiness of multimodal models.

%\subsection*{Extended Visualizations of Model Behavior}
%\label{subsec:extended_visualizations}

\begin{figure}[H]
\centering
\begin{vlmcasebox}

\begin{center}
  \begin{minipage}{0.6\linewidth}
    \centering
    \includegraphics[width=\linewidth]{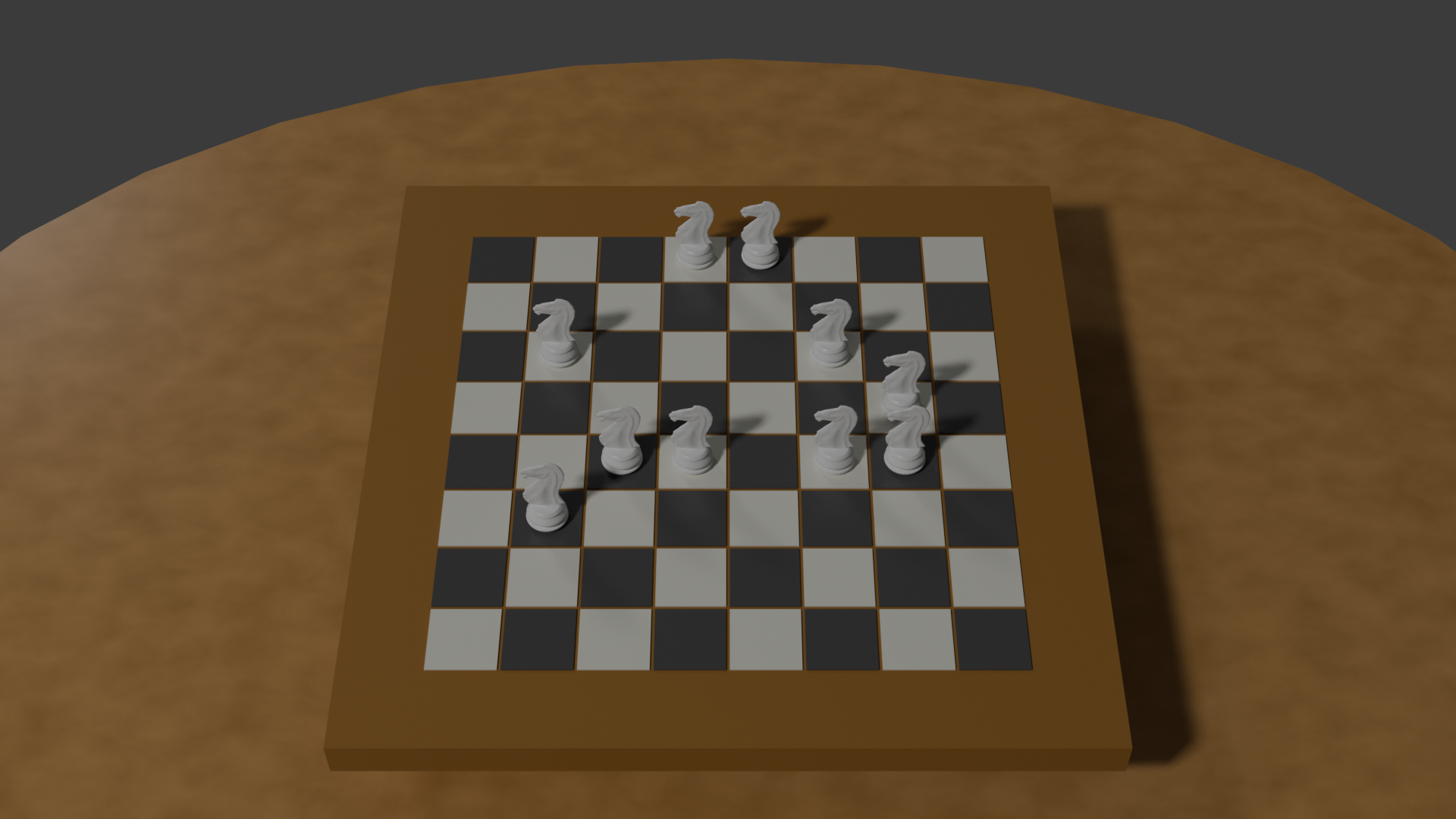}
    \captionof*{figure}{\textbf{Input Image}}
  \end{minipage}
\end{center}

\vspace{0.8em}

\textbf{Task: Number Pieces Counting in Synthetic Chessboard Image}

\textbf{Question:} Analyze the image carefully, be aware that this image may differ from a classical chessboard scene. How many pieces are there in the image? 
Respond in a declarative format: \textit{“The number of pieces in the image is:”}

\vspace{0.6em}

\begin{center}
\begin{tabular}{@{}lll@{}}
\toprule
\textbf{Model} & \textbf{Response} &  \\
\midrule
GPT-4.1 & The number of pieces in the image is: \textbf{10} & \cmark \\
GPT-4.1-mini & The number of pieces in the image is: \textbf{10} & \cmark \\
LLaMA-4-scout & The number of pieces in the image is: \textbf{8} & \xmark \\
LLaMA-4-maverick & The number of pieces in the image is: \textbf{9} & \xmark \\
Gemma-4b & The number of pieces in the image is: \textbf{8} & \xmark \\
Gemma-12b & The number of pieces in the image is: \textbf{8} & \xmark \\
LLaMA3.2-vision & The number of pieces in the image is: \textbf{12} & \xmark \\
Mistral-small3.1 & The number of pieces in the image is: \textbf{10} & \cmark \\
\bottomrule
\end{tabular}
\end{center}

\vspace{0.5em}
\textbf{Ground Truth:} 10 \\
%\textbf{Result:} All models answered correctly.

\end{vlmcasebox}
\caption{VLM responses to a number pieces counting question based on a synthetic chess scene.}
\label{fig:chess_counting_vertical_case}
\end{figure}

We include below additional heatmaps generated for each model across specific diagnostic task variants. These highlight the spatial structure of predictions and common failure modes under controlled synthetic conditions. Each heatmap corresponds to a specific combination of prompt strategy (Helpful, Neutral, Chain-of-Thought~\cite{wei2022chain}) and prompt reformulation (Declarative or Missing Word). Figure~\ref{fig:chess_counting_vertical_case} illustrates model outputs for a representative example of the counting task.

\subsection{Count Blur}

\begin{figure}[h]
  \centering
  \includegraphics[width=0.30\linewidth]{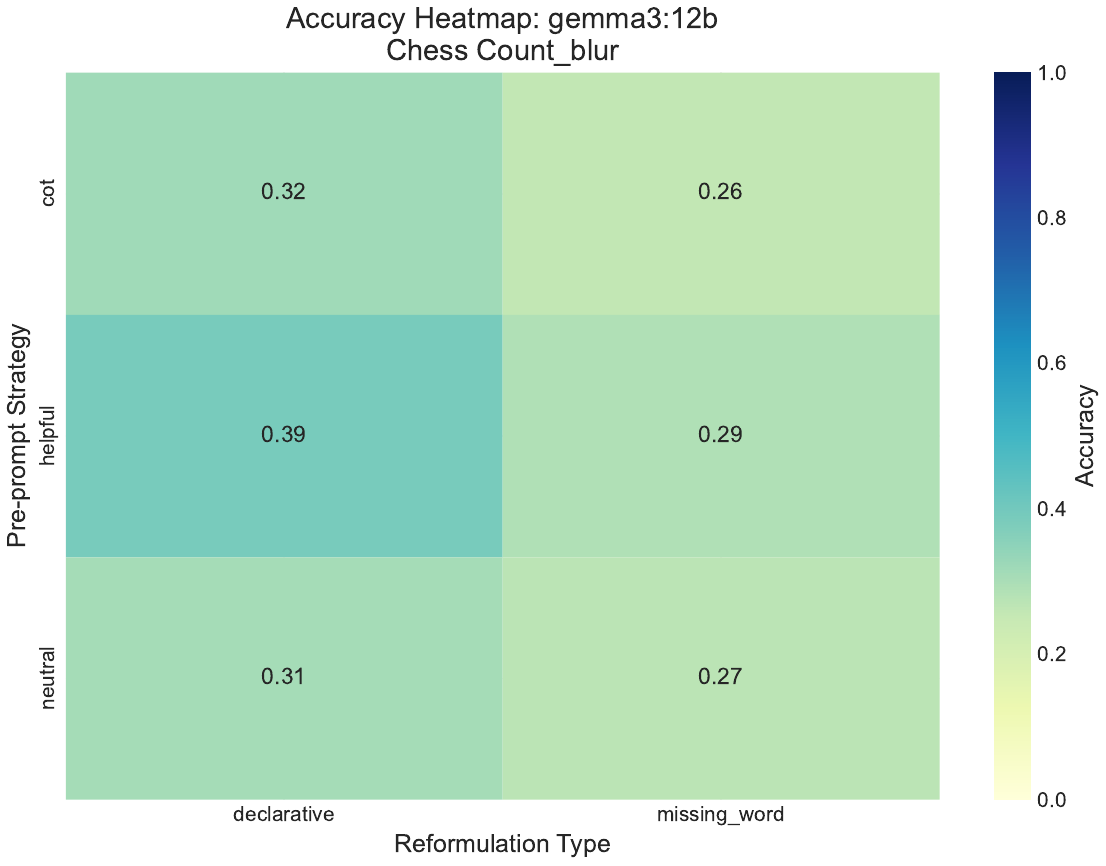}
  \includegraphics[width=0.30\linewidth]{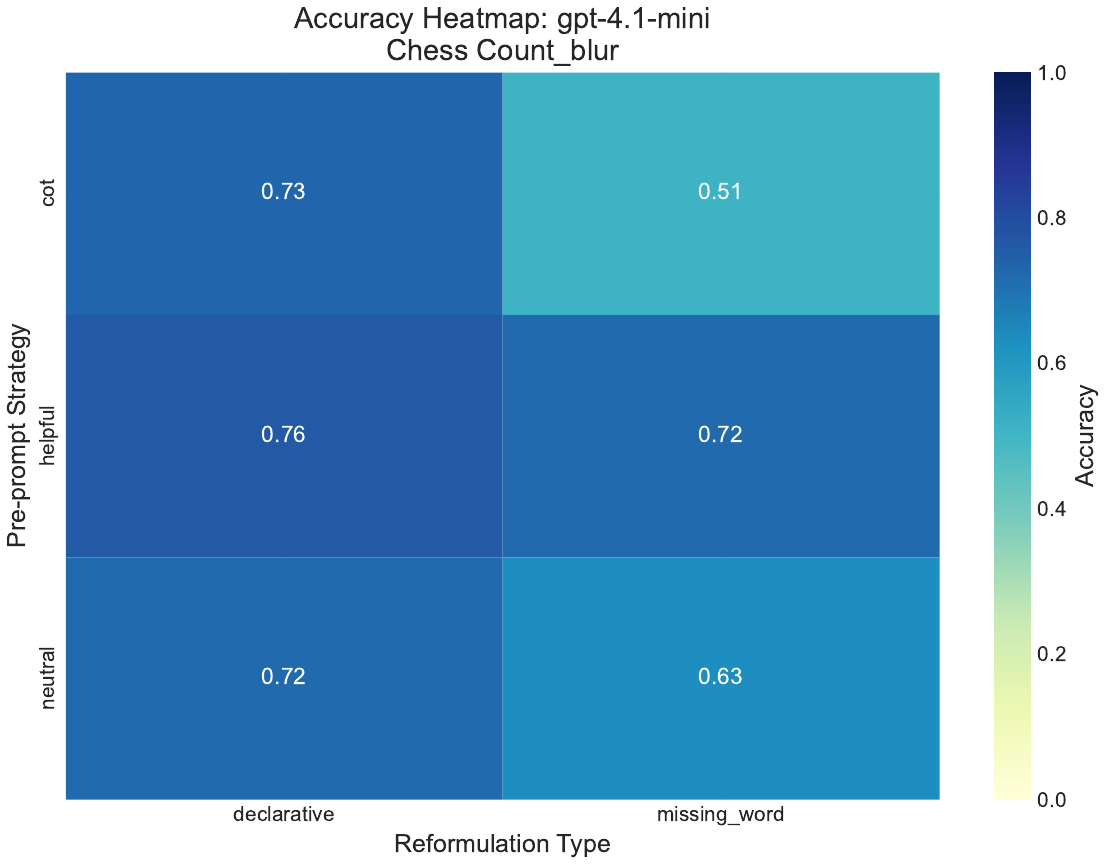}
  \includegraphics[width=0.30\linewidth]{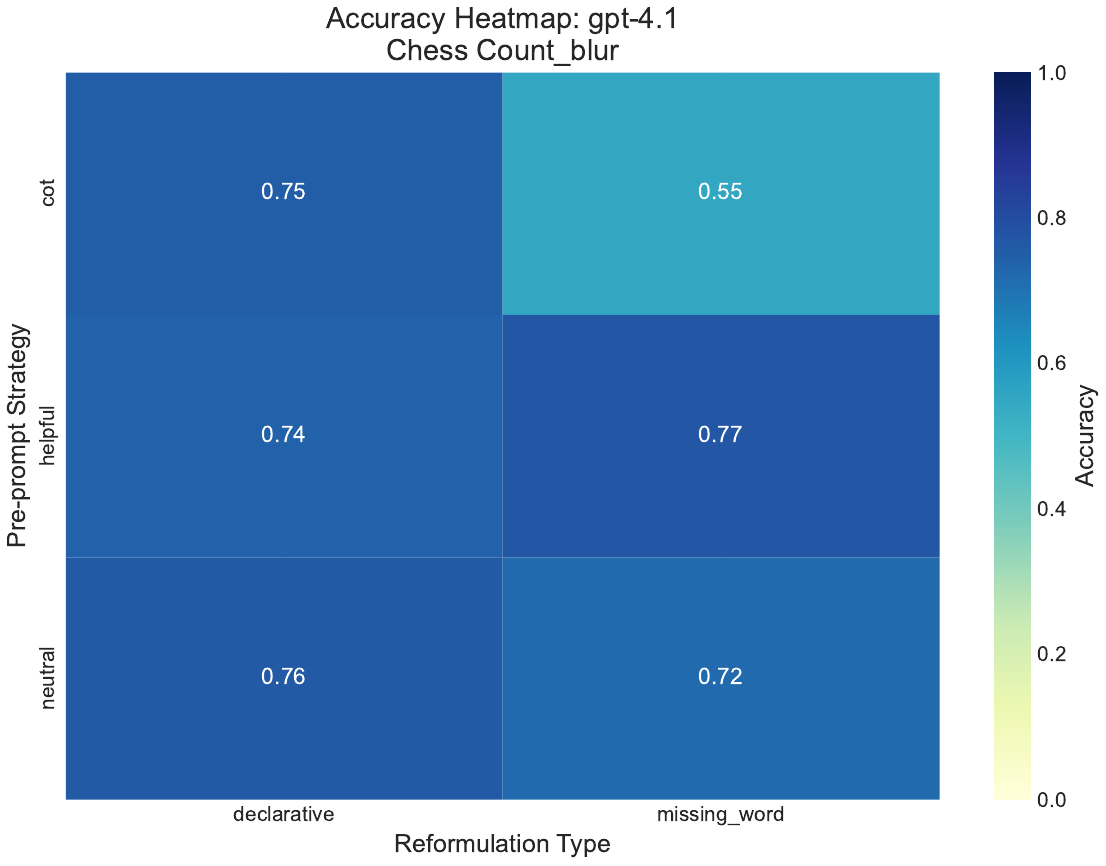}
  \\
  \includegraphics[width=0.30\linewidth]{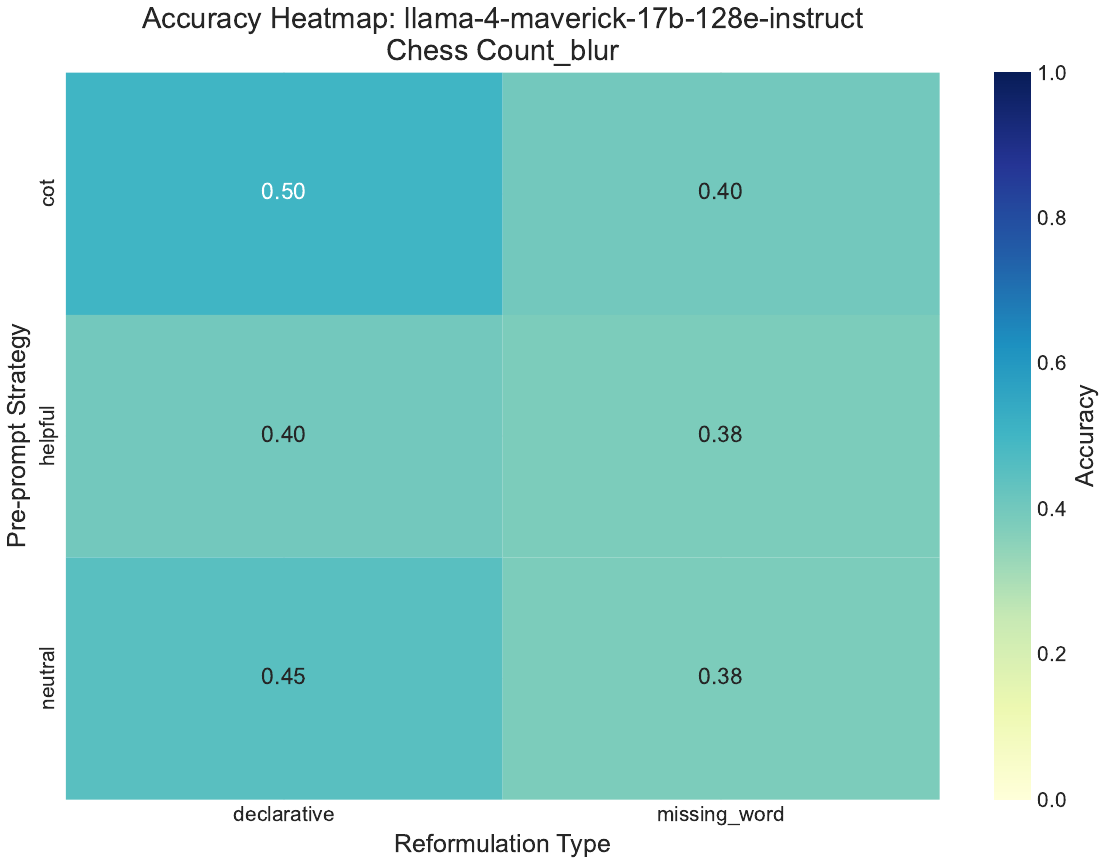}
  \includegraphics[width=0.30\linewidth]{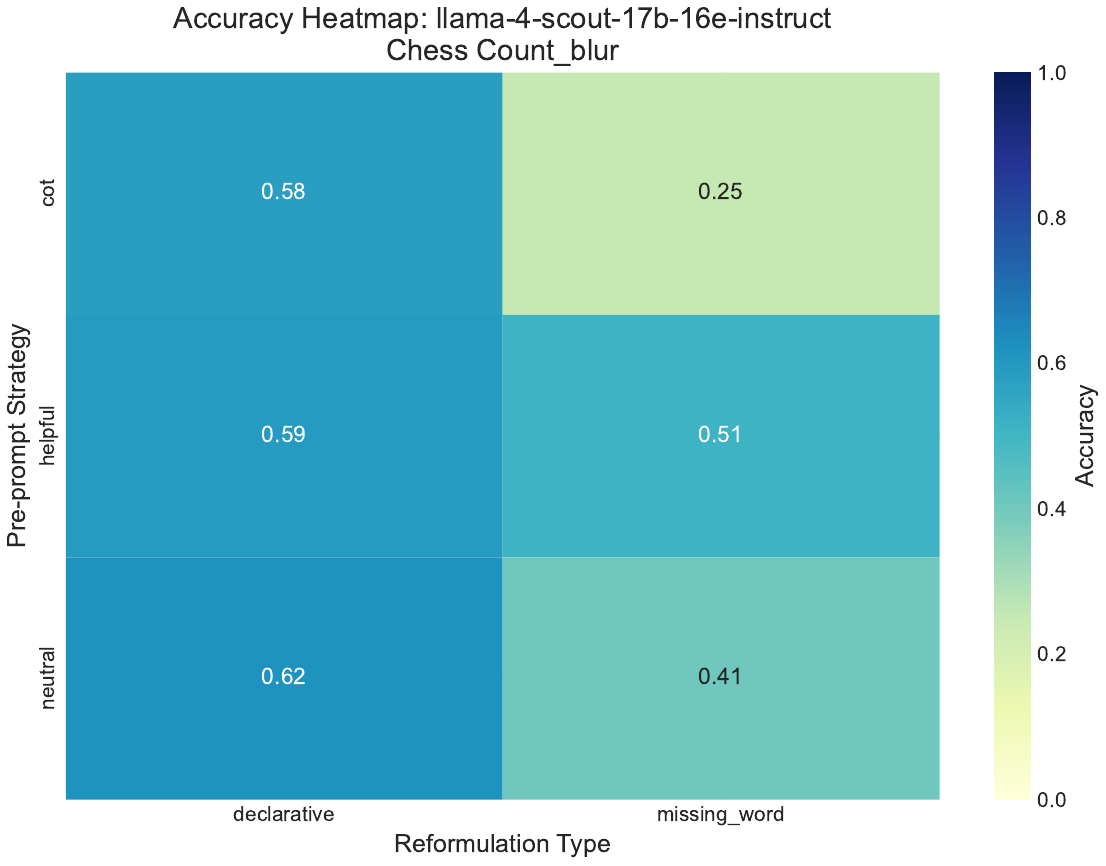}

  \caption{Spatial accuracy heatmaps for the Chess Count Blur task across multiple models and prompting strategies, averaged over 191 synthetic chess scenes with localized camera blur. Declarative prompts paired with Helpful strategies yield the most spatially stable and robust performance under visual degradation.}
  \label{fig:chess_count_blur_heatmap}
\end{figure}

\begin{figure}[ht]
  \centering
  \includegraphics[width=0.30\linewidth]{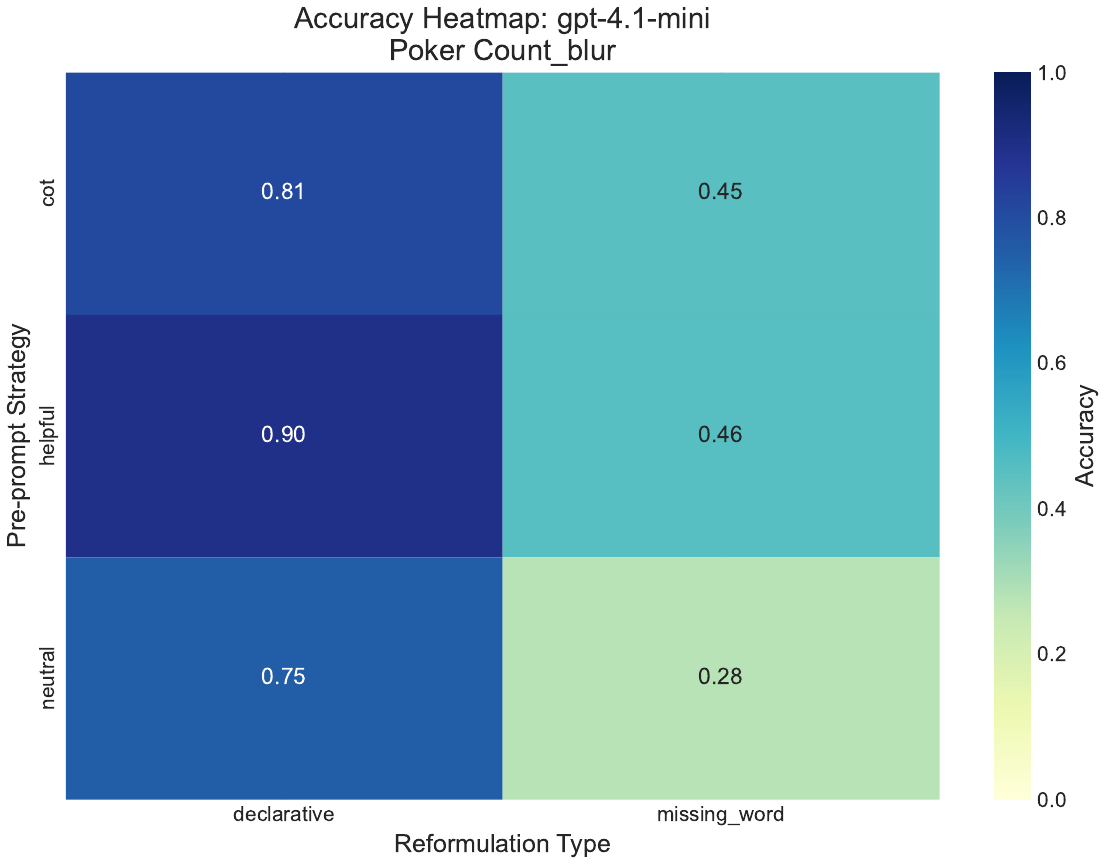}
  \includegraphics[width=0.30\linewidth]{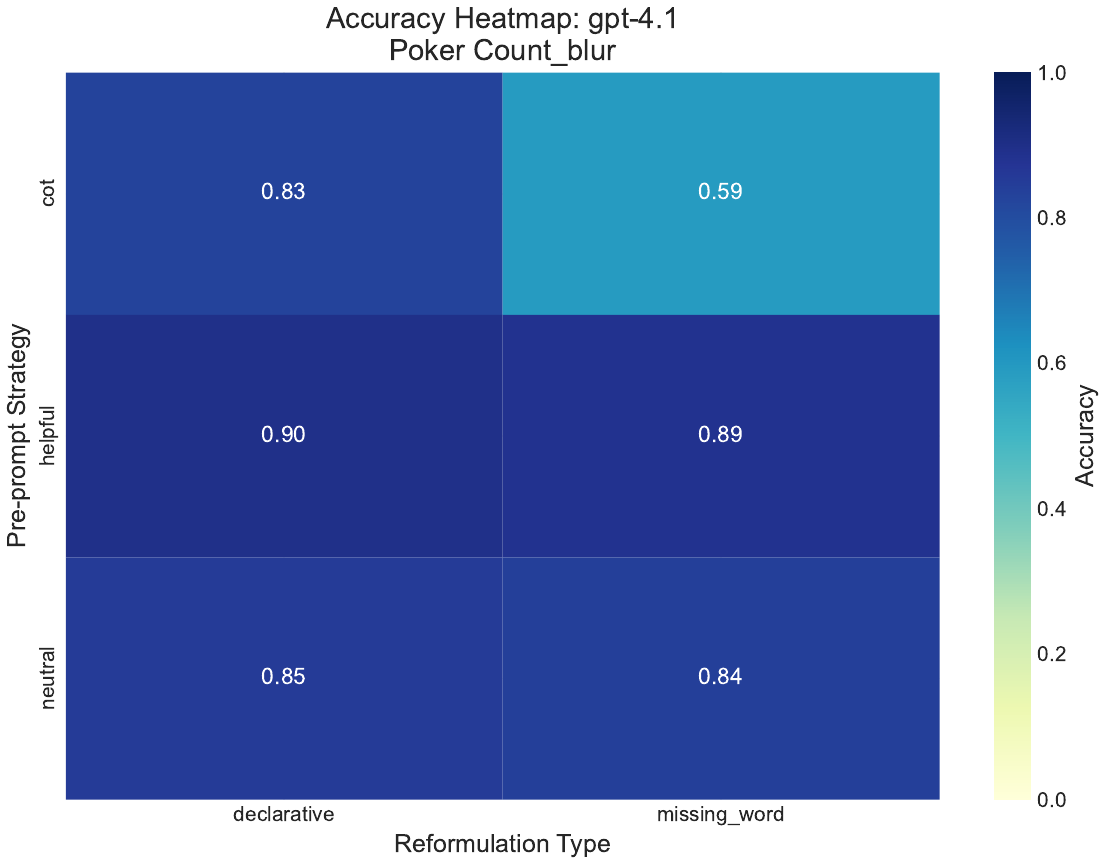}
  \includegraphics[width=0.30\linewidth]{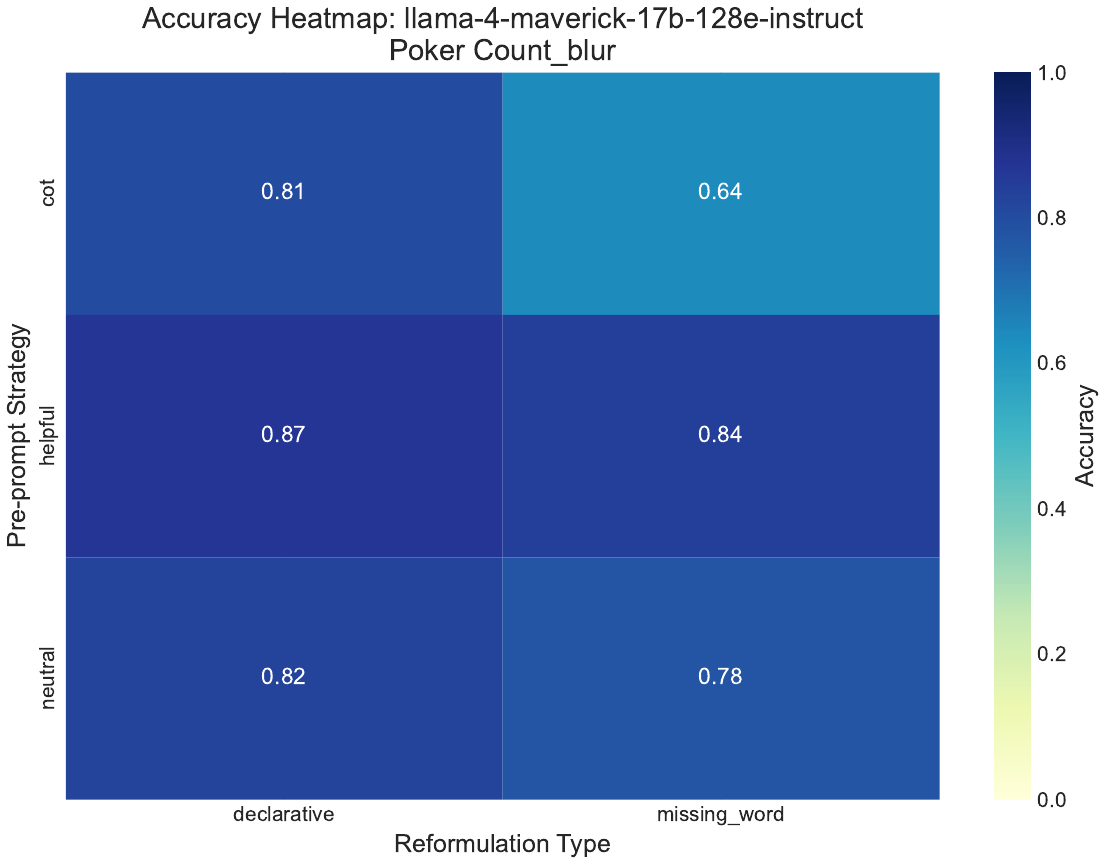}
  \\
  \includegraphics[width=0.30\linewidth]{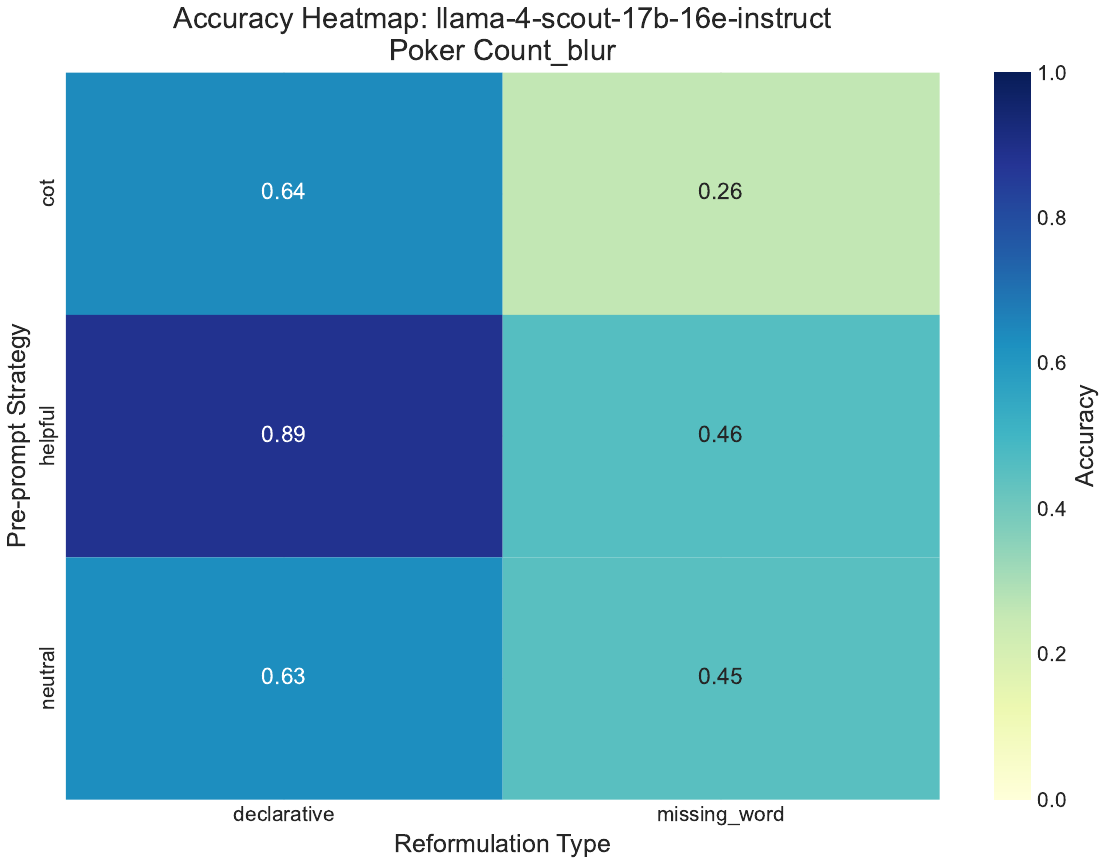}
  \includegraphics[width=0.30\linewidth]{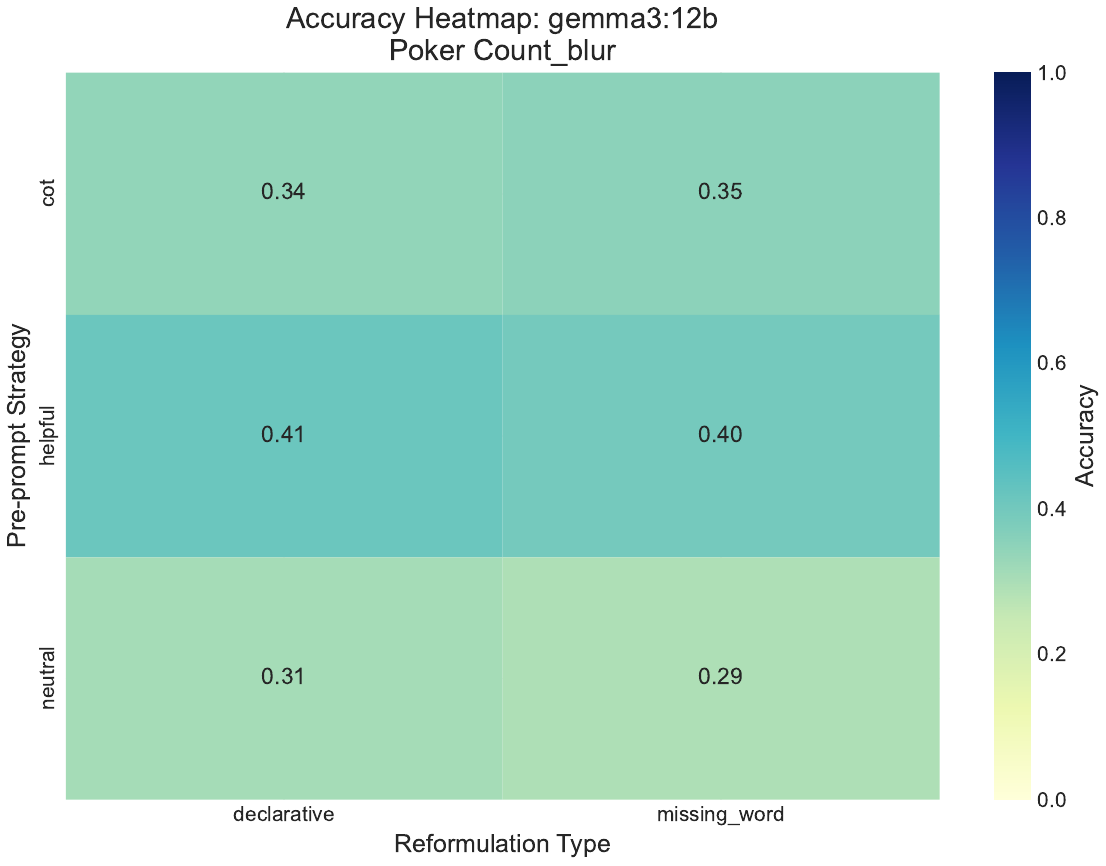}

  \caption{Spatial accuracy heatmaps for the Poker Count Blur task across multiple models and prompting strategies, averaged over 50 synthetic poker scenes with localized camera blur. Declarative prompts paired with Helpful strategies yield the most spatially stable and robust performance under visual degradation.}
  \label{fig:poker_heatmap_count_blur}
\end{figure}

This diagnostic setup introduces the \textbf{count blur} variation, where we simulate depth-of-field effects by applying spatially localized camera blur to the input images. This mimics real-world conditions such as defocus, motion blur, or background noise, which often reduce the clarity of visual cues necessary for accurate object enumeration. The primary objective of this variant is to assess the robustness of Vision-Language Models (VLMs) in scenarios where object boundaries become ambiguous or occluded due to degraded image quality. By varying the intensity and location of the blur (e.g., central vs. peripheral regions), we can analyze whether models rely on holistic scene understanding or overfit to sharp, high-frequency regions. This setup serves as a stress test for the model's spatial reasoning and ability to generalize counting abilities beyond clean synthetic. Figures ~\ref{fig:chess_count_blur_heatmap} and \ref{fig:poker_heatmap_count_blur} illustrate respectively the results for Chess and Poker datasets. The Table~\ref{tab:count_blur_task} highlights the results of the models on different blur conditions. 

\begin{table}[ht]
\centering
\caption{VLM scores over the counting task on Chess dataset (↑: higher is better, ↓: lower is better). Declarative reformulation, helpful preprompt. Piece count ranges from $1$ to $21$.}
\label{tab:counting_declarative_helpful}
\renewcommand{\arraystretch}{1.1}
\setlength{\tabcolsep}{3pt}
\small
\vspace{4pt}
\begin{tabular}{lccccccccc}
\toprule
\textbf{Model} & \textbf{Acc↑} & \textbf{F1↑} & \textbf{Prec↑} & \textbf{Rec↑} & \textbf{MAE↓} & \textbf{MSE↓} & \textbf{NMAE↓} & \textbf{Rel↑} & \textbf{Cons↑} \\
\midrule
gpt-4.1         & 0.743 & 0.853 & 0.793 & 0.922 & 0.319 & 0.476 & 0.160 & 0.743 & 0.262 \\
gpt-4.1-mini    & \textbf{0.806} & \textbf{0.893} & \textbf{0.828} & \textbf{0.969} & \textbf{0.209} & \textbf{0.251} & \textbf{0.084} & \textbf{0.806} & 0.277 \\
LLaMA-4-scout   & 0.639 & 0.780 & 0.678 & 0.917 & 0.466 & 0.707 & 0.202 & 0.639 & 0.267 \\
LLaMA-4-maverick& 0.445 & 0.616 & 0.494 & 0.817 & 0.670 & 0.921 & 0.318 & 0.445 & 0.330 \\
gemma3:4b       & 0.476 & 0.645 & 0.503 & 0.901 & 0.838 & 1.906 & 0.317 & 0.476 & 0.419 \\
gemma3:12b       & 0.495 & 0.645 & 0.543 & 0.916 & 0.862 & 1.893 & 0.320 & 0.487 & 0.459\\
LLaMA3.2-vision  & 0.400 & 0.571 & 0.455 & 0.857 & 0.932 & 2.356 & 0.590 & 0.403 & 0.343 \\
mistral-small3.1  & 0.520 & 0.684 & 0.520 & 0.923 & 0.880 & 1.802 & 0.224 & 0.536 & \textbf{0.557} \\
\bottomrule
\end{tabular}
\end{table}

%through camera blur. 
%The goal is to test whether models can maintain accurate object counting under reduced visual clarity.

Each heatmap reflects the outcome of a specific preprompt and reformulation strategy combination. We compare:
\begin{itemize}
    \item \textbf{Prompt strategies:} Helpful, Neutral, and Chain-of-Thought (CoT).
    \item \textbf{Reformulations:} Declarative vs. Missing Word.
\end{itemize}

\textbf{Notable observations:}
\begin{itemize}
    \item Helpful + Declarative combinations consistently yield the most accurate and spatially uniform predictions, especially for GPT-4.1.
    \item Missing Word formulations are more brittle under blur, leading to degraded performance even for strong models.
    \item LLaMA and Gemma models show inconsistent spatial activation, especially near the image center, indicating struggles with object separation when blur is present.
\end{itemize}

In the figure~\ref{fig:chess_count_blur_heatmap} GPT models demonstrate strong robustness under blur, achieving the best accuracy under Declarative reformulation and Helpful preprompt. Their performance remains stable even with central blur, highlighting effective abstraction over degraded input features. LLaMA-4-Scout achieves surprisingly strong results with Declarative reformulation and Helpful preprompt but collapses under Missing Word reformulation, indicating high sensitivity to linguistic scaffolding. LLaMA-4-Maverick and Gemma3 exhibit irregular spatial activation patterns and struggle with object separation under blur. Their heatmaps often show fragmented or inconsistent predictions, especially near the image center, where blur is typically applied.

In contrast to the chess setup, the Poker Count Blur results (Figure~\ref{fig:poker_heatmap_count_blur}) reveal more pronounced variability across models and prompting strategies. While GPT models again lead overall performance, their spatial activation maps suggest slightly reduced robustness compared to the chess scenes, likely due to the more irregular and less structured layout of poker elements. GPT-4.1, in particular, maintains high spatial accuracy across all prompt variations, with strong resilience even under significant central blur.

LLaMA-4-Maverick performs surprisingly well in this setup, showing more consistent activation than in the chess domain, and demonstrating better object separation under blur. This suggests the model benefits from the sparser spatial distribution of poker tokens. LLaMA-4 Scout also achieves strong performance when guided by Declarative + Helpful prompts, but, similar to the chess task, degrades sharply under Missing Word configurations.

Gemma3:12b continues to underperform across all prompting strategies, with heatmaps revealing scattered and incoherent predictions, particularly in blurred central zones. Compared to chess, the poker domain further exposes the model’s limited robustness and poor generalization under visual uncertainty.

As with the chess task, prompting strategy plays a pivotal role. Declarative prompts combined with Helpful scaffolding consistently yield the most spatially stable predictions. In contrast, Chain-of-Thought reasoning again appears to destabilize performance—likely due to irrelevant or hallucinated intermediate reasoning steps that fail to compensate for degraded visual cues.

Prompting strategies play a critical role. Declarative prompts consistently outperform Missing Word formulations across all models. Chain-of-Thought tends to destabilize performance in this setup, likely due to added reasoning steps that are less effective when visual cues are ambiguous. In Table~\ref{tab:counting_declarative_helpful} we illustrate the results for the Counting task on 800 samples of the Chess Dataset.

\begin{figure}[ht]
  \centering

  \begin{minipage}[t]{0.48\linewidth}
    \centering
    \includegraphics[width=\linewidth]{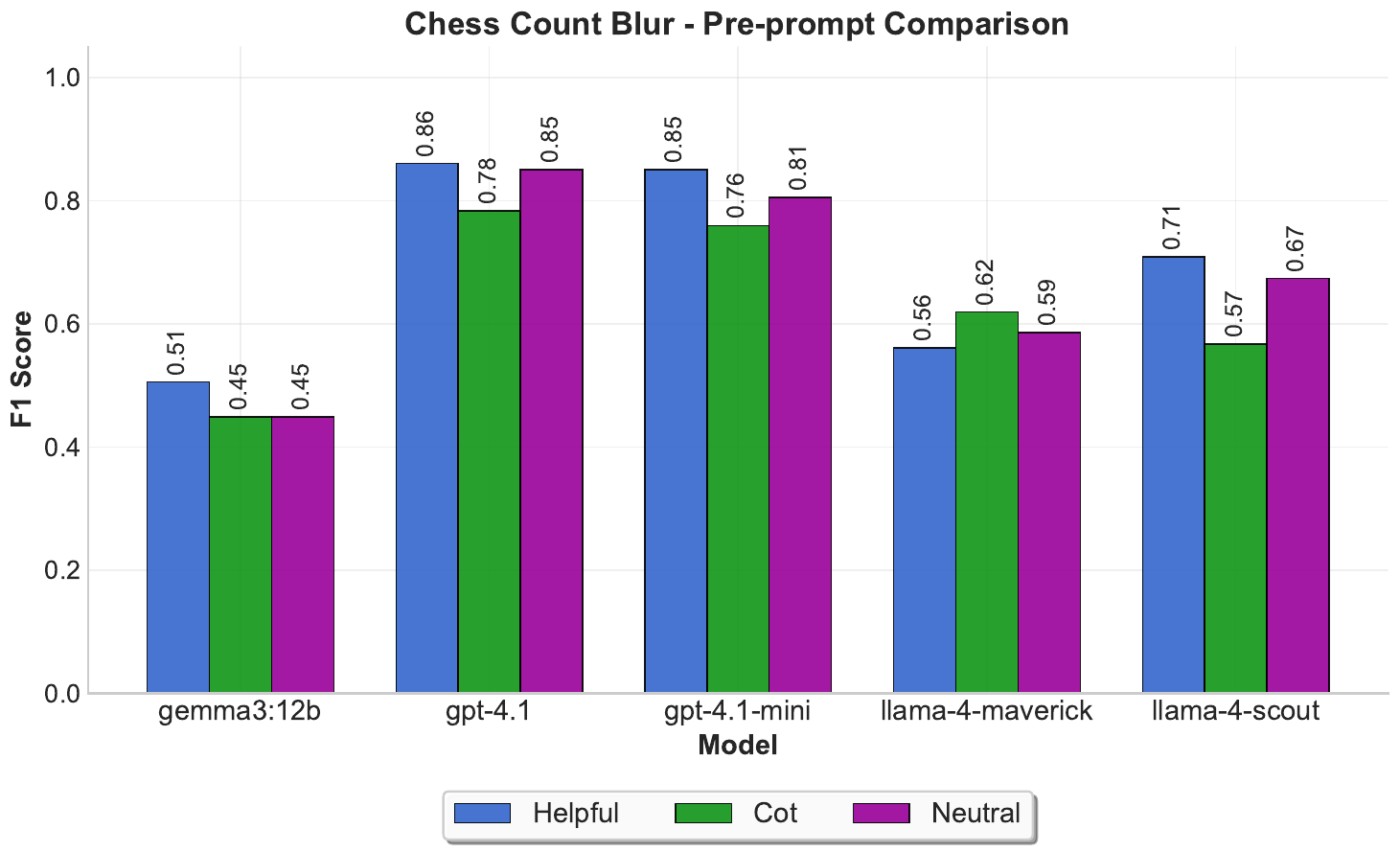}
    \textbf{(a)} Chess – preprompt Comparison
  \end{minipage}
  \hfill
  \begin{minipage}[t]{0.48\linewidth}
    \centering
    \includegraphics[width=\linewidth]{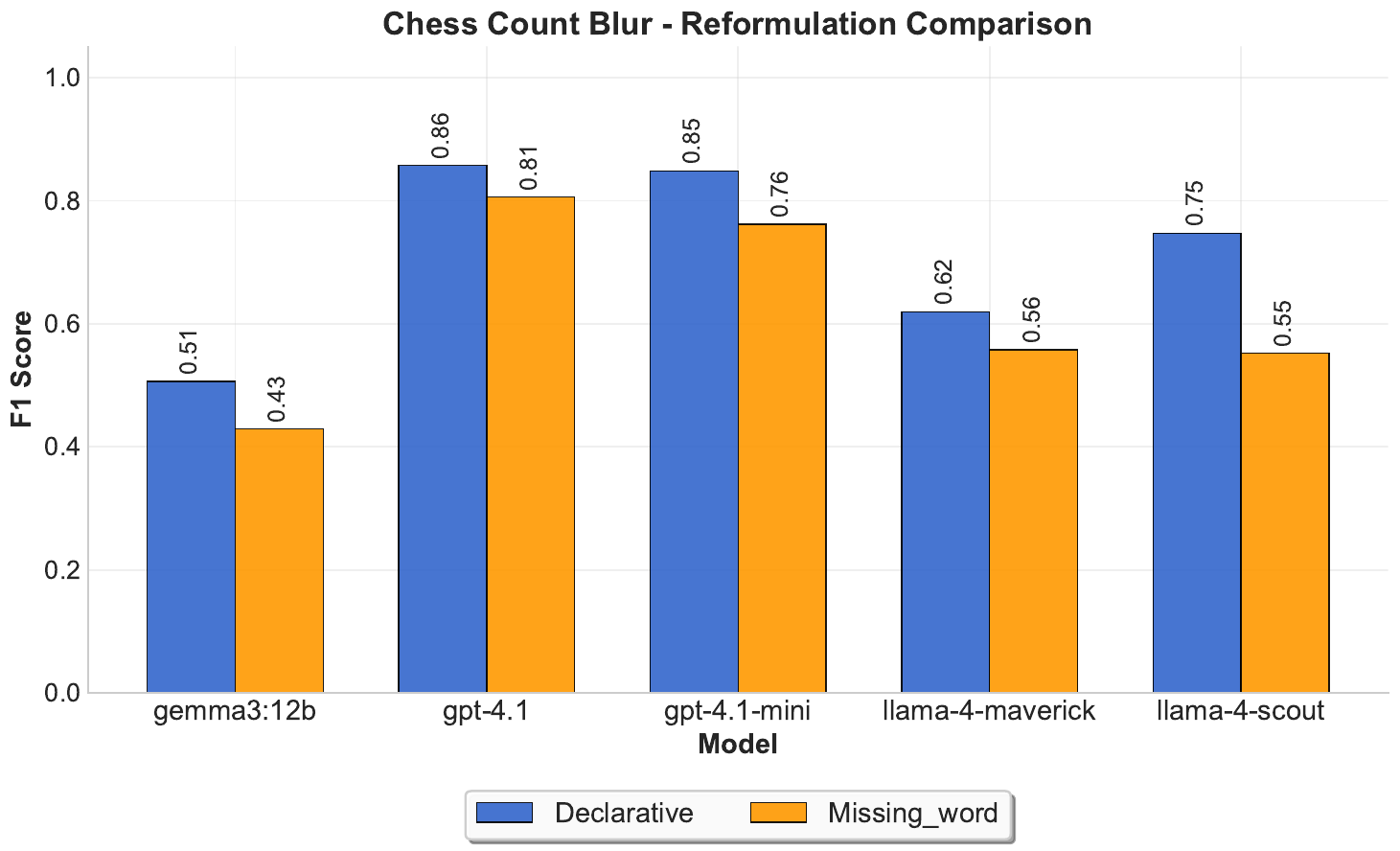}
    \textbf{(b)} Chess – Reformulation Comparison
  \end{minipage}
  \vspace{4mm}

  \begin{minipage}[t]{0.48\linewidth}
    \centering
    \includegraphics[width=\linewidth]{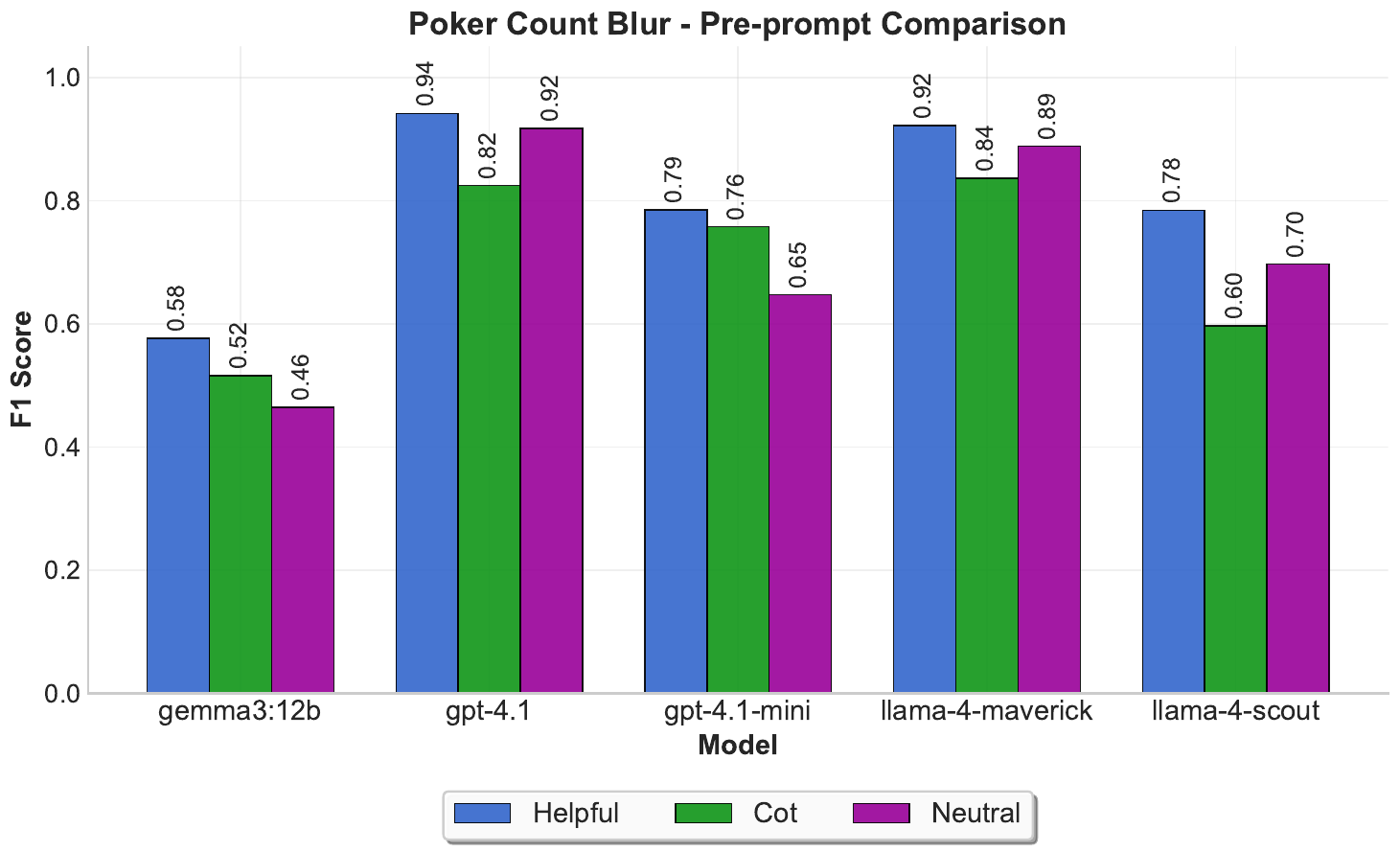}
    \textbf{(c)} Poker – preprompt Comparison
  \end{minipage}
  \hfill
  \begin{minipage}[t]{0.48\linewidth}
    \centering
    \includegraphics[width=\linewidth]{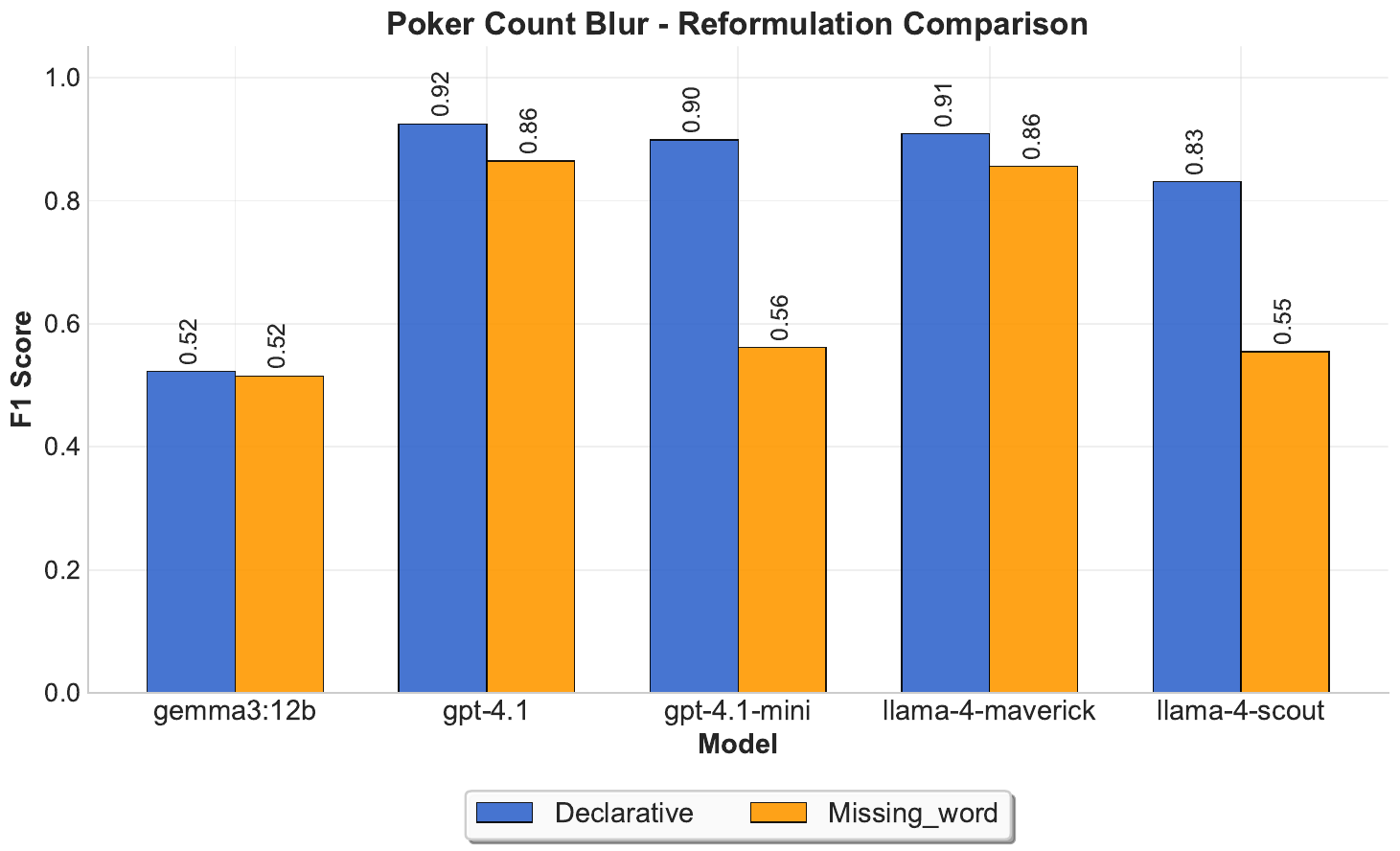}
    \textbf{(d)} Poker – Reformulation Comparison
  \end{minipage}

  \caption{
    \textbf{F1 Score Comparison for Count Blur Tasks in Chess and Poker.}
    This figure shows how linguistic strategies affect model performance under blurred visual conditions.
    (a) and (c) compare preprompt types (Helpful, Chain-of-Thought, Neutral), while (b) and (d) compare reformulation styles (Declarative vs. Missing Word).
    GPT models consistently benefit from explicit linguistic scaffolding, while LLaMA and Gemma variants show significant variance depending on prompt configuration.
  }
  \label{fig:f1_comparison_chess_poker}
\end{figure}

The results presented in Figure~\ref{fig:f1_comparison_chess_poker} offer a detailed breakdown of model performance under the Count Blur setting across two tasks (chess and poker), and two linguistic dimensions: preprompt type and reformulation style.

Across all conditions, GPT-based models (GPT-4.1 and GPT-4.1-mini) demonstrate superior F1 scores, particularly under the Helpful + Declarative combination. This pattern is consistent in both structured (chess) and unstructured (poker) scenes, suggesting strong robustness to visual degradation when linguistic scaffolding is explicit. GPT-4.1-mini slightly trails GPT-4.1 but generally follows similar trends, indicating that model size may affect resilience only marginally when prompts are clear.

In contrast, open-source models such as LLaMA and Gemma display greater variance in F1 performance, underscoring their higher sensitivity to prompt phrasing. For example, LLaMA-4-Scout performs competitively in declarative setups but exhibits sharp drops with Missing Word reformulations, particularly in poker scenes where spatial regularity is lacking. This drop reveals a prompt-induced brittleness in models without strong pretrained instruction alignment.

Interestingly, LLaMA-4-Maverick shows more consistent performance in the poker task than in chess. One hypothesis is that in less structured layouts (like poker), the model may rely more on global object counting rather than exact spatial localization, reducing its reliance on precise blur-resilient visual grounding.

The bar plots also reinforce that Chain-of-Thought reasoning is not universally helpful: while it benefits certain configurations, it often destabilizes performance in blur-sensitive contexts. This is likely due to additional reasoning steps introducing error propagation when visual inputs are already degraded.

These findings reinforce a key insight from our main experiments: prompting design plays a pivotal role in model robustness, especially under partial information scenarios. Declarative prompts and helpful framing not only improve raw accuracy but also contribute to spatial consistency and reduced confusion under blur conditions.

\begin{figure}[ht]
  \centering
    \centering
    \includegraphics[width=\columnwidth]{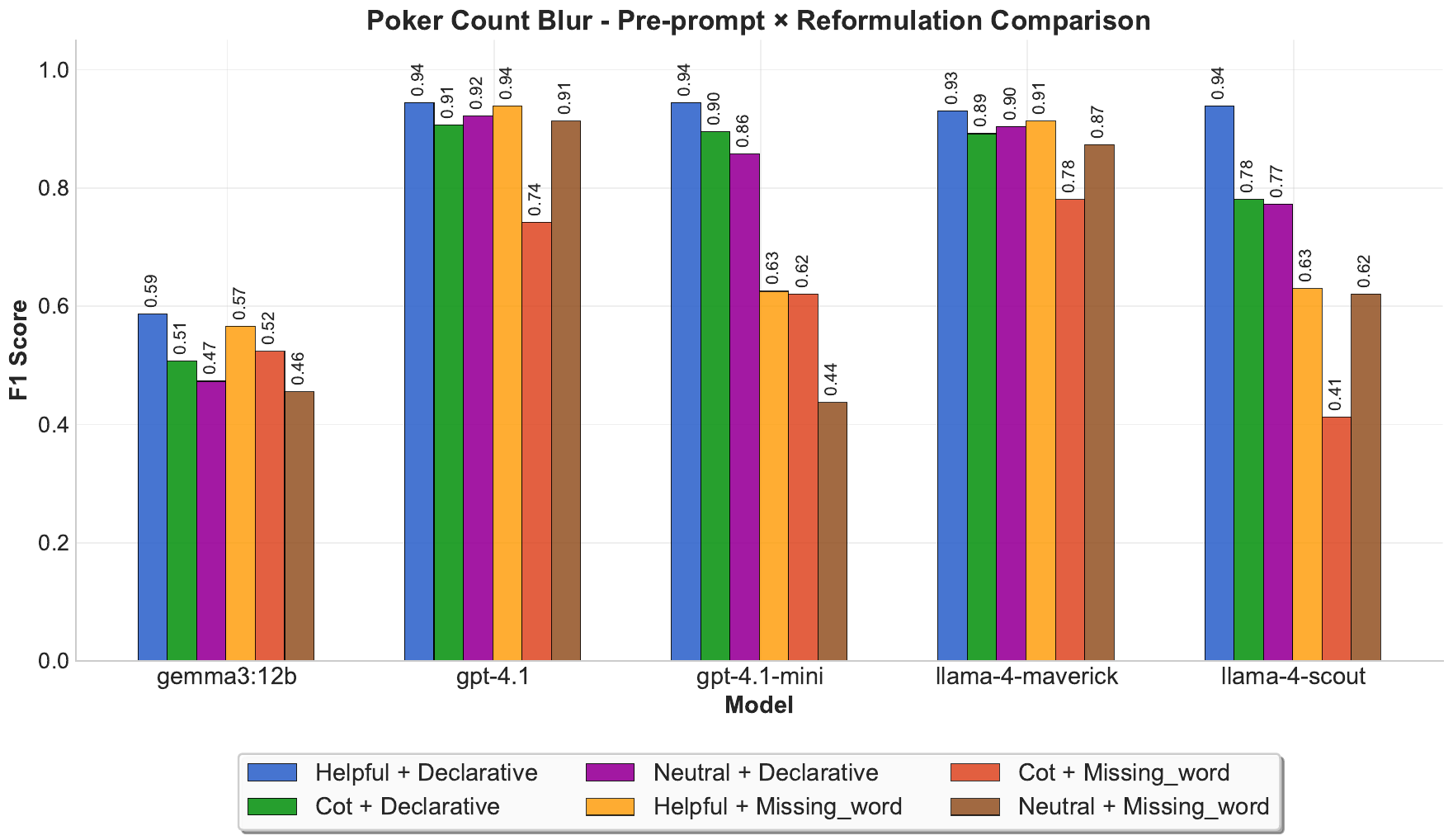}
  \caption{\textbf{F1 Score Comparison for Count Blur (Poker) across Prompt × Reformulation Combinations.}
    The interaction of prompting strategies reveals similar trends as in the chess setting, but with even sharper performance gaps between configurations.
    GPT-4.1 achieves near-ceiling F1 scores under Helpful + Declarative input, while open-source models such as Gemma and LLaMA degrade significantly when phrasing is ambiguous or under-specified (e.g., Neutral + Missing Word).
    The broader spread across combinations emphasizes the importance of carefully aligned instructions for models operating under visual degradation.
  }
  \label{fig:poker_count_blur_full_interaction}
\end{figure}

\begin{figure}[ht]
  \centering
    \centering
    \includegraphics[width=\columnwidth]{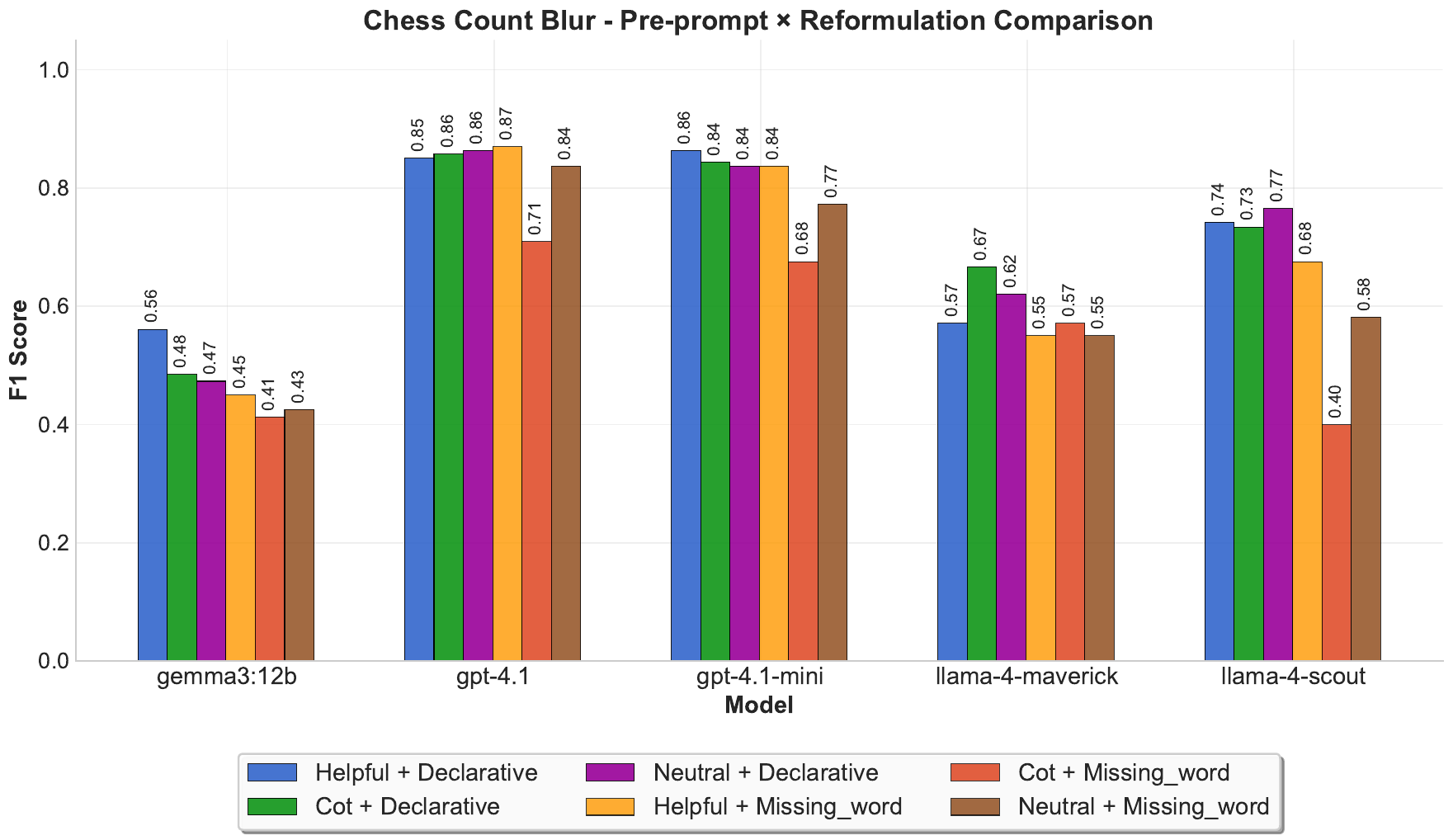}
  \caption{\textbf{F1 Score Comparison for Count Blur (Chess) across Prompt × Reformulation Combinations.}
    Each bar represents model performance under a unique combination of preprompt style (Helpful, CoT, Neutral) and reformulation strategy (Declarative, Missing Word). 
    GPT models (GPT-4.1, GPT-4.1-mini) perform robustly across most configurations, especially with Helpful + Declarative prompting.
    In contrast, LLaMA and Gemma exhibit strong variability—particularly under Missing Word reformulations—showing reduced reliability when linguistic clarity decreases.
  }
  \label{fig:chess_count_blur_full_interaction}

\end{figure}

\begin{table}[h]
\centering
\caption{Performance Metrics for Count Blur Task. Performance on the counting task with different levels of blur. Low, Medium and High blur conditions progressively reduce visibility.}
\vspace{4pt}
\label{tab:count_blur_task}
\setlength{\tabcolsep}{4pt}
\small
\begin{tabular}{l|l|cc}
\toprule
Blur Condition & Model & Accuracy (\%) & MAE \\\addlinespace[0.5ex]
%\multicolumn{4}{l}{\scriptsize Blur values: Low: 1.0, High: 5.0} 
%\midrule
\hline
\multirow{5}{*}{Low Blur: 1} 
 & gpt-4.1 & 74.3 & 0.32 \\
 & gpt-4.1-mini & \textbf{80.6} & \textbf{0.21} \\
 & LLaMA-4-scout-17b-16e-instruct & 63.9 & 0.47 \\
 & LLaMA-4-maverick-17b-128e-instruct & 44.5 & 0.67 \\
 & gemma3:12b & 45.0 & 1.41 \\
\midrule
\multirow{5}{*}{Medium Blur: 3} 
 & gpt-4.1 & 74.0 & 0.33 \\
 & gpt-4.1-mini & \textbf{76.0} & \textbf{0.25} \\
 & LLaMA-4-scout-17b-16e-instruct & 59.0 & 0.54 \\
 & LLaMA-4-maverick-17b-128e-instruct & 40.0 & 0.69 \\
 & gemma3:12b & 39.0 & 1.59 \\
\midrule
\multirow{5}{*}{High Blur: 5} 
 & gpt-4.1 & \textbf{55.0} & \textbf{0.69} \\
 & gpt-4.1-mini & 51.0 & 1.40 \\
 & LLaMA-4-scout-17b-16e-instruct & 25.0 & 8.23 \\
 & LLaMA-4-maverick-17b-128e-instruct & 40.0 & 5.74 \\
 & gemma3:12b & 26.0 & 4.58 \\

\bottomrule
\end{tabular}

\end{table}

%%% SUmmaaryyyy 

The full interaction plots in Figures~\ref{fig:poker_count_blur_full_interaction} and ~\ref{fig:chess_count_blur_full_interaction} expose how prompt type and reformulation style interact to amplify or mitigate performance gaps under Blur conditions. In both domains, Helpful preprompt with Declarative reformulation combination remains the clear optimum, driving GPT to near-ceiling F1 scores. However, the penalty for moving away from this configuration is notably task-dependent: 
\begin{itemize}
    \item \textbf{Poker scenes} (Figure~\ref{fig:poker_count_blur_full_interaction}) show the widest spread across bars, with open-source models collapsing under Neutral preprompt with Missing-Word reformulation, or even CoT preprompt with Missing Word reformulation, highlighting their reliance on explicit guidance when object layout is irregular. 
    \item \textbf{Chess scenes} (Figure~\ref{fig:chess_count_blur_full_interaction}) exhibit a narrower spread, yet still reveal a consistent drop for LLaMA and Gemma variants whenever either the preprompt is less informative or the reformulation omits key lexical cues. 
\end{itemize}
Taken together, these interaction plots underscore the compounding effect of linguistic scaffolding on model robustness. Our findings suggest that both preprompt framing and reformulation structure contribute synergistically to performance under visual degradation. The absence of either component, particularly in open-source models with limited instruction tuning, leads to a marked decline in task reliability, highlighting the necessity of jointly optimizing prompt clarity and linguistic structure in VLM evaluation protocols. 
\subsection{Identification}

%In this setup, objects are closely packed and must be uniquely identified based on type or label. This task emphasizes fine-grained perception and name resolution.
This task assesses a model's ability to recognize and classify individual chess pieces in cluttered scenes where objects are tightly packed. Unlike counting or localization, the identification task demands precise type-level resolution (e.g., distinguishing a white pawn from a black rook) based on both visual appearance and spatial cues. The setup is particularly challenging as it stresses the model's fine-grained visual understanding, symbol-to-label mapping, and resistance to object occlusion or visual overlap. 

Models are evaluated across variations in prompt formulation (Declarative vs. Missing Word) and instruction strategy (helpful, Chain-of-Thought, Neutral). This allows us to test how linguistic scaffolding influences the model's ability to resolve ambiguous visual identities. 

\textbf{Insights:}
\begin{itemize}
    \item Declarative reformulations enhance type discrimination, especially when paired with explicit prompts (e.g., "Identify all...").
    \item Chain-of-Thought prompts do not consistently improve results in this task, possibly due to hallucinated intermediate steps.
    \item Open models (e.g., Gemma, Mistral) exhibit frequent misidentifications in cluttered areas regardless of prompting, indicating limited visual resolution or weak grounding.
\end{itemize}

\begin{figure}[ht]
  \centering
  \includegraphics[width=0.30\linewidth]{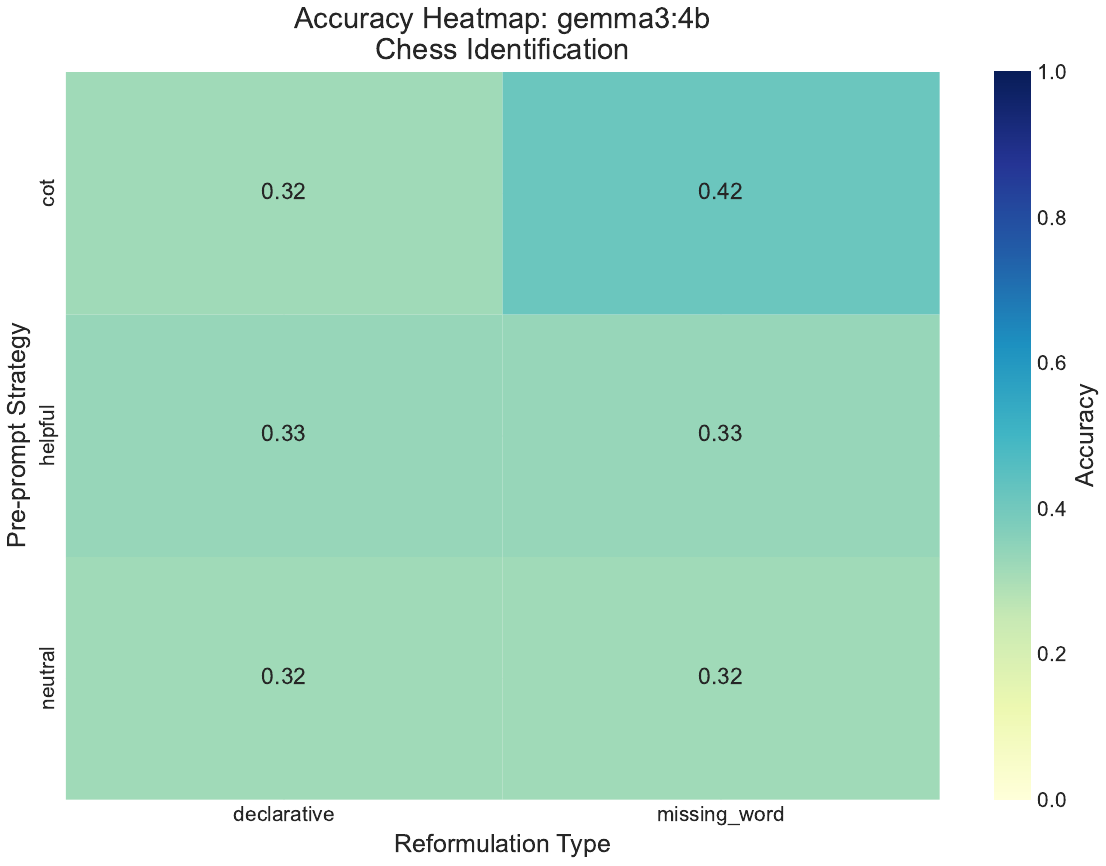}
  \includegraphics[width=0.30\linewidth]{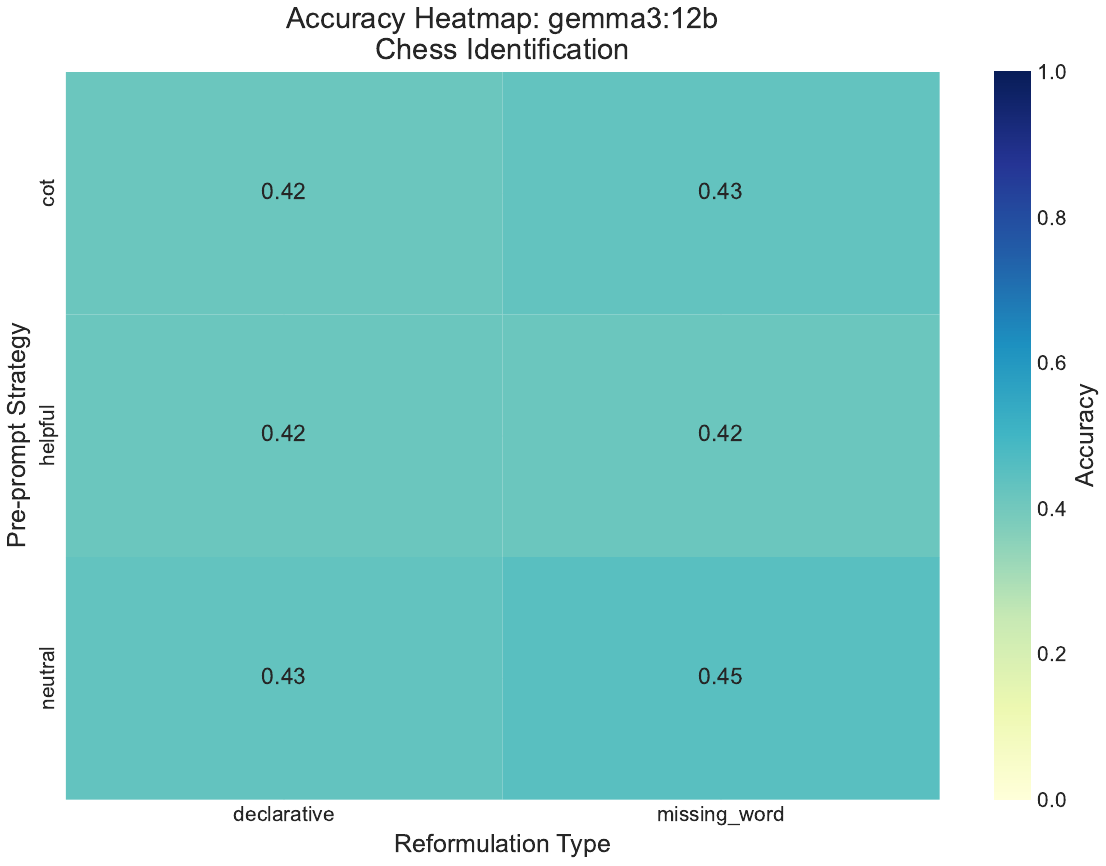}
  \includegraphics[width=0.30\linewidth]{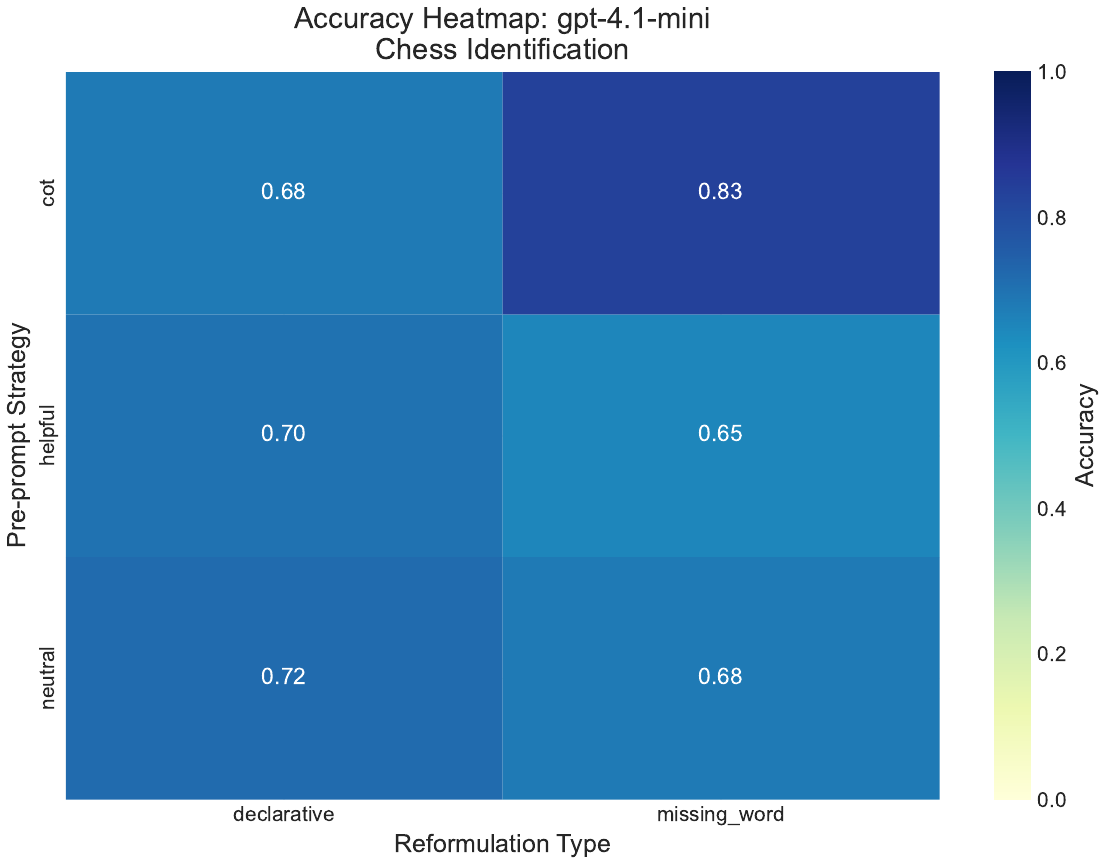}
  \\
  \includegraphics[width=0.30\linewidth]{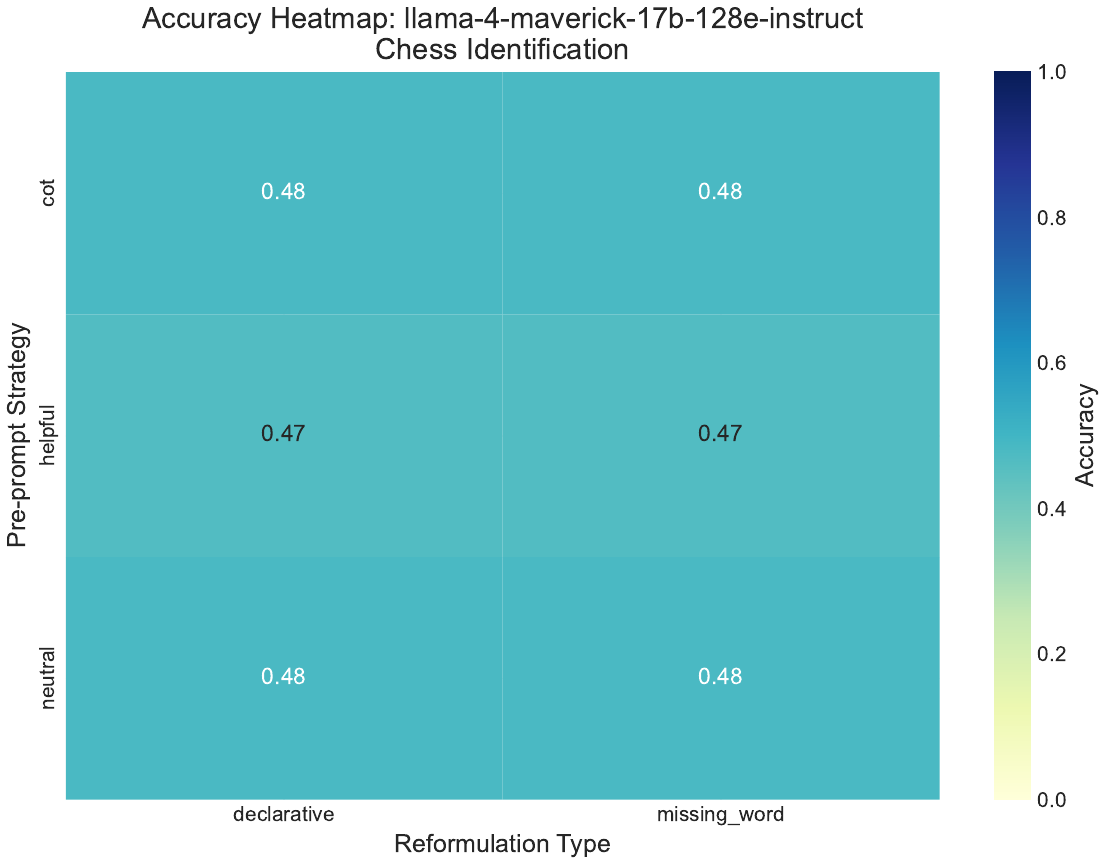}
  \includegraphics[width=0.30\linewidth]{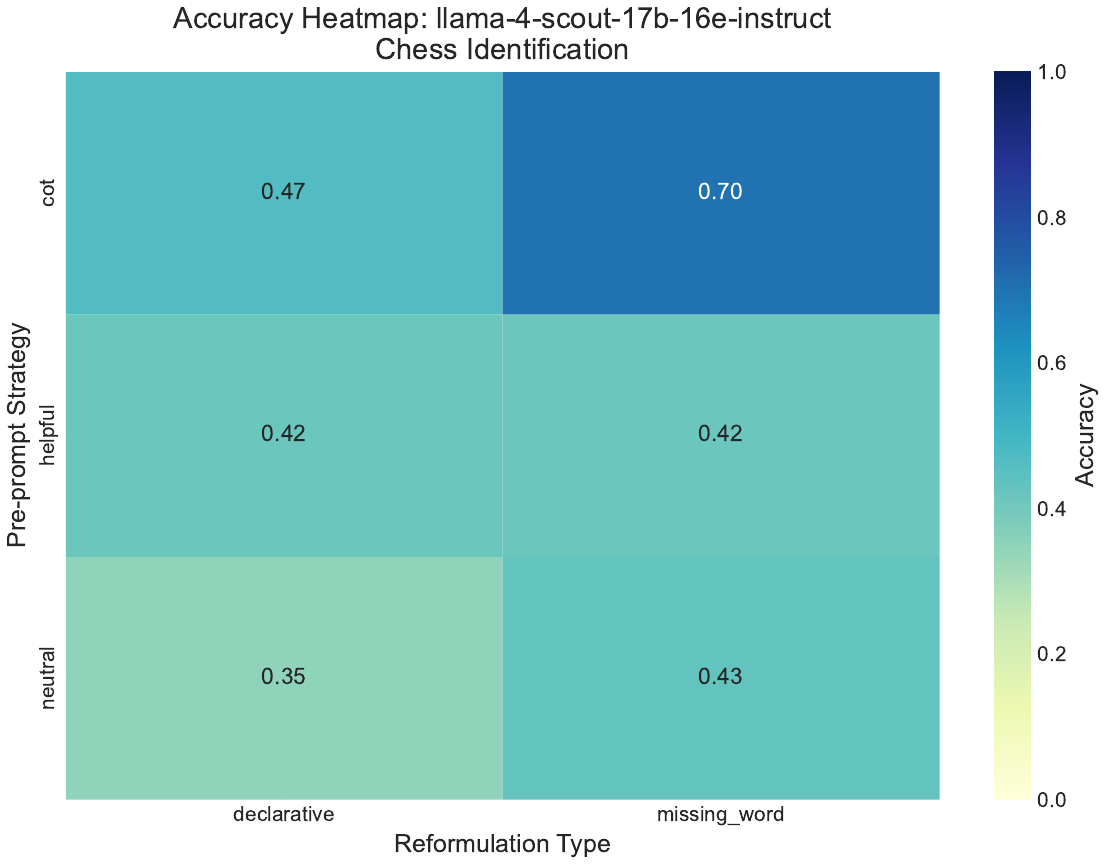}
  \includegraphics[width=0.30\linewidth]{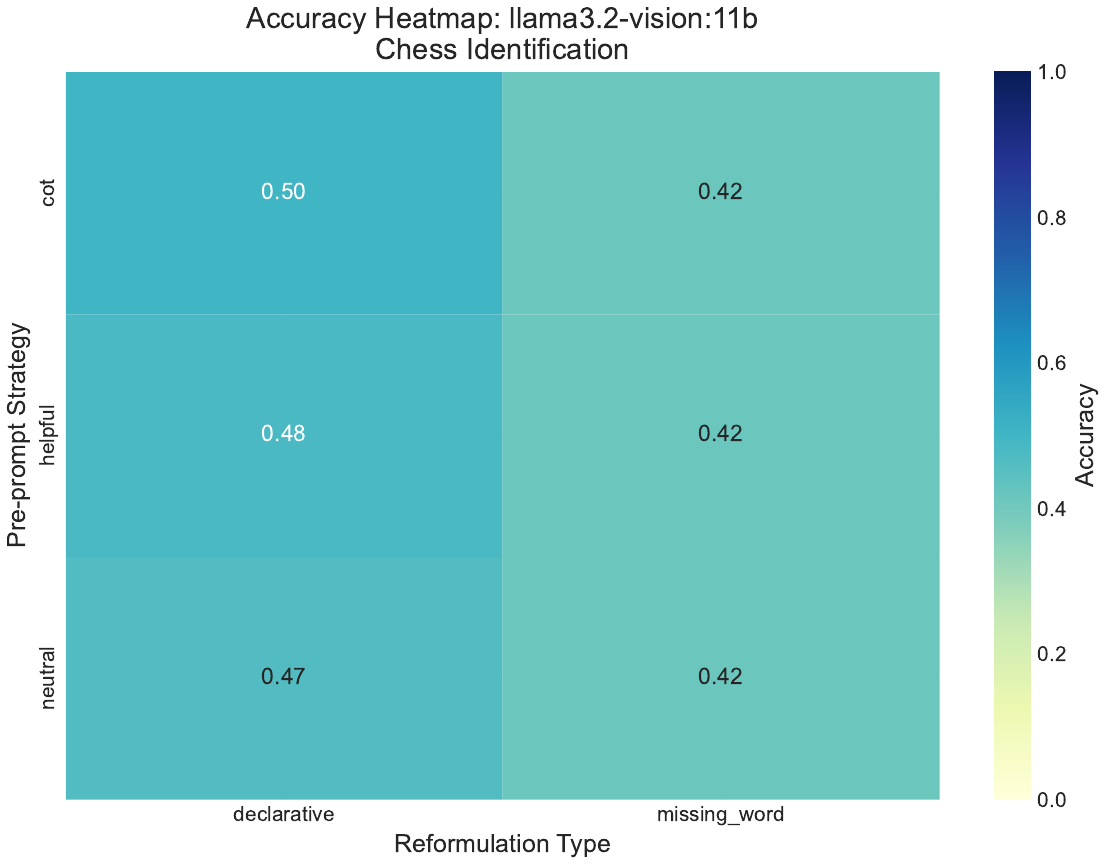}
  \\
  \includegraphics[width=0.30\linewidth]{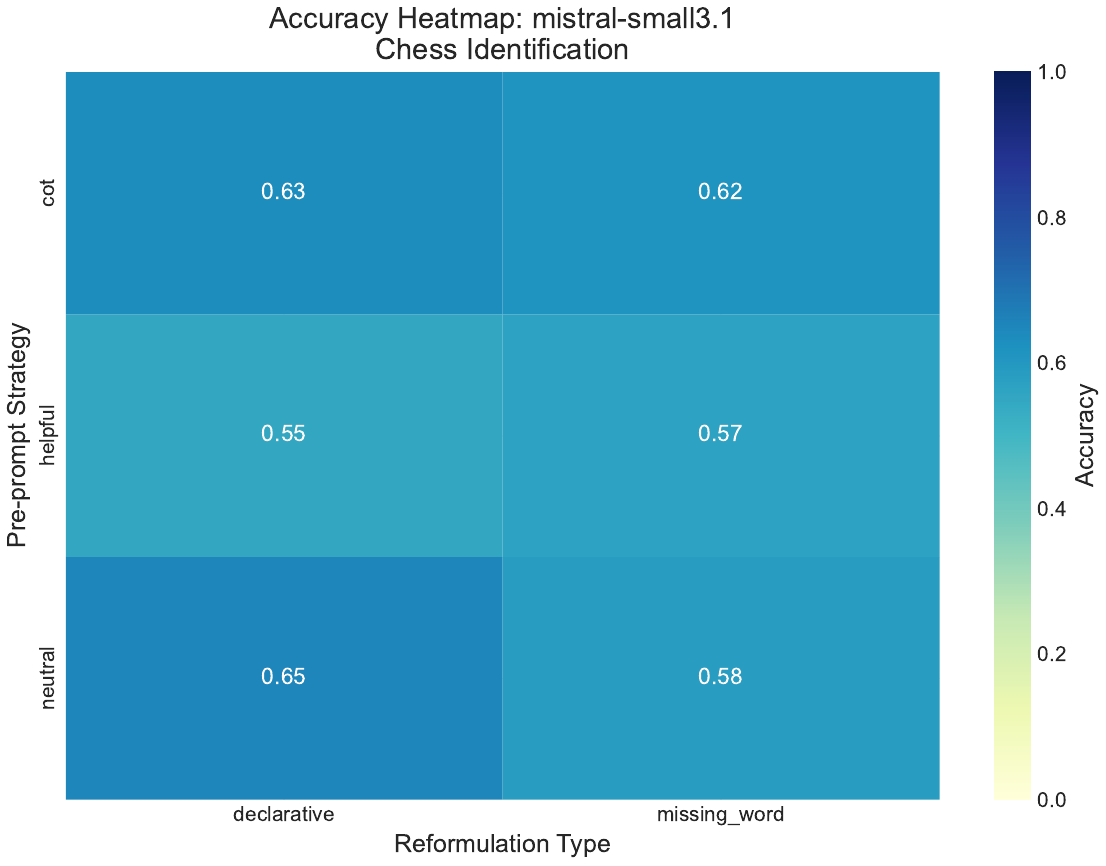}

  \caption{Heatmaps for the Identification task across multiple models and prompting strategies, averaged over 30 densely packed synthetic chess scenes.}
  \label{fig:identification_heatmap}
\end{figure}
Overall, GPT-4.1-mini demonstrates the most consistent performance across all prompting styles, with the highest accuracy observed with Declarative reformulation and Helpful preprompt. Its heatmap shows well-defined predictions with minimal confusion between piece types. 
Mistral-small3.1 performs moderately well suggesting relatively stable visual grounding, though it suffers slightly more in scenes with high visual congestion. LLaMA variants show mixed behavior. Accuracy range depends on prompt configuration. Notably, LLaMA-4 Scout variant briefly peaks with Declarative reformulation and Helpful preprompt, indicating some sensitivity to instruction quality. Gemma models underperform consistently, likely due to limited visual fidelity and weaker object-label grounding. They frequently confuse similar-looking pieces and fail to disambiguate based on position or color cues. Across all models, Declarative reformulations with Helpful preprompting result in the most accurate and spatially stable identification. Chain-of-Thought prompts often lead to hallucinated intermediate reasoning, especially for open-source models, reducing final accuracy.

%%%%%%%%%%%%%%%%%%%%%%%%%%%%%%%%%%%%%%%%%%%%%%%%%%%%%%%%%%%%% Identification Distance 

\subsection{Identification Distance}

This variation probes the ability of models to identify object types when the objects are spaced apart across the scene. Unlike the standard Identification setup where objects are densely packed, this version introduces increasing spatial distance between entities. This setting reduces visual occlusion but increases the need for robust spatial reference and positional grounding.

It simulates real-world perception scenarios such as autonomous navigation or robotic sorting tasks, where entities must be recognized from a distance or across sparsely populated environments.

\begin{figure}[ht]
  \centering
  \includegraphics[width=0.30\linewidth]{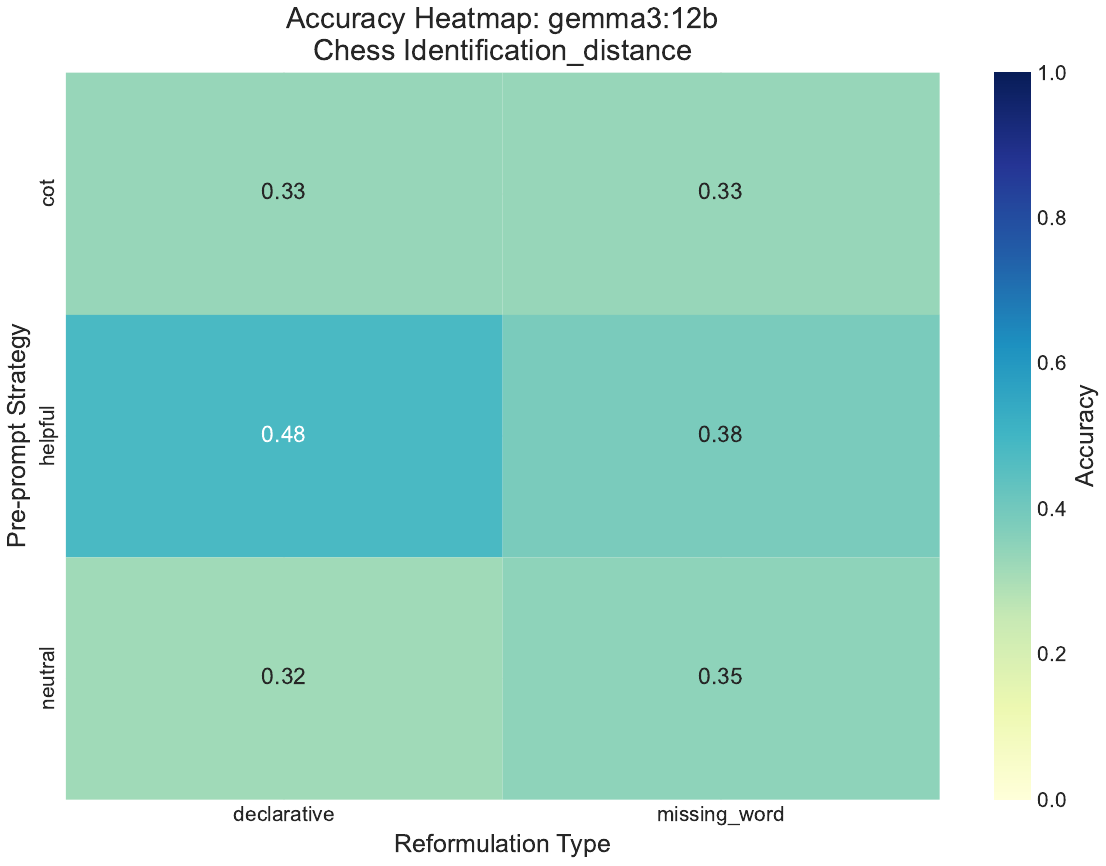}
  \includegraphics[width=0.30\linewidth]{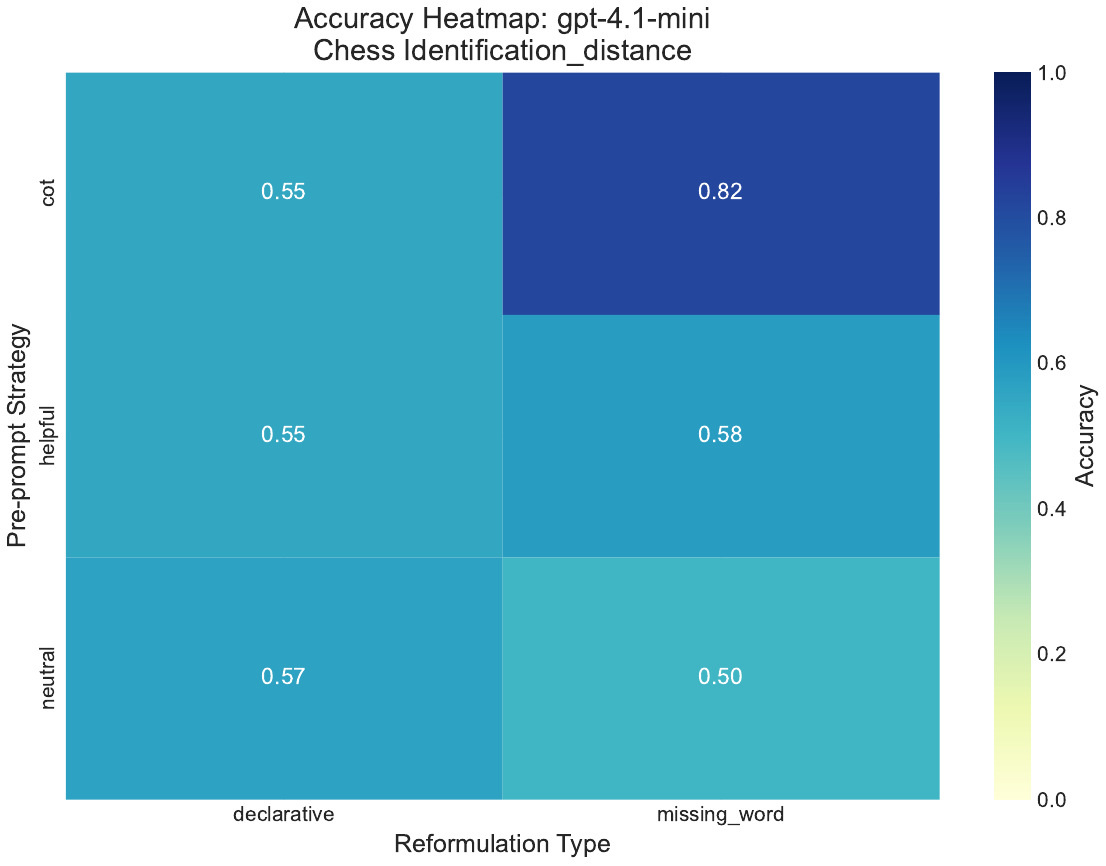}
  \includegraphics[width=0.30\linewidth]{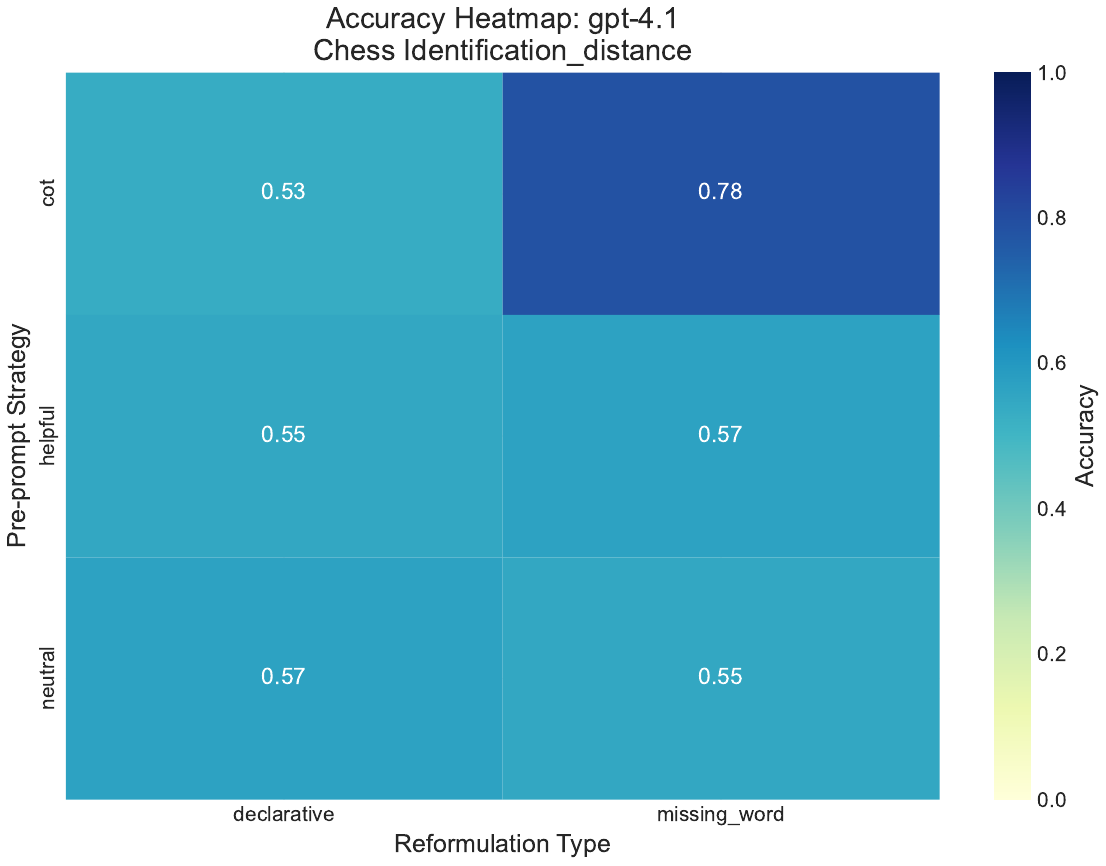}
  \\
  \includegraphics[width=0.30\linewidth]{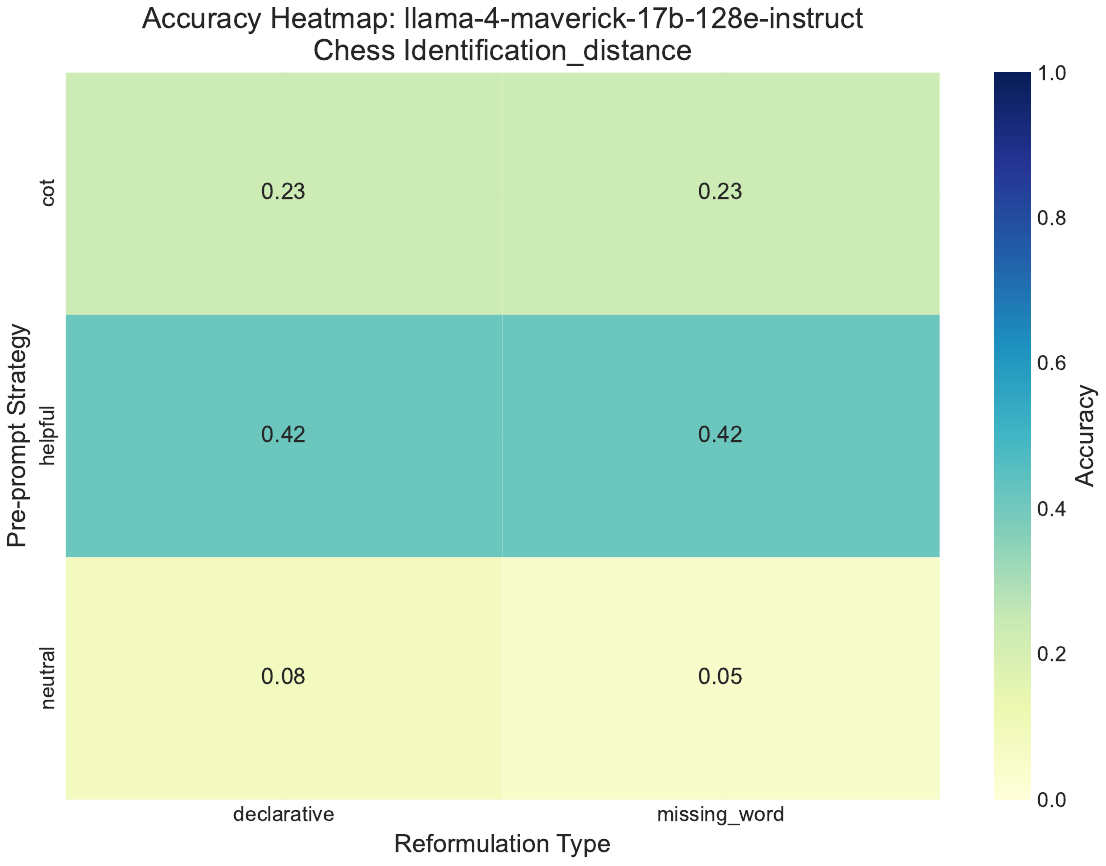}
  \includegraphics[width=0.30\linewidth]{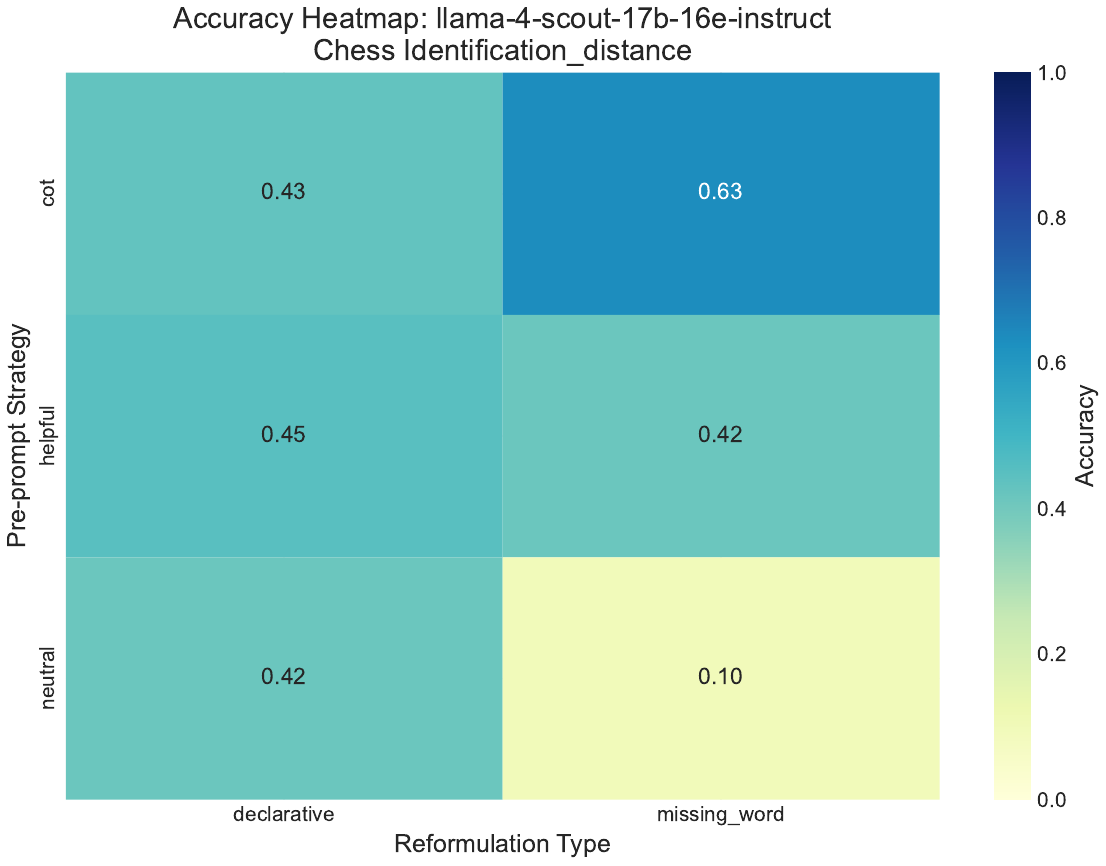}
  \includegraphics[width=0.30\linewidth]{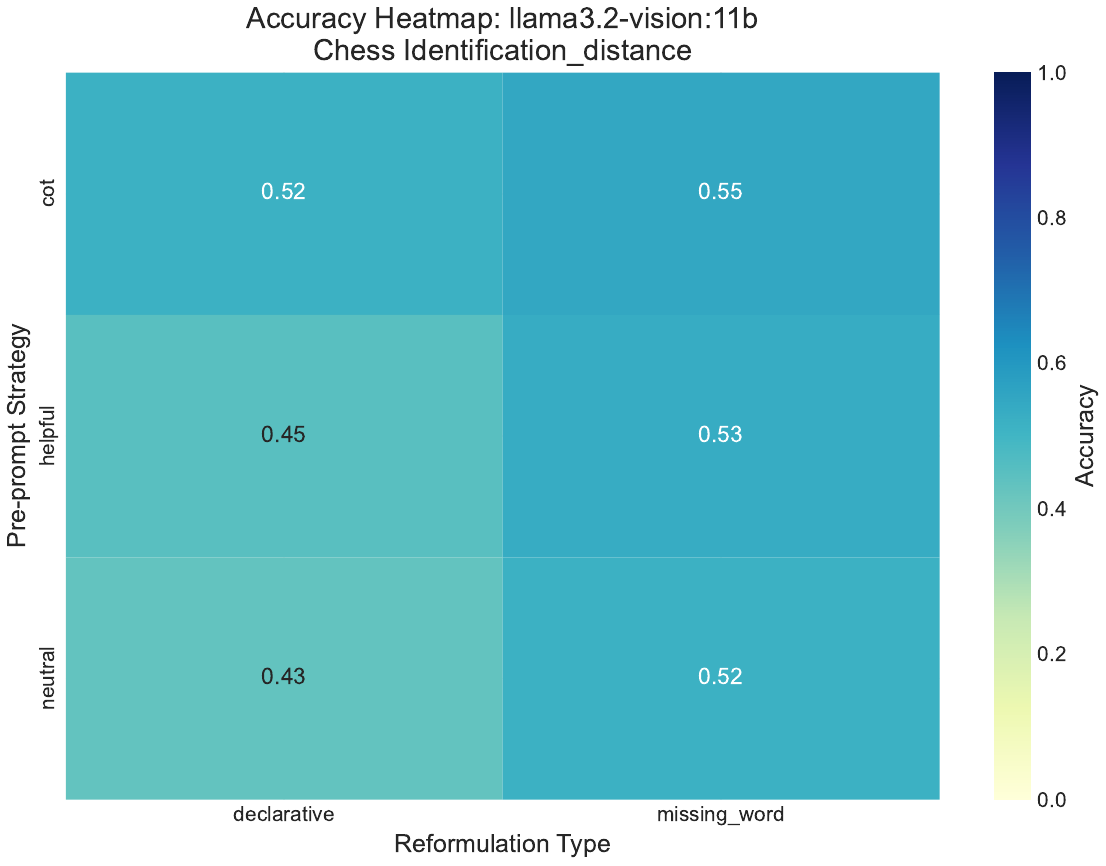}
  \\
  \includegraphics[width=0.30\linewidth]{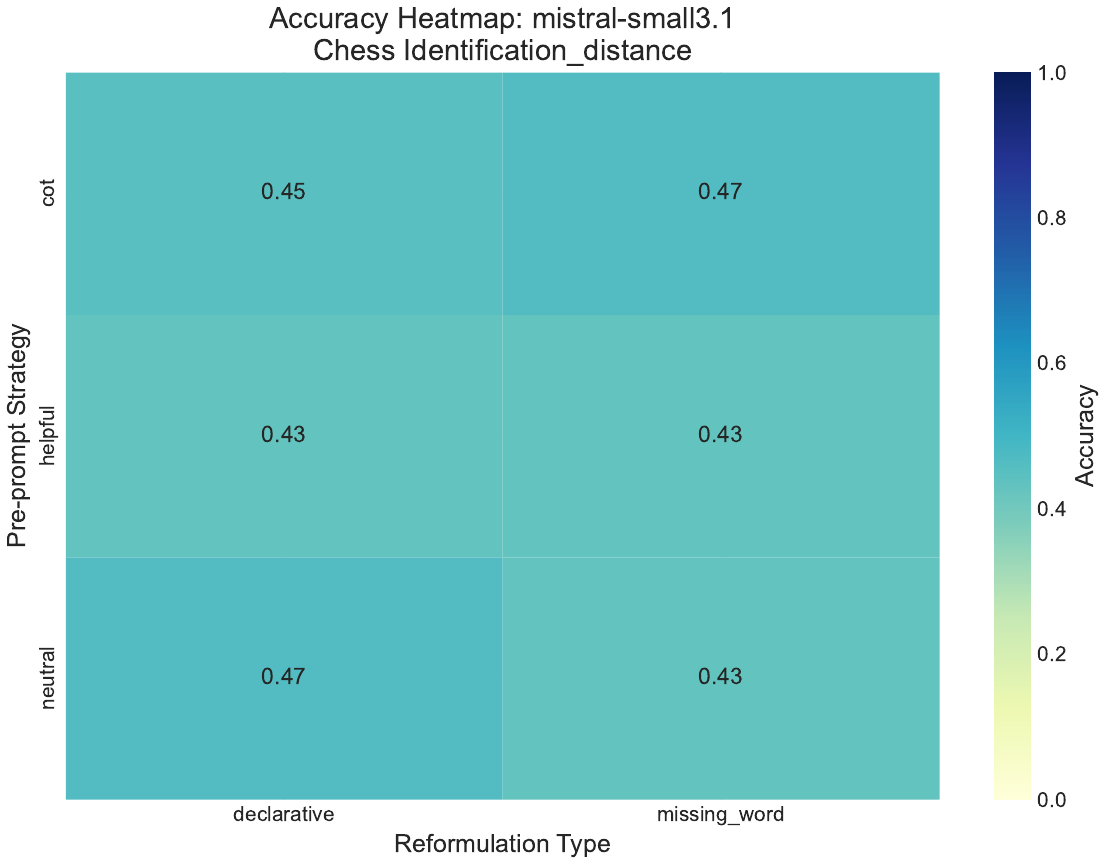}
 
  \caption{Heatmaps for the Identification Distance task across multiple models and prompting strategies, averaged over 140 spatially-separated synthetic scenes. This variation isolates long-range visual discrimination capacity.}
  \label{fig:identification_distance_heatmap}
\end{figure}

We again compare different prompting styles (Helpful, Chain-of-Thought, Neutral) and reformulation templates (Declarative vs. Missing Word), focusing on how they modulate model attention and symbol-to-object mapping under reduced visual clutter. Figure~\ref{fig:identification_distance_heatmap} illustrates the different results. 

\begin{figure}[ht]
  \centering
  \begin{minipage}[t]{0.48\linewidth}
    \centering
    \includegraphics[width=\linewidth]{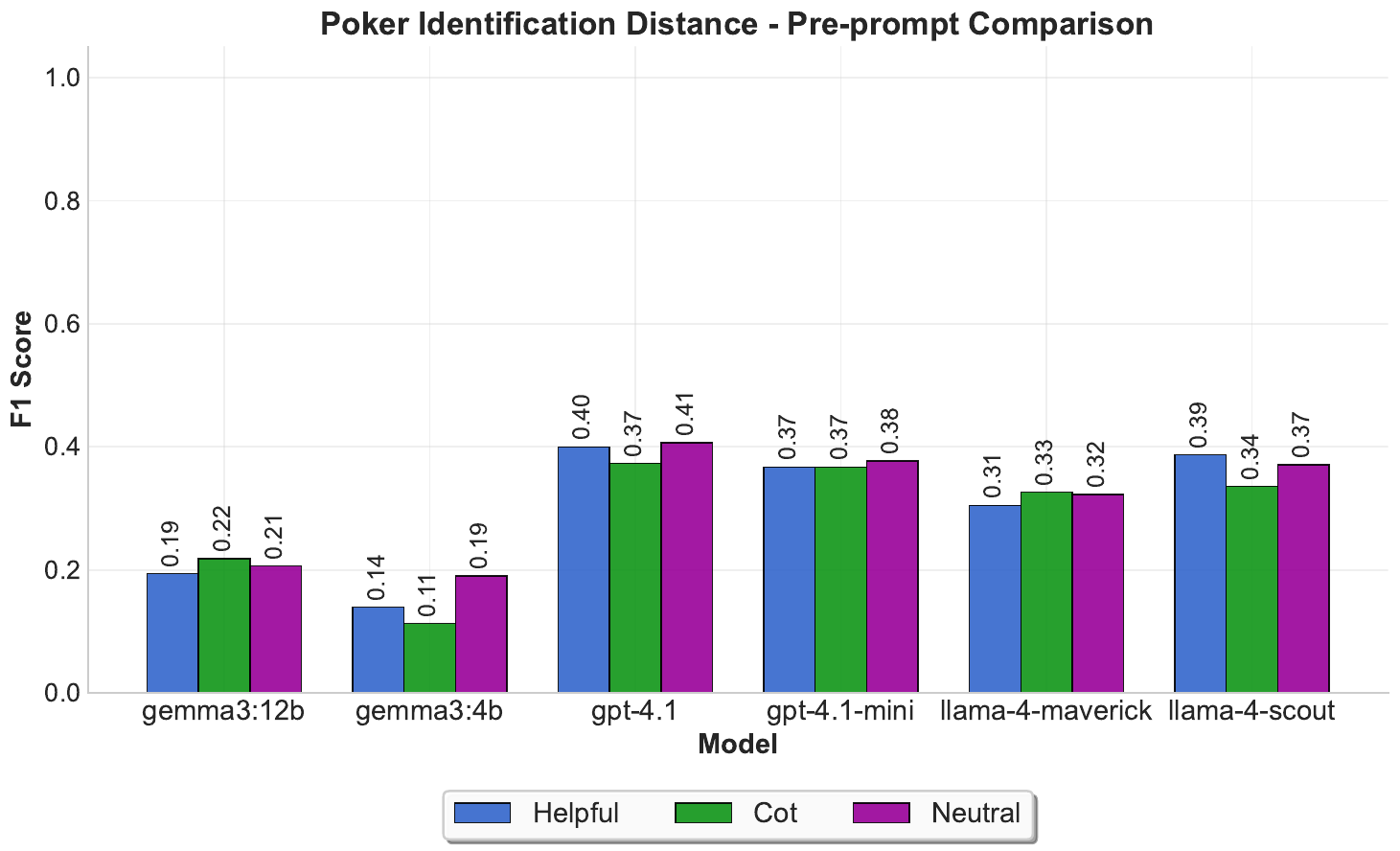}
    \textbf{(a)} Preprompt Comparison
  \end{minipage}
  \hfill
  \begin{minipage}[t]{0.48\linewidth}
    \centering
    \includegraphics[width=\linewidth]{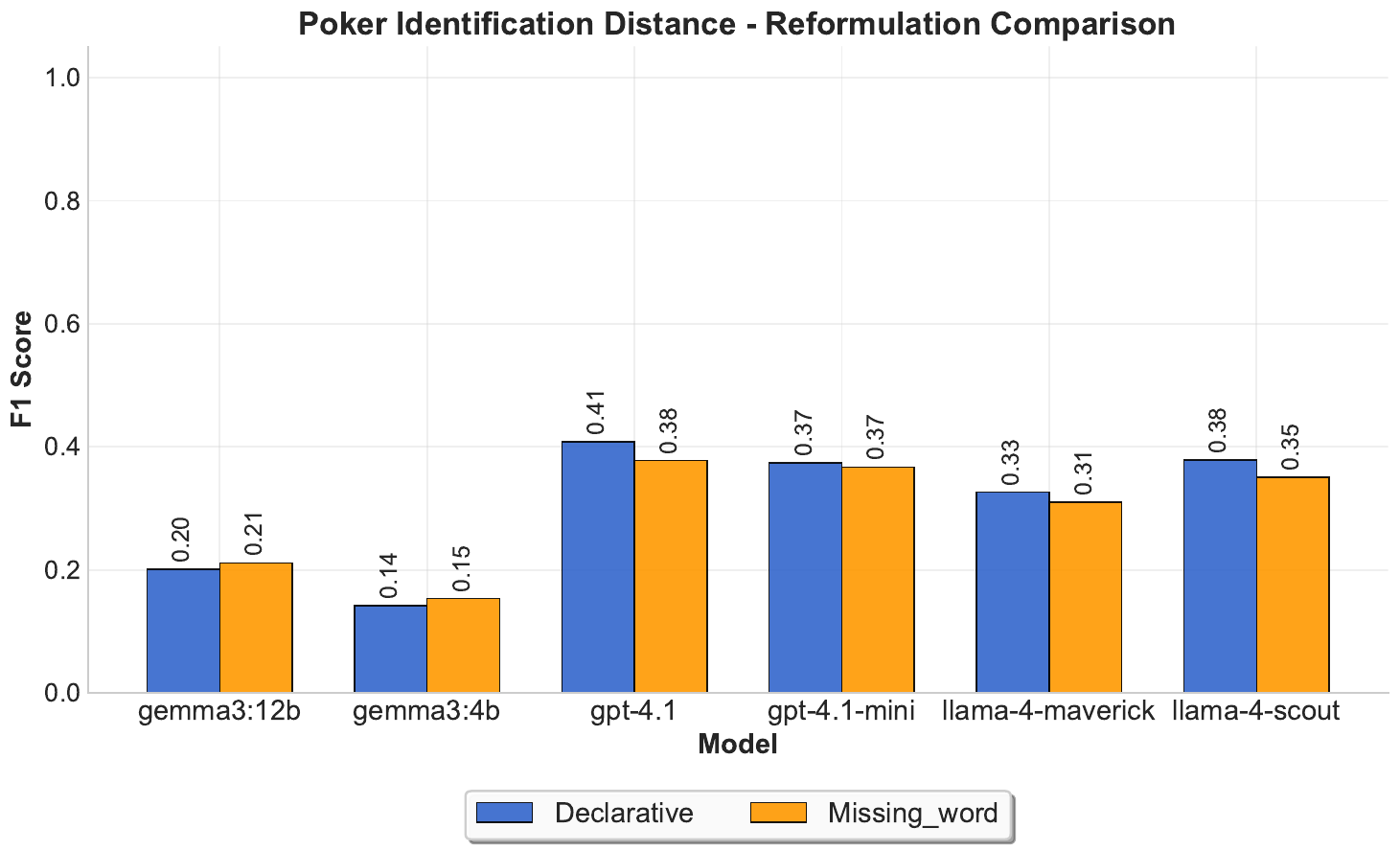}
    \textbf{(b)} Reformulation Comparison
  \end{minipage}

  \caption{\textbf{F1 Score Comparison for the Poker Identification Distance Task.}
  This figure presents a detailed comparison of model performance when identifying spatially-separated objects in poker scenes under varied linguistic conditions.
  Subfigure (a) compares preprompt types (Helpful, Chain-of-Thought, Neutral), while subfigure (b) contrasts reformulation styles (Declarative vs. Missing Word).
  GPT models consistently benefit from explicit linguistic scaffolding, particularly the Declarative + Helpful configuration, achieving higher F1 scores and more stable outputs.
  In contrast, open-source models (LLaMA, Gemma) exhibit greater sensitivity to prompt clarity and structure, with sharp performance drops under ambiguous phrasing.
  These results reinforce that even in low-occlusion, high-separation contexts, prompt design remains essential for achieving reliable identification performance.
  }
  \label{fig:f1_poker_identification_distance}
\end{figure}

The Poker Identification Distance results further highlight the role of linguistic scaffolding in modulating VLM behavior under low-density, spatially separated settings. As shown in Figure~\ref{fig:f1_poker_identification_distance}, GPT-based models maintain strong performance across all prompting styles, with the best F1 scores observed under the Declarative + Helpful configuration. This suggests that these models are capable of leveraging both spatial context and linguistic cues to maintain object identity across distributed scenes, a property especially useful in real-world applications like robotic vision or inventory inspection.

Interestingly, LLaMA-4 Maverick outperforms Scout in most configurations for this task, reflecting improved stability under distance-based separation. However, both models remain highly sensitive to prompt phrasing: performance drops significantly when the prompt becomes less explicit, as in the Missing Word condition. Gemma’s results confirm this fragility; it shows erratic responses and reduced F1 scores across both prompting and reformulation variations.

\begin{figure}[H]
\centering
\begin{vlmcasebox}

\begin{center}
  \begin{minipage}{0.6\linewidth}
    \centering
    \includegraphics[width=\linewidth]{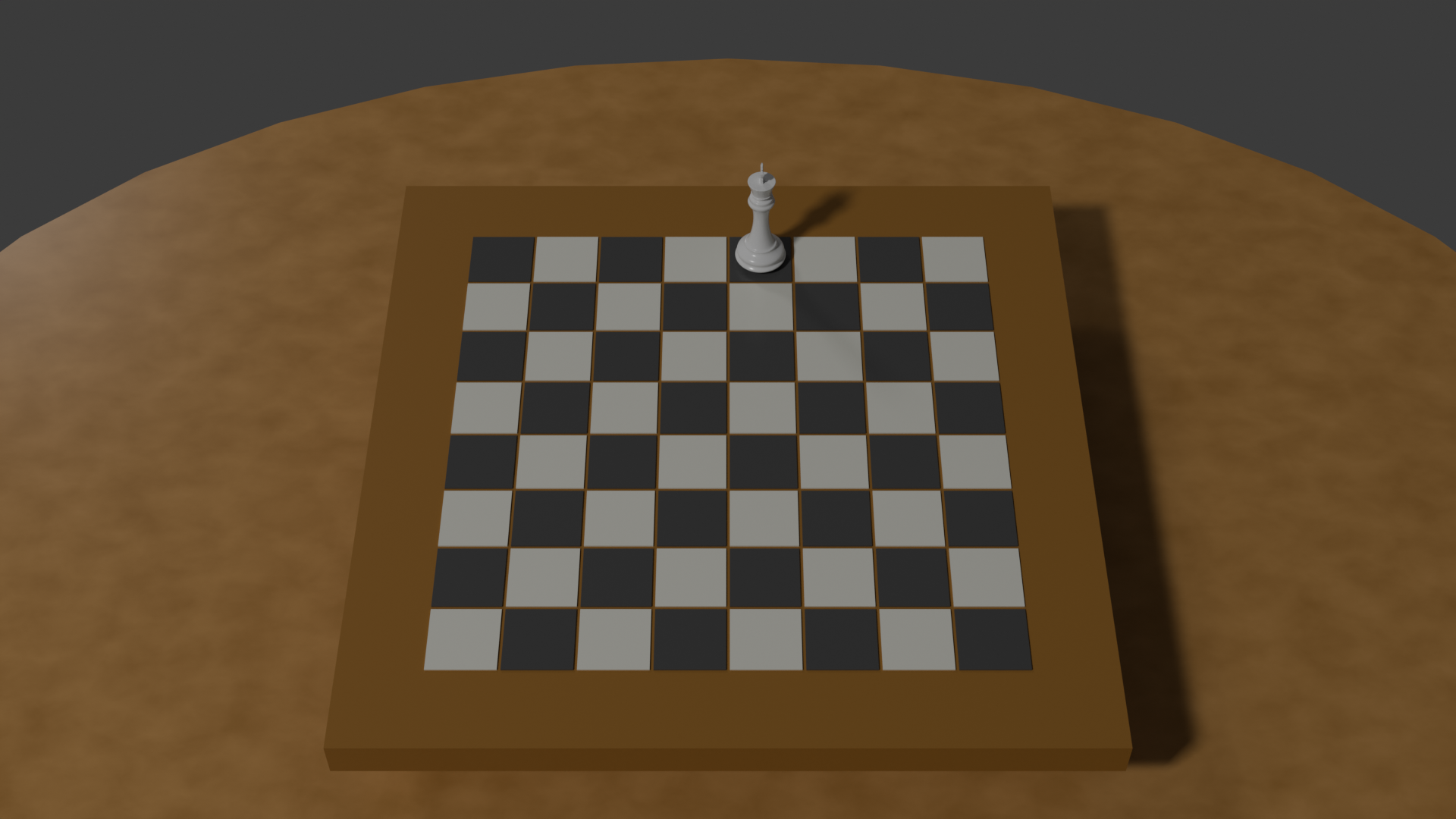}
    \captionof*{figure}{\textbf{Input Image}}
  \end{minipage}
\end{center}

\vspace{0.8em}

\textbf{Task: Color Identification in Synthetic Chessboard Image}

\textbf{Question:} Analyze the image carefully. What color is the piece on the board?  
Respond in a declarative format: \textit{“The color of the piece on the board is:”}

\vspace{0.6em}

\begin{center}
\begin{tabular}{@{}lll@{}}
\toprule
\textbf{Model} & \textbf{Response} &  \\
\midrule
GPT-4.1 & The color of the piece on the board is: \textbf{white} & \cmark \\
GPT-4.1-mini & The color of the piece on the board is: \textbf{white} & \cmark \\
LLaMA-4-scout & The color of the piece on the board is: \textbf{white} & \cmark \\
LLaMA-4-maverick & The color of the piece on the board is:  \textbf{white} & \cmark \\
Gemma-4b & The color of the piece on the board is:  \textbf{white} & \cmark \\
Gemma-12b & The color of the piece on the board is: \textbf{white} & \cmark \\
LLaMA3.2-vision & The color of the piece on the board is: \textbf{white} & \cmark \\
Mistral-small3.1 & The color of the piece on the board is: \textbf{white} & \cmark \\
\bottomrule
\end{tabular}
\end{center}

\vspace{0.5em}
\textbf{Ground Truth:} White \\
\textbf{Result:} All models answered correctly.

\end{vlmcasebox}
\caption{VLM responses to a color identification question based on a synthetic chess scene.}
\label{fig:chess_vertical_case}
\end{figure}

Notably, while the overall visual complexity is lower in this task compared to densely packed scenes, prompting style still significantly affects performance. Chain-of-Thought reasoning, in particular, continues to be a double-edged sword: in some cases, it helps weaker models by injecting reasoning structure, but often it introduces noise that leads to reduced confidence in symbol-to-object resolution. These findings confirm that even when objects are well-separated, linguistic design remains a critical lever for guiding model precision in identity-based tasks. Figure~\ref{fig:chess_vertical_case} presents a representative example of model predictions for the identification task.

\textbf{Findings:}
\begin{itemize}
    \item GPT-family models leverage spatial separation well, especially under Helpful prompts with Declarative phrasing.
    \item LLaMA-based models underperform in Declarative + CoT combinations, possibly due to misalignment between reasoning chains and visual grounding.
    \item Missing Word prompts reduce recall at peripheral regions, indicating sensitivity to template ambiguity.
\end{itemize}

%%%%%%%%%%%%%%%%%%%%%%%%%%%%%%%%%%%%%%%%%%%%%%%%%%% Localization 

\subsection{Localization}

This diagnostic task evaluates the ability of the models to localize individual objects within compact and discrete spatial layouts. In the \textbf{Chess} setting, models are required to identify the position of specific pieces on a board where even minor spatial drift leads to errors. This setup emulates constrained applications such as tabletop manipulation or industrial assembly tasks, where visual clutter is minimal but spatial precision is critical. 

In the \textbf{Poker} setup, we use a 3x3 grid containing a single visible card per scene. This configuration evaluates localization under lower density but higher ambiguity due to card orientation, subtle visual differences, and less structured spatial priors. 

\begin{figure}[H]
\centering
\begin{vlmcasebox}

\begin{center}
  \begin{minipage}{0.6\linewidth}
    \centering
    \includegraphics[width=\linewidth]{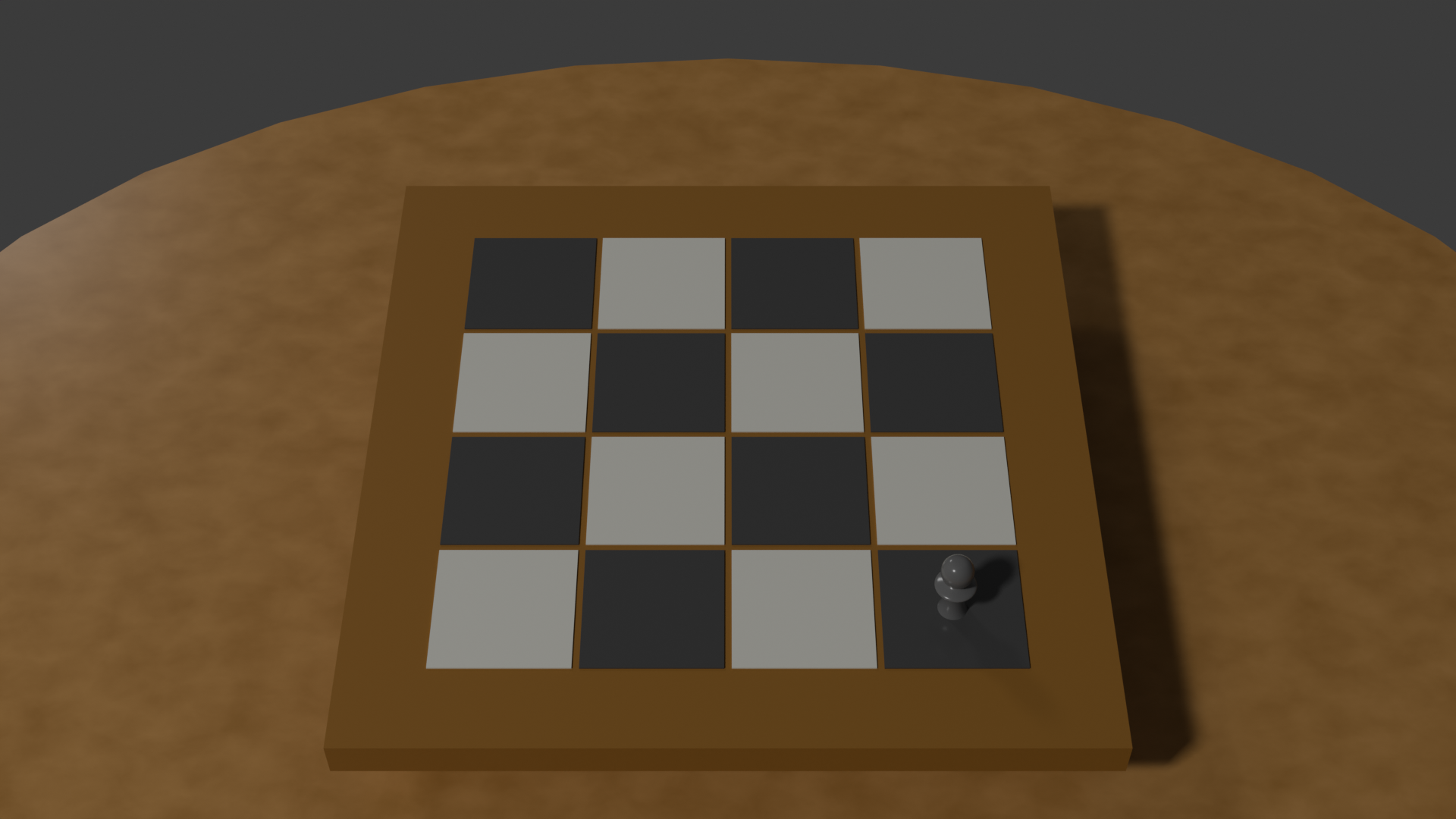}
    \captionof*{figure}{\textbf{Input Image}}
  \end{minipage}
\end{center}

\vspace{0.8em}

\textbf{Task: Piece Localization in Synthetic Chessboard Image}

\textbf{Column Question:} Analyze the image carefully, be aware that this image may differ from a classical chessboard scene. Numbering the columns from left to right, starting with 0, on which column is the piece on the board? 
Respond in a declarative format: \textit{“The column on which the piece is on the board is:”}  

\textbf{Row Question:} Analyze the image carefully, be aware that this image may differ from a classical chessboard scene. Numbering the rows from top to bottom, starting with 0, on which row is the piece on the board? Respond in a declarative format: \textit{“The row on which the piece is on the board is:”}
\vspace{0.6em}

\begin{center}
\begin{tabular}{@{}lll@{}}
\toprule
\textbf{Model} & \textbf{Response} &  \\
\midrule
GPT-4.1 & The column on which the piece is on the board is: \textbf{3} & \cmark \\
GPT-4.1 & The row on which the piece is on the board is: \textbf{3} & \cmark \\
GPT-4.1-mini & The column on which the piece is on the board is: \textbf{3} & \cmark \\
GPT-4.1-mini & The row on which the piece is on the board is: \textbf{3} & \cmark \\
LLaMA-4-scout & The column on which the piece is on the board is: \textbf{2} & \xmark \\
LLaMA-4-scout & The row on which the piece is on the board is: \textbf{3} & \cmark \\
LLaMA-4-maverick & The column on which the piece is on the board is: \textbf{3} & \cmark \\
LLaMA-4-maverick & The row on which the piece is on the board is: \textbf{3} & \cmark \\
Gemma-4b & The column on which the piece is on the board is: \textbf{2} & \xmark \\
Gemma-4b & The row on which the piece is on the board is: \textbf{1} & \xmark \\
Gemma-12b & The column on which the piece is on the board is: \textbf{3} & \cmark \\
Gemma-12b & The row on which the piece is on the board is: \textbf{3} & \cmark \\
LLaMA3.2-vision & The column on which the piece is on the board is: \textbf{2} & \xmark \\
LLaMA3.2-vision & The row on which the piece is on the board is: \textbf{2} & \xmark \\
Mistral-small3.1 & The column on which the piece is on the board is: \textbf{3} & \cmark \\
Mistral-small3.1 & The row on which the piece is on the board is: \textbf{3} & \cmark \\

\bottomrule
\end{tabular}
\end{center}

\vspace{0.5em}
\textbf{Ground Truth Column:} 3 \\
\textbf{Ground Truth Row:} 3

\end{vlmcasebox}
\caption{VLM responses to a color localization question based on a synthetic chess scene.}
\label{fig:chess_localization_vertical_case}
\end{figure}

Unlike standard object detection tasks, this diagnostic focuses on fine-grained spatial grounding: the model must correctly associate linguistic expressions (e.g., “the red king in the bottom left”) with discrete grid locations. We vary both \textit{preprompt strategies} (Helpful, Chain-of-Thought, Neutral) and \textit{reformulation types} (Declarative vs. Missing Word) to assess how instruction framing affects spatial accuracy. Figure~\ref{fig:chess_localization_vertical_case} presents an example showcasing the predictions of different models on a localization task sample.
\textbf{Key findings:}
\begin{itemize}
    \item \textbf{GPT-based models} (particularly GPT-4.1 and GPT-4.1-mini) demonstrate high spatial precision and uniform activation across grid cells, especially when prompted with Declarative + Helpful combinations.
    \item \textbf{Chain-of-Thought prompting} introduces variance in localization, often disrupting spatial alignment by triggering non-grounded reasoning pathways.
    \item \textbf{Open-source models} (e.g., LLaMA, Gemma) are more sensitive to instruction type. LLaMA variants exhibit degraded and inconsistent localization under blur or ambiguous phrasing; 
    Gemma shows highly noisy and scattered predictions across all prompting styles.
\end{itemize}

\begin{figure}[ht]
  \centering
  \includegraphics[width=0.30\linewidth]{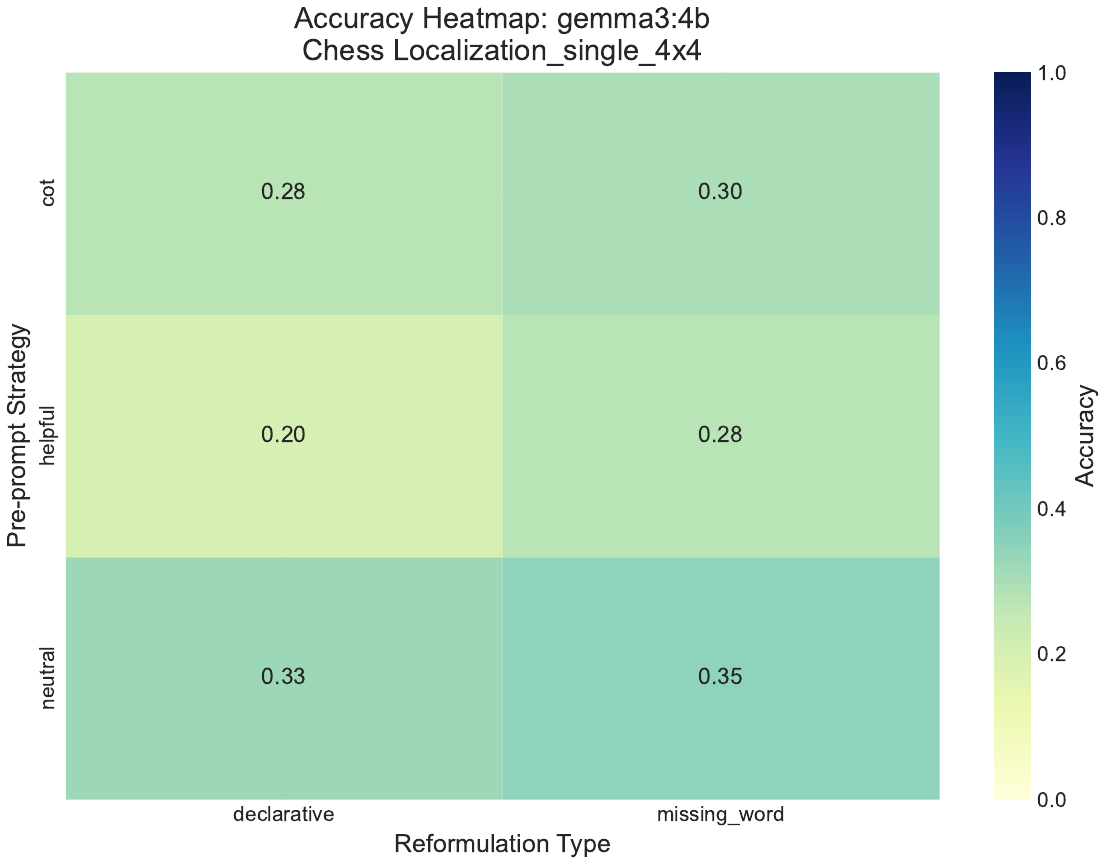}
  \includegraphics[width=0.30\linewidth]{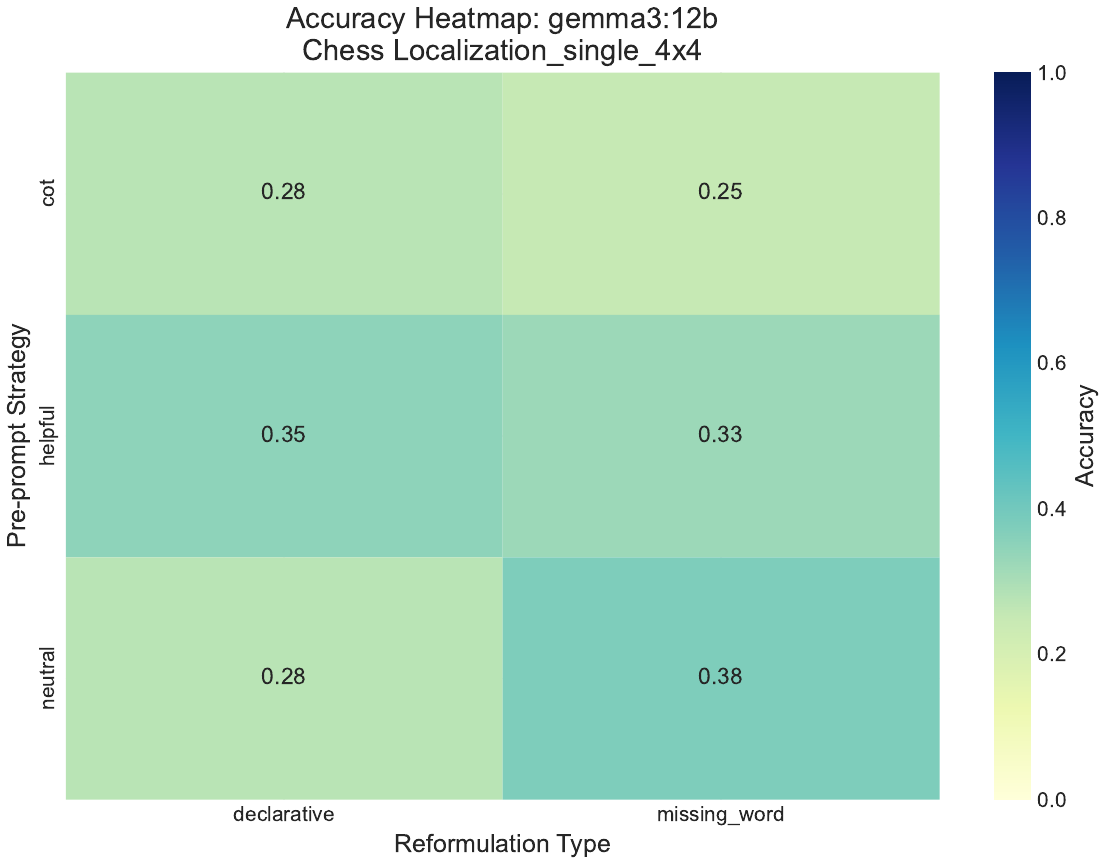}
  \includegraphics[width=0.30\linewidth]{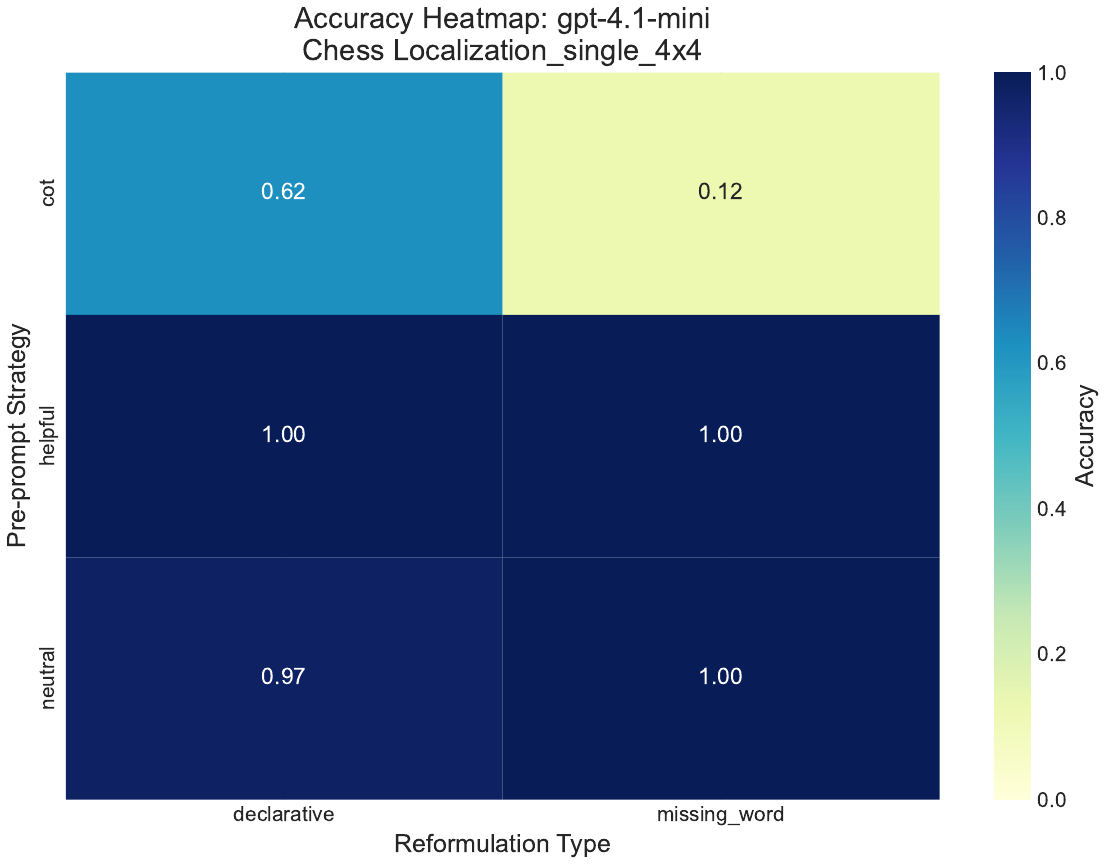}
  \\
  \includegraphics[width=0.30\linewidth]{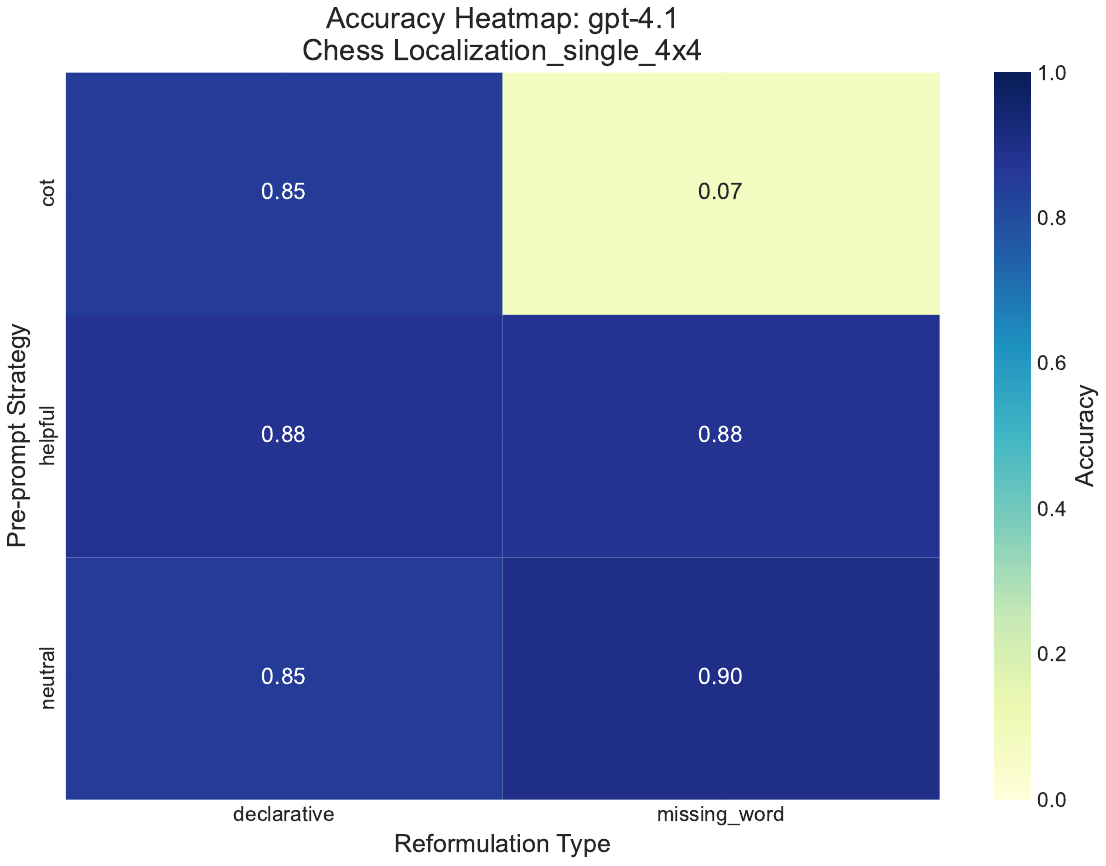}
  \includegraphics[width=0.30\linewidth]{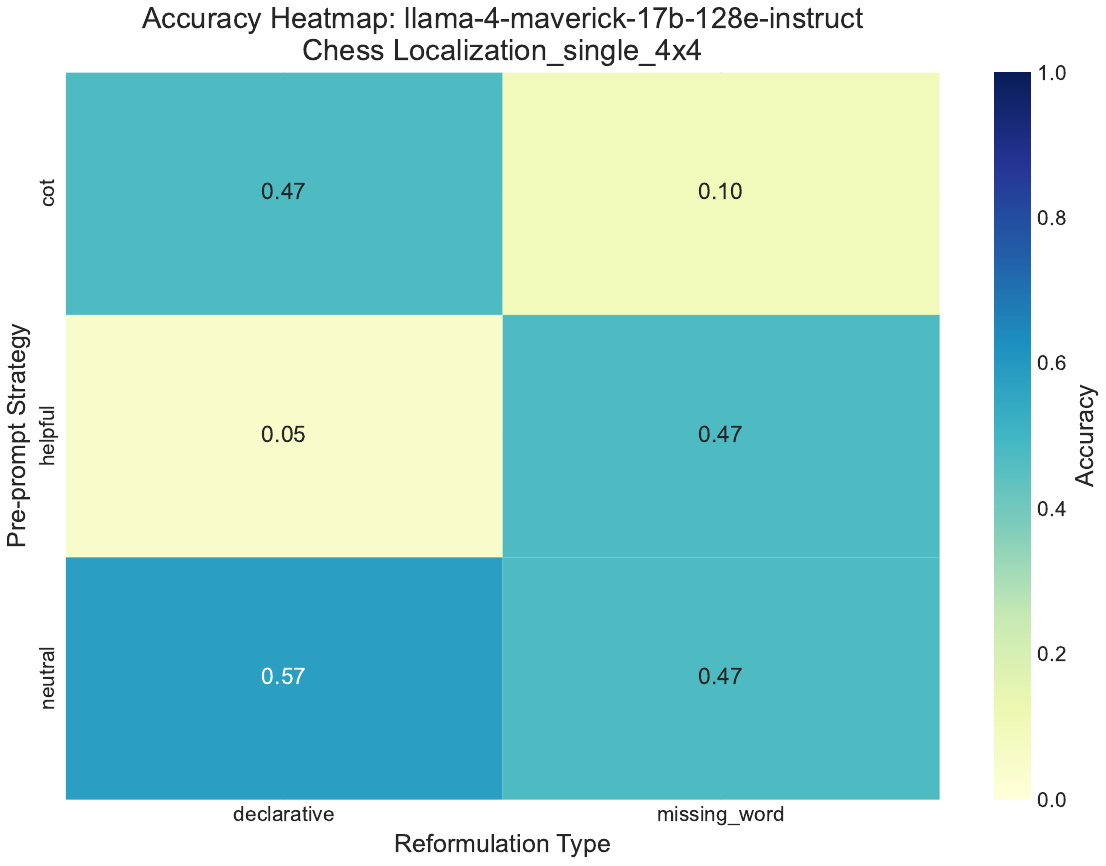}
  \includegraphics[width=0.30\linewidth]{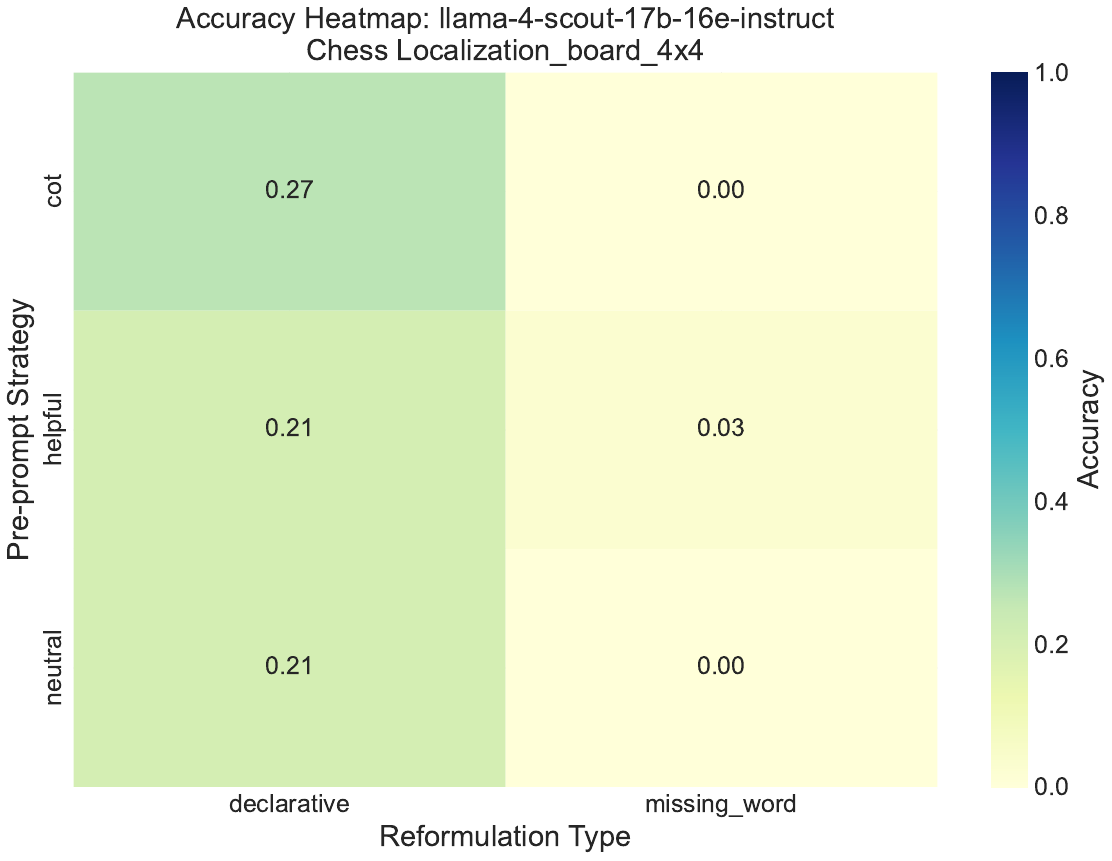}
  \\ 
  \includegraphics[width=0.30\linewidth]{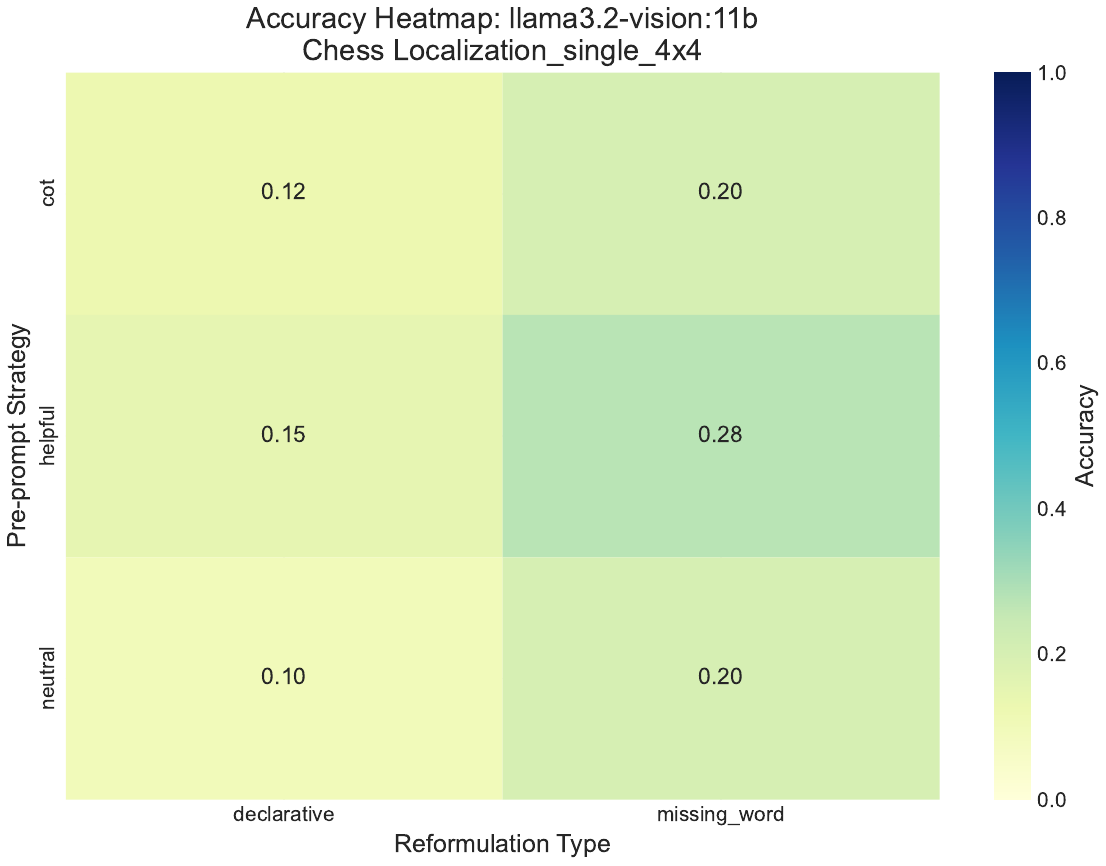}
  \includegraphics[width=0.30\linewidth]{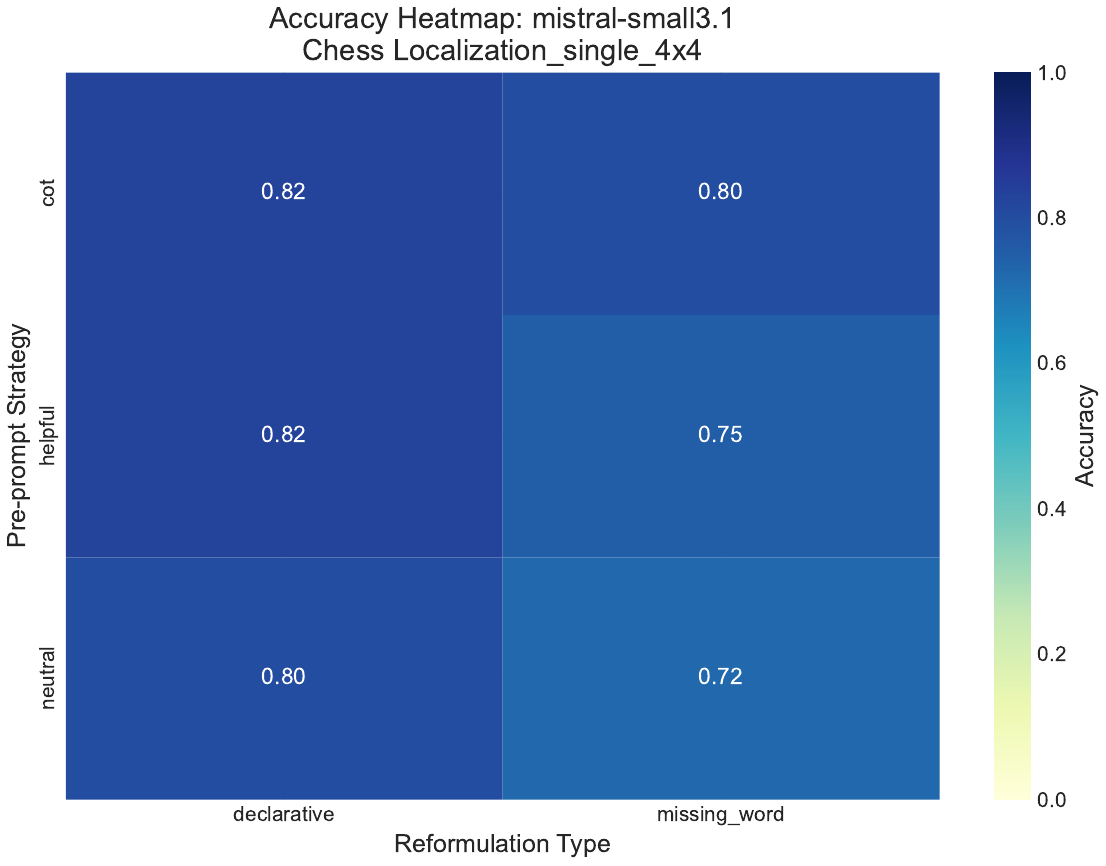}

  \caption{Heatmaps for the Localization Board 4x4 task across multiple models and prompting strategies, averaged over 50 synthetic chess scenes. Bright areas indicate accurate localization; darker areas highlight regions of frequent misprediction. This diagnostic reveals how well models spatially ground objects under minimal layout constraints.}
  \label{fig:localization_board_4x4_heatmap}
\end{figure}

The Figure~\ref{fig:localization_board_4x4_heatmap} illustrates the different results. 
Despite its smaller scale relative to GPT-4 models, \textbf{Mistral-small3.1} delivers surprisingly competitive performance on the 4×4 localization task. Its heatmaps reveal coherent spatial activation with low positional drift across most prompting conditions. While it does not achieve the spatial uniformity of GPT-based models, Mistral demonstrates reliable grounding under Declarative + Helpful prompts and maintains acceptable coverage even with Neutral inputs. However, mild instability emerges under Chain-of-Thought prompting, where reasoning steps appear to interfere with direct spatial mapping—occasionally resulting in diagonal mislocalizations. These results suggest that Mistral possesses a moderately strong spatial prior and can generalize across linguistic formats, though it remains more sensitive to prompt structure than instruction-tuned proprietary models.

By contrast, \textbf{LLaMA-4 Scout and Maverick} exhibit degraded and inconsistent spatial activation. Their performance is highly contingent on explicit prompt framing, with significant accuracy drops under less structured instructions. The \textbf{LLaMA-3.2 Vision} variant performs worst overall: its heatmaps frequently contain empty or incoherent activations, indicating a fundamental deficiency in visual-spatial alignment despite its multimodal architecture. 

\begin{table}[ht]
\centering
\caption{VLM scores over the relative localization task on the Chess dataset, with declarative instruction and helpful preprompt (↑: higher is better, ↓: lower is better. The grid is a $4{\times}4$ Board.}
\label{tab:localization_missing_word}
\renewcommand{\arraystretch}{1.1}
\setlength{\tabcolsep}{3pt}
%\scriptsize
\vspace{5pt}
\small	
\begin{tabular}{lccccccc}
\toprule
\textbf{Model} & \textbf{Acc↑} & \textbf{F1↑} & \textbf{Prec↑} & \textbf{Rec↑} & \textbf{MAE↓} & \textbf{MSE↓} & \textbf{NMAE↓} \\
\midrule
gpt-4.1         & \textbf{0.790} & \textbf{0.883} & \textbf{0.798} & \textbf{0.988} & \textbf{0.210} & \textbf{0.210} & \textbf{0.097} \\
gpt-4.1-mini    & 0.570 & 0.726 & 0.704 & 0.750 & 0.480 & 0.600 & 0.382  \\
LLaMA-4-scout   & 0.290 & 0.450 & 0.382 & 0.547 & 2.580 & 13.96 & 1.693  \\
LLaMA-4-maverick& 0.510 & 0.675 & 0.671 & 0.680 & 0.550 & 0.690 & 0.440  \\
gemma3          & 0.420 & 0.592 & 0.447 & 0.875 & 0.760 & 1.120 & 0.445  \\
gemma3:12b      & 0.170 & 0.291 & 0.283 & 0.298 & 1.320 & 2.520 & 0.947  \\
LLaMA3.2-Vision & 0.230 & 0.374 & 0.411 & 0.343 & 0.940 & 1.280 & 0.665  \\
mistral-3.1     & 0.540 & 0.701 & 0.675 & 0.730 & 0.520 & 0.660 & 0.382  \\
\bottomrule
\end{tabular}
\end{table}

%%%%%%%%%%%%%%%%%%%%%%%%%%%%%%%%%%%%%%%%%%%%%%%%%%%%%%%%%%%%%%%%%%%%%%%%%%%%%%%

\begin{figure}[h]
  \centering
    \centering
    \includegraphics[width=\columnwidth]{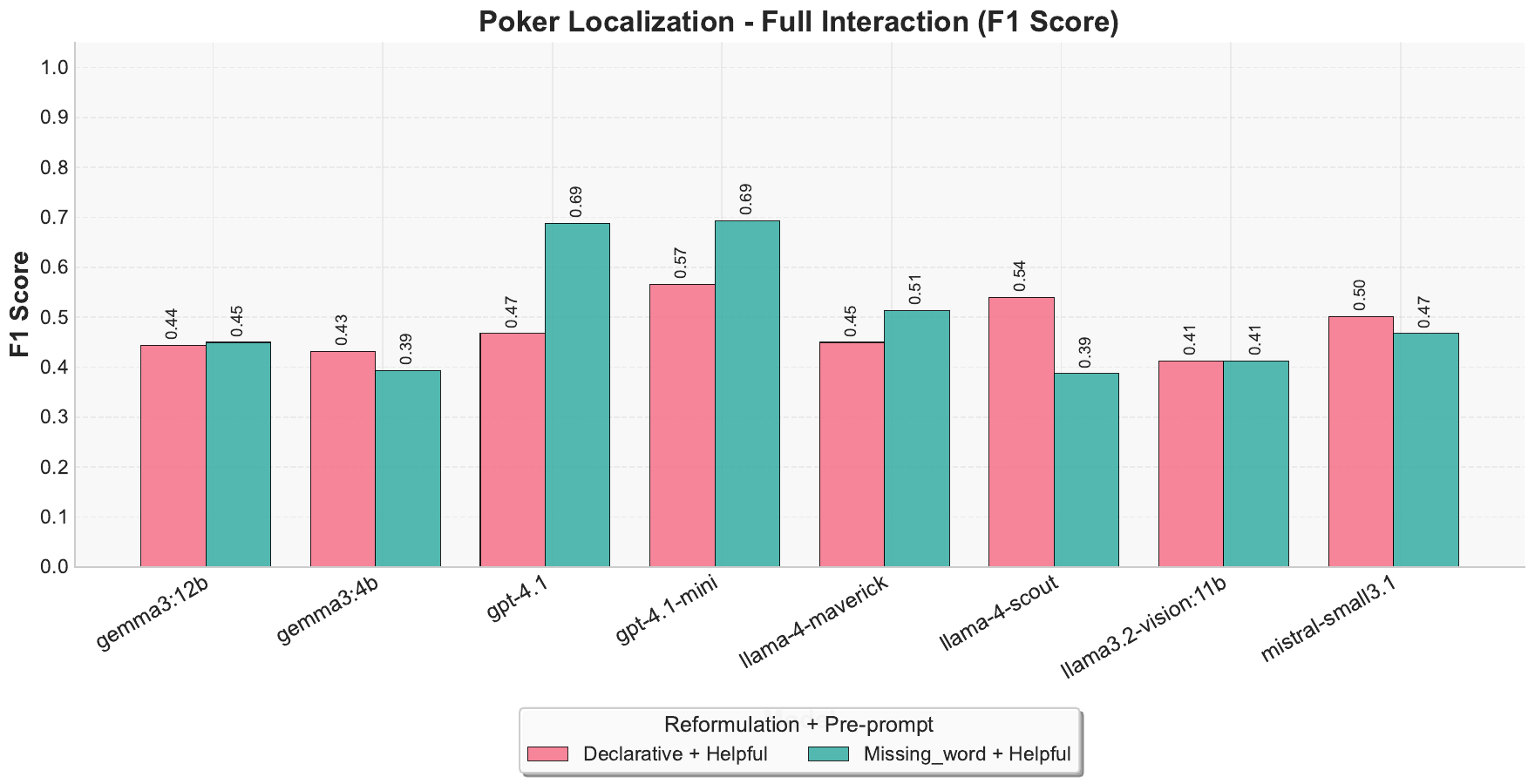}
  \caption{
     \textbf{F1 Score for Poker Localization Task across Prompt × Reformulation Combinations.}
    This figure visualizes the effect of linguistic scaffolding on model performance in the poker localization setting (3×3 grid). Bars represent F1 scores for each model under two prompting strategies: Declarative + Helpful and Missing Word + Helpful. GPT models (notably GPT-4.1 and GPT-4.1-mini) achieve the highest accuracy, while open-source models exhibit varying sensitivity to reformulation style.
  }
  \label{fig:combined_preprompt_poker}
\end{figure}

In Figure~\ref{fig:combined_preprompt_poker}, we focus on two representative prompting configurations: \textbf{Declarative + Helpful} and \textbf{Missing Word + Helpful}. This choice reflects prior findings across our evaluation suite, where Declarative + Helpful consistently yielded the highest spatial accuracy, while Missing Word + Helpful exposed prompt-induced variability.

Results on the poker localization task confirm this trend. GPT-based models (GPT-4.1 and GPT-4.1-mini) achieve top-tier F1 scores under both conditions, demonstrating robustness to reformulation style. Mistral-small3.1 performs reasonably well under explicit prompting but exhibits degradation with Missing Word formulations, suggesting limited resilience to implicit language structures.

In contrast, LLaMA variants particularly Scout and Vision experience sharp performance drops when explicit lexical cues are removed, reinforcing their reliance on surface prompt clarity. Gemma models perform weakest overall, struggling even under the most favorable conditions. These results emphasize that even in low-density visual environments, prompt design remains a dominant factor influencing spatial reasoning, especially for models with weaker instruction tuning.

%\begin{figure}[ht]
%  \centering
%    \centering
%    \includegraphics[width=\columnwidth]{figures/Appendix_C_barplots/poker/poker_count_blur_bar_f1_score.pdf}
%  \caption{ \blue{Add Caption}
%  }
%  \label{fig:}
%\end{figure}

%\begin{figure}[ht]
%  \centering
%    \centering
%    \includegraphics[width=\columnwidth]{figures/poker/poker_combined_preprompt_reformulation.pdf}
%  \caption{
%    Impact of the choice of preprompt and instructions on accuracy (mean over all poker tasks).
%  }
%  \label{fig:combined_preprompt_poker}
%\end{figure}

%\begin{figure}[ht]
%  \centering
%    \centering
%    \includegraphics[width=\columnwidth]{figures/Barplot/chess_preprompt_reformulation-3.pdf}
%  \caption{
%    Impact of the choice of preprompt and instructions on accuracy (mean over all Chess tasks).
%  }
%  \label{fig:combined_preprompt_poker}
%\end{figure}
\clearpage
\section{Detailed Framework}
\label{sec:detailed_framework}

This section provides a comprehensive description of the dataset generation framework developed for structured visual reasoning tasks, with a particular focus on chess and poker scenes. Our framework is designed for maximal modularity, reproducibility, and flexibility: it supports precise configuration of scene elements, systematic variation of experimental parameters, and rigorous annotation for downstream evaluation. We first detail the process for image dataset generation, from scene construction in Blender through to the integration of controlled noise and domain-specific logic for chess and poker. We then describe the formulation and handling of questions and answers, and finally the mechanisms for result retrieval and model evaluation.

\subsection{Image dataset generation}
\label{sec:image_dataset_generation}

The generation of image datasets in our framework is anchored in a modular, scriptable pipeline built on Blender. We begin by designing a scene configuration that specifies all key visual and physical parameters—ranging from object placement, camera setup, lighting, and materials, to controlled sources of variability and noise. Scene elements are modeled as strongly-typed configuration objects, ensuring both clarity and extensibility. The framework accommodates both procedural content generation (directly via Blender's Python API) and the integration of pre-designed assets, and it exposes multiple layers for customization—from low-level mesh manipulation to high-level experiment orchestration. The following subsections describe the main components of this pipeline, starting with the foundational capabilities and scripting interfaces provided by Blender.

\subsubsection{Blender in general}
\label{sec:blender_general}

Blender is a powerful, open-source 3D creation software that supports the full pipeline of 3D content production, including modeling, rigging, animation, simulation, rendering, compositing, and video editing. It features a physically based rendering engine advanced geometry manipulation tools, and support for scripting through a Python API (\texttt{bpy}).

\paragraph{bpy} Through \texttt{bpy}, one can either procedurally generate 3D content (e.g., meshes, lights, cameras) directly in Python or import pre-designed assets from \texttt{.blend} files, such as those downloaded from online asset libraries. The core data structure of Blender is organized around key modules such as \texttt{bpy.data} (for accessing all objects), \texttt{bpy.context} (for active elements), and \texttt{bpy.ops} (for operator-based actions, such as \texttt{bpy.ops.object}, \texttt{bpy.ops.mesh}, etc.). Scenes can be configured via \texttt{scene.render} settings (e.g., resolution, output format), while materials and shaders are managed using node trees (\texttt{material.node\_tree.nodes} and \texttt{...links}). Objects can be manipulated via their transforms (\texttt{location}, \texttt{rotation\_euler}, \texttt{scale}), and geometries can be edited through mesh operations. This modular and scriptable architecture allows for flexible automation of complex 3D workflows, enabling precise control over content generation.

\paragraph{bpy.data} 
The \texttt{bpy.data} module provides access to all data blocks in a Blender project, including objects, meshes, materials, cameras, lights, scenes, and more. These data blocks represent the raw components of a 3D scene and are not bound to any particular context. For example, \texttt{bpy.data.objects} retrieves the list of all objects in the project, while \texttt{bpy.data.materials} provides access to defined materials. This module is essential for querying, modifying, or creating persistent elements in a scene and is typically used when full control over the data structure is needed, independent of what is currently selected or active in the user interface.

\paragraph{bpy.context} 
The \texttt{bpy.context} module refers to the current state of the Blender interface and holds references to the active scene, selected objects, active camera, and more. Unlike \texttt{bpy.data}, which is global, \texttt{bpy.context} is transient and dynamic—useful when writing scripts that depend on user selection or UI actions. For example, \texttt{bpy.context.active\_object} accesses the object currently selected, allowing for context-sensitive manipulation such as transforming geometry or editing meshes in the right mode. Many Blender operators rely on context to function properly.

\paragraph{bpy.ops}
The \texttt{bpy.ops} module provides access to Blender's built-in operators, which are high-level actions that simulate interactive operations performed in the UI. This includes functionality for object manipulation (\texttt{bpy.ops.object}), mesh editing (\texttt{bpy.ops.mesh}), rendering, importing/exporting, and more. For example, \texttt{bpy.ops.mesh.primitive\_cube\_add()} creates a new cube in the scene. Operators are context-sensitive and often require that the relevant object or mode be active.

\paragraph{scene.render}
The \texttt{scene.render} settings control output and rendering configurations for a Blender scene. This includes image resolution, file format, frame rate, output file paths, and rendering engine parameters. For example, \texttt{scene.render.resolution\_x} and \texttt{scene.render.image\_settings.file\_format} are used to specify the dimensions and format (e.g., PNG, JPEG) of rendered images.

\paragraph{Materials and Node Trees} 
Materials in Blender define the visual appearance of surfaces and can range from simple color definitions to complex shader graphs. These materials are composed using node trees, accessible via \texttt{material.node\_tree.nodes} and \texttt{material.node\_tree.links}. Nodes define operations (e.g., shaders, textures, math functions), while links connect their inputs and outputs to form a graph that defines the material behavior. Materials can be assigned to objects via their data (e.g., \texttt{object.data.materials}) and dynamically created or modified via Python, enabling procedural control over appearance.

\paragraph{Object Transformation and Manipulation} 
Individual objects in Blender can be manipulated via their transformation attributes: \texttt{location}, \texttt{rotation\_euler}, and \texttt{scale}. These properties are vectors that define an object’s position, orientation, and size within the scene. For instance, setting \texttt{obj.location = (1.0, 0.0, 0.0)} translates the object along the x-axis. Objects also have a \texttt{data} field that links them to their geometric representation (e.g., mesh or camera) and can be duplicated, hidden, or animated.

\paragraph{Mesh Editing} 
Meshes are editable geometric structures consisting of vertices, edges, and faces. Mesh editing can be performed using both \texttt{bpy.ops.mesh} operators (e.g., extrude, subdivide) and direct manipulation via the mesh data structure (e.g., \texttt{obj.data.vertices}). The former is suitable for higher-level transformations, while the latter allows low-level procedural generation of geometry. Switching to edit mode and updating the mesh with \texttt{bpy.ops.object.mode\_set(mode='EDIT')} is often required for certain operations.

%%%%%%%%%%%%%%%%%%%%%%%%%%%%%%%%%%%%%%%%%%%%%%%%%%%%%%%%%%%%%%%%%%%%%
\subsubsection{Dataset General Setup}
\label{sec:dataset_general_setup}

\paragraph{Overview}
The dataset generation process is built on a modular and extensible framework that separates configuration, construction, and orchestration of all scene elements. This structure allows precise control, high reproducibility, and straightforward extensibility for synthetic data production.

\paragraph{Models Layer}
All scene parameters are structured as strongly-typed data classes (models) defined in \texttt{models.py} files. Each model captures the configuration and validation logic for a specific scene component.

\begin{itemize}
    \item \textbf{MaterialModel}: Defines appearance properties for surfaces.
    \begin{description}
      \item[\texttt{color}] Default: (0.8, 0.8, 0.8, 1.0). RGBA tuple or string (e.g., \texttt{"white"}).
      \item[\texttt{roughness}] Default: 0.5. Controls micro-surface scattering (0.0 = mirror, 1.0 = diffuse).
      \item[\texttt{material\_name}] Default: None. Optional string identifier for reuse or tracking.
      \item[\texttt{custom\_material}] Default: None. Optional Blender reference.
    \end{description}

    \item \textbf{CameraModel}: Controls camera position, orientation, and variability.
    \begin{description}
      \item[\texttt{distance}] Default: \texttt{"medium"}. Preset or float; distance from scene center.
      \item[\texttt{angle}] Default: \texttt{"medium"}. Preset or float; vertical angle (90° = top-down).
      \item[\texttt{horizontal\_angle}] Default: 0.0. Azimuthal rotation (0–360°).
      \item[\texttt{randomize\_distance}] Default: False. Toggles distance randomization.
      \item[\texttt{randomize\_distance\_percentage}] Default: 0.1. Range for distance variation.
      \item[\texttt{randomize\_angle}] Default: False. Toggles angle randomization.
      \item[\texttt{randomize\_angle\_percentage}] Default: 0.1. Range for angle variation.
    \end{description}

    \item \textbf{TableModel}: Configures the table supporting objects in the scene.
    \begin{description}
      \item[\texttt{shape}] Default: RECTANGULAR. Options: RECTANGULAR, CIRCULAR, ELLIPTIC.
      \item[\texttt{length}] Default: 2.0. Length along x-axis (in Blender units).
      \item[\texttt{width}] Default: 1.0. Width along y-axis.
      \item[\texttt{height}] Default: 0.9. Table height.
      \item[\texttt{texture}] Default: WOOD. Options: WOOD, MARBLE, METAL.
      \item[\texttt{material}] Nested \texttt{MaterialModel}. Defines surface appearance.
    \end{description}

    \item \textbf{FloorModel}: Configures the ground surface below the table.
    \begin{description}
      \item[\texttt{color}] Default: (0.8, 0.8, 0.8, 1.0). Floor base color.
      \item[\texttt{roughness}] Default: 0.5. Controls reflection properties.
      \item[\texttt{material}] Nested \texttt{MaterialModel}. Full material definition.
    \end{description}

    \item \textbf{BackgroundModel}: Controls the visual and lighting background of the scene.
    \begin{description}
      \item[\texttt{color}] Default: (0.5, 0.5, 0.5, 1.0). Used when HDRI is disabled.
      \item[\texttt{use\_hdri}] Default: False. Enables HDRI-based lighting.
      \item[\texttt{hdri\_path}] Default: None. File path to HDRI texture (if enabled).
    \end{description}

    \item \textbf{LightingModel}: Configures directional and ambient scene lighting.
    \begin{description}
      \item[\texttt{lighting}] Default: \texttt{"medium"}. Presets: very\_low (0.3) to very\_high (2.0), or float.
      \item[\texttt{key\_light\_power}] Default: 300.0. Primary light intensity.
      \item[\texttt{fill\_light\_power}] Default: 50.0. Fills shadows.
      \item[\texttt{back\_light\_power}] Default: 50.0. Enhances object separation.
    \end{description}

    \item \textbf{ResolutionModel}: Defines the output image resolution and quality.
    \begin{description}
      \item[\texttt{width}] Default: 1920. Output width (pixels).
      \item[\texttt{height}] Default: 1080. Output height (pixels).
      \item[\texttt{resolution\_percentage}] Default: 100. Renders at percentage of full resolution.
      \item[\texttt{pixel\_aspect\_x}, \texttt{pixel\_aspect\_y}] Default: 1.0. Pixel shape scaling.
      \item[\texttt{randomize}] Default: False. Toggles resolution variation.
      \item[\texttt{randomize\_percentage}] Default: 0.1. Range for randomized resolution.
    \end{description}
    Presets values:
    \begin{itemize}[nosep]
      \item \texttt{"low"}: 640$\times$480
      \item \texttt{"medium"}: 1280$\times$720
      \item \texttt{"high"}: 1920$\times$1080
    \end{itemize}

    \item \textbf{RenderModel}: Specifies render engine and output parameters.
    \begin{description}
      \item[\texttt{engine}] Default: \texttt{"CYCLES"}. Options: CYCLES, EEVEE, WORKBENCH.
      \item[\texttt{samples}] Default: 128. Number of render samples (higher = better quality).
      \item[\texttt{exposure}] Default: 0.0. Global brightness adjustment (in stops).
      \item[\texttt{file\_format}] Default: \texttt{"PNG"}. Options: PNG, JPEG, TIFF, EXR, etc.
      \item[\texttt{resolution}] Nested \texttt{ResolutionModel}. Render size configuration.
      \item[\texttt{output\_path}] Default: None. Output directory path.
      \item[\texttt{gpu\_enabled}] Default: True. Enables GPU rendering when available.
      \item[\texttt{gpus}] Default: None. GPU device indices (e.g., [0, 1]).
    \end{description}

    \item \textbf{SceneSetupModel}: Aggregates all sub-models into a single, hierarchical configuration.
\end{itemize}

\paragraph{Builder Layer (Scene Construction Logic)}
The construction of the scene is managed by a coordinated set of builder modules:
\begin{itemize}
    \item \texttt{general\_setup.py}: Orchestrates the end-to-end construction pipeline, calling individual builders in the correct order (scene clearing, resolution, render setup, environment, table, floor, camera, lighting).
    \item \texttt{camera.py}: Implements camera placement logic, including support for controlled randomization.
    \item \texttt{rendering.py}: Handles render engine configuration and scene cleanup.
    \item \texttt{resolution.py}: Defines and applies resolution presets or explicit overrides.
    \item Additional modules manage table, floor, background, and lighting setup.
\end{itemize}

\paragraph{Integration Layer (Pipeline Usage)}
The configuration for a scene is typically defined as a JSON or YAML dictionary, parsed and passed to the scene setup functions. The integration layer ensures:
\begin{itemize}
    \item Loading and validating configuration files.
    \item Sequential invocation of all builder modules to assemble the scene.
    \item Full serialization/deserialization for reproducibility.
    \item Support for parameter overrides and randomization (for experiment diversity).
\end{itemize}

An example of configuration is given below:

\begin{mycase}{light_grey}
\begin{lstlisting}[language=Python, label={lst:scene_config}]
config = {
    "resolution": {"resolution": "high"},
    "camera": {
        "distance": "medium",
        "angle": 60.0,
        "randomize_angle": True,
        "randomize_angle_percentage": 0.1
    },
    "table": {
        "shape": "rectangular",
        "length": 2.0,
        "width": 1.0,
        "material": {"color": (0.8, 0.7, 0.6, 1.0)}
    },
    "lighting": {"lighting": "high"}
}
\end{lstlisting}
\end{mycase}

%%%%%%%%%%%%%%%%%%%%%%%%%%%%%%%%%%%%%%%%%%%%%%%%%%%%%%%%%%%%%%%%%%%%%
\subsubsection{Dataset Noise}
\label{sec:dataset_noise}

\paragraph{Overview}
The \texttt{noise} module introduces systematic, configurable perturbations into the rendering pipeline to enhance dataset diversity and realism. It focuses on three main types of noise: (i) \emph{blur noise}, simulating camera depth-of-field effects; (ii) \emph{light noise}, introducing controlled variations in lighting intensity and direction; and (iii) \emph{table texture noise}, varying the complexity of the material applied to the table surface.

\paragraph{Models Layer (\texttt{noise/models.py})}
The configuration is built upon a set of strongly-typed dataclasses defined in \texttt{models.py}. A common abstract base, \texttt{BaseNoiseModel}, defines shared attributes across noise types, and specific subclasses extend this base:

\begin{itemize}
    \item \textbf{BaseNoiseModel}: Abstract base class for all noise types.
    \begin{description}
        \item[\texttt{enabled}] Default: \texttt{True}. Toggles whether the noise effect is applied.
        \item[\texttt{intensity}] Default: \texttt{1.0}. Float in range 0.0--2.0, defines effect strength.
        \item[\texttt{seed}] Default: \texttt{None}. Optional integer seed for reproducibility.
        \item[\texttt{blend\_mode}] Default: \texttt{"MIX"}. Blend type: \texttt{"MIX"}, \texttt{"ADD"}, \texttt{"MULTIPLY"}, etc.
        \item[\texttt{opacity}] Default: \texttt{1.0}. Float in range 0.0--1.0, determines visual strength of the effect.
    \end{description}

    \item \textbf{BlurNoiseModel}: Simulates depth-of-field blur by manipulating camera aperture (f-stop) settings.
    \begin{description}
        \item[\texttt{intensity}] Default: \texttt{"none"}. Accepts preset or float value.\\
        \textbf{Presets:}
        \begin{itemize}[nosep]
            \item \texttt{"none"}: No blur (disabled)
            \item \texttt{"very\_low"}: f/9.0
            \item \texttt{"low"}: f/4.0
            \item \texttt{"medium"}: f/2.0
            \item \texttt{"high"}: f/1.0
            \item \texttt{"very\_high"}: f/0.5
        \end{itemize}
    \end{description}

    \item \textbf{LightNoiseModel}: Applies global lighting variation while maintaining a balanced three-point setup.
    \begin{description}
        \item[\texttt{lighting}] Default: \texttt{"medium"}. Controls global brightness via scalar presets.\\
        \textbf{Presets:}
        \begin{itemize}[nosep]
            \item \texttt{"very\_low"}: 0.3$\times$
            \item \texttt{"low"}: 0.6$\times$
            \item \texttt{"medium"}: 1.0$\times$ (default)
            \item \texttt{"high"}: 1.5$\times$
            \item \texttt{"very\_high"}: 2.0$\times$
        \end{itemize}
        \item[Base values (medium):]
        \begin{itemize}[nosep]
            \item Key Light: 400
            \item Fill Light: 200
            \item Back Light: 300
        \end{itemize}
    \end{description}

    \item \textbf{TableTextureNoiseModel}: Adjusts the complexity of table surface textures using procedural shaders.
    \begin{description}
        \item[\texttt{table\_texture}] Default: \texttt{"medium"}. Controls texture entropy.\\
        \textbf{Presets:}
        \begin{itemize}[nosep]
            \item \texttt{"low"}: Monochrome surface (simple color + roughness)
            \item \texttt{"medium"}: Color noise and variation
            \item \texttt{"high"}: Multi-layer textures with bump mapping
        \end{itemize}
        \item[Color Palettes (low entropy):]
        \begin{itemize}[nosep]
            \item Light gray: (0.8, 0.8, 0.8, 1.0)
            \item Medium gray: (0.6, 0.6, 0.6, 1.0)
            \item Dark gray: (0.4, 0.4, 0.4, 1.0)
            \item Light wood: (0.8, 0.7, 0.6, 1.0)
            \item Medium wood: (0.6, 0.5, 0.4, 1.0)
            \item Dark wood: (0.4, 0.3, 0.2, 1.0)
        \end{itemize}
    \end{description}

    \item \textbf{NoiseConfigModel}: Serves as the master container for all noise-related configuration models.
    \begin{description}
        \item[\texttt{blur}] Optional \texttt{BlurNoiseModel} instance.
        \item[\texttt{light}] Optional \texttt{LightNoiseModel} instance.
        \item[\texttt{table\_texture}] Optional \texttt{TableTextureNoiseModel} instance.
    \end{description}
\end{itemize}

\paragraph{Builder and Implementation Layer}
Each noise type has a dedicated implementation module responsible for its operational logic:
\begin{itemize}
    \item \texttt{blur.py}: Sets depth-of-field parameters on the active camera. F-stop values define the blur intensity, and focus distance is automatically computed when not explicitly set.
    \item \texttt{light.py}: Applies a standardized three-point lighting configuration, scaling intensity uniformly across presets and maintaining key/fill/back ratios.
    \item \texttt{table\_texture.py}: Generates shader node trees with increasing entropy levels, from uniform color to layered procedural patterns and bump maps.
    \item Additional modules handle distractors or future noise types.
\end{itemize}

\paragraph{Integration Layer}
The module \texttt{set\_noise\_config.py} orchestrates the application of noise by parsing the unified configuration, instantiating relevant model classes, delegating to builder functions, and applying the results directly to the scene. The integration returns a composite summary of the applied noise elements and parameters for inspection or reuse.

\paragraph{Configuration System}
The noise system supports both human-readable and programmatic usage. Named presets such as \texttt{"low"}, \texttt{"medium"}, and \texttt{"high"} map to technical parameters (e.g., f-stop, light multiplier, entropy level). All configuration models provide explicit defaults and can be serialized into JSON or YAML.

An example of configuration is given below:

\begin{mycase}{light_grey}
\begin{lstlisting}[language=Python, label={lst:noise_config}]
noise_config = {
    "blur": "medium",      # 2.0 f-stop
    "lighting": "high",    # 150% intensity
    "table_texture": "low" # Simple monochrome surface
}
\end{lstlisting}
\end{mycase}

%%%%%%%%%%%%%%%%%%%%%%%%%%%%%%%%%%%%%%%%%%%%%%%%%%%%%%%%%%%%%%%%%%%%%
\subsubsection{Chess Base Content}
\label{sec:chess}

\paragraph{Overview}
The chess module provides specialized components for generating parametrizable chess boards and pieces in Blender. It supports both procedural and asset-based (e.g., \texttt{.blend} file) geometry and material assignment, with a focus on reproducibility, modularity, and flexibility for synthetic vision tasks.

\paragraph{Models Layer (\texttt{chess/models.py})}
The chess dataset generation is structured around two primary models: the \textbf{BoardModel} and the \textbf{PieceModel}, both implemented as dataclasses.

\begin{itemize}
    \item \textbf{BoardModel}: Configures the geometry, layout, and appearance of the chess board.
    \begin{description}
        \item[\texttt{length}] Default: 0.7. Board length.
        \item[\texttt{width}] Default: 0.7. Board width.
        \item[\texttt{thickness}] Default: 0.05. Thickness of the board.
        \item[\texttt{border\_width}] Default: 0.05. Width of the board border.
        \item[\texttt{location}] Default: (0, 0, 0.9). Board position in world coordinates.
        \item[\texttt{rows}] Default: 8. Number of rows (must be a power of 2).
        \item[\texttt{columns}] Default: 8. Number of columns (must be a power of 2).
        \item[\texttt{random\_pattern}] Default: False. Enables random square pattern.
        \item[\texttt{pattern\_seed}] Default: None. Seed for deterministic pattern generation.
        \item[\texttt{board\_material}] Nested \texttt{MaterialModel}. Material for board frame.
        \item[\texttt{white\_material}] Nested \texttt{MaterialModel}. Material for white squares.
        \item[\texttt{black\_material}] Nested \texttt{MaterialModel}. Material for black squares.
    \end{description}

    \item \textbf{PieceModel}: Configures geometry, placement, and material for a chess piece.
    \begin{description}
        \item[\texttt{piece\_type}] Type of piece (e.g., king, queen, rook, etc.).
        \item[\texttt{location}] Default: (0, 0, 0). World-space placement.
        \item[\texttt{material}] Nested \texttt{MaterialModel}. Controls appearance.
        \item[\texttt{geometry}] Nested \texttt{GeometryModel}. Scale, rotation, randomness.
    \end{description}

    \item \textbf{GeometryModel} (for pieces):
    \begin{description}
        \item[\texttt{scale}] Default: 0.1. Piece size scaling.
        \item[\texttt{random\_rotation}] Default: False. Enables random rotation.
        \item[\texttt{max\_rotation\_angle}] Default: 15.0. Maximum rotation angle (degrees).
        \item[\texttt{seed}] Optional integer for reproducibility.
    \end{description}

    \item \textbf{MaterialModel} (shared, see previous section): Color, roughness, material name/reference, and optional custom Blender material.
\end{itemize}

\paragraph{Board Builder Layer}
\begin{itemize}
    \item Implemented in \texttt{board.py}. Handles geometry generation, procedural or custom material assignment, pattern creation (checkerboard or random), and spatial tracking of all cell positions.
    \item Board is built in layers: frame $\rightarrow$ squares $\rightarrow$ collection assignment.
    \item Exposes methods to query or iterate over cell positions for piece placement.
    \item Includes validation (power-of-2 grid) and scene cleanup.
\end{itemize}

\paragraph{Piece Builder Layer}
\begin{itemize}
    \item Implemented in \texttt{factories.py}. Pieces are instantiated from configurations, either procedurally or by extracting meshes from \texttt{.blend} files.
    \item Abstract base factory, with implementations for:
    \begin{itemize}
        \item \textbf{DefaultPieceFactory}: Procedural.
        \item \textbf{OldSchoolPieceFactory}: Imports from \texttt{.blend} assets.
        \item \textbf{StonesColorPieceFactory}: Imports from \texttt{.blend} assets.
    \end{itemize}
    \item Piece classes encapsulate geometry creation, positioning, material assignment, and (if needed) mesh caching for performance.
\end{itemize}

\paragraph{Integration Layer}
The chess board and pieces are integrated within the main dataset generation scripts (e.g., \texttt{generate\_chess\_image.py}):
\begin{itemize}
    \item Instantiate \texttt{ChessBoard} with configuration, build board, track cell positions.
    \item For each piece, use cell position and piece configuration to instantiate and place with the chosen factory.
    \item Supports switching styles by factory selection, and parameterization by passing different configs.
\end{itemize}

An example of board configuration is given below:

\begin{mycase}{light_grey}
\begin{lstlisting}[language=Python, label={lst:board_config}]
board_config = {
    "length": 0.7,
    "width": 0.7,
    "thickness": 0.05,
    "location": (0, 0, 0.9),
    "border_width": 0.05,
    "rows": 8,
    "columns": 8,
    "random_pattern": False,
    "board_material": {
        "color": (0.4, 0.3, 0.2, 1.0),
        "roughness": 0.5
    },
    "white_material": {
        "color": (0.9, 0.9, 0.9, 1.0),
        "roughness": 0.3
    },
    "black_material": {
        "color": (0.1, 0.1, 0.1, 1.0),
        "roughness": 0.3
    }
}
\end{lstlisting}
\end{mycase}

An example of piece configuration is given below:

\begin{mycase}{light_grey}
\begin{lstlisting}[language=Python, label={lst:piece_config}]
piece_config = {
    "type": "king",
    "location": (0, 4),
    "color": (0.9, 0.9, 0.9, 1.0),
    "scale": 0.08,
    "random_rotation": True,
    "max_rotation_angle": 10.0,
    "roughness": 0.3,
    "material_name": "KingMaterial"
}
\end{lstlisting}
\end{mycase}

%%%%%%%%%%%%%%%%%%%%%%%%%%%%%%%%%%%%%%%%%%%%%%%%%%%%%%%%%%%%%%%%%%%%%
\subsubsection{Chess Generation Logic}
\label{sec:chess_generation_logic}

\paragraph{Overview}
The chess dataset generation system orchestrates the production of diverse and configurable chess scene images. It is built around a layered, modular architecture supporting both high-level experiment definition and low-level control over piece, board, and noise configurations.

\paragraph{Core Components}

\begin{itemize}
    \item \textbf{Dataset Configuration}: High-level \texttt{yaml} files (\texttt{dataset\_configs/*.yml}) define global experiment parameters and variable selection strategies.
    \item \textbf{Variable Management}: \texttt{generate\_dataset.py} processes variable definitions and manages generation of configuration combinations.
    \item \textbf{Advanced Models}: \texttt{advanced\_models.py} includes data models for piece count, type, and position, enabling expressive and precise dataset definition.
    \item \textbf{Configuration Generation}: \texttt{advanced\_generator.py} provides generators to build concrete board and piece configurations from high-level specs.
    \item \textbf{Image Generation}: \texttt{generate\_img\_from\_yaml.py} instantiates configurations, selects piece style factories, and coordinates the rendering pipeline.
\end{itemize}

\subparagraph{Dataset Configurations}
\texttt{yaml} files specify the experimental setup, with explicit variable control and experiment metadata. For example:

\begin{mycase}{light_grey}
\begin{lstlisting}[]
dataset:
  name: chess_identification
  output_dir: ../data/chess_ident_dataset
  seed: 42
  piece_set: old_school

variables:
  chess.count_config:
    variate_type: fixed
    variate_levels:
      type: fixed
      value: 3
      randomization: false

  chess.type_config:
    variate_type: varying_random
    variate_levels: [pawn, rook, knight, bishop, queen, king]
    n_images: 10

  noise.blur:
    variate_type: varying_all
    variate_levels: [none, very_low, low, medium, 
                      high, very_high]
    n_images: 5
\end{lstlisting}
\end{mycase}

Variable types supported:
\begin{itemize}
    \item Fixed (\texttt{fixed}): Single constant value
    \item All possible values (\texttt{varying\_all}): Test all enumerated values
    \item Random selection (\texttt{varying\_random}): Sample randomly from a list
    \item Range (\texttt{varying\_among\_range}): Sample from a continuous range
\end{itemize}

\subparagraph{Variable Management}
Variable handling is managed by data classes and utility classes:

\begin{mycase}{light_grey}
\begin{lstlisting}[language=Python]
@dataclass
class VariableConfig:
    variate_type: str
    variate_levels: Union[Any, List[Any], Tuple[Any, Any]]
    n_images: int = 1
    randomize: bool = False
    randomize_percentage: float = 0.2
\end{lstlisting}
\end{mycase}

Core utility classes:
\begin{itemize}
    \item \textbf{VariableCombinationGenerator}: Builds all combinations of configuration parameters from high-level experiment definitions.
    \item \textbf{DatasetConfig} and \textbf{ConfigConverter}: Aggregate and convert variables into structured configurations for downstream generators.
\end{itemize}

\subparagraph{Advanced Models (\texttt{advanced\_models.py})}

\textbf{PieceCountModel}: Specifies how many pieces to place on the board.
\begin{mycase}{light_grey}
\begin{lstlisting}[language=Python]
@dataclass
class PieceCountModel:
    spec_type: CountSpecificationType = \
                CountSpecificationType.PRESET
    preset: str = "medium"
    count: int = 10
    min_count: Union[int, str] = 5
    max_count: Union[int, str] = 15
\end{lstlisting}
\end{mycase}

\textbf{PieceTypeModel}: Specifies which piece types to use.
\begin{mycase}{light_grey}
\begin{lstlisting}[language=Python]
@dataclass
class PieceTypeModel:
    spec_type: PieceTypeSpecification = \
                PieceTypeSpecification.PRESET
    preset: str = "medium"
    types: List[str] = field(default_factory=list)
    n_types: int = 3
\end{lstlisting}
\end{mycase}

\textbf{PiecePosition}: Controls spatial distribution and placement constraints.
\begin{mycase}{light_grey}
\begin{lstlisting}[language=Python]
@dataclass
class PiecePosition:
    allowed_positions: List = field(default_factory=list)
    spread_level: SpreadLevel = SpreadLevel.MEDIUM
    start_point: Union[StartingPoint, Tuple] = \
                               StartingPoint.CENTER
\end{lstlisting}
\end{mycase}

\subparagraph{Configuration Generation (\texttt{advanced\_generator.py})}
Generators for each major component:
\begin{itemize}
    \item \textbf{PieceCountGenerator}, \textbf{PieceTypeGenerator}, \textbf{PiecePositionGenerator}: Convert high-level specs into detailed configurations.
    \item \textbf{ChessConfigGenerator}: Orchestrates board and piece config creation.
\end{itemize}

\subparagraph{Image Generation (\texttt{generate\_img\_from\_yaml.py})}
The \texttt{ChessImageGenerator} class connects configuration and rendering:

\begin{mycase}{light_grey}
\begin{lstlisting}[language=Python]
class ChessImageGenerator:
    def __init__(self, config_path=None, 
                 config_dict=None, piece_set=None):
        # Initialize with config source...

    def load_config(self):
        # Load and validate configuration...

    def _create_chess_config(self):
        # Process chess-specific configuration...

    def generate_image(self, output_dir=None,
                       base_filename="chess_scene"):
        # Generate the chess image...
\end{lstlisting}
\end{mycase}

\paragraph{Configuration and Execution Flow}
\begin{enumerate}
    \item \textbf{Define dataset configuration} (YAML): Select variables, types, values, and randomization.
    \item \textbf{Generate variable combinations}: VariableCombinationGenerator produces exhaustive or sampled experimental setups.
    \item \textbf{Build chess configuration}: ChessConfigGenerator converts high-level variables into board and pieces configs.
    \item \textbf{Generate images}: ChessImageGenerator runs the rendering, using selected style factories and noise configs.
    \item \textbf{Trace metadata}: Outputs both images and accompanying legend files for reproducibility.
\end{enumerate}

\paragraph{Example Application Scenarios}
\begin{itemize}
    \item \textbf{Piece Identification Tasks}: Varying piece types with controlled counts
    \begin{lstlisting}[]
chess.type_config:
  variate_type: varying_all
  variate_levels: [pawn, rook, knight, bishop, queen, king]
    \end{lstlisting}

    \item \textbf{Position Localization Tasks}: Distributing pieces with different spread patterns
    \begin{lstlisting}[]
chess.position_config.spread_level:
  variate_type: varying_all
  variate_levels: [low, medium, high]
    \end{lstlisting}

    \item \textbf{Visual Robustness Testing}: Adding controlled noise to images
    \begin{lstlisting}[]
noise.blur:
  variate_type: varying_all
  variate_levels: [none, low, medium, high]
    \end{lstlisting}

    \item \textbf{Count Estimation Tasks}: Varying the number of pieces
    \begin{lstlisting}[]
chess.count_config:
  variate_type: varying_among_range
  variate_levels: [5, 25]
    \end{lstlisting}
\end{itemize}

\noindent
This sophisticated architecture enables researchers to create precisely controlled, reproducible, and diverse chess image datasets for training and evaluating computer vision models on tasks like object identification, counting, and localization.
%%%%%%%%%%%%%%%%%%%%%%%%%%%%%%%%%%%%%%%%%%%%%%%%%%%%%%%%%%%%%%%%%%%%%

%%%%%%%%%%%%%%%%%%%%%%%%%%%%%%%%%%%%%%%%%%
\subsubsection{Poker Generation Logic}
\label{sec:poker_generation_logic}

\paragraph{Overview}
The poker dataset generation framework closely mirrors the modular architecture of the chess system. It consists of layered components for configuration, variable management, and scene construction. However, the poker framework introduces domain-specific logic for player management, river (community cards) handling, and flexible card/chip distribution.

\paragraph{Core Architecture}

\begin{itemize}
    \item \textbf{Dataset Configuration Layer}: High-level YAML files (\texttt{dataset\_configs/*.yml}) define global parameters and variable strategies.
    \item \textbf{Variable Management Layer}: (\texttt{generate\_dataset.py}) Generates all combinations of experimental variables, tracking metadata for reproducibility.
    \item \textbf{Scene Generation Layer}: (\texttt{scene\_generator.py}) Systematically builds up poker scenes according to input configuration.
    \item \textbf{Component Building Layer}: Modular builders for cards, players, chip stacks, and community (river) cards.
\end{itemize}

\paragraph{Configuration System}
As with chess, the configuration is driven by YAML files specifying the dataset parameters and experimental variables. For example:

\begin{mycase}{light_grey}
\begin{lstlisting}[]
dataset:
  name: poker_chip_variations
  output_dir: /home/...
  seed: 5678

variables:
  n_players:
    variate_type: fixed
    variate_levels: 4

  card_distribution_inputs.overall_cards:
    variate_type: varying_all_range
    variate_levels: [2, 15]
    n_images: 2

  chip_distribution_inputs.color_options:
    variate_type: varying_among
    variate_levels: [ [1.0, 0.0, 0.0, 1.0], [0.0, 0.0, 1.0, 1.0]]
\end{lstlisting}
\end{mycase}

The system supports variable types such as \texttt{fixed}, \texttt{varying\_all}, \texttt{varying\_random}, and \texttt{varying\_all\_range}, for both card and chip configurations, as well as camera and noise parameters.

\paragraph{Poker-Specific Logic}
While the general workflow is analogous to chess, poker dataset generation introduces:

\begin{itemize}
    \item \textbf{Player Logic}: Each player receives a configurable hand of cards and chip stacks, with control over card selection, arrangement, face-up/face-down mix, and chip pile details.
    \item \textbf{River (Community) Logic}: The framework allows specification of board/river cards, including precise arrangement, card identities, and spread.
    \item \textbf{Card Distribution Logic}: Flexible mechanisms for dealing cards to players and the river, including control over overlap, layout (horizontal, vertical, auto), and visual spread.
\end{itemize}

\noindent Example of player and river configuration:
\begin{mycase}{light_grey}
\begin{lstlisting}[language=Python]
{
  'player_id': 'Alice',
  'hand_config': {
      'card_names': ['AS', 'AH', 'AD', 'AC', '2S'],
      'n_cards': 5,
      'location': (-0.6, 0.0, table_height + 0.01),
      'scale': 0.1,
      'spread_factor_h': 0.2,
      'spread_factor_v': 0.05,
      'n_verso': 0,
      'random_seed': 101
  },
  'chip_area_config': {
      "base_pile_config": {
          "n_chips": 8,
          "base_chip_config": {"chip_object_name": "Cylinder001",
                               "scale": 0.06, 
                               "color": (0.1, 0.2, 0.8, 1)},
          "spread_factor": 0.1
      },
      "n_piles": 2,
      "n_chips_per_pile": [8, 10],
      "pile_colors": [None, (0.2, 0.8, 0.2, 1)],
      "pile_spreads": [None, 0.3],
      "random_seed": 1001
  }
}

"community_cards": {
    'card_names': ['4C', '4H', '4D', '4S', '5C'],
    'n_cards': 5,
    'start_location': (-0.3, 0, 0.9 + 0.01),
    'scale': 0.1,
    'n_verso': 0,
    'card_gap': {'base_gap_x': 0.15, 'base_gap_y': 0.005,
                 'random_gap': False}
}
\end{lstlisting}
\end{mycase}

\paragraph{Key Features and Variations}
\begin{itemize}
    \item \textbf{Flexible Player Setup}: Any number of players, each with distinct hands and chip stacks.
    \item \textbf{Customizable Card Layouts}: Support for horizontal, vertical, or auto layout, and explicit overlap/spread control.
    \item \textbf{Community/River Cards}: Positioning and spread of board cards, face-up or face-down.
    \item \textbf{Chips}: Color, pile count, and scale randomized or specified per player.
    \item \textbf{Noise, Camera, and Lighting}: Full control over noise (e.g., blur), camera angles, and lighting as in chess.
\end{itemize}

\paragraph{Summary}
In summary, the poker dataset generation logic generalizes the chess framework to card-based games, with additional modules for modeling players, community cards, and complex card/chip arrangements. The design enables researchers to create richly annotated, reproducible datasets for AI tasks such as card identification, counting, and arrangement analysis.

%%%%%%%%%%%%%%%%%%%%%%%%%%%%%%%%%%%%%%%%%%%%%%%%%%%%%%%%%%%%%%%%%%%%%
%%%%%%%%%%%%%%%%%%%%%%%%%%%%%%%%%%%%%%%%%%%%%%%%%%%%%%%%%%%%%%%%%%%%%
\subsection{Questions}
\label{sec:questions}

The question system provides a structured, extensible framework for generating and formatting questions to assess visual reasoning abilities on chess and poker scenes. This system supports a wide spectrum of cognitive tasks, question formulations, preprompt strategies, and instruction combinations.

\subsubsection{Main content}
\label{sec:question_main_content}

Questions are defined per domain in a central \texttt{all\_questions.json} file, covering both chess and poker. Each game includes a rich variety of question types, covering counting, identification, localization, and compositional reasoning.

\paragraph{Categories and Examples}
\begin{itemize}
    \item \textbf{Counting}:\\
    \texttt{count\_pieces} --- ``How many pieces are there in the image?''\\
    \texttt{count\_total\_cards} --- ``How many cards are present in the entire scene (including all hands and community cards)?''
    \item \textbf{Identification}:\\
    \texttt{identify\_type\_one\_piece} --- ``What piece is on the board?''\\
    \texttt{identify\_cards} --- ``What are the cards on the table?''
    \item \textbf{Localization}:\\
    \texttt{localize\_column\_one\_piece} --- ``On which column is the piece on the board?''\\
    \texttt{count\_community\_cards} --- ``How many community cards are visible on the table?''
    \item \textbf{Combined/Complex}:\\
    \texttt{count\_identification\_white\_pieces} --- ``How many white pieces are there on the board?''\\
    \texttt{count\_identify\_face\_up\_cards} --- ``How many cards are face up on the table?''
\end{itemize}

Questions are selected and instantiated by key (e.g., \texttt{count\_pieces}), with category and format determined by the handler logic.

\subsubsection{preprompts}
\label{sec:question_preprompts}

preprompts are optional instructional phrases prepended to the main question to guide model reasoning or control for dataset biases. Supported preprompts include:

\begin{itemize}
    \item \textbf{Debiased:}\\
    \textit{Chess}: ``This is not a real chess game. The number of each piece and their position can vary arbitrarily. Just focus on answering the following question based on the visual content.''\\
    \textit{Poker}: ``This is not a real poker game. The number of each card and their position can vary arbitrarily. Just focus on answering the following question based on the visual content.''
    \item \textbf{Chain-of-Thought (CoT):}\\
    ``First, think carefully, step by step, about the question being asked and the relevant elements of the image to which the question refers. End your message with \{answer : <answer>\}''
    \item \textbf{Debiased CoT:}\\
    Combines both debiased and chain-of-thought instructions.
\end{itemize}

preprompts can be flexibly combined with any question or instruction format. For example:

\begin{quote}
    \textit{Debiased CoT for chess}:\\
    ``This is not a real chess game. The number of each piece and their position can vary arbitrarily. Just focus on answering the following question based on the visual content. First, think carefully, step by step, about the question being asked and the relevant elements of the image to which the question refers. End your message with \{answer : <answer>\}''
\end{quote}

\subsubsection{Instructions}
\label{sec:question_instructions}

We can automatically add to the main question instructions to obtain various answer styles. The system provides:

\begin{itemize}
    \item \textbf{Direct question} (default):\\
    ``How many pieces are there in the image?''
    \item \textbf{Declarative statement}:\\
    ``The number of pieces in the image is:''
    \item \textbf{Fill-in-the-blank}:\\
    ``There are \_\_\_\_ pieces in the image.''
\end{itemize}

\subsubsection{Overall (combinations)}
\label{sec:question_overall}

The system enables systematic generation of question variants for robust evaluation. For each question, the handler can:

\begin{itemize}
    \item Select domain (chess or poker)
    \item Specify question key(s) or task category
    \item Choose instruction format (direct, declarative, fill-in-blank)
    \item Add any preprompt(s) (debiased, CoT, debiased\_cot)
    \item Optionally substitute variables (e.g., position X,Y) from image metadata
\end{itemize}

\paragraph{Example Combinations}

\begin{itemize}
    \item \textbf{Basic, direct:}\\
    ``How many pieces are there in the image?''
    \item \textbf{With CoT preprompt:}\\
    ``First, think carefully, step by step, about the question being asked and the relevant elements of the image to which the question refers. How many pieces are there in the image?''
    \item \textbf{Declarative, with debiased prompt:}\\
    ``This is not a real poker game. The number of each card and their position can vary arbitrarily. Just focus on answering the following question based on the visual content. The number of cards present in the entire scene (including all hands and community cards) is:''
    \item \textbf{Fill-in-blank with CoT:}\\
    ``First, think carefully, step by step, about the question being asked and the relevant elements of the image to which the question refers. There are \_\_\_\_ cards in the entire scene (including all hands and community cards).''
\end{itemize}

%%%%%%%%%%%%%%%%%%%%%%%%%%%%%%%%%%%%%%%%%%%%%%%%%%%%%%%%%%%%%%%%%%%%%
\subsection{Legend and answers}
\label{sec:legend_answers}

\subsubsection{Legends}
\label{sec:legends}

The legend generation system documents key metadata for each generated image, supporting traceability, dataset evaluation, and model training. Chess and poker domains use parallel architectures, adapted for their respective scene elements.

\paragraph{Process Overview}

\begin{enumerate}[leftmargin=2em]
    \item \textbf{Convert Configurations}: Parse internal configuration dictionaries (scene, board/table, pieces/cards, camera, noise) into a unified intermediate structure.
    \item \textbf{Build Hierarchical Legends}: Construct structured dictionaries capturing all relevant parameters and object details.
    \item \textbf{Format Legends}: Output both human-readable text files and machine-readable JSON files.
    \item \textbf{Write to Disk}: Save legends alongside rendered images for downstream use.
\end{enumerate}

\paragraph{Chess Legends}

\begin{itemize}
    \item \textbf{Board}: Dimensions, pattern, position, material, and colors
    \item \textbf{Pieces}: Type, position (board/world), color, scale, and other properties
    \item \textbf{Camera}: Distance, angles, world coordinates
    \item \textbf{Noise}: Blur, lighting, table texture
\end{itemize}

\paragraph{Poker Legends}

\begin{itemize}
    \item \textbf{Scene Setup}: Table, camera, lighting, render parameters
    \item \textbf{Cards}: Player hands (cards, position, face-up/down), community cards, layout
    \item \textbf{Players}: IDs, hand details, chip stacks
    \item \textbf{Noise}: Visual effects applied
\end{itemize}

\paragraph{Example Output (Excerpt)}
\begin{quote}
\texttt{CHESS PIECES LEGEND}\\
KING PIECES (2):\\
- KING\_1: Board Position: row 0, col 4; Color: RGBA(0.9, 0.9, 0.9, 1.0); Scale: 0.08\\

\texttt{COMMUNITY CARDS (POKER):}\\
Cards: 7D, 3C, 9H\\
Player: Player\_1; Hand Cards: JC, KS
\end{quote}

\paragraph{Key Features}

\begin{itemize}
    \item \textbf{Dual Output}: Text for readability, JSON for automation
    \item \textbf{Comprehensive}: Covers all scene parameters and objects
    \item \textbf{Domain-Specific}: Adapted for both chess and poker elements
    \item \textbf{Error Robust}: Handles missing or invalid data gracefully
\end{itemize}

\vspace{1em}
This system ensures every image is accompanied by a complete, interpretable record, supporting reliable evaluation of visual reasoning tasks.

\subsubsection{Answer extraction}
\label{sec:answer_extraction}

The \texttt{AnswerExtractor} class implements a principled, automated approach for extracting ground-truth answers from structured image legends. This process is fundamental for evaluating visual question answering (VQA) systems, as it guarantees alignment between the generated images, the underlying parameters, and the target answers used for benchmarking.

\paragraph{Core Extraction Mechanism.}
At its core, the extractor operates on JSON-formatted legends generated during image synthesis. The main extraction method is driven by the \textit{question key} and \textit{game type}, optionally leveraging the full question text for context. Upon receiving a legend dictionary, the extractor first routes the request according to the question type, applies any necessary context-aware logic (e.g., parsing the question to identify parameters like piece or suit), and formats the output in the appropriate structure (string, list, or dictionary). This systematic approach supports a wide range of question and answer formats.

\paragraph{Domain-Specific Logic: Chess.}
For chess images, the extractor covers a variety of tasks, including:
\begin{itemize}
    \item \textbf{Simple Counting:} Returns the number of pieces present.
    \item \textbf{Board Properties:} Computes the number of squares from board dimensions.
    \item \textbf{Piece Localization:} Retrieves the row or position of a given piece.
    \item \textbf{Piece Identification:} Identifies the type of a single piece.
    \item \textbf{Spatial Reasoning:} Calculates the spatial distance (e.g., rows apart) between pieces.
    \item \textbf{Color-Based Questions:} Counts the number of pieces by color (e.g., white pieces).
\end{itemize}
Each question type is mapped to dedicated extraction logic, ensuring accurate and reliable retrieval from nested legend structures.

\paragraph{Domain-Specific Logic: Poker.}
For poker images, the extraction logic is adapted to card game semantics and includes:
\begin{itemize}
    \item \textbf{Card Counting:} Aggregates the total number of cards by summing community, player, and overlap cards.
    \item \textbf{Card Identification:} Lists the community cards or retrieves specific cards shown.
    \item \textbf{Player-Based Questions:} Determines which player has the most cards in hand.
    \item \textbf{Attribute-Based Questions:} Parses the question text (e.g., via regex) to count cards matching a target suit or attribute.
\end{itemize}
This flexible, parameter-driven logic ensures coverage of the broad spectrum of poker VQA tasks.

%%%%%%%%%%%%%%%%%%%%%%%%%%%%%%%%%%%%%%%%%%%%%%%%%%%%%%%%%%%%%%%%%%%%%
\subsection{Result retrieval}
\label{sec:result_retrieval}

For each visual reasoning task, model predictions are retrieved using a dedicated function that interfaces with the selected Vision Language Model (VLM) provider. For every image-question pair, the system submits both the image and the formulated question to the VLM via an API call. This process is encapsulated in the central function \texttt{evaluate\_VLM\_on\_task()}, which handles provider-specific formatting, request dispatch, and response parsing. The returned prediction may take various forms depending on the provider; hence, post-processing is applied to extract the actual answer text. For instance, if the response is a structured dictionary, the answer is extracted from the relevant field. In cases where numeric values are expected, regular expressions are used to robustly extract numerical information from free-form model outputs. All predictions are collected and stored for downstream evaluation, including accuracy calculation and error analysis.

\section{Correlation between synthetic and real data}
\label{sec:correlation_synthetic_real}

\begin{figure}[H]
    \centering
    \includegraphics[width=\linewidth]{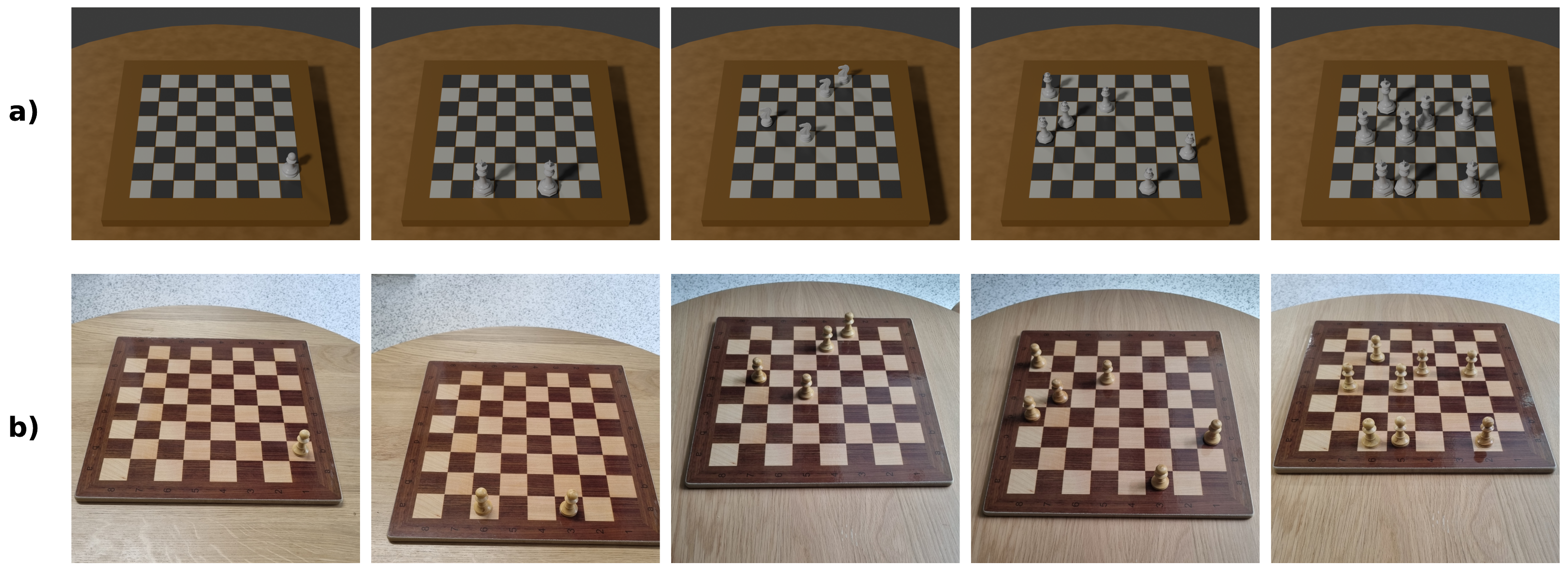}
    \caption{
        Examples of synthetic images (a) and their real reproductions (b).  
    }
    \label{fig:synth_vs_real}
\end{figure}

To assess the robustness and real-world applicability of Vision-Language Models (VLMs) on visual reasoning tasks, we evaluated the correlation between their predictions on synthetic chess images and on matched real-world reproductions. Each model was tasked with counting chess pieces ranging from 1 to 8, with synthetic and real images covering the same set of target counts. For each number of pieces, we reproduce in real life 10 chess scenes taken from the synthetic data, using their legend as ground truth (see Figure \ref{fig:synth_vs_real}).

\begin{figure}[H]
    \centering
    \includegraphics[width=0.8\linewidth]{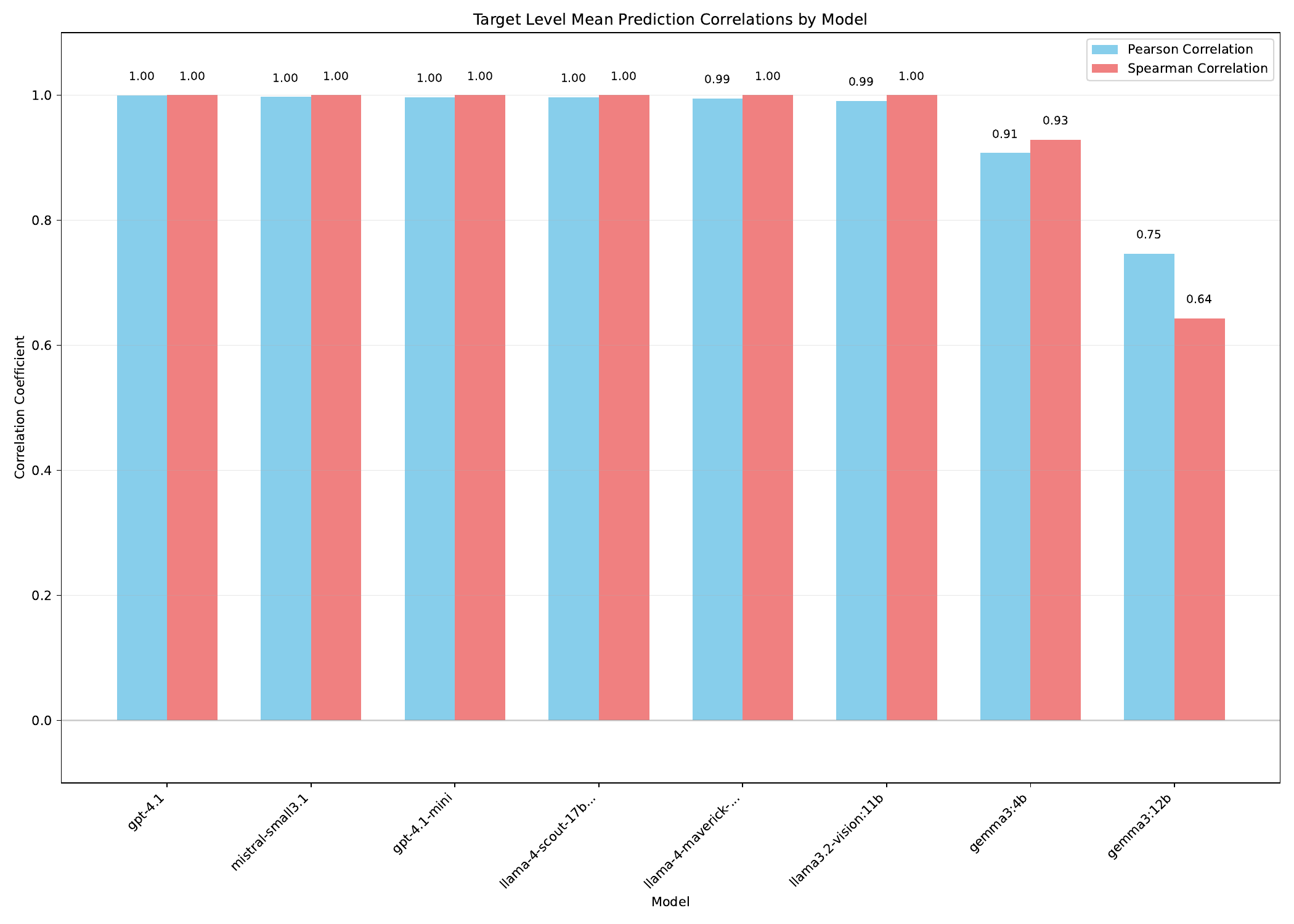}
    \caption{
        Correlation between model performance on synthetic and real images, aggregated by target level. For each number of pieces (1 to 8), we average the accuracy across all 10 samples and 6 preprompt-instruction pairs, then compute both Pearson and Spearman correlation coefficients over the 8 aggregated results.
    }
    \label{fig:target_level_correlation_summary}
\end{figure}

\begin{figure}[H]
    \centering
    \includegraphics[width=0.8\linewidth]{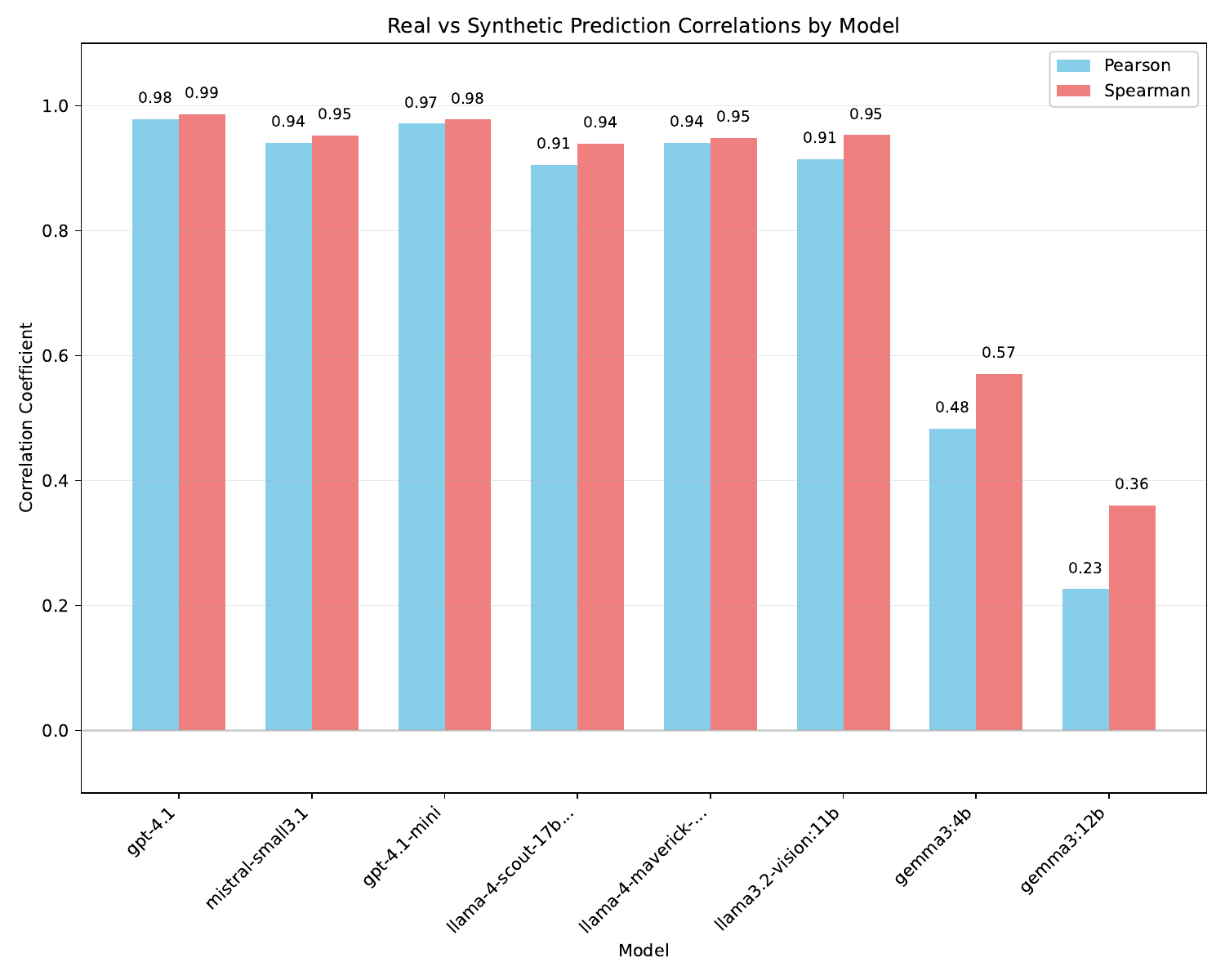}
    \caption{
        Correlation between model performance on synthetic and real images, computed at the individual prediction level. For each number of pieces (1 to 8), each preprompt-instruction pair, and each image scene (10 per level), we calculate both Pearson and Spearman correlation coefficients for accuracy. The average across all these scores is presented.
    }
    \label{fig:correlation_all_predictions_by_model}
\end{figure}

\begin{figure}[H]
    \centering
    \includegraphics[width=0.9\linewidth]{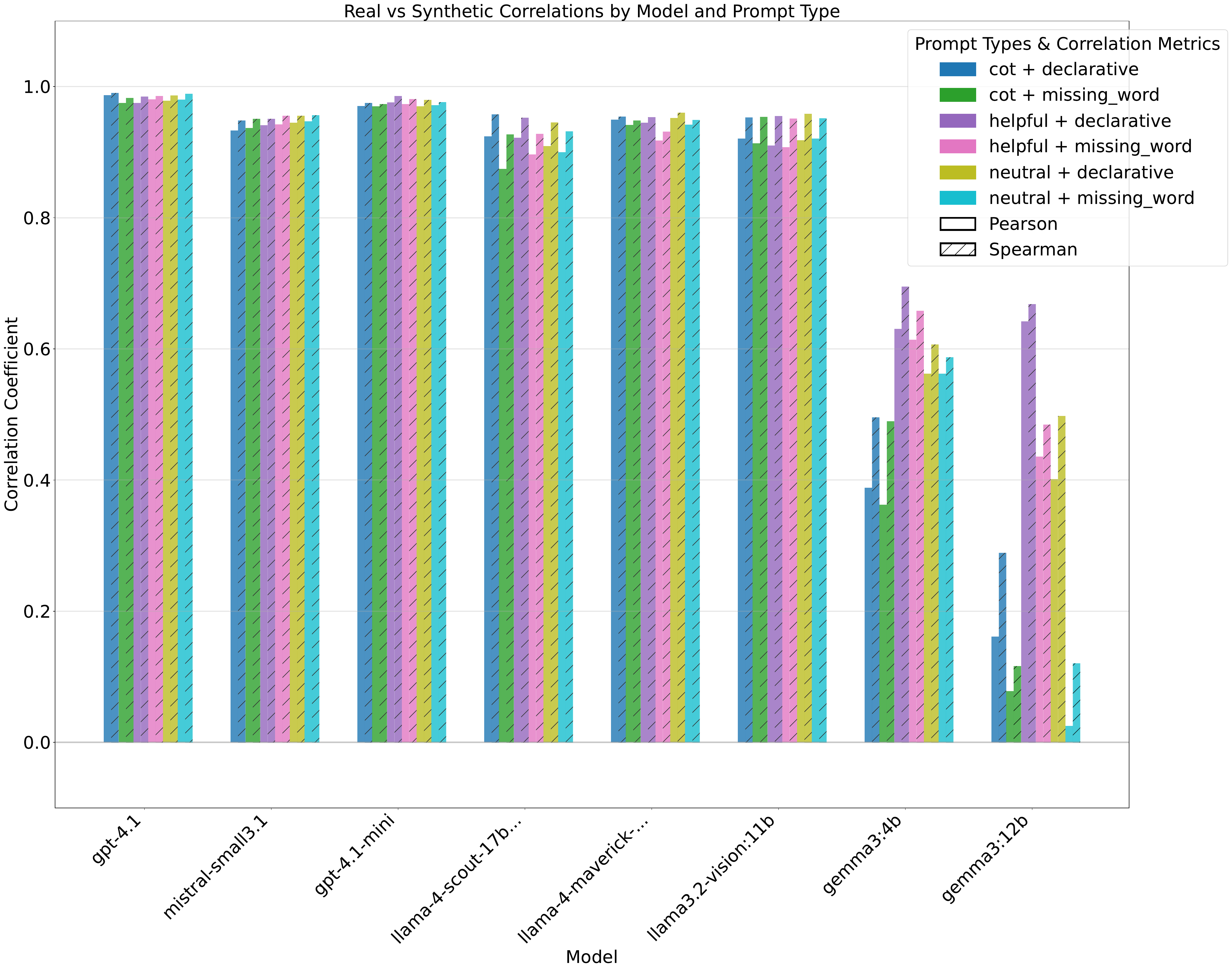}
    \caption{
        Correlation between model performance on synthetic and real images by instruction and preprompt type. For each number of pieces (1 to 8), we compute Pearson and Spearman correlation coefficients for accuracy across 10 images, then report the mean correlation for each prompt type.
    }
    \label{fig:correlations_by_prompt_type}
\end{figure}

As shown in Figures~\ref{fig:target_level_correlation_summary}\ref{fig:correlation_all_predictions_by_model}\ref{fig:correlations_by_prompt_type}, most models achieve high alignment between synthetic and real predictions. When we average over all ten images per level and all preprompt and reformulation results, Pearson and Spearman correlation coefficients reach above 0.99 on the aggregated accuracies for the strongest models (GPT-4.1, GPT-4.1-mini, Mistral-small-3.1, and both LLaMA-4 models). However, performance drops are observed for smaller or less capable models, with correlations as low as 0.64 (Gemma3:12b), indicating substantial degradation in real-world generalization for these architectures. This high overall correlation for advanced models suggests that performance on synthetic diagnostic datasets is a strong proxy for real-world counting ability in structured visual tasks.

%%%%%%%%%%%%%%%%%%%%%%%%%%%%%%%%%%%%%%%%%%%%%%%%%%%%%%%%%%%%

\end{document}